%% file: main.tex
\documentclass{article}
\usepackage{iclr2027_conference,times}
\usepackage{enumitem,graphicx,layout,lipsum,pdfcomment,soul,xcolor,xspace,tcolorbox,amsmath,amssymb, booktabs}
\usepackage{colortbl}
\definecolor{xamber}{HTML}{ECD4AA}
\usepackage{pifont}
\usepackage[normalem]{ulem}
\usepackage{hyperref}
\usepackage{wrapfig}
\input{jay_colors_util}

\def\rvx{{\mathbf{x}}}

\def\rve{{\boldsymbol{\epsilon}}}
\newcommand{\xsup}[1]{\rlap{\textsuperscript{\normalfont#1}}}
\makeatletter
\def\xthanks#1{\protected@xdef\@thanks{\@thanks\protect\footnotetext{#1}}}
\makeatother
\iclrfinalcopy
\title{Are We Overthinking One-Step Generation?}
\author{Woobin Park\xsup{$1$,$\star$}
\hspace{2.1em}
Jaeyeon Kim\xsup{$2$,$\star$}
\hspace{2.1em}
Yilun Du\xsup{$2$}
\hspace{2.1em}
Sitan Chen\xsup{$2$}
\hspace{2.1em}
Bohyung Han\xsup{$1$}
\\[1.0ex]
~$^1$Seoul National University\hspace{0.5em}
~$^2$Harvard University\hspace{0.5em}
~$^\star$Equal contribution\hspace{0.5em}
}
\date{Summer 2026}
\begin{document}
\maketitle
\vspace{-0.15in}
\begin{abstract}
Diffusion and flow matching are now dominant frameworks for generative modeling in continuous domains, but they typically require multiple network evaluations at inference. 
This has motivated extensive research on one-step generative modeling primarily built on diffusion and flow, though existing approaches often rely on complex training machinery. In this work, we introduce \textbf{Priming Models}, a simple two-stage framework for training one-step generators. Motivated by the difficulty of distributional training from scratch, Priming Models are first trained on a reconstruction objective, e.g., pixel-space denoising at a \emph{fixed} noise level, before undergoing feature-space distributional post-training.
Our central scientific finding is that this simple reconstruction stage is already enough to entirely alleviate the difficulty of distributional training.
Empirically, across class-conditional image generation, text-to-image generation, and world modeling, Priming Models significantly advance the Pareto frontier of training compute vs. generation quality. On ImageNet at $256 \times 256$ resolution, we achieve a state-of-the-art FID of $1.19$ among one-step generators using approximately one-tenth the training compute of prior methods with comparable FID. We open-source our codebase at \url{https://github.com/JaeyeonKim01/priming}.
\end{abstract}
\input{1_Introduction}
\input{2_Method}
\input{3_Experiment}

\input{5_Additional_experiments}
\input{6_Conclusion}
% \newpage
% \subsection*{AI use statement}
% We used generative AI tools to assist with language editing and rephrasing. We did not use generative AI tools to generate experimental data, develop methods, interpret experimental results, or formulate the scientific conclusions of this work. All AI-assisted text and presentation suggestions were reviewed and revised by the authors, and all references identified with AI assistance were independently verified against the original sources. The authors take full responsibility for the final content of this work.
\bibliography{main}
\bibliographystyle{iclr2027_conference}
\appendix
\input{Appendix_A_related_works}
\input{Appendix_method_details}
\input{Appendix_C_exp_details}
\input{Appendix_D_additional_results}
\input{Appendix_B_qualitative}
\end{document}

%% file: jay_colors_util.tex
\definecolor{xblue}{HTML}{4169E1}
\definecolor{xgreen}{HTML}{036C3A}
\definecolor{xpurple}{HTML}{9838B1}
\definecolor{xslategray}{HTML}{70818F}
\definecolor{xorange}{HTML}{FF8C00}
\definecolor{xcyan}{HTML}{06AEEF}
\definecolor{xred}{HTML}{FF0000}
\definecolor{xgray}{HTML}{808080}
\definecolor{xxgreen}{HTML}{009F86}
\definecolor{xsienna}{HTML}{8B4512}
\definecolor{xxpurple}{HTML}{623E99}
\newcommand{\xblue}[1]{\textcolor{xblue}{#1}}

\newcommand{\xxgreen}[1]{\textcolor{xxgreen}{#1}}
\newcommand{\xred}[1]{\textcolor{xred}{#1}}
\newcommand{\coloredhl}[2]{{\textcolor{#1}{\sethlcolor{#1!10}\hl{#2}}}\xspace}
\newcommand{\coloredul}[2]{\textcolor{#1}{\underline{#2}}\xspace}

\definecolor{ablation1}{HTML}{EEEAF7}
\definecolor{ablation2}{HTML}{E8F0F8}
\definecolor{ablation3}{HTML}{E5F4F1}
\definecolor{ablation4}{HTML}{ECF4E5}

\newcommand{\ablationhl}[2]{{\setlength{\fboxsep}{0.5pt}\colorbox{#1}{#2}}}
 \colorlet{tableHighlight}{xsienna!15}
\newcommand{\tablecolor}[1]{\rowcolor{#1}}
\tcbuselibrary{skins}
\newcommand{\mybox}[2][xblue]{%
    \begin{tcolorbox}[
        enhanced,
        colback=#1!5!white,
        frame hidden,
        boxrule=0pt,
        borderline west={2.5pt}{0pt}{#1},
        sharp corners,
        boxsep=0pt,
        left=8pt, right=8pt,
        top=6pt, bottom=6pt,
        before skip=7pt, after skip=7pt,
        fontupper=\normalsize,
        before upper={\colorlet{boxaccent}{#1}}
    ]
    #2
    \end{tcolorbox}%
}

%% file: 1_Introduction.tex
\vspace{-0.15in}
\begin{figure}[!ht]
    \centering    \includegraphics[width=0.496\textwidth,trim=16bp 0bp 3bp 0bp,clip]{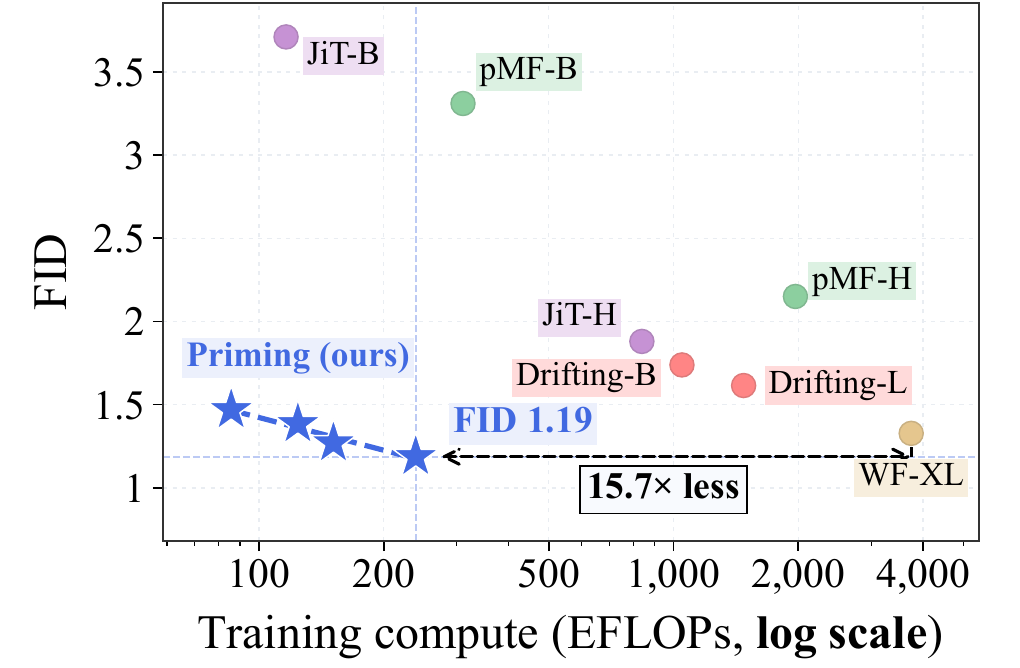}\hspace{0.008\textwidth}%
    \includegraphics[width=0.496\textwidth,trim=16bp 0bp 3bp 0bp,clip]{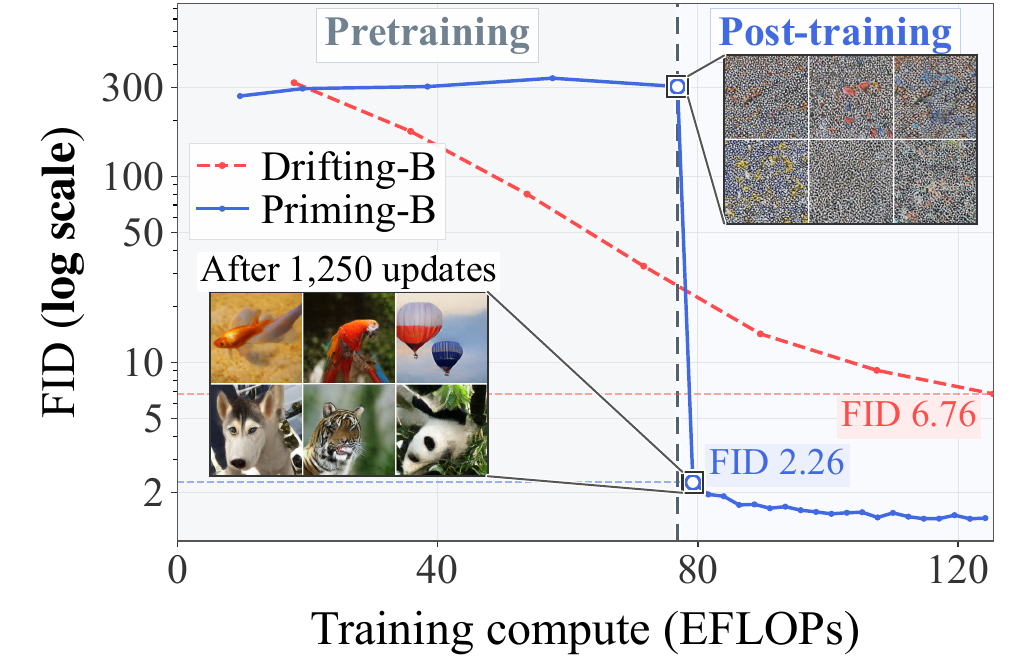}\par
    \setlength{\abovecaptionskip}{3pt}
    \vspace{-0.05in}
    \caption{\textbf{Priming Models} dramatically improve the training compute vs. FID trade-off on ImageNet (\textbf{left}). Simple reconstruction pretraining provides an effective initialization that makes subsequent distributional post-training substantially more efficient (\textbf{right}).}
    \label{fig:main}
\end{figure}
\vspace{-0.15in}
\section{Introduction}
A central goal of generative modeling is to achieve high sample quality with efficient training and inference. Diffusion and flow-based models~\citep{song2020score,ho2020denoising,lipman2022flow,albergo2025stochastic} have demonstrated remarkable generative capabilities and now serve as backbones for several domains~\citep{ho2020denoising,abramson2024accurate,chi2025diffusion}, but they rely on iterative sampling with multiple network evaluations. 

This has motivated considerable effort toward one- and few-step generation~\citep{salimans2022progressive}. Much of this progress builds on diffusion and flow formulations, including consistency-based modeling~\citep{song2023consistency,lu2025simplifying}, flow-map learning~\citep{boffi2024flow,geng2026mean,geng2026improved}, and distributional distillation~\citep{luo2023diff,yin2024improved,yin2024one}. While promising, these approaches require a considerable amount of extra machinery to work, like specialized stabilization strategies, Jacobian–vector products, classifier-free guidance, a teacher to distill from, etc.

A complementary approach is to train a generator by directly matching its output distribution to the data distribution. This naturally supports generation in a single network evaluation and has a long history in GANs~\citep{goodfellow2014generative} and maximum mean discrepancy-based generators~\citep{li2015generative,li2017mmd}. Translating these objectives into effective training, however, requires significant engineering effort~\citep{gulrajani2017improved,mescheder2018training}.

This latter approach has seen a recent resurgence of interest, with Drifting Models~\citep{deng2026generative} and W-Flow~\citep{han2026one} demonstrating that direct distributional learning can produce high-quality one-step generators at scale. Their training recipes nevertheless rely on longer training-time schedules and subtle tricks like style-conditioning~\citep{karras2019style}. Thus, the simplicity of one-step inference still comes at the cost of substantial complexity in getting these models to train.

In this work, we therefore ask: \coloredhl{xblue}{\textbf{Does training a one-step generator need to be this complicated?}} We introduce \coloredhl{xblue}{\textbf{Priming Models}}, a drastically more straightforward framework for one-step generative modeling with just two basic components:

\vspace{-0.03in}
\mybox[xblue]{
    \textbf{\textcolor{boxaccent}{
        [Stage 1] Reconstruction pretraining:}} We pretrain the network with a simple reconstruction objective \--- in fact, even denoising at a \emph{fixed} corruption level suffices.
    \vspace{0.05in}
    
    \textbf{\textcolor{boxaccent}{
        [Stage 2] Distributional post-training:}}
    We use the pretrained network directly as a one-step
    generator that maps Gaussian noise to samples, then
    refine it by matching generated and real distributions
    in feature space, e.g., using an optimal transport loss.
}
\vspace{-0.03in}
Empirically, \textbf{\xblue{Priming Models}} achieve strong performance with substantially reduced training compute across three visual domains:
\vspace{-0.03in}
\mybox[xxgreen]{
 \begin{itemize}[leftmargin=0.8em]\item \xxgreen{\textbf{ImageNet}}: Priming Models achieve state-of-the-art FID (\xred{1.19})\footnotemark and Fr\'echet distances across other feature spaces among one-step generators, outperforming Pixel MeanFlow~\citep{lu2026one}, Drifting~\citep{deng2026generative}, and even multistep diffusion baseline JiT~\citep{li2026back}, with approximately $10\times$ less compute than Drifting (Figure~\ref{fig:main}, left).
 \item \xxgreen{\textbf{Text-to-Image}}: Priming Models substantially outperform one-step MeanFlow on GenEval and DPG, while reducing training compute by a factor of \xred{1.9}.
  \item \xxgreen{\textbf{World models}}: Priming Models outperform the Drifting baseline~\citep{lu2026driftworld} with approximately half the training compute, while also achieving performance comparable to the (multi-step) flow matching baseline. \end{itemize}
}
\vspace{-0.03in}
\footnotetext{FD-loss~\citep{yang2026representation} reports FID 0.72 for pixel MeanFlow by directly optimizing Fr\'echet Distance. We exclude this configuration from comparison and revisit this in Section~\ref{sec:posttraining}.}
A key factor behind our strong empirical results is that \textbf{reconstruction pretraining substantially accelerates distributional post-training} (Figure~\ref{fig:main}, right). Without the former, even when the distributional objective provides useful correction directions, the network's Jacobian may hinder how effectively the parameters update to realize these improvements. Through controlled experiments, we demonstrate that reconstruction pretraining ameliorates this by aligning the Jacobian and shaping how readily the model's outputs can be refined.

Together, our findings suggest a surprisingly minimal design principle for generative modeling: use reconstruction pretraining to \emph{prepare} a generator for distributional post-training. Priming Models realize this principle, enabling high-quality one-step generators to be efficiently trained from scratch.
\begin{figure}[h]
    \centering
    \vspace{-15pt}
    \includegraphics[width=\textwidth]{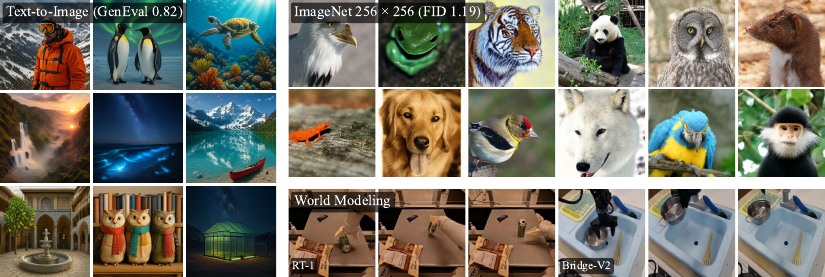}\par
    \setlength{\abovecaptionskip}{2pt}
    \setlength{\belowcaptionskip}{0pt}
    \caption{One-step samples from Priming Models across three visual domains.}
    \label{fig:main_examples}
    \vspace{-3pt}
\end{figure}

%% file: 2_Method.tex
\newpage
\section{Priming Models} \label{sec:method}
We first present our motivation behind \emph{\xblue{\textbf{Priming Models}}}. Distributional objectives, such as Sinkhorn divergence or Wasserstein distance, offer a natural approach to one-step generative modeling by directly matching generated and real distributions. The central challenge, however, lies in translating these objectives into effective training procedures, a difficulty that helped motivate interest in alternatives such as diffusion models~\citep{dhariwal2021diffusion}. Indeed, early maximum mean discrepancy (MMD)-based generators were demonstrated on relatively small datasets~\citep{li2015generative,li2017mmd,genevay2018learning}, while GANs have achieved high-quality generation at scale~\citep{goodfellow2014generative,arjovsky2017wasserstein,brock2018large,karras2019style,sauer2022stylegan} but remain sensitive to optimization~\citep{gulrajani2017improved,mescheder2018training}.
\vspace{-0.1in}
\begin{wrapfigure}[20]{r}{0.38\textwidth}
    \centering
    \begin{minipage}{\linewidth}
        \raggedright
        {\scriptsize\bfseries (a) Drifting-L}\par\vspace{1pt}
        \includegraphics[width=\linewidth]{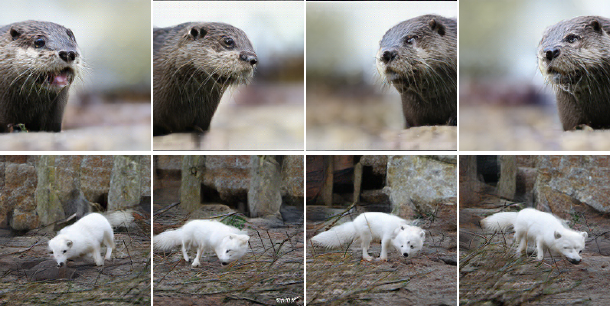}\par
        {\scriptsize\bfseries (b) W-Flow XL/2}\par\vspace{1pt}
        \includegraphics[width=\linewidth]{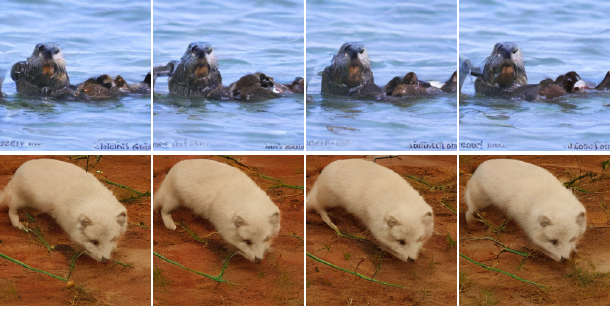}\par\vspace{-2pt}
        \setlength{\abovecaptionskip}{0pt}
        \setlength{\belowcaptionskip}{0pt}
        \caption{Fixing style embedding limits diversity (ImageNet `otter', `arctic fox'); more samples in Appendix~\ref{appendix:style_embedding}.}
        \label{fig:drifing_style_embedding}
    \end{minipage}
\end{wrapfigure}

More recently, Drifting Models~\citep{deng2026generative} and W-Flow~\citep{han2026one} have achieved high-quality one-step generation on ImageNet using velocity fields trained based on kernel- and entropic OT-based distributional distances, respectively. Their reported training compute, however, remains costly; both L-size models are trained for $1{,}280$ epochs, taking approximately \emph{three days on 64 NVIDIA H100 GPUs}. Both also augment Gaussian inputs with learned style embeddings inspired by StyleGAN~\citep{karras2019style}.

We find that fixing these style embeddings yields only marginal differences between images \emph{across different Gaussian noises} (Figure~\ref{fig:drifing_style_embedding}), although varying the style embeddings still produces diverse samples. This suggests that style conditioning is an important source of sample diversity in this setting. Taken together, these observations suggest that optimizing distributional objectives from scratch, especially without these tweaks, remains challenging. This leads us to ask:

\mybox[xblue]{
\textbf{\xblue{Question:}} Can a suitable initialization make distributional post-training more effective?
}
Our intuition is that distributional training is challenging from scratch because at the start, the network has no information about the structure of natural images. To that end, we seek to first learn a related capability: recovering clean-image structure from corrupted inputs. Denoising reconstruction provides a natural way to learn this capability. Corrupted inputs retain information about the underlying images to guide recovery, while their clean counterparts provide direct supervision.

Our hypothesis is that learning to reconstruct provides a useful initialization for subsequent distributional optimization. This motivates \emph{reconstruction pretraining} as a \xblue{\textbf{first stage}} toward learning the full noise-to-image mapping.
\vspace{-0.05in}
\subsection{Priming Models: Training recipe} \label{sec:attractor}
\vspace{-0.05in}
Building on this intuition, Priming Models train a one-step generator in two stages: reconstruction pretraining and distributional post-training. Both stages optimize the same network $f_\theta$, conditioned on an input $c$, such as a class label or text prompt. We denote the empirical training distribution by $p_{\mathrm{data}}$; here our notation suppresses the implicit dependence on $c$.

\textbf{\xblue{[Stage 1] Reconstruction pretraining.}} We first pretrain the network to reconstruct clean samples from corrupted inputs. Our main instantiation uses a conventional denoising objective~\citep{vincent2008extracting,vincent2011connection}, with the noise level $s$ sampled from a prescribed distribution $p(\cdot)$. In particular, we minimize
\begin{equation} \label{eq:pretrain_loss}
    \mathcal{L}_\theta = \mathbb{E}_{\rvx_0,\rve} \|f_\theta(\rvx_{s},c) - \rvx_0\|^2 , \quad \rvx_s = (1-s)\rvx_0 + s \rve,
\end{equation}
where $\rvx_0\sim p_{\mathrm{data}}$, $\rve\sim\mathcal{N}(0,\mathbf{I})$, $s\sim p$ are sampled independently. 
Surprisingly, we find that even taking $p$ to be a point mass at a single noise level suffices; our simplest instantiation sets $p = \delta_{0.8}$, fixing the noise level at $s=0.8$ throughout. This choice follows the intuition above: a high noise level brings the pretraining inputs closer to the pure Gaussian inputs used for one-step generation ($s=1$), while preserving information about the clean targets to guide reconstruction. We systematically examine this choice in Section~\ref{sec:pretraining}.

\coloredul{xblue}{{This stage is quite flexible}}: Our pretraining stage also accommodates other forms, for instance, masking each patch, as well as 
reconstruction in the representation space of a pretrained RAE~\citep{singh2026raev2}. We revisit these alternatives in Section~\ref{sec:pretraining}.

\textbf{\xblue{[Stage 2] Distributional post-training.}} After pretraining, we use $f_\theta$ \emph{directly as a one-step generator}: Given Gaussian noise $\rve\sim\mathcal{N}(0,\mathbf{I})$, the network yields $\hat{\rvx}_0 = f_\theta(\rve,c)$. We denote the resulting pushforward distribution by $q_\theta$, i.e., $f_\theta(\rve,c) \sim q_\theta$, with dependence on $c$ suppressed as above. We then post-train this generator by matching generated and real distributions in feature space. Given a frozen feature encoder $\phi$ and a differentiable distributional loss $\mathcal{F}$, we optimize
\begin{equation} \label{eq:post_train}
    \mathcal{L}_\theta = \mathcal{F}(\phi_{\#} q_\theta , \phi_\# p_{\mathrm{data}}),
\end{equation}
where $\phi_{\#}(\cdot)$ denotes the pushforward distribution via $\phi$. \coloredul{xblue}{The choice of $\mathcal{F}$ is simple but also flexible}, accommodating several established objectives: Sinkhorn divergence~\citep{sinkhorn1964relationship,cuturi2013sinkhorn} and maximum mean discrepancy (MMD)~\citep{glaunes2004diffeomorphic,gretton2012kernel}. We implement these objectives through the following surrogate
\begin{equation*}
\mathcal{F}(a,b) = \mathcal{D}(a,b) - \mathcal{D}(a, \xred{\mathrm{sg}}[a]),
\end{equation*}
where $\mathcal{D}$ is a functional on pairs of feature distributions and $\xred{\mathrm{sg}}$ denotes the stop-gradient. The first term encourages alignment between generated and real samples, while the second provides a repulsive self-interaction among generated samples.

Our main choice of $\mathcal{D}$ is an optimal transport (OT) cost, yielding a surrogate for the Sinkhorn divergence. Alternatively, a Laplacian kernel $\mathcal{D}(a,b) = \mathbb{E}_{x\sim a, y\sim b}[-\exp(-\left\|x-y\right\|)]$ gives us a surrogate for squared MMD. We find that even the simpler mean pairwise distance, $\mathcal{D}(a,b) = \mathbb{E}_{x\sim a, y\sim b}[\left\|x-y\right\|]$, is effective.
During training, we estimate these objectives using minibatches of samples and backpropagate through $\phi \circ f_\theta$ to update $\theta$. Appendix~\ref{appendix:distributional_posttraining} provides the justification for these surrogates and implementation details. Below we briefly review optimal transport.

\textbf{Optimal transport.} Given a function measuring the \emph{cost} of moving between samples, OT seeks the \emph{least costly} way to redistribute mass from one distribution to another. For the minibatch of generated and real samples $(\{a_i\}_{i=1}^m, \{b_j\}_{j=1}^n)$, respectively, we use the squared distance in a frozen feature space and compute an entropy-regularized transport plan:
\begin{equation*}
     C(x,y) = \frac{1}{2}\left\|x - y \right\|_2^2,\quad    \xblue{\pi^\star_\epsilon} = \mathrm{argmin}_{\pi}
\left[\sum\nolimits_{i,j} \xblue{\pi_{i,j}}C(a_i,b_j)+\epsilon\sum\nolimits_{i,j}\xblue{\pi_{ij}\log\pi_{ij}}\right],
\end{equation*}
where $\pi$ is minimized over all feasible \emph{transports}, parametrized by a nonnegative $m \times n$ matrix with row sums $1/m$ and column sums $1/n$. A \emph{transport} specifies a coupling between endpoint measures, in this case uniform over $m$ generated samples and $n$ real samples respectively, and $\pi_{ij}$ specifies how much mass is transported from $a_i$ to $b_j$. Thus, the marginal constraints ensure that the two distributions' mass is preserved. 

For $\epsilon>0$, the entropically regularized problem has a unique minimizer $\xblue{\pi^\star_\epsilon}$, which we approximate as $\xblue{\hat{\pi}_\epsilon}$ using the Sinkhorn algorithm~\citep{cuturi2013sinkhorn}. This proxy $\xblue{\hat{\pi}_\epsilon}$ yields the transport cost
\begin{equation} \label{eq:ot_def}
     \mathrm{OT}_\epsilon(a,b)  := \sum\nolimits_{i,j} \mathrm{\xred{sg}}[\xblue{\hat{\pi}_\epsilon}]_{i,j}\, C(a_i,b_j).
\end{equation}
\textbf{Comparison with other prior work.} Mainstream approaches to one-step generation build on diffusion or flow-matching trajectories~\citep{liu2023rectified,song2023consistency,boffi2024flow,geng2026mean,salimans2022progressive,yin2024one}. In contrast, Priming Models require neither prescribed trajectories nor a pretrained diffusion or flow model.

Recent work~\citep{feng2026rdm,yang2026representation} uses distributional losses to improve already strong one-step generators. In Priming Models, reconstruction pretraining produces poor initial one-step samples (Figure~\ref{fig:main}, right), and distributional post-training enables high-quality generation.

As noted at the beginning of Section~\ref{sec:method}, Drifting~\citep{deng2026generative} and W-Flow~\citep{han2026one} train models from scratch using distributional losses. Priming is more efficient than these methods and does not rely on additional modeling components such as classifier-free guidance or style embeddings; refer to Section~\ref{sec:exp}.

Most closely related to our work, D2O~\citep{zheng2025d2o} and \citet[Section~4.3]{yang2026representation} demonstrate the conversion of multi-step diffusion models into one-step generators. Priming Models build on this perspective, but our results show that starting from a multi-step model is unnecessary: even denoising at a single fixed noise level without time conditioning, as well as alternative pretraining objectives (Section~\ref{sec:pretraining}), can provide an effective initialization for distributional post-training. Appendix~\ref{app:related_work} provides further discussion.

\subsection{Priming Models: Understanding}  \label{sec:intuition}
\begin{wrapfigure}[23]{R}{0.38\textwidth}
    \centering
    \vspace{-0.2in}
\includegraphics[width=\linewidth,trim=32bp 0bp 0bp 0bp,clip]{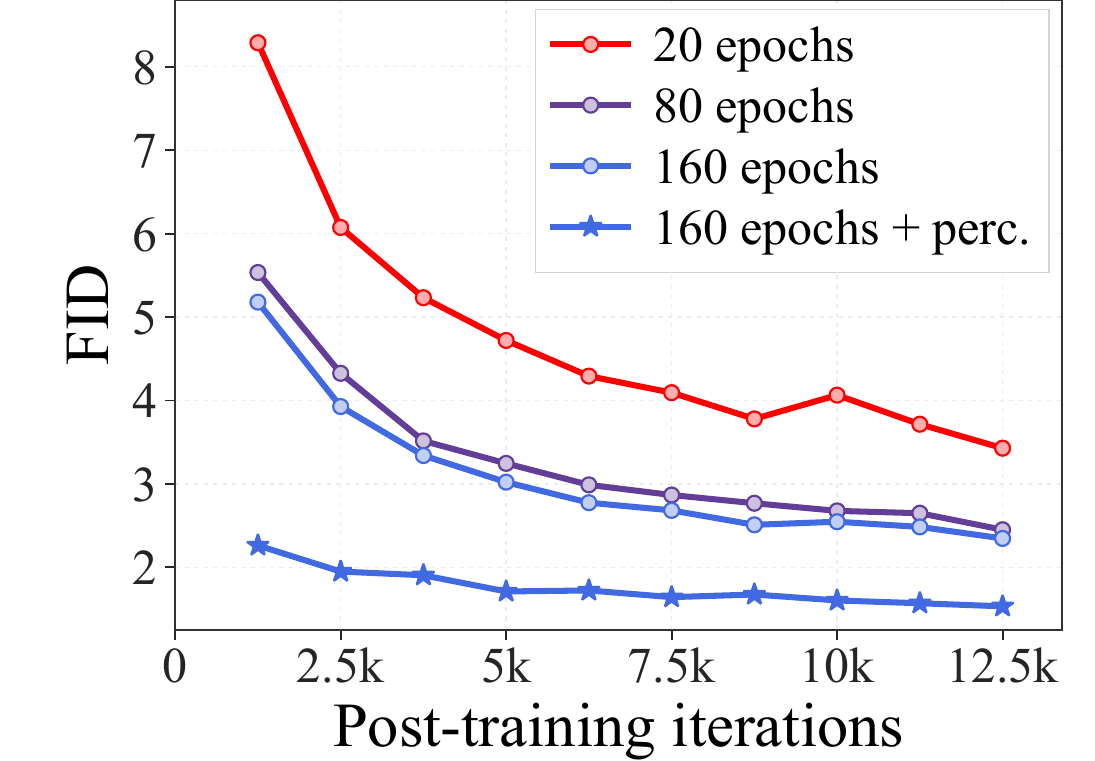}\par
    \vspace{0.01in}
    \includegraphics[width=\linewidth,trim=32bp 0bp 0bp 0bp,clip]{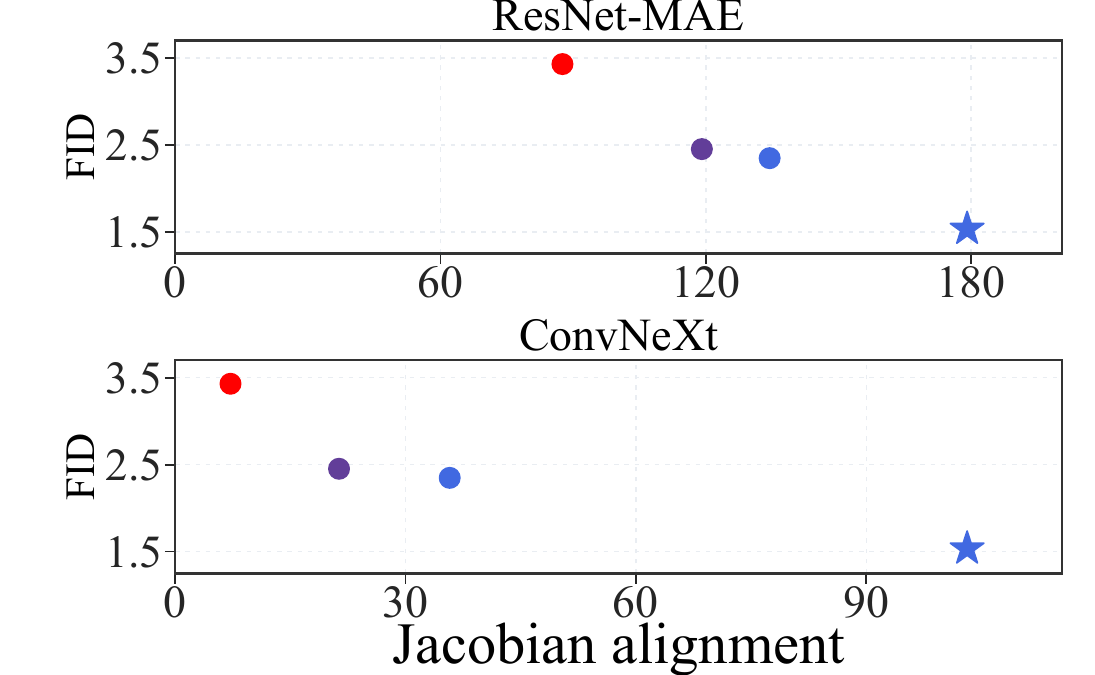}\par
    \setlength{\abovecaptionskip}{0pt}
    \setlength{\belowcaptionskip}{0pt}
    \caption{Pretraining strengthens the Jacobian alignment, improving post-training efficiency.}
    \label{fig:intuition_results}
\end{wrapfigure}

\vspace{-0.05in}
In this section, we systematically examine our intuition behind the two-stage training design of Priming Models. Figure~\ref{fig:intuition_results} shows the online FID trajectories during post-training for models pretrained on ImageNet 
for $20$, $80$, and $160$ epochs of reconstruction pretraining. It also includes a model pretrained for $160$ epochs with a perceptual loss (Section~\ref{sec:exp_imagenet}), which achieves a lower FID. At a matched number of post-training updates, longer pretraining is associated with lower FID.
\mybox[xblue]{
\textbf{\xblue{Intuition.}} We posit that reconstruction pretraining captures useful structure of the data manifold, making generated samples readily correctable through parameter updates.
}
\textbf{Jacobian alignment.} To probe this intuition, we examine how parameter updates move generated samples in feature space. Let $h_\theta$ denote the feature representations of samples, and let $v$ be the gradient of the distributional loss in \eqref{eq:post_train} with respect to $h_\theta$.
A small gradient-descent step with learning rate $\eta$ induces the feature update
\vspace{-0.04in}
\begin{equation*}
\mbox{\small$\displaystyle\Delta h_\theta \approx -\eta J_\theta J_\theta^\top v,\qquad J_\theta:=\frac{\partial h_\theta}{\partial\theta}$},
\end{equation*}
where $J_\theta$ is the Jacobian of the generated features with respect to the network parameters. Thus, $J_\theta J_\theta^\top$ determines how the loss gradient translates into a feature update. When the gradient aligns with directions to which $J_\theta J_\theta^\top$ responds strongly, the resulting feature update is larger. We examine this relationship through controlled experiments below.

\textbf{Setup.} We evaluate the four models in Figure~\ref{fig:intuition_results}, all before post-training. For each ImageNet class and in each of the ResNet-MAE and ConvNeXt feature spaces used for distributional post-training, we collect gradients of the OT loss with respect to generated features. We then apply PCA and retain the top $r=64$ directions. Writing $U$ for an orthonormal basis of this subspace, we measure its normalized squared Jacobian response as $d\| J_\theta^\top U\|_F^2/r\|J_\theta\|_F^2$, where $d$ is the total dimension of the feature vector being differentiated. Normalizing by the total squared Frobenius norm controls for differences in overall Jacobian scale across models.

\textbf{Result.} Figure~\ref{fig:intuition_results} reports the average over eight classes. Across the four models, higher normalized response before post-training (x-axis) is associated with a lower online FID after post-training, consistent with our hypothesis. We discuss the detailed setup in Appendix~\ref{appendix:exp_detail_jacobian}.

%% file: 3_Experiment.tex
\vspace{-0.1in}
\section{Experiments} \label{sec:exp}
\vspace{-0.1in}
We now demonstrate that Priming Models \coloredhl{xxgreen}{\textbf{significantly advance the Pareto frontier of training efficiency vs. generation quality}} across three visual domains: class-conditional generation on ImageNet (Section~\ref{sec:exp_imagenet}), text-to-image generation (Section~\ref{sec:t2i}), and world modeling (Section~\ref{sec:world_model}).

\vspace{-0.05in}
\subsection{ImageNet: Class-conditional generation} \label{sec:exp_imagenet}
\textbf{Setup.} We study class-conditional generation on ImageNet-1K~\citep{deng2009imagenet} at $256 \times 256$ resolution. We train Priming Models directly in pixel space at two scales: Priming-B ($16$ blocks, width $768$, and $115.1$M parameters) and Priming-L ($32$ blocks, width $1024$, and $405.1$M parameters), both using $16 \times 16$ patches. For each size, we use checkpoints after $80$ and $160$ epochs of pretraining, yielding four configurations in total. All four follow the same fixed-noise reconstruction pretraining and then receive $25{,}000$ updates of the same OT post-training procedure.

\textbf{Pretraining.} As described in Section~\ref{sec:attractor}, we pretrain the network on the reconstruction loss, at the fixed noise level $s = 0.8$. Motivated by \citet{lu2026one}, we add an auxiliary perceptual loss, where we minimize the mean squared error between the output and the real image in the feature space (\coloredul{xxgreen}{green term}), i.e.,
\begin{equation*}
    \mathcal{L}_\theta = \mathbb{E}_{\rvx_0,\rve} \left[\|f_\theta(\rvx_{s},c) - \rvx_0\|^2 + \xxgreen{\lambda  \|\phi(f_\theta(\rvx_{s},c)) - \phi(\rvx_0)\|^2}\right], \quad \lambda > 0.
\end{equation*}
In particular, we use the LPIPS~\citep{zhang2018unreasonable} and ConvNeXt~\citep{woo2023convnext} with weights $\lambda = 0.4$ and $\lambda = 0.1$, respectively. We use a global batch size of $1{,}024$ throughout.

\begin{wrapfigure}[12]{r}{0.39\textwidth}
    \centering
    \vspace{-\intextsep}
    \includegraphics[width=\linewidth,trim=12bp 0bp 3bp 0bp,clip]{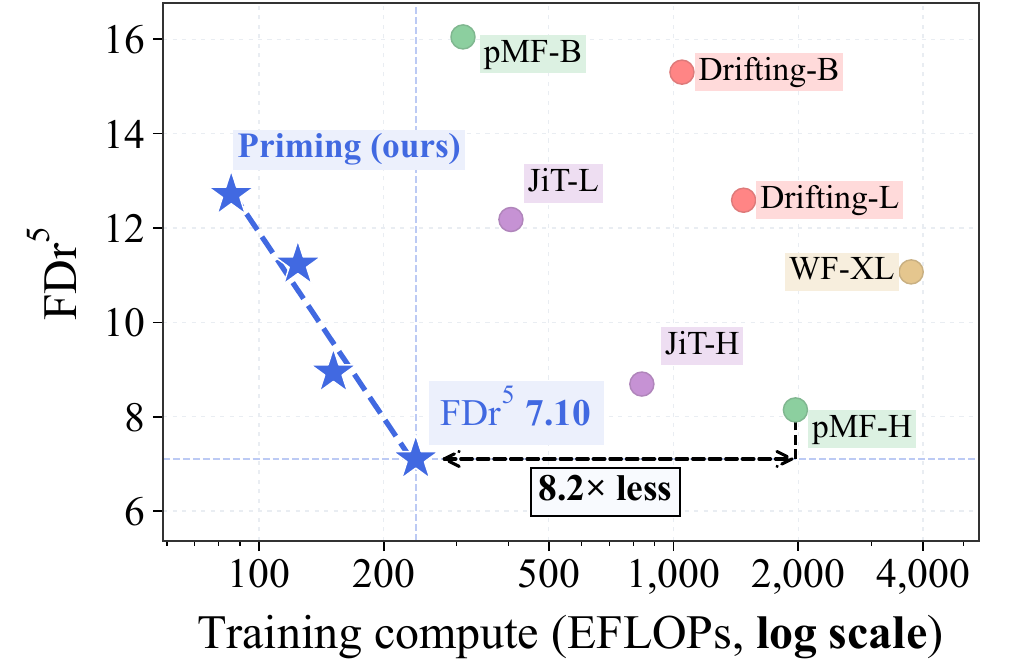}
    \setlength{\abovecaptionskip}{3pt}
    \vspace{-.2in}
    \caption{\textbf{Priming Models} on $\text{FDr}^5$.}
    \label{fig:imagenet_fdr5}
\end{wrapfigure}
\textbf{Post-training.} We now explain how we instantiate the OT setup from~\eqref{eq:ot_def} in our post-training. The regularization strength $\epsilon$ controls transport sharpness: smaller values concentrate mass on low-cost pairs, while larger values spread it more broadly. We combine both complementary signals in a weighted sum
\begin{equation*}
    \mathcal{D}(a,b)=0.75\,\mathrm{OT}_{0.001}(a,b)
+0.25\,\mathrm{OT}_{0.01}(a,b),
\end{equation*}
In practice, for each sampled ImageNet class at each training step, we generate $128$ samples online, use $64$ of them for the stop-gradient term, and draw $128$ training samples.

We apply the OT objective in the feature spaces of the ResNet-MAE and ConvNeXt encoders from Drifting Models~\citep{deng2026generative}. For each feature space, we normalize the features using a fixed, space-specific scale, then compute a separate transport plan and OT loss as described above. Post-training runs for $25{,}000$ iterations with an EMA decay of $0.999$. Further training details are provided in Appendix~\ref{appendix:exp_detail_imagenet}.

\textbf{Baselines.}
We compare against two strong recent one-step generators in pixel space: pixel Mean Flow (pMF)~\citep{lu2026one} and Drifting Models~\citep{deng2026generative}. We also include JiT~\citep{li2026back}, a diffusion model trained in pixel space, as our multi-step baseline. For completeness, we include recent one-step generators trained in the SD-VAE latent space~\citep{rombach2022high}, including W-Flow~\citep{han2026one}, and improved MeanFlow (iMF)~\citep{geng2026improved}. We omit other methods, such as (improved) Consistency Models~\citep{song2024improved}, vanilla MeanFlow~\citep{geng2026mean}, Transition Matching~\citep{wang2026transition}, as prior work has reported stronger performance with the baselines included above.

\begin{wraptable}[16]{r}{0.50\textwidth}
    \centering
    \small
    \setlength{\tabcolsep}{4pt}
    \vspace{-0.15in}
    \begin{tabular}{lrrr}
        \toprule
        Method & FID $\downarrow$ & $\text{FDr}^5$ $\downarrow$ & EFLOPs \\
        \midrule
        pMF-L & 2.55 & 10.85 & 821.75 \\
        pMF-H & 2.15 & \underline{8.14} & 1968.54 \\
        JiT-H & 1.88 & 8.69 & 839.00 \\
        iMF-XL/2 & 1.74 & 9.75 & 3201.12 \\
        Drifting-B & 1.74 & 15.30 & 1048.23 \\
        \tablecolor{tableHighlight}
        \textbf{Priming-B} (80 ep.) & 1.47 & 12.71 & 85.75 \\
        Drifting-L & 1.61 & 12.59 & 1476.39 \\
        \tablecolor{tableHighlight}
        \textbf{Priming-B} (160 ep.) & 1.38 & 11.23 & 124.18 \\
        WF-XL/2 & 1.33 & 11.07 & 3744.43 \\
        \tablecolor{tableHighlight}
  \textbf{Priming-L} (80 ep.) & 
  \underline{1.27} & 8.94 & 151.29 \\
        \tablecolor{tableHighlight}
        \textbf{Priming-L} (160 ep.) & \textbf{1.19} & \textbf{7.10} & 238.95 \\
        \bottomrule
    \end{tabular}
    \vspace{-0.1in}
    \caption{\textbf{ImageNet}. Priming Models improve the quality--compute trade-off. All methods are one-step except JiT-H (CFG scale $2.2$ and $200$ NFEs).}
    \label{tab:imagenet}
\end{wraptable}

\textbf{Quantitative results.} We report FID~\citep{heusel2017gans} as our primary metric. We evaluate our final EMA checkpoints using 50,000 generated images, balanced across the 1,000 ImageNet classes, with each image produced in a single network pass. The evaluation is conducted using three different random seeds. To assess the distributional agreement across other representation spaces, we additionally report $\text{FDr}^5$, defined as the mean normalized Fr\'echet distance across five feature spaces: Inception-V3~\citep{szegedy2016rethinking}, DINOv2~\citep{oquab2023dinov2}, ViT-MAE~\citep{he2022masked}, SigLIP~\citep{zhai2023sigmoid}, and CLIP~\citep{radford2021learning}.\footnote{$\text{FDr}^5$ follows $\text{FDr}^6$~\citep{yang2026representation}, excluding ConvNeXt because both Drifting Models and our method use it during training.}. These five evaluation encoders are not used in the training of either Priming Models or the compared baselines. We recompute FID for all methods in Table~\ref{tab:imagenet} under a common \texttt{torch-fidelity} evaluation implementation.

Alongside FID and $\text{FDr}^5$, we report training compute in EFLOPs, i.e., $10^{18}$ floating-point operations. For Priming Models, this accounts for both pretraining and distributional post-training, including feature-encoder computations. Priming-L pretrained for $160$ epochs achieves the lowest FID ($1.19$) and $\text{FDr}^5$ ($7.10$) in Table~\ref{tab:imagenet}, at a total training cost of $238.95$ EFLOPs. Moreover, all four Priming Model configurations substantially improve the Pareto frontier for training compute versus both quality metrics compared to the evaluated baselines, as shown in Figure~\ref{fig:main}(left) and Figure~\ref{fig:imagenet_fdr5}.

\textbf{Additional results.}
Figure~\ref{fig:main}(right) shows that FID falls sharply early in post-training and continues to improve through $25{,}000$ updates. Although the pretrained model initially produces noisy outputs, it generates clean images within $1{,}250$ post-training updates. Figure~\ref{fig:main_examples} presents samples from Priming-L pretrained for $160$ epochs, with more examples in Appendix~\ref{appendix:examples_imagenet}, which exhibit clear visual quality. Appendix~\ref{appendix:image_additional} reports precision and recall comparable to those of the baselines. It also includes nearest-neighbor comparisons with the ImageNet training set, which show no obvious signs of memorization.

\vspace{-0.1in}
\subsection{Text-to-Image Generation} \label{sec:t2i}
We now evaluate Priming Models on text-to-image generation at $512 \times 512$ resolution to assess their effectiveness beyond ImageNet-scale.

\textbf{Setup.} We build on MiniT2I~\citep{minit2i2026}, an open-sourced codebase for pixel-space text-to-image generation with publicly available training data. Our primary baseline is the MiniT2I MeanFlow checkpoint. For our model, we adopt the same $258$M-parameter MM-JiT-B/16 generator backbone. Text captions are encoded by the $341$M-parameter encoder of FLAN-T5-Large~\citep{chung2024scaling}, which remains frozen throughout training. Following MiniT2I, we scale the Gaussian noise by a factor of 2, i.e., $\rve \sim \mathcal{N}(0, 2^2\mathbf{I})$, throughout training and inference.

\textbf{Pretraining.} We first train with the reconstruction objective on CC12M~\citep{changpinyo2021conceptual}, then continue reconstruction training on a curated mixture of $120$k images from the codebase. Both phases use a global batch size of $1024$. We use two pretraining noise configurations: a fixed corruption level $s=0.8$ and corruption levels sampled from a reweighted logit-normal distribution $p(s) \propto s^{-2} \operatorname{LogNorm}(s; 0.8, 0.8^2)$.

\textbf{Post-training.} We then post-train the model using the OT objective in~\eqref{eq:ot_def} with $\epsilon=0.05$. At each update, we sample $64$ distinct text captions. For each caption, we generate $32$ samples online, applying stop-gradient to $16$ of them. We compute the objective independently for each caption group using globally pooled features from SigLIP, ViT-MAE, and Inception-V3. For each feature branch, we normalize features by their within-batch pairwise RMS distance. Since each caption is paired with a single real image, the generated--real OT term reduces to the average squared feature distance to that image. Post-training runs for $4{,}000$ updates without EMA.

\textbf{Multi-Step Extension.} Inspired by Consistency Models~\citep{song2023consistency}, we extend Priming Model to support multi-step generation. Specifically, during post-training, we define a finite set of noise levels $\mathcal{S}\subset(0,1]$ that includes $s=1$ and train the network to predict a clean image $f_\theta(\rvx_s,c)$ from inputs at each noise level $s\in\mathcal{S}$. We apply the same OT objective with $\rvx_s$ as the input, enabling the model to generate clean samples from both pure noise and partially noisy inputs. The multi-step post-training schedule is detailed in Appendix~\ref{appendix:exp_detail_t2i}.

At inference time, we repeatedly apply the same network along a decreasing noise schedule $1=s_0>\cdots>s_{K-1}>s_K= 0$, with $s_k\in\mathcal{S}$. We draw independent noise $\rve_0,\ldots,\rve_{K-1}\sim\mathcal{N}(0,2^2\mathbf{I})$, initialize $\rvx_{s_0}=\rve_0$, and set $\rve_K=\mathbf{0}$ for notational simplicity. Then we alternate clean-image prediction with re-noising at the next, lower noise level:
\begin{equation*}
     \hat{\rvx}_0^{(k)} =f_\theta(\rvx_{s_k},c), \quad \rvx_{s_{k+1}}=(1-s_{k+1})\hat{\rvx}_0^{(k)}+s_{k+1}\rve_{k+1},\quad k=0,\ldots,K-1.
\end{equation*}
The final step returns a clean-image prediction without re-noising. Thus, we return $\rvx_{s_K}=\rvx_0$ after $K$ network evaluations.

\textbf{Evaluation.}
We evaluate text--image alignment using GenEval~\citep{ghosh2023geneval} and DPG-Bench~\citep{hu2024ella}. We report the MiniT2I MeanFlow baseline's performance with both one- and four-step sampling, using its default guidance scale of $12.0$ and guidance interval of $[0.2, 0.8]$. We also include a multistep pixel-space flow-matching baseline. Additional training and evaluation details are provided in Appendix~\ref{appendix:exp_detail_t2i}.

\begin{wraptable}[14]{r}{0.55\textwidth}
    \centering
    \setlength{\tabcolsep}{3pt}
    \resizebox{\linewidth}{!}{%
    \begin{tabular}{lrrr}
        \toprule
        Method & GenEval $\uparrow$ & DPG-Bench $\uparrow$ & EFLOPs \\
        \midrule
        Flow matching (100 steps)$^\dagger$ & \textbf{0.873} & \textbf{84.20} & 408.04 \\
        MeanFlow (4 steps)$^\dagger$ & 0.842 & 83.23 & 546.64 \\
        \tablecolor{tableHighlight}
        \textbf{Priming} (2 steps) & 0.844 & 82.81 & 455.02\\
        \midrule
        MeanFlow (1 step) & 0.548 & 76.49 & 546.64 \\
       \tablecolor{tableHighlight}
        \textbf{Priming} ($s=0.8$, 200k) & 0.740 & 81.30 & 293.53 \\
        \tablecolor{tableHighlight}
        \textbf{Priming} (logit-normal, 290k) & \textbf{0.818} & \textbf{82.16} & 420.16 \\
        \bottomrule
    \end{tabular}}
    \vspace{-0.1in}
    \caption{\textbf{Text-to-Image.} Priming Models outperform one-step MeanFlow; the logit-normal variant approaches four-step MeanFlow with less training compute. Priming extended to two-step is competitive with four-step MeanFlow. The scores of $^\dagger$ are from \citet{minit2i2026}.}
    \label{tab:t2i}
\end{wraptable}
\textbf{Results.} As shown in Table~\ref{tab:t2i}, Priming pretrained for $290$k updates with a logit-normal noise schedule achieves the best results among one-step generators: $0.818$ on GenEval and $82.16$ on DPG-Bench, outperforming one-step MeanFlow on both metrics. Despite requiring only one network evaluation per image, its scores approach those of four-step MeanFlow and the $100$-step flow-matching baseline. 
With a fixed corruption level of $s=0.8$ and $200$k pretraining updates, Priming-B reaches $0.740$ on GenEval and $81.30$ on DPG-Bench at a lower total training cost of $293.530$ EFLOPs, $1.86\times$ reduction relative to the MeanFlow baseline. With multi-step extension, the 2-step Priming achieves $0.844$ on GenEval and $82.81$ on DPG-Bench, achieving a competitive result to four-step MeanFlow. We show several generated samples in Figure~\ref{fig:main_examples}, with further examples in Appendix~\ref{appendix:examples_t2i}.

\subsection{World Modeling} \label{sec:world_model}
\vspace{-.15in}
\begin{table*}[h]
    \centering
    \caption{\textbf{World modeling.} Priming Model outperforms both the one-step DriftWorld baseline and a 32-step flow-matching baseline across all quality metrics while using less compute than DriftWorld.}
    \label{tab:world_modeling}
    \small
    \setlength{\tabcolsep}{3.5pt}
    \resizebox{\textwidth}{!}{%
    \begin{tabular}{llc ccc cc cc}
        \toprule
        & & & \multicolumn{5}{c}{Generation quality} & \multicolumn{2}{c}{Efficiency} \\
        \cmidrule(lr){4-8} \cmidrule(lr){9-10}
        Dataset & Method & NFE & SSIM $\uparrow$ & PSNR $\uparrow$ & LPIPS $\downarrow$ & FID $\downarrow$ & FVD $\downarrow$ & Time/frame (ms) $\downarrow$ & Train (EFLOPs) $\downarrow$ \\
        \midrule
        & Flow Matching & $32$ & $0.85$ & $25.428$ & $0.076$ & $7.8$ & $\mathbf{27.24}$ & $322$ & 12.37 \\
        \textbf{RT-1} & DriftWorld~\citep{lu2026driftworld} & $1$ & $0.841$ & $24.203$ & $0.097$ & $8.99$ & $62.84$ & $26$ & $30.71$ \\
        \tablecolor{tableHighlight}
        \cellcolor{white} & \textbf{Priming} & $1$ & $\mathbf{0.878}$ & $\mathbf{26.765}$ & $\mathbf{0.063}$ & $\mathbf{3.16}$ & $30.96$ & $\mathbf{19}$ & $\mathbf{12.37}$ \\
        \midrule
        & Flow Matching & $32$ & $0.841$ & $23.37$ & $0.079$ & $11.02$ & $39.87$ & $445$ & 9.55 \\
        \textbf{Bridge-V2} & DriftWorld~\citep{lu2026driftworld} & $1$ & $0.835$ & $22.691$ & $0.091$ & $4.86$ & $46.05$ & $30$ & $37.05$ \\
        \tablecolor{tableHighlight}
        \cellcolor{white} & \textbf{Priming} & $1$ & $\mathbf{0.857}$ & $\mathbf{24.549}$ & $\mathbf{0.067}$ & $\mathbf{2.82}$ & $\mathbf{32.58}$ & $\mathbf{23}$ & $\mathbf{9.50}$ \\
        \bottomrule
    \end{tabular}
    }
    \vspace{-0.1in}
\end{table*}
Finally, we apply Priming Models to action-conditioned world models on two robot-manipulation domains: RT-1~\citep{brohan2023rt1} and 
Bridge-V2~\citep{walke2023bridgedatav2}. In particular, we train a Priming Model to predict the next frame conditioned on recent observation and action histories.

\textbf{Setup.}
We follow the data processing pipeline and the architecture choice of U-Net with action conditioning from DriftWorld~\citep{lu2026driftworld}. For both datasets, frames are resized to $256 \times 256$ and then encoded into the SD3-VAE~\citep{esser2024scaling} latent space, in which our Priming Model operates. Our primary baseline is DriftWorld and $32$-step Flow Matching, trained with approximately the same EFLOPs as Priming.

\textbf{Training.}
We first perform reconstruction pretraining for $64$ epochs using the reweighted logit-normal distribution from Section~\ref{sec:t2i}, with a global batch size of $64$ and an EMA decay of $0.9999$. The distributional post-training begins from these weights. At each update, we process $32$ conditions with $16$ live and $16$ stop-gradient generated samples, together with one real next frame per condition. We use the OT objective~\eqref{eq:ot_def} with $\epsilon=0.05$, and pairwise-RMS feature scaling. Post-training runs for $4{,}000$ updates.

\textbf{Evaluation.}
We evaluate EMA checkpoints on the held-out validation sets, with a single network evaluation per predicted frame. SSIM, PSNR, and LPIPS metrics are used to compare generated frames to the corresponding ground-truth frames; FID and FVD~\citep{unterthiner2018towards} measure distributional agreement.
Both datasets are evaluated in 8-frame autoregressive chunks, re-anchoring on the real trajectory between chunks. FVD uses the first 16 predicted frames of each trajectory. We additionally report the time required to generate a frame and EFLOPs, including both pretraining and post-training.
Complete conditioning, training, and rollout details are provided in Appendix~\ref{appendix:exp_detail_wm}.

\textbf{Results.} Table~\ref{tab:world_modeling} reports generation quality and training compute on RT-1 and Bridge-V2. Despite generating each sample in a single step, Priming Models outperform both the $32$-step flow-matching baseline and DriftWorld on every reported quality metric, except FVD on RT-1. Compared with DriftWorld, they use $59.7\%$ less training compute on RT-1 ($12.37$ versus $30.71$ EFLOPs) and $74.4\%$ less on Bridge-V2 ($9.50$ versus $37.05$ EFLOPs). Notably, we do not rely on additional techniques used by DriftWorld, such as motion weighting or self-forcing. We provide generated samples in Figure~\ref{fig:main_examples} and Appendix~\ref{appendix:examples_wm}.

%% file: 5_Additional_experiments.tex
\vspace{-0.1in}
\section{Additional Experiments}
\vspace{-0.05in}
In this section, we further investigate the training design of Priming Models using the ImageNet setup described in Section~\ref{sec:exp_imagenet}. We first examine pretraining design choices, including alternatives to denoising, in Section~\ref{sec:pretraining}, then compare post-training objectives beyond OT in Section~\ref{sec:posttraining}.
\vspace{-0.1in}
\subsection{Examining pretraining design choices} \label{sec:pretraining}
\begin{table*}[h]
    \centering
    \vspace{-0.2in}
     \caption{\textbf{Ablation study}. \protect\ablationhl{ablation1}{w/o perc.} removes perceptual loss; \mbox{\protect\ablationhl{ablation1}{Masking/RAE}} change pretraining. \protect\ablationhl{ablation2}{Energy/Laplacian} replace OT; \protect\ablationhl{ablation3}{SigLIP/MAE/Inception} changes the post-training feature space.}
     \vspace{0.03in}
    \label{tab:ablation}
    \small
    \setlength{\tabcolsep}{2pt}
    \begin{tabular*}{\textwidth}{@{\extracolsep{\fill}}l
        >{\columncolor{tableHighlight}}c
        *{3}{>{\columncolor{ablation1}}c}
        *{2}{>{\columncolor{ablation2}}c}
        >{\columncolor{ablation3}}c@{}}
        \toprule
        & \multicolumn{1}{>{\columncolor{tableHighlight}}c}{\textbf{Reference}}
        & \multicolumn{3}{>{\columncolor{ablation1}}c}{\textbf{Pretraining}}
        & \multicolumn{2}{>{\columncolor{ablation2}}c}{\textbf{Post-training (loss)}}
        & \multicolumn{1}{>{\columncolor{ablation3}}c}{\textbf{Post-training (feature)}} \\
        \cmidrule(lr){2-2}\cmidrule(lr){3-5}\cmidrule(lr){6-7}\cmidrule(l){8-8}
        & \textbf{Priming-B (160 ep.)} & w/o perc. & Masking & RAE
        & Energy & Laplacian & \footnotesize{SigLIP/MAE/Inception} \\
        FID $\downarrow$ & \textbf{1.38} & 2.07 & 1.87 & 2.34
        & 1.95 & 1.85 & 1.65 \\
        $\text{FDr}^5$ $\downarrow$ & \textbf{11.23} & 16.13 & 16.14 & 14.27
        & 11.55 & 11.22 & 6.13 \\
        \bottomrule
    \end{tabular*}
\end{table*}
\textbf{Pretraining with masking.}
We substitute the denoising corruption with patch-wise masking: randomly selecting $80\%$ of the image patches and replacing them with independently sampled Gaussian noise, leaving the remaining patches uncorrupted. We train Priming-B to reconstruct $\rvx_0$ using the same reconstruction and perceptual losses as in the default pretraining. After $160$ epochs of pretraining followed by $12{,}500$ steps of distributional post-training, the model achieves an FID of $1.87$ with an EMA decay of $0.999$ (Table~\ref{tab:ablation}; \mbox{\protect\ablationhl{ablation1}{Masking}}).

\textbf{Pretraining as distillation from RAE.}
We next demonstrate that representation distillation provides an alternative initialization for distributional post-training. Using the frozen encoder of a pretrained RAEv2~\citep{singh2026raev2} as a teacher, we train a class-conditioned student encoder with 300M parameters to recover clean-image representations from corrupted inputs. Specifically, we minimize $\mathbb{E}|f_\theta(\rvx_s,c)-\mathrm{\xred{sg}}[E_{\mathrm{teacher}}(\rvx_0)]|_2^2$, matching the student's output to the teacher's representation of the corresponding clean image. We then compose the trained student encoder with the pretrained, frozen RAEv2 decoder to produce images in pixel space. After $12{,}500$ steps of distributional post-training, the model achieves an FID of $2.34$ with an EMA decay of $0.999$ (Table~\ref{tab:ablation}; \mbox{\protect\ablationhl{ablation1}{RAE}}).

\textbf{Ablation study: Perceptual loss.}
We compare Priming-B-160 (Section~\ref{sec:exp_imagenet}) with the same 160-epoch pretraining without perceptual loss. After 25,000 matched OT updates, removing perceptual loss worsens terminal EMA FID from $1.38$ to $2.07$ and $\text{FDr}^5$ from $11.23$ to $16.13$ (Table~\ref{tab:ablation}; \mbox{\protect\ablationhl{ablation1}{w/o perc.}}). We hypothesize that supervision in external feature spaces helps the model capture useful structure of the data manifold.

\textbf{Ablation study: Noise level.}
We next examine the effect of the noise level used during pretraining. Using the same Priming-B pipeline, we pretrain each model for $80$ epochs at a fixed noise level, comparing our default $s=0.8$ with $s\in\{1.0,0.9,0.7,0.5\}$.
After $12{,}500$ OT post-training updates, $s=0.8$ yields the lowest terminal EMA FID ($1.65$), followed by $s=0.7$ ($1.76$), $s=0.9$ ($1.97$), $s=0.5$ ($2.46$), and $s=1.0$ ($65.08$). These results support our intuition that the effective noise level should balance retaining the clean image's information with keeping the noise level close to pure Gaussian noise.
\vspace{-0.05in}
\subsection{Examining post-training design choices} \label{sec:posttraining}
\textbf{Other choices of $\mathcal{D}$.} We recall our post-training objective $\mathcal{F}(a,b) = \mathcal{D}(a,b) - \mathcal{D}(a, \xred{\mathrm{sg}}[a])$. Starting from the same Priming-B checkpoint after 160 epochs of pretraining, we evaluate two alternatives to OT in the full ResNet-MAE and ConvNeXt feature spaces: the pairwise distance $\mathcal{D}(a,b) = \mathbb{E}_{x\sim a, y\sim b}[\left\|x-y \right\|_2]$ and negative Laplacian-kernel similarity $\mathcal{D}(a,b) = \mathbb{E}_{x\sim a, y\sim b}[-\exp(-\left\|x-y \right\|_2)]$. At $25{,}000$ updates, their terminal EMA FIDs are $1.95$ and $1.85$, respectively, compared with $1.38$ for OT (Table~\ref{tab:imagenet}). The $\text{FDr}^5$ are $11.55$ and $11.22$, compared with $11.23$ for OT (Table~\ref{tab:ablation}; \protect\ablationhl{ablation2}{Energy/Laplacian}). These results show that Priming Models work with multiple distributional losses, with OT yielding the best FID among the choices tested. 

\textbf{Other choices of feature space.} We also replace the full ResNet-MAE and ConvNeXt feature spaces with simpler, lightweight spaces, SigLIP, Inception-V3, and ViT-MAE, keeping the same perceptual Priming-B-160 initialization and $25{,}000$ OT post-training setup. The terminal EMA checkpoint reaches FID $1.65$ and $\text{FDr}^5$ $6.13$, versus $1.38$ and $11.23$ for the full-feature OT reference (Table~\ref{tab:imagenet}, and Table~\ref{tab:ablation}; \protect\ablationhl{ablation3}{SigLIP/MAE/Inception}). Inception, SigLIP, and ViT-MAE are components of $\text{FDr}^5$, so the improvement in $\text{FDr}^5$ is not an independent held-out-feature result.

\textbf{FD-loss.} Fr\'echet distance~\citep{dowson1982frechet} is widely used to evaluate generative models, but directly optimizing it was believed to be challenging, as reliable estimation typically requires tens of thousands of samples. FD-loss~\citep{yang2026representation} addresses this challenge by accumulating feature statistics across training steps through a feature queue or exponential moving averages, while backpropagating only through the current batch, and has demonstrated consistent improvements in pretrained one-step generators. 
We show that FD-loss also fits within the Priming Model framework as an alternative distributional post-training objective. Applied to the pretrained Priming-B checkpoint, it yields efficient post-training; we provide the setup and results in Appendix~\ref{appendix:fd_loss}.

%% file: 6_Conclusion.tex
\section{Conclusion}
In this work, we introduce Priming Models, a simple two-stage framework for training one-step generative models end-to-end. Reconstruction pretraining prepares the generator for efficient distributional post-training, enabling strong empirical performance across multiple visual domains.

That said, our work leaves several open questions. Priming Models rely on pretrained feature encoders. Although we demonstrate our method with multiple sets of feature encoders, this dependence raises the question of whether it can be eliminated. The relationship between reconstruction pretraining and Jacobian alignment also warrants deeper investigation. Finally, closing the remaining performance gap with multi-step diffusion and flow models remains an important challenge.

\textbf{Outlook.} More broadly, our results suggest that we may be overthinking the problem: the specialized formulations and elaborate training machinery used in prior work are not the only path to train one-step generators. We hope our work opens the door to simpler and more scalable approaches to one-step generative modeling.

\section*{Acknowledgments.}
The authors acknowledge support from the Kempner Institute. JK acknowledges support from the Jump Trading Fellowship and Kempner Graduate Fellowship. SC acknowledges support from NSF CAREER award CCF-2441635 and the Harvard Dean's Competitive Fund for Promising Scholarship.

%% file: Appendix_A_related_works.tex
\newpage
\section{Related Work}
\label{app:related_work}
In this section, we review prior work on one- and few-step generative modeling, organized by their methodological category.

\textbf{\protect\ablationhl{ablation1}{Category 1: Trajectory straightening and flow-map learning.}}
These lines of work obtain few-step generative models by learning from a diffusion or probability-flow trajectory. Progressive distillation repeatedly compresses a teacher sampler into fewer steps~\citep{salimans2022progressive}. Consistency Models instead enforce agreement among predictions made at different points along the same trajectory~\citep{song2023consistency}, while subsequent work improves their stability and scaling or extends consistency to trajectory segments~\citep{geng2025easy,song2024improved,lu2025simplifying,kim2024consistency}. Rectified Flow and reflow~\citep{liu2023rectified,liu2023instaflow} seek straighter transport paths that are easier to integrate in few steps. These straightening methods modify the coupling and resulting trajectory so that coarse numerical integration can accurately reach its endpoint.

A related but distinct family learns the finite-time transition across an interval of a flow-matching trajectory, such as a flow map or average velocity. This includes Flow Map Matching, Shortcut Models, MeanFlow, Improved MeanFlow, and Inductive Moment Matching~\citep{boffi2024flow,frans2025shortcut,geng2026mean,geng2026improved,zhou2025inductive}. These methods differ in their targets and training estimators, but generally supervise a particular time-dependent transport or consistency relation.

\coloredhl{xblue}{\textbf{Priming:}} Priming Models do not attempt to recover a prescribed trajectory or flow map: reconstruction pretraining provides an initialization for the network, which we then directly use as a one-step generator.

\textbf{\protect\ablationhl{ablation2}{Category 2: Teacher-guided distillation.}}
Another line of work distills a pretrained diffusion teacher into a one-step student, with much of the focus on designing effective distillation objectives. Diff-Instruct~\citep{luo2023diff} uses score-based distributional objectives derived from a diffusion teacher, while DMD and DMD2~\citep{yin2024one,yin2024improved} optimize the student using the difference between teacher and student score estimates. Score Identity Distillation~\citep{zhou2024sid} provides a related score-based formulation. Adversarial Diffusion Distillation combines score distillation with an adversarial objective to improve fidelity in the low-step regime~\citep{sauer2024add}.

\coloredhl{xblue}{\textbf{Priming:}} These approaches inherit the cost and machinery required to obtain and query that teacher. In contrast, Priming Models are trained from scratch and therefore do not require a pretrained teacher.

\textbf{\protect\ablationhl{ablation3}{Category 3:Distributional loss (from scratch).}}
Directly matching a generator distribution to the data distribution has a long history. Generative Adversarial Networks (GANs) learn a one-pass generator through an adversarially trained discriminator~\citep{goodfellow2014generative}, with Wasserstein objectives and regularization improving the geometry and stability of training~\citep{arjovsky2017wasserstein,gulrajani2017improved,mescheder2018training}. Generative Moment Matching Networks use maximum mean discrepancy (MMD) in place of adversarial discrimination, while MMD-GANs further learn the MMD kernel adversarially~\citep{li2015generative,li2017mmd}. \citet{genevay2018learning} trains implicit generative models directly using a Sinkhorn divergence. These works, however, are limited in empirical scope and, more importantly, differ from Priming Models in that we apply the distributional objective after a reconstruction pretraining stage.

Recent works revisit direct one-step learning at a more practical scale. YOSO~\citep{luo2024yoso} stabilizes adversarial one-step training through a self-cooperative diffusion GAN, while Adversarial Flow~\citep{lin2025adversarial} trains with an adversarial objective but constrains the generator to learn the deterministic transport map associated with flow matching. 

\coloredhl{xblue}{\textbf{Priming:}} Priming Models share the goal of direct distribution matching, but do not optimize it from random initialization: reconstruction pretraining first teaches the generator to correct corrupted inputs toward clean images, after which a distributional objective refines it into a one-step generator.

\textbf{\protect\ablationhl{ablation3}{Category 3-1: Drifting Models.}}
Drifting Models~\citep{deng2026generative} construct an attraction-repulsion field using a normalized Laplacian kernel and train the generator to match stop-gradient targets obtained by displacing its generated features along this field. W-Flow~\citep{han2026one} derives a training-time velocity field from the Wasserstein gradient flow of the Sinkhorn divergence, using barycentric projections computed from entropic optimal transport plans. \citet{van2026generate} additionally examine how the choice of frozen feature space affects transport stability in distribution matching.

\coloredhl{xblue}{\textbf{Priming:}} Priming Models introduce reconstruction pretraining before distributional optimization, and this stage is key to our strong compute vs. quality trade-off.
Our implementation also seeks to dispense with the additional engineering machinery used in these works. Both reported ImageNet architectures include learned random style embeddings, and both frameworks support classifier-free guidance; Drifting additionally normalizes its constructed drift field. Priming does not use these additional mechanisms and instead optimizes an OT-based distributional surrogate. All four models reported in Section~\ref{sec:exp} share exactly the same post-training configuration, further illustrating the simplicity of our approach.

This suggests that reconstruction pretraining can make distributional optimization more efficient while supporting a simpler training recipe. We hypothesize that this initialization reduces the need for additional training mechanisms, although our experiments do not isolate their individual contributions.

\textbf{\protect\ablationhl{ablation4}{Category 4: Distributional loss (fine-tuning).}}
Distributional objectives have also been used to adapt pretrained generative models. MMD-DDM~\citep{aiello2023mmd} fine-tunes pretrained diffusion models with MMD to improve quality under a restricted sampling-step budget. DiffRatio~\citep{chen2025diffratio} estimates the density ratio between noisy real and generated distributions to train a one-step model without a teacher score. FD-loss~\citep{yang2026representation} makes feature-space Fr\'echet distance practical as a training objective by maintaining population-level feature statistics, and applies it to pretrained one- and multi-step generators. More recently, iRDM~\citep{feng2026rdm} scales MMD-based post-training across multiple pretrained representation spaces. 

Most closely related to our work, D2O~\citep{zheng2025d2o} shows that adversarial fine-tuning alone can convert a pretrained diffusion model into a one-step generator, while \citet[Section~4.3]{yang2026representation} show that FD-loss can convert JiT, a multi-step diffusion model, into a competitive one-step generator.

\coloredhl{xblue}{\textbf{Priming:}} Priming Models build on the perspective that pretraining can prepare a network for subsequent distributional optimization. We investigate how simple this pretraining task can be, showing that reconstruction—even denoising at a single fixed noise level without noise-level conditioning—can provide an effective initialization for our distributional training, despite poor initial one-step samples (Figure~\ref{fig:main}, right). We further examine alternative pretraining tasks (Section~\ref{sec:pretraining}) within the same framework. 

Finally, viewing reconstruction pretraining and distributional post-training as a complete training procedure also enables comparisons with prior methods based on total training compute across both stages, measured in EFLOPs.

%% file: Appendix_method_details.tex
\section{Distributional Post-Training Details}
\label{appendix:distributional_posttraining}
\newcounter{posttrainalgorithm}

This section specifies the objective and implementation used for distributional post-training. It first establishes why the stop-gradient loss in Section~\ref{sec:attractor} minimizes a standard distributional discrepancy, then gives its finite-batch form, conditional feature construction, normalization, and optimization algorithm. Dataset- and model-specific choices are provided in Appendix~\ref{appendix:exp_details}.

\subsection{Why the stop-gradient objective minimizes a divergence}
\label{appendix:stop_gradient_justification}

Let $q_\theta$ and $p$ denote the generated and real feature distributions, respectively, and let $\mathcal D$ be a symmetric, differentiable functional of two distributions. Consistent with the main text, write the practical stop-gradient objective as
\begin{equation*}
    \mathcal F(q_\theta,p)
    :=
    \mathcal D(q_\theta,p)
    -\mathcal D(q_\theta,\mathrm{sg}[q_\theta]).
    \label{eq:appendix_stop_gradient_objective}
\end{equation*}
The real self-interaction is constant with respect to $\theta$, while symmetry makes the derivatives through the two arguments of $\mathcal D(q_\theta,q_\theta)$ equal. The chain rule therefore gives
\begin{equation}
    \nabla_\theta\mathcal F(q_\theta,p)
    =
    \nabla_\theta\!\left[
        \mathcal D(q_\theta,p)
        -\tfrac12\mathcal D(q_\theta,q_\theta)
        -\tfrac12\mathcal D(p,p)
    \right].
    \label{eq:appendix_stop_gradient_identity}
\end{equation}
Thus, $\mathcal F$ need not equal the expression inside the brackets in forward value, but it has the same parameter gradient. This is the mathematical justification for the objective in the main text: the stop-gradient form removes a constant term and avoids evaluating two identical derivatives of the generated self-interaction.

For $\mathcal D=\mathrm{OT}_\epsilon(\cdot,\cdot)$, we first distinguish the full entropy-regularized OT objective from the cost-only quantity. Define the regularized optimal plan, the full OT objective value, and its cost-only counterpart by
\begin{equation}
\begin{aligned}
    \pi_\epsilon^\star(r,s)
    &=\underset{\pi\in\Pi(r,s)}{\operatorname{argmin}}
      \left\{
        \langle \pi,C\rangle
        +\epsilon\,\operatorname{KL}(\pi\,\|\,r\otimes s)
      \right\}, \\
    \mathcal T_\epsilon(r,s)
    &=\left\langle \pi_\epsilon^\star(r,s),C\right\rangle
      +\epsilon\,\operatorname{KL}
       \!\left(\pi_\epsilon^\star(r,s)\,\|\,r\otimes s\right), \\
    \mathrm{OT}_\epsilon(r,s)
    &=\left\langle
       \operatorname{sg}[\pi_\epsilon^\star(r,s)],C
      \right\rangle.
\end{aligned}
    \label{eq:appendix_regularized_ot_value}
\end{equation}
The corresponding Sinkhorn divergence is
\begin{equation}
    S_\epsilon(q_\theta,p)
    =\mathcal T_\epsilon(q_\theta,p)
     -\tfrac12\mathcal T_\epsilon(q_\theta,q_\theta)
     -\tfrac12\mathcal T_\epsilon(p,p).
    \label{eq:appendix_sinkhorn_identity}
\end{equation}
The cost-only quantity $\mathrm{OT}_\epsilon$ does not equal $\mathcal T_\epsilon$ in forward value. For fixed empirical marginal weights and an exact optimal coupling $\pi_\epsilon^\star$, however, calculating the gradient gives
\begin{equation}
    \nabla_\theta\mathcal T_\epsilon(q_\theta,p)
    =\nabla_\theta\!\left\langle
        \operatorname{sg}[\pi_\epsilon^\star(q_\theta,p)],C
    \right\rangle
    =\nabla_\theta \mathrm{OT}_\epsilon(q_\theta,p).
    \label{eq:appendix_ot_gradient_equivalence}
\end{equation}
and therefore $\nabla_\theta\mathcal F(q_\theta,p) =\nabla_\theta S_\epsilon(q_\theta,p)$. Likewise, for a symmetric kernel $k$, define $\mathcal D_k(r,s)=-\,\mathbb E_{X\sim r,Y\sim s}[k(X,Y)]$. Substitution into Equation~\eqref{eq:appendix_stop_gradient_identity} yields
\begin{equation*}
    \frac{1}{2}\operatorname{MMD}_k^2(q_\theta,p)
    =\frac{1}{2}\mathbb E[k(X,X')]+\frac{1}{2}\mathbb E[k(Y,Y')]
     -\mathbb E[k(X,Y)].
    \label{eq:appendix_mmd_identity}
\end{equation*}
where $X,X'\overset{\mathrm{iid}}{\sim}q_\theta$ and $Y,Y'\overset{\mathrm{iid}}{\sim}p$. Hence $\nabla_\theta\mathcal F(q_\theta,p) =\nabla_\theta\frac{1}{2}\operatorname{MMD}_k^2(q_\theta,p)$. For both choices, $\mathcal F$ therefore performs gradient descent on the corresponding distributional discrepancy.

\subsection{Finite-batch gradient surrogates}
\label{appendix:finite_batch_surrogates}

We implement the detached occurrence of $q_\theta$ using an independently generated reference batch. For one matching group, let $G=\{g_i\}_{i=1}^{B}$, $G^{\mathrm{ref}}=\{g_i^{\mathrm{ref}}\}_{i=1}^{B}$, and $R=\{r_j\}_{j=1}^{N}$ denote the active generated, reference generated, and real feature sets after a suitable normalization (see Appendix~\ref{appendix:pair_rms} for details). The two generated sets use independent latent samples, gradients propagate only through $G$, and both $G^{\mathrm{ref}}$ and $R$ are detached.

\textbf{Sinkhorn surrogate.} For the finite sets $G$ and $R$, the exact optimizer $\pi_\epsilon^\star(G,R)$ from Equation~\eqref{eq:appendix_regularized_ot_value} is represented by a nonnegative $B\times N$ matrix with row sums $1/B$ and column sums $1/N$. Following the notation in Section~\ref{sec:attractor}, we denote its Sinkhorn approximation by $\widehat\pi_\epsilon(G,R)\approx\pi_\epsilon^\star(G,R)$. Substituting this approximation into the cost-only definition gives
\begin{equation*}
    \mathrm{OT}_\epsilon(G,R)
    =\sum_{i,j}
      \mathrm{sg}[\widehat\pi_\epsilon(G,R)]_{ij}
      \frac{\lVert g_i-r_j\rVert_2^2}{2}.
    \label{eq:appendix_sinkhorn_cost}
\end{equation*}
Cross and generated-reference plans are solved independently, giving
\begin{equation}
    \widehat{\mathcal F}^{\mathrm{OT}}(G,G^{\mathrm{ref}},R)
    =\mathrm{OT}_\epsilon(G,R)
     -\mathrm{OT}_\epsilon(G,\mathrm{sg}[G^{\mathrm{ref}}]).
    \label{eq:appendix_grouped_sinkhorn}
\end{equation}

\textbf{MMD surrogate.} For a symmetric kernel $k$, the corresponding finite-batch loss is
\begin{equation}
    \widehat{\mathcal F}^{\mathrm{MMD}}(G,G^{\mathrm{ref}},R)
    =\frac{1}{B^2}\sum_{i=1}^{B}\sum_{i'=1}^{B}
       k(g_i,\mathrm{sg}[g^{\mathrm{ref}}_{i'}])
     -\frac{1}{BN}\sum_{i=1}^{B}\sum_{j=1}^{N}
       k(g_i,r_j).
    \label{eq:appendix_grouped_mmd}
\end{equation}
The Laplacian and energy-distance variants reported in the paper use $k(a,b)=\exp(-\lVert a-b\rVert_2)$ and $k(a,b)=-\lVert a-b\rVert_2$, respectively.

Equations~\eqref{eq:appendix_grouped_sinkhorn} and~\eqref{eq:appendix_grouped_mmd} are the finite-batch objectives used in our implementation.

\subsection{Conditional and multi-feature construction}
\label{appendix:conditional_feature_construction}

Let $\{c_k\}_{k=1}^{C}$ denote the conditions selected for one update. Depending on the experiment, a condition is an ImageNet class, a text caption, or an observation--action history. For every condition, we draw $N$ real targets and two independent batches of $B$ latent variables:
\begin{align*}
    y_{kj} &\sim p_{\mathrm{data}}(\cdot\mid c_k),
    &&j=1,\ldots,N, \notag\\
    x_{ki} &= f_\theta(z_{ki},c_k),
    &x'_{ki} &= f_\theta(z'_{ki},c_k),
    &&i=1,\ldots,B,
    % \label{eq:appendix_conditional_samples}
\end{align*}
where $z_{ki}$ and $z'_{ki}$ are mutually independent. Gradients propagate through the active samples $x_{ki}$ only; the reference samples $x'_{ki}$ and real samples $y_{kj}$ are detached.

For feature branch $r$, the frozen encoder $\phi_r$ returns $P_r$ vectors of dimension $d_r$ per image. For condition $c_k$ and feature position $p$, define the three matching sets directly:
\begin{align}
    G^r_{k,p}
    &:=\{\phi_r(x_{ki})_p\}_{i=1}^{B}, \notag\\
    G^{r,\mathrm{ref}}_{k,p}
    &:=\mathrm{sg}
       [\{\phi_r(x'_{ki})_p\}_{i=1}^{B}], \notag\\
    R^r_{k,p}
    &:=\mathrm{sg}
       [\{\phi_r(y_{kj})_p\}_{j=1}^{N}].
    \label{eq:appendix_feature_populations}
\end{align}
Pooled representations have $P_r=1$, whereas spatial or token-valued representations have $P_r>1$. Each pair $(k,p)$ is matched separately, so conditions and feature positions are never pooled. The three sets above have shapes $[B,d_r]$, $[B,d_r]$, and $[N,d_r]$, respectively. Feature encoders are evaluated online; no feature bank or precomputed real-feature cache is used.

Let $s_r$ be the branch scale defined below. We evaluate either Equation~\eqref{eq:appendix_grouped_sinkhorn} or Equation~\eqref{eq:appendix_grouped_mmd}; denoting the selected finite-batch objective by $\widehat{\mathcal F}$, define
\begin{equation}
    \widehat{\mathcal F}_{r,k,p}
    :=
    \widehat{\mathcal F}\!\left(
      G^r_{k,p}/s_r,\;
      G^{r,\mathrm{ref}}_{k,p}/s_r,\;
      R^r_{k,p}/s_r
    \right).
    \label{eq:appendix_normalized_group}
\end{equation}
The complete post-training loss is
\begin{equation}
    \mathcal L_\theta
    =\sum_{r\in\mathcal R}\alpha_r
      \frac{1}{CP_r}
      \sum_{k=1}^{C}\sum_{p=1}^{P_r}
      \widehat{\mathcal F}_{r,k,p}.
    \label{eq:appendix_multifeature_loss}
\end{equation}
Each branch is treated independently. Branch weights $\alpha_r$ are applied as written and are not subsequently renormalized. For the ImageNet ResNet-MAE/ConvNeXt combination, weights are uniform within each backbone and sum to $1/2$ over the ResNet-MAE branches and $1/2$ over the ConvNeXt branches and multiply the overall loss by 5.

\subsection{Feature scaling}
\label{appendix:pair_rms}

Feature branches can have substantially different scales, so every set in branch $r$ is divided by a common branch scale $s_r$ as in Equation~\eqref{eq:appendix_normalized_group}. We use two scaling strategies.

\textbf{Fixed calibration for ImageNet.}
For ImageNet, we set $s_r=\sigma_r$, where $\sigma_r$ is calibrated once as the median pairwise Euclidean distance among the branch features of $2{,}000$ training images sampled with fixed seed $0$. This scale remains fixed throughout post-training; see Appendix~\ref{appendix:exp_detail_imagenet}.

\textbf{Pair-RMS scaling for text-to-image and world modeling.}
For text-to-image and world-model post-training, we compute one detached scale for each branch from the current minibatch. For two feature sets $S$ and $T$, define
\begin{equation*}
    d^2(S,T)
    :=\frac{1}{|S||T|}
      \sum_{s\in S}\sum_{t\in T}\lVert s-t\rVert_2^2.
\end{equation*}
Branch $r$ uses
\begin{equation}
    s_r
    =\mathrm{sg}\!\left[
      \sqrt{\max\!\left\{
        \frac{1}{2CP_r}\sum_{k=1}^{C}\sum_{p=1}^{P_r}
          d^2(G^r_{k,p},R^r_{k,p})
        +\frac{1}{2CP_r}\sum_{k=1}^{C}\sum_{p=1}^{P_r}
          d^2(G^r_{k,p},G^{r,\mathrm{ref}}_{k,p}),\,
        \delta^2
      \right\}}
    \right],
    \label{eq:appendix_pair_rms}
\end{equation}
where $\delta>0$ is a small numerical floor. Every active, reference, and real set in branch $r$ is divided by this same minibatch-dependent $s_r$.

\subsection{Algorithm and distributed execution}
\label{appendix:posttraining_algorithm}

Algorithm~\ref{alg:distributional_posttraining} summarizes one optimization step.
\refstepcounter{posttrainalgorithm}
\label{alg:distributional_posttraining}
\begin{center}
\begin{minipage}{0.97\linewidth}
\hrule
\vspace{3pt}
\textbf{Algorithm \theposttrainalgorithm: One distributional post-training update}
\vspace{3pt}
\hrule
\vspace{3pt}
\textbf{Input:} Generator $f_\theta$; frozen branches $\{\phi_r\}$; branch weights $\{\alpha_r\}$; selected finite-batch objective $\widehat{\mathcal F}$.
\begin{enumerate}[leftmargin=1.6em,itemsep=1pt,topsep=3pt]
    \item Sample conditions $c_{1:C}$, real targets $y_{k,1:N}$, and independent active and reference noises $z_{k,1:B},z'_{k,1:B}$.
    \item Generate $x=f_\theta(z,c)$ and $x'=\mathrm{sg}[f_\theta(z',c)]$; initialize $\mathcal L_\theta=0$.
    \item For each feature branch $r$:
    \begin{enumerate}[label=(\alph*),leftmargin=1.5em,itemsep=0pt,topsep=1pt]
        \item Construct $G^r_{k,p}$, $G^{r,\mathrm{ref}}_{k,p}$, and $R^r_{k,p}$ for every $(k,p)$ using Equation~\eqref{eq:appendix_feature_populations}.
        \item Compute the detached $s_r$ following Section~\ref{appendix:pair_rms}, then divide every set by $s_r$.
        \item Evaluate $\widehat{\mathcal F}_{r,k,p}$ using Equation~\eqref{eq:appendix_grouped_sinkhorn} or Equation~\eqref{eq:appendix_grouped_mmd}, then set \\ 
        $\mathcal L_\theta\leftarrow\mathcal L_\theta+ \alpha_r(CP_r)^{-1}\sum_{k,p} \widehat{\mathcal F}_{r,k,p}$.
    \end{enumerate}
    \item Update $\theta$ using $\nabla_\theta\mathcal L_\theta$.
\end{enumerate}
\vspace{-2pt}
\hrule
\end{minipage}
\end{center}

%% file: Appendix_C_exp_details.tex
\newpage
\section{Experimental Details}
\label{appendix:exp_details}
\subsection{Class-conditional generation on ImageNet} \label{appendix:exp_detail_imagenet}
\textbf{Training configurations.}
For each model described in Section~\ref{sec:exp_imagenet}, Priming-B and Priming-L, we initialize the $80$- and $160$-epoch variants from two checkpoints of the same reconstruction run and post-train them separately. All four configurations share the settings in Table~\ref{tab:appendix_imagenet_hparams}. Class conditioning uses eight learned class tokens without label dropout, as Priming Models do not deploy classifier-free guidance. We use 8 NVIDIA H200 GPUs for both stages of the training.

\noindent\begin{minipage}{\textwidth}
    \centering
    \small
    \setlength{\tabcolsep}{4pt}
    \begin{tabular}{lllccc}
        \toprule
        Stage & Optimizer & Duration & Batch & Learning rate & EMA decay \\
        \midrule
        Stage 1 (Reconstruction loss) & Muon & $80$ / $160$ epochs & $1{,}024$ & $10^{-3}$ & $0.9999$ \\
        Stage 2 (Distributional loss) & AdamW & $25$k updates & $16$ classes & $10^{-5}$ & $0.999$ \\
        \bottomrule
    \end{tabular}

    \vspace{0.05in}
    \refstepcounter{table}
    \label{tab:appendix_imagenet_hparams}
    \small
    \textbf{Table~\thetable: ImageNet training hyperparameters.} Settings apply to all four Priming-B/L configurations.
\end{minipage}
\par\smallskip

\textbf{Reconstruction pretraining.}
We use the fixed corruption level and perceptual-loss weights described in Section~\ref{sec:exp_imagenet}. Muon uses momentum $0.95$, while auxiliary AdamW updates use $(\beta_1,\beta_2)=(0.9,0.95)$. The learning rate remains constant, with no warmup or weight decay.

\textbf{Distributional post-training.}
We initialize the generator from the online reconstruction weights and retain the pretrained EMA shadow, changing its decay to $0.999$. We initialize a fresh AdamW optimizer with $(\beta_1,\beta_2)=(0.9,0.95)$, zero weight decay, and gradient-norm clipping at $1.0$. We use a linear learning-rate warmup schedule, increasing the learning rate linearly to $10^{-5}$ over the first $125$ updates and keeping it constant thereafter.

Each update samples $16$ distinct ImageNet classes, with two classes assigned to each GPU. We compute the per-class distributional losses locally within each GPU. In total, each update uses $1{,}024$ differentiable generated images, $1{,}024$ independently generated stop-gradient references from the current generator, and $2{,}048$ real images.

\textbf{Feature spaces and transport computation.}
We follow the feature choice setup of \citep{deng2026generative}, using the $84$ feature outputs from the frozen ResNet-MAE encoder and $13$ from ConvNeXt. This includes spatial activations and pooled mean and standard-deviation features. Each encoder family receives total weight $1/2$, distributed equally among its outputs. 

Before post-training, for each feature space, we calibrate a fixed scale $\sigma$ using the median pairwise Euclidean distance among $2{,}000$ training images sampled once with a fixed seed and divide each feature value by this $\sigma$.

For every class and feature output, we solve separate generated--real and generated--generated transport problems at each regularization strength specified in Section~\ref{sec:exp_imagenet}. We use log-domain Sinkhorn iterations with a maximum of $1{,}000$ iterations and marginal-error tolerance $10^{-6}$, checked every $10$ iterations. 
We average the resulting objective over classes and apply an overall loss multiplier of $5$.  

\textbf{Evaluation.}
We evaluate the final post-training EMA checkpoint for Priming Model instances. Each evaluation generates $50$ images per class, totaling $50{,}000$ images. We report the mean over sampling seeds $1$, $2$, and $3$. These are repeated sampling evaluations of one trained checkpoint per configuration.

\subsection{Text-to-image generation}
\label{appendix:exp_detail_t2i}

\textbf{Architecture and data.}
We conduct the text-to-image experiments at $512\times512$ resolution in pixel space using the MiniT2I MM-JiT-B/16 architecture~\citep{minit2i2026}. The generator has $258$M parameters. We encode captions with the $341$M-parameter FLAN-T5-Large encoder~\citep{chung2024scaling}, which remains frozen in every training stage. Reconstruction pretraining uses CC12M~\citep{changpinyo2021conceptual}, followed by reconstruction fine-tuning on the curated $120$k-image mixture released with MiniT2I. We sample $\rve \sim \mathcal{N}(0, 2^2\mathbf{I})$ for reconstruction pretraining and as the generator input during distributional post-training and inference. Reconstruction pretraining on CC12M and subsequent reconstruction fine-tuning
on the curated mixture both use $16$ NVIDIA H200 GPUs. Distributional
post-training uses $8$ NVIDIA H200 GPUs.

\noindent\begin{minipage}{\textwidth}
    \centering
    \normalsize
    \setlength{\tabcolsep}{4pt}
    \begin{tabular}{lllcc}
        \toprule
        Stage & Corruption or objective & Duration & Batch & Learning rate \\
        \midrule
        Reconstruction & reweighted logit-normal & $250$k $+$ $40$k & $1{,}024$ & $4\times10^{-4}\rightarrow4\times10^{-5}$ \\
        Reconstruction & $s=0.8$ & $160$k $+$ $40$k & $1{,}024$ & $4\times10^{-4}\rightarrow4\times10^{-5}$ \\
        Distributional & OT, $\epsilon=0.05$ & $4$k & $64$ captions & $10^{-5}$ \\
        \bottomrule
    \end{tabular}

    \vspace{0.05in}
    \refstepcounter{table}
    \label{tab:appendix_t2i_hparams}
    \small
    \textbf{Table~\thetable: T2I training hyperparameters.} Durations are in updates. Reconstruction entries list CC12M and curated-mixture values; the distributional row describes one-step post-training, with batch size measured in captions.
\end{minipage}
\par\smallskip

\textbf{Reconstruction pretraining.}
We evaluate the two corruption schedules summarized in Table~\ref{tab:appendix_t2i_hparams}. For the reweighted logit-normal configuration, we use the normalized effective density $p(s) \propto s^{-2} \operatorname{LogNorm}(s; 0.8, 0.8^2)$ for $s \in (0,1)$, where $\operatorname{LogNorm}(\cdot; \mu, \sigma^2)$ denotes the logit-normal density defined by $\operatorname{logit}(s) \sim \mathcal{N}(\mu, \sigma^2)$.
The logit-normal model uses the sampled corruption schedule throughout CC12M pretraining and curated-mixture fine-tuning, whereas the fixed-noise model reported in Table~\ref{tab:t2i} uses $s=0.8$ in both stages. We use AdamW for all reconstruction training.

\textbf{Distributional post-training.}
For the one-step models, starting from each reconstruction checkpoint, we optimize the OT objective in~\eqref{eq:ot_def} with the settings in Table~\ref{tab:appendix_t2i_hparams}. Each update contains $64$ distinct captions. For every caption, we generate $32$ samples: gradients are propagated through $16$ samples, and the other $16$ form the stop-gradient generated reference set. Each caption is associated with one real image. Consequently, the generated--real transport term for a caption reduces to the average squared feature distance between its generated samples and that image.

We compute a separate objective using globally pooled SigLIP, ViT-MAE, and Inception-V3 features, using $10$ Sinkhorn iterations for each OT computation. Before computing OT, each feature branch is normalized by its within-batch pairwise RMS distance so that the branches have comparable scales. We combine the three feature-space objectives with equal weights and optimize them with a freshly initialized AdamW optimizer and no warmup. All reported Priming Model scores use the final online checkpoint.

\textbf{Multi-step distributional post-training.}
We use the ordered noise levels $(1.0,0.75,0.5,0.25)$ and train in two consecutive phases. For the first $4{,}000$ updates, the sample counts per caption at these levels are $(16,8,4,2)$, respectively. These counts apply separately to the live generated samples, through which gradients are propagated, and the stop-gradient generated reference samples. Starting from the resulting $4{,}000$-update checkpoint, we continue for another $4{,}000$ updates with counts $(8,8,8,8)$ for each generated set. Thus, both phases use matching live and stop-gradient sample counts at every noise level, and multi-step post-training lasts $8{,}000$ updates in total. We use the same feature-space OT objective as above.

\textbf{Evaluation.} GenEval~\citep{ghosh2023geneval} uses all $553$ prompts and four images per prompt ($2{,}212$ images in total), while DPG-Bench~\citep{hu2024ella} uses all $1{,}065$ prompts. We use random seed $0$. The one-step MeanFlow baseline uses the CFG scale $12.0$ and guidance interval $[0.2,0.8]$; the corresponding four-step result uses the same released model and guidance configuration. The training-compute totals in Table~\ref{tab:t2i} include reconstruction training and the subsequent method-specific stage.

\subsection{World modeling}
\label{appendix:exp_detail_wm}

\textbf{Data representation and conditioning.}
We follow the preprocessing and action-conditioned U-Net architecture of DriftWorld~\citep{lu2026driftworld} for both RT-1~\citep{brohan2023rt1} and Bridge-V2~\citep{walke2023bridgedatav2}. Frames are resized to $256\times256$ and encoded with the SD3-VAE~\citep{esser2024scaling}; the Priming Model operates in this latent space. Both datasets use the same prediction task. To predict $o_{t+1}$, the model receives four consecutive observations and their four aligned actions,
\begin{equation*}
    \bigl(o_{t-3},o_{t-2},o_{t-1},o_t\bigr),
    \qquad
    \bigl(a_{t-3},a_{t-2},a_{t-1},a_t\bigr).
\end{equation*}
Here, $a_t$ is directly associated with the transition $o_t\rightarrow o_{t+1}$, while the preceding three actions provide action history.

\noindent\begin{minipage}{\textwidth}
    \centering
    \normalsize
    \setlength{\tabcolsep}{3pt}
    \begin{tabular}{llcccc}
        \toprule
        Stage & Corruption or objective & Duration & Batch & Learning rate & EMA decay \\
        \midrule
        Reconstruction & reweighted logit-normal & $64$ epochs & $64$ & $10^{-4}$ & $0.9999$ \\
        Distributional & OT, $\epsilon=0.05$ & $4$k updates & $32$ conditions & $10^{-5}$ & $0.999$ \\
        \bottomrule
    \end{tabular}

    \vspace{0.05in}
    \refstepcounter{table}
    \label{tab:appendix_wm_hparams}
    \small
    \textbf{Table~\thetable: World-model training hyperparameters.} Settings are shared by RT-1 and Bridge-V2; the post-training batch counts conditioning sequences.
\end{minipage}
\par\smallskip

\textbf{Reconstruction pretraining.}
We pretrain each model with the settings in Table~\ref{tab:appendix_wm_hparams}, using AdamW.

\textbf{Distributional post-training.}
Post-training starts from the EMA checkpoint of the reconstruction-pretrained model and uses the settings in Table~\ref{tab:appendix_wm_hparams}. Each update contains $32$ distinct conditioning sequences. For every condition, we draw $16$ generated samples through which gradients are propagated, $16$ stop-gradient generated samples, and one real next-frame target. We use $10$ Sinkhorn iterations. The feature branches comprise unpooled patch tokens from blocks $2$, $5$, and $8$ of a frozen DINOv3-B/16 encoder together with the spatial SD3-VAE latent representation. We normalize each branch by its within-batch pairwise RMS distance before computing its OT objective. Post-training uses AdamW. Both reconstruction pretraining and distributional post-training use $8$
NVIDIA H200 GPUs.

\textbf{Autoregressive evaluation.}
We evaluate the EMA checkpoints on the held-out validation sets without CFG. Priming Models and DriftWorld use one network evaluation per predicted frame, whereas the flow-matching baseline uses $32$. During autoregressive evaluation, a predicted frame is inserted into the subsequent four-frame visual history, while the provided action sequence remains the conditioning sequence. We generate in $8$-frame autoregressive chunks and re-anchor the visual context on the real trajectory between chunks. All image-quality metrics compare decoded predictions against the corresponding SD3-VAE reconstructions of the target frames rather than the original input pixels. SSIM, PSNR, and LPIPS compare paired frames, and FID measures distributional agreement using all predicted frames. FVD is computed on $2{,}048$ validation trajectories that contain at least $16$ predicted frames, using the first $16$ predictions from each trajectory. Runtime is reported per generated frame, and the Priming Model training-compute totals include both reconstruction pretraining and distributional post-training.

\subsection{Experimental details for Section~\ref{sec:intuition}}
\label{appendix:exp_detail_jacobian}

\textbf{Models and correction subspaces.}
We study four Priming-B initializations before distributional post-training: reconstruction pretraining for $20$, $80$, or $160$ epochs without perceptual loss, and $160$ epochs with perceptual loss. For each model, we generate $512$ images in each of eight fixed ImageNet classes using matched Gaussian noise inputs across models. 

We then collect feature-space OT gradients in batches of $64$ queries, using three reference draws per batch, each with $128$ real images and $64$ independently generated stop-gradient references.
The analysis covers $13$ frozen feature outputs: $11$ from ResNet-MAE and two from ConvNeXt.
For each checkpoint, class, and feature output pair, we normalize each gradient to a unit vector and fit an uncentered PCA basis with $r=64$.

\textbf{Jacobian response and comparison with post-training.}
For a generated sample $i$, let $J_i=\partial h_\theta(\rve_i,c)/\partial\theta$ and let $U_i$ contain the orthonormal PCA directions fitted on the opposite fold. Within each class, we estimate the normalized response $\mathcal A=\frac{d}{r}\frac{\mathbb E_i\|J_i^\top U_i\|_F^2}{\mathbb E_i\|J_i\|_F^2}$, where $d$ is the flattened feature dimension. 

We compute vector--Jacobian products using two random unit directions with signed coefficients in the PCA subspace and four normalized Gaussian directions in the full feature space per image, to approximate the Frobenius norm. We average the squared response within each group before taking their ratio, then average the resulting classwise scores equally over the eight classes. 

Thus, a large $\mathcal A$ indicates stronger Jacobian response along the estimated OT correction subspace relative to the model's overall Jacobian scale. As mentioned in Section~\ref{sec:intuition}, we compare these alignments with online-weight FID after a common $12{,}500$ post-training updates.

\begin{figure}[htbp]
    \centering
    \includegraphics[width=\textwidth]{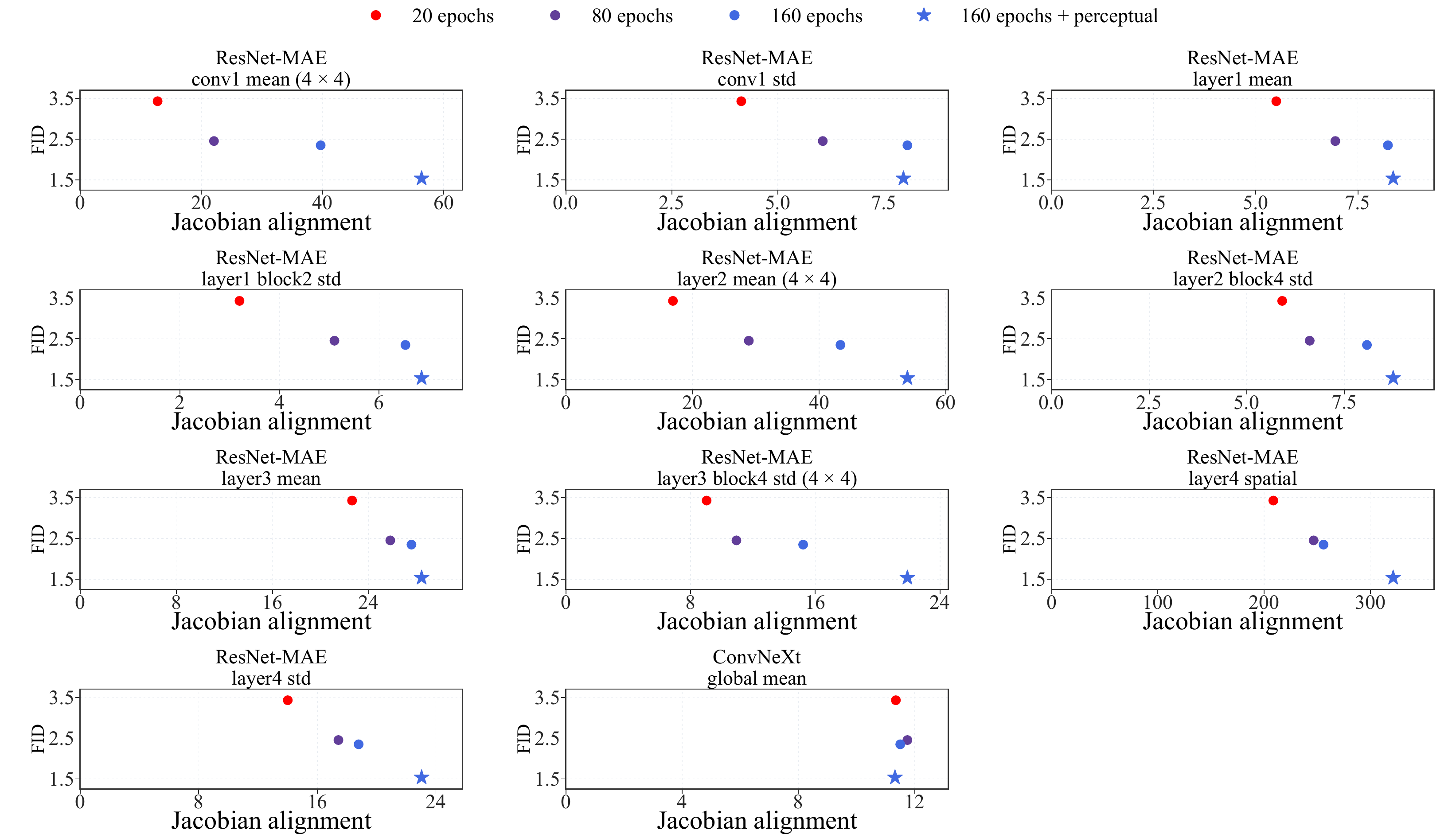}
    \caption{\textbf{Jacobian alignment in additional feature spaces.} The $11$ outputs complement the two shown in Figure~\ref{fig:intuition_results}, covering all $13$ analyzed ResNet-MAE and ConvNeXt outputs.  Colors and markers match the main-text figure; horizontal axes are linear with feature-specific ranges; all panels share the same FID range.}
    \label{fig:jacobian_additional_features}
\end{figure}

\textbf{Additional feature-space results.} Figure~\ref{fig:jacobian_additional_features} reports the alignment results for $11$ ResNet-MAE and ConvNeXt feature space, complementing Figure~\ref{fig:intuition_results}. Across all ten additional ResNet-MAE outputs, alignment increases from $20$ to $80$ to $160$ epochs without perceptual loss, alongside lower post-training FID. ConvNeXt global-mean alignment varies little and does not follow the FID ordering. Thus, the association extends across multiple feature outputs, but is not uniform across representations.

%% file: Appendix_D_additional_results.tex
\newpage
\section{Additional results on imagenet} \label{appendix:image_additional}
For completeness, we report precision and recall for the Priming Model variants as well as those of the baselines in Table~\ref{tab:imagenet} and assess potential memorization through nearest-neighbor retrieval against the training set.

\textbf{Precision and recall.} As shown in Table~\ref{tab:imagenet_precision_recall}, Priming Model instances exhibit comparable precision and recall to the baselines.
\begin{table}[htbp]
    \centering
    \small
    \setlength{\tabcolsep}{10pt}
    \begin{tabular}{lrr}
        \toprule
        Method & Precision $\uparrow$ & Recall $\uparrow$ \\
        \midrule
        JiT-H$^{\dagger}$ & 0.78 & 0.62 \\
        Drifting-B & 0.806 & 0.601 \\
        \tablecolor{tableHighlight}
        \textbf{Priming-B} (80 ep.) & 0.786 & 0.617 \\
        Drifting-L & 0.815 & 0.602 \\
        \tablecolor{tableHighlight}
        \textbf{Priming-B} (160 ep.) & 0.783 & 0.624 \\
        WF-XL/2 & 0.762 & 0.674 \\
        \tablecolor{tableHighlight}
        \textbf{Priming-L} (80 ep.) & 0.781 & 0.634 \\
        \tablecolor{tableHighlight}
        \textbf{Priming-L} (160 ep.) & 0.781 & 0.649 \\
        \bottomrule
    \end{tabular}
    \caption{\textbf{Precision and recall on ImageNet.} All methods use one-step generation except JiT-H. $^{\dagger}$JiT-H values are taken from Table~13 of \citet{li2026back}.}
    \label{tab:imagenet_precision_recall}
\end{table}

\input{figures/imagenet_neighbors_grouped_upload/imagenet_neighbors}
\section{Fd-loss}  
\label{appendix:fd_loss}

\subsection{FD-loss Objective}
FD-loss~\citep{yang2026representation} directly compares the generated and real feature distributions through their first two moments. Let $\phi$ be a frozen feature encoder, and let $q_\theta^\phi:=\phi_\#q_\theta$ and $p^\phi:=\phi_\#p$ denote the generated and real feature distributions, respectively. Denoting their means and covariances as $(\mu_q,\Sigma_q)$ and $(\mu_p,\Sigma_p)$, 
the objective is the squared Fr\'echet distance between the corresponding moment-matched Gaussian distributions~\citep{dowson1982frechet}:
\begin{equation}
    \mathcal F_{\mathrm{FD}}(q_\theta^\phi,p^\phi)
    :=\lVert\mu_q-\mu_p\rVert_2^2
      +\operatorname{Tr}\!\left(
        \Sigma_q+\Sigma_p
        -2\bigl(\Sigma_p^{1/2}\Sigma_q\Sigma_p^{1/2}\bigr)^{1/2}
      \right).
    \label{eq:appendix_fd_objective}
\end{equation}
Equivalently, Equation~\eqref{eq:appendix_fd_objective} is the squared $2$-Wasserstein distance between $\mathcal N(\mu_q,\Sigma_q)$ and $\mathcal N(\mu_p,\Sigma_p)$.

Unlike the objectives for Sinkhorn divergence or MMD, $\mathcal F_{\mathrm{FD}}$ is minimized directly: there is no generated self-interaction term or independent generated reference batch. Gradients propagate through the generated moments $(\mu_q,\Sigma_q)$ to $q_\theta$, whereas the real moments $(\mu_p,\Sigma_p)$ are fixed. 

In practice, FD-loss replaces the population moments with accumulated estimates so that the objective can be optimized without placing the full feature population in a single minibatch~\citep{yang2026representation}. FD-loss performs marginal matching instead of conditional distribution matching.

\subsection{FD-loss ImageNet experiments}
\label{appendix:fd_loss_experiments}

\textbf{Setup.}
We evaluate FD-loss as an alternative distributional post-training objective for the pixel-space Priming Models from Section~\ref{sec:exp_imagenet}. Priming-B has $115.1$M parameters and is used for the pretraining ablations, while Priming-L has $405.1$M parameters and is used for the pretraining-duration study. Both models use $16\times16$ patches and generate ImageNet-1K images at $256\times256$ resolution in one network evaluation. These experiments are separate from the OT-post-trained results in the main text.

Reconstruction pretraining uses Muon with learning rate $10^{-3}$ and global batch size $1{,}024$. Unless stated otherwise, we use the fixed corruption level $s=0.8$ and the pixel, LPIPS, and ConvNeXt reconstruction objective described in Section~\ref{sec:exp_imagenet}. FD post-training lasts $20$ epochs and uses AdamW with learning rate $10^{-5}$, cosine decay, five epochs of linear warmup, and global batch size $256$. We follow the multi-representation FD-SIM configuration of \citet{yang2026representation}: the training objective combines equally weighted FD terms from frozen SigLIP, Inception-V3, and MAE encoders. Real feature moments are precomputed, while generated moments are accumulated across updates with decay $0.999$.

We evaluate the final online weights; the accumulated feature statistics are used only during FD post-training. Each result uses $50{,}000$ generated images with uniformly sampled labels. We report FID, precision, recall, and $\mathrm{FDr}^{6}$ following the FD-loss evaluation protocol. Here, $\mathrm{FDr}^{6}$ is the validation-normalized Fr\'echet distance averaged across six representations and differs from $\mathrm{FDr}^{5}$ used in the main experiments.

\textbf{Priming-B pretraining ablations.}
We first pretrain Priming-B for $80$ epochs and apply the same $20$-epoch FD-loss stage to every configuration. Table~\ref{tab:fdloss_priming_b} compares pixel reconstruction alone with the addition of LPIPS and ConvNeXt perceptual supervision under fixed and logit-normal corruption schedules. Perceptual supervision lowers $\mathrm{FDr}^{6}$ from $7.49$ to $4.95$ for fixed $s=0.8$ and from $8.30$ to $5.57$ for the reweighted logit-normal distribution. Fixing $s=0.8$ also outperforms the tested logit-normal schedule under either reconstruction objective.

Keeping perceptual reconstruction fixed, we then sweep $s\in\{0.6,0.7,0.8,0.9,1.0\}$. Among these values, $s=0.8$ gives the lowest $\mathrm{FDr}^{6}$. At $s=1$, the corrupted input is independent of its paired clean image, and performance deteriorates sharply. Thus, simply matching the eventual Gaussian generator input is insufficient; retaining some clean-image information during reconstruction provides a substantially better initialization.

\noindent\begin{minipage}{\textwidth}
    \centering
    \begin{minipage}[t]{0.47\textwidth}
        \centering
        \textbf{(a) Reconstruction objective and schedule}\par\smallskip
        \small
        \setlength{\tabcolsep}{5pt}
        \begin{tabular}{lcc}
            \toprule
            Schedule & Pixel & + Perceptual \\
            \midrule
            Fixed $s=0.8$ & 7.49 & \textbf{4.95} \\
            Logit-normal & 8.30 & 5.57 \\
            \bottomrule
        \end{tabular}
    \end{minipage}
    \hfill
    \begin{minipage}[t]{0.47\textwidth}
        \centering
        \textbf{(b) Fixed-corruption sweep with perceptual reconstruction}\par\smallskip
        \setlength{\tabcolsep}{12pt}
        \begin{tabular}{cc}
            \toprule
            Fixed $s$ & $\mathrm{FDr}^{6}\downarrow$ \\
            \midrule
            0.6 & 6.13 \\
            0.7 & 5.86 \\
            0.8 & \textbf{4.95} \\
            0.9 & 5.95 \\
            1.0 & 50.06 \\
            \bottomrule
        \end{tabular}
    \end{minipage}
    \par\vspace{0.05in}
    \refstepcounter{table}
    \label{tab:fdloss_priming_b}
    \small
    \textbf{Table~\thetable: Priming-B FD-loss ablations.} All configurations use $80$ epochs of reconstruction pretraining followed by $20$ epochs of FD post-training. Values are $\mathrm{FDr}^{6}$; lower is better.
\end{minipage}
\par\smallskip

\textbf{Scaling Priming-L pretraining.}
Finally, we vary fixed-$s=0.8$ perceptual pretraining from $20$ to $160$ epochs while holding FD post-training at $20$ epochs. Table~\ref{tab:fdloss_priming_l} shows that longer reconstruction pretraining consistently improves the principal distributional metrics: FID decreases from $1.07$ to $0.94$, $\mathrm{FDr}^{6}$ decreases from $5.46$ to $3.29$, and recall increases from $0.53$ to $0.60$. Precision remains approximately constant. This result further supports reconstruction pretraining as an effective initialization for a direct Fr\'echet-distance objective, rather than only for the OT objective used in the main experiments.

\noindent\begin{minipage}{\textwidth}
    \centering
    \small
    \setlength{\tabcolsep}{5pt}
    \begin{tabular}{lrrrrrr}
        \toprule
        Model & Pretrain ep. & FD ep. & FID$\downarrow$ & $\mathrm{FDr}^{6}\downarrow$ & Prec.$\uparrow$ & Recall$\uparrow$ \\
        \midrule
        Priming-L & 20  & 20 & 1.07 & 5.46 & 0.77 & 0.53 \\
        Priming-L & 40  & 20 & 1.01 & 4.40 & 0.77 & 0.58 \\
        Priming-L & 80  & 20 & 0.96 & 3.64 & 0.77 & 0.59 \\
        Priming-L & 160 & 20 & \textbf{0.94} & \textbf{3.29} & 0.77 & \textbf{0.60} \\
        \bottomrule
    \end{tabular}
    \par\vspace{0.05in}
    \refstepcounter{table}
    \label{tab:fdloss_priming_l}
    \small
    \textbf{Table~\thetable: Scaling reconstruction pretraining before FD-loss.} Priming-L uses fixed $s=0.8$ with perceptual reconstruction, followed by a fixed $20$-epoch FD post-training stage. Each row is one training run evaluated using the final online weights.
\end{minipage}
\par\smallskip

%% file: figures/imagenet_neighbors_grouped_upload/imagenet_neighbors.tex
\textbf{Memorization test.}
Each group shows one generated image (blue border) followed by its four nearest ImageNet training images, ordered from nearest to farthest. Retrieval uses CLIP ViT-L/14 cosine similarity within the same class, searching all $1{,}300$ training images per class. We show $20$ visually selected queries per model, fixed before retrieval. We do not observe the signs of memorization in either Priming-B or Priming-L-160.

\par\smallskip\begingroup\graphicspath{{figures/imagenet_neighbors_top4/}}
\input{figures/imagenet_neighbors_grouped_upload/figures/imagenet_neighbors_top4/rows}
\endgroup

%% file: figures/imagenet_neighbors_grouped_upload/figures/imagenet_neighbors_top4/rows.tex
\newpage
\begingroup\setlength{\parindent}{0pt}\setlength{\parskip}{0pt}
\setlength{\fboxsep}{0pt}\setlength{\fboxrule}{0.6pt}
\noindent\begin{minipage}{\linewidth}
\textbf{Priming-B (160 ep.)}\par\vspace{3pt}
\begin{minipage}[t]{0.485\linewidth}\centering
{\fontsize{7}{8}\selectfont goldfish (class 1)\par}\vspace{1pt}
\includegraphics[width=\linewidth]{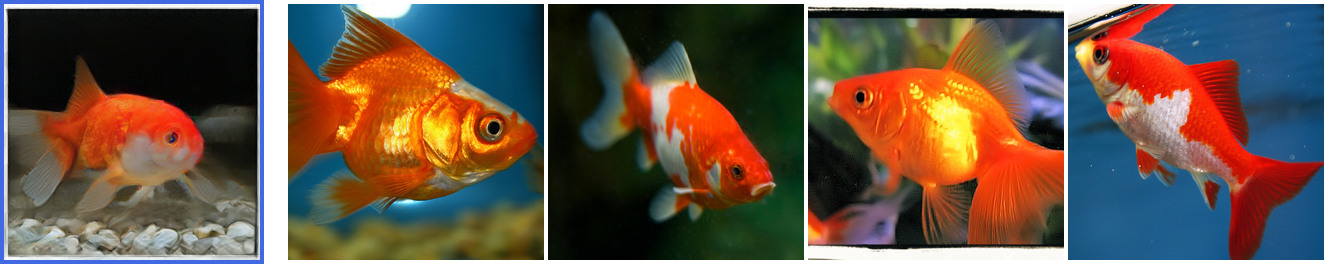}\par\vspace{1pt}
\makebox[\linewidth][l]{\fontsize{5.5}{6}\selectfont \makebox[0.1993958\linewidth][c]{\textcolor{xblue}{\textbf{Generated}}}%
\hspace{0.0181269\linewidth}%
\makebox[0.1933535\linewidth][c]{NN 1}%
\hspace{0.0030211\linewidth}%
\makebox[0.1933535\linewidth][c]{NN 2}%
\hspace{0.0030211\linewidth}%
\makebox[0.1933535\linewidth][c]{NN 3}%
\hspace{0.0030211\linewidth}%
\makebox[0.1933535\linewidth][c]{NN 4}}
\end{minipage}\hfill%
\begin{minipage}[t]{0.485\linewidth}\centering
{\fontsize{7}{8}\selectfont bald eagle (class 22)\par}\vspace{1pt}
\includegraphics[width=\linewidth]{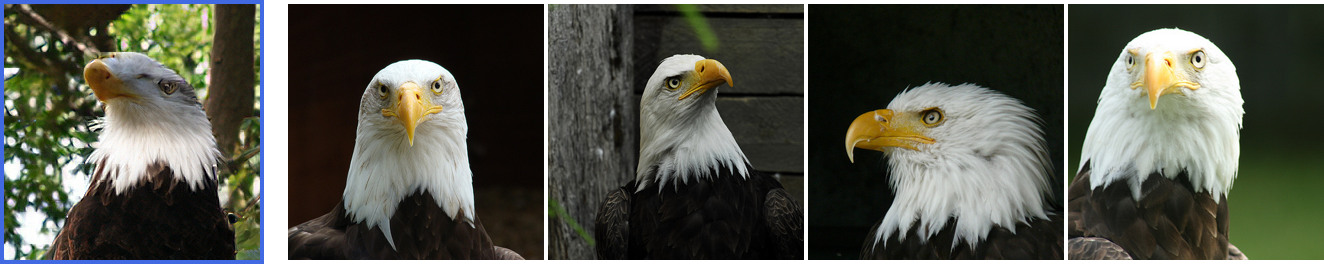}\par\vspace{1pt}
\makebox[\linewidth][l]{\fontsize{5.5}{6}\selectfont \makebox[0.1993958\linewidth][c]{\textcolor{xblue}{\textbf{Generated}}}%
\hspace{0.0181269\linewidth}%
\makebox[0.1933535\linewidth][c]{NN 1}%
\hspace{0.0030211\linewidth}%
\makebox[0.1933535\linewidth][c]{NN 2}%
\hspace{0.0030211\linewidth}%
\makebox[0.1933535\linewidth][c]{NN 3}%
\hspace{0.0030211\linewidth}%
\makebox[0.1933535\linewidth][c]{NN 4}}
\end{minipage}\end{minipage}\par\vspace{4pt}
\noindent\begin{minipage}{\linewidth}
\begin{minipage}[t]{0.485\linewidth}\centering
{\fontsize{7}{8}\selectfont macaw (class 88)\par}\vspace{1pt}
\includegraphics[width=\linewidth]{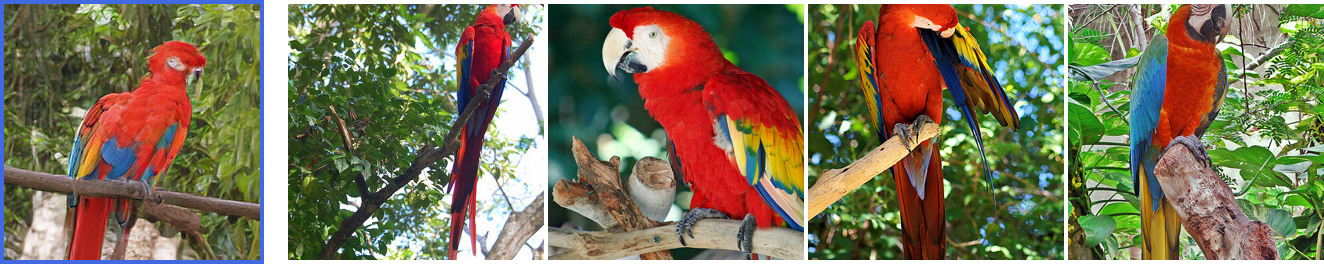}\par\vspace{1pt}
\makebox[\linewidth][l]{\fontsize{5.5}{6}\selectfont \makebox[0.1993958\linewidth][c]{\textcolor{xblue}{\textbf{Generated}}}%
\hspace{0.0181269\linewidth}%
\makebox[0.1933535\linewidth][c]{NN 1}%
\hspace{0.0030211\linewidth}%
\makebox[0.1933535\linewidth][c]{NN 2}%
\hspace{0.0030211\linewidth}%
\makebox[0.1933535\linewidth][c]{NN 3}%
\hspace{0.0030211\linewidth}%
\makebox[0.1933535\linewidth][c]{NN 4}}
\end{minipage}\hfill%
\begin{minipage}[t]{0.485\linewidth}\centering
{\fontsize{7}{8}\selectfont flamingo (class 130)\par}\vspace{1pt}
\includegraphics[width=\linewidth]{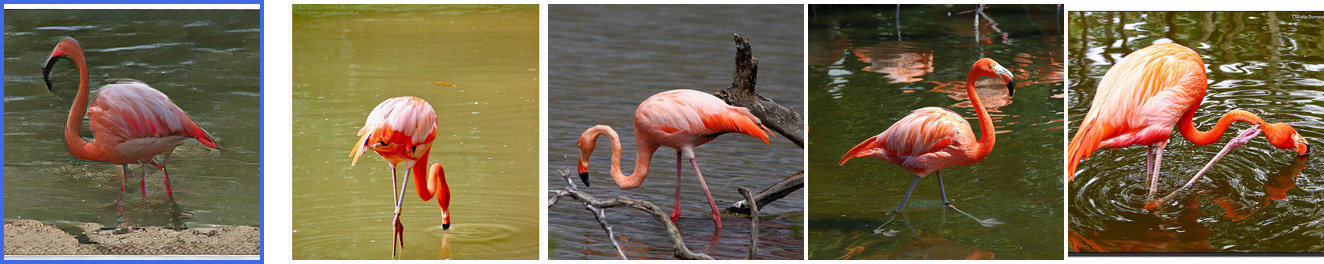}\par\vspace{1pt}
\makebox[\linewidth][l]{\fontsize{5.5}{6}\selectfont \makebox[0.1993958\linewidth][c]{\textcolor{xblue}{\textbf{Generated}}}%
\hspace{0.0181269\linewidth}%
\makebox[0.1933535\linewidth][c]{NN 1}%
\hspace{0.0030211\linewidth}%
\makebox[0.1933535\linewidth][c]{NN 2}%
\hspace{0.0030211\linewidth}%
\makebox[0.1933535\linewidth][c]{NN 3}%
\hspace{0.0030211\linewidth}%
\makebox[0.1933535\linewidth][c]{NN 4}}
\end{minipage}\end{minipage}\par\vspace{4pt}
\noindent\begin{minipage}{\linewidth}
\begin{minipage}[t]{0.485\linewidth}\centering
{\fontsize{7}{8}\selectfont golden retriever (class 207)\par}\vspace{1pt}
\includegraphics[width=\linewidth]{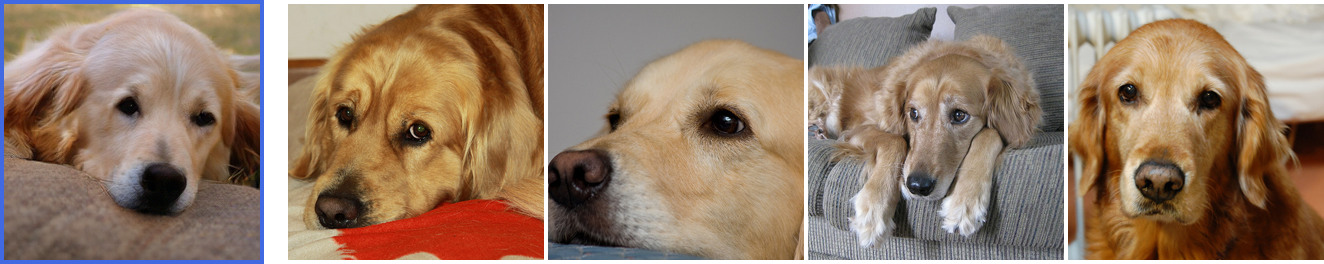}\par\vspace{1pt}
\makebox[\linewidth][l]{\fontsize{5.5}{6}\selectfont \makebox[0.1993958\linewidth][c]{\textcolor{xblue}{\textbf{Generated}}}%
\hspace{0.0181269\linewidth}%
\makebox[0.1933535\linewidth][c]{NN 1}%
\hspace{0.0030211\linewidth}%
\makebox[0.1933535\linewidth][c]{NN 2}%
\hspace{0.0030211\linewidth}%
\makebox[0.1933535\linewidth][c]{NN 3}%
\hspace{0.0030211\linewidth}%
\makebox[0.1933535\linewidth][c]{NN 4}}
\end{minipage}\hfill%
\begin{minipage}[t]{0.485\linewidth}\centering
{\fontsize{7}{8}\selectfont Siberian husky (class 250)\par}\vspace{1pt}
\includegraphics[width=\linewidth]{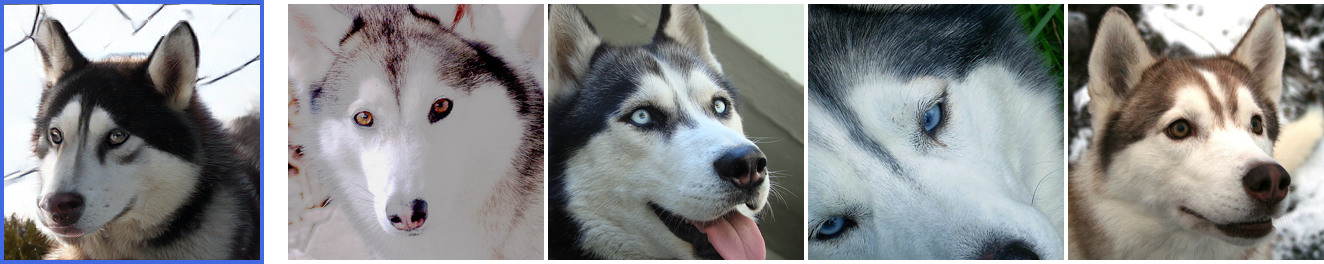}\par\vspace{1pt}
\makebox[\linewidth][l]{\fontsize{5.5}{6}\selectfont \makebox[0.1993958\linewidth][c]{\textcolor{xblue}{\textbf{Generated}}}%
\hspace{0.0181269\linewidth}%
\makebox[0.1933535\linewidth][c]{NN 1}%
\hspace{0.0030211\linewidth}%
\makebox[0.1933535\linewidth][c]{NN 2}%
\hspace{0.0030211\linewidth}%
\makebox[0.1933535\linewidth][c]{NN 3}%
\hspace{0.0030211\linewidth}%
\makebox[0.1933535\linewidth][c]{NN 4}}
\end{minipage}\end{minipage}\par\vspace{4pt}
\noindent\begin{minipage}{\linewidth}
\begin{minipage}[t]{0.485\linewidth}\centering
{\fontsize{7}{8}\selectfont tabby (class 281)\par}\vspace{1pt}
\includegraphics[width=\linewidth]{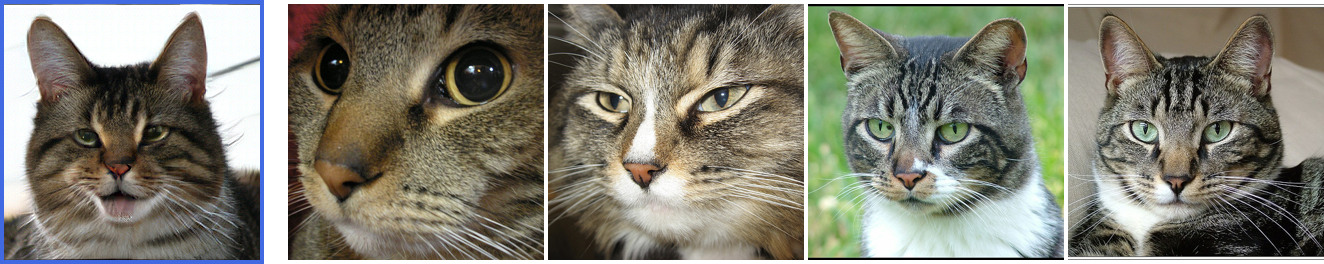}\par\vspace{1pt}
\makebox[\linewidth][l]{\fontsize{5.5}{6}\selectfont \makebox[0.1993958\linewidth][c]{\textcolor{xblue}{\textbf{Generated}}}%
\hspace{0.0181269\linewidth}%
\makebox[0.1933535\linewidth][c]{NN 1}%
\hspace{0.0030211\linewidth}%
\makebox[0.1933535\linewidth][c]{NN 2}%
\hspace{0.0030211\linewidth}%
\makebox[0.1933535\linewidth][c]{NN 3}%
\hspace{0.0030211\linewidth}%
\makebox[0.1933535\linewidth][c]{NN 4}}
\end{minipage}\hfill%
\begin{minipage}[t]{0.485\linewidth}\centering
{\fontsize{7}{8}\selectfont lion (class 291)\par}\vspace{1pt}
\includegraphics[width=\linewidth]{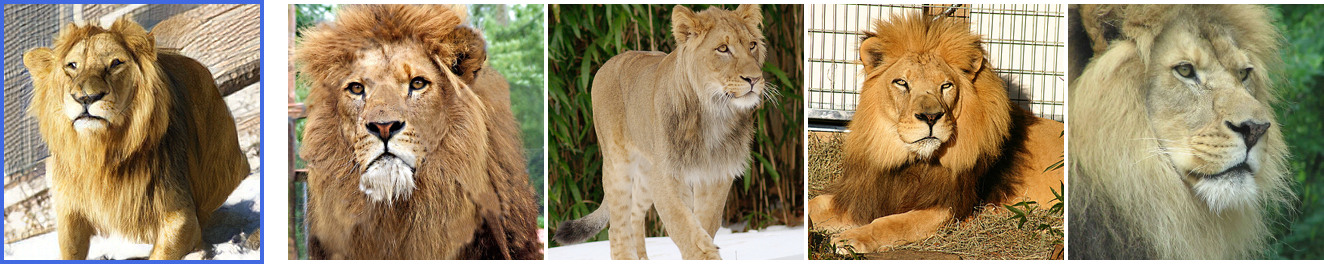}\par\vspace{1pt}
\makebox[\linewidth][l]{\fontsize{5.5}{6}\selectfont \makebox[0.1993958\linewidth][c]{\textcolor{xblue}{\textbf{Generated}}}%
\hspace{0.0181269\linewidth}%
\makebox[0.1933535\linewidth][c]{NN 1}%
\hspace{0.0030211\linewidth}%
\makebox[0.1933535\linewidth][c]{NN 2}%
\hspace{0.0030211\linewidth}%
\makebox[0.1933535\linewidth][c]{NN 3}%
\hspace{0.0030211\linewidth}%
\makebox[0.1933535\linewidth][c]{NN 4}}
\end{minipage}\end{minipage}\par\vspace{4pt}
\noindent\begin{minipage}{\linewidth}
\begin{minipage}[t]{0.485\linewidth}\centering
{\fontsize{7}{8}\selectfont tiger (class 292)\par}\vspace{1pt}
\includegraphics[width=\linewidth]{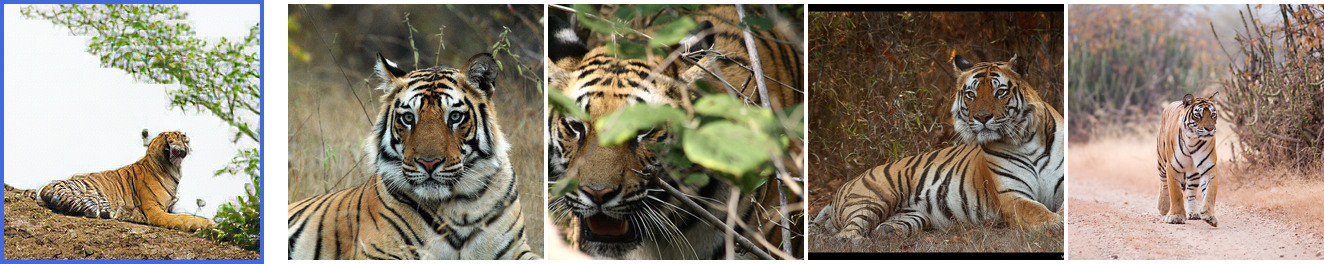}\par\vspace{1pt}
\makebox[\linewidth][l]{\fontsize{5.5}{6}\selectfont \makebox[0.1993958\linewidth][c]{\textcolor{xblue}{\textbf{Generated}}}%
\hspace{0.0181269\linewidth}%
\makebox[0.1933535\linewidth][c]{NN 1}%
\hspace{0.0030211\linewidth}%
\makebox[0.1933535\linewidth][c]{NN 2}%
\hspace{0.0030211\linewidth}%
\makebox[0.1933535\linewidth][c]{NN 3}%
\hspace{0.0030211\linewidth}%
\makebox[0.1933535\linewidth][c]{NN 4}}
\end{minipage}\hfill%
\begin{minipage}[t]{0.485\linewidth}\centering
{\fontsize{7}{8}\selectfont ice bear (class 296)\par}\vspace{1pt}
\includegraphics[width=\linewidth]{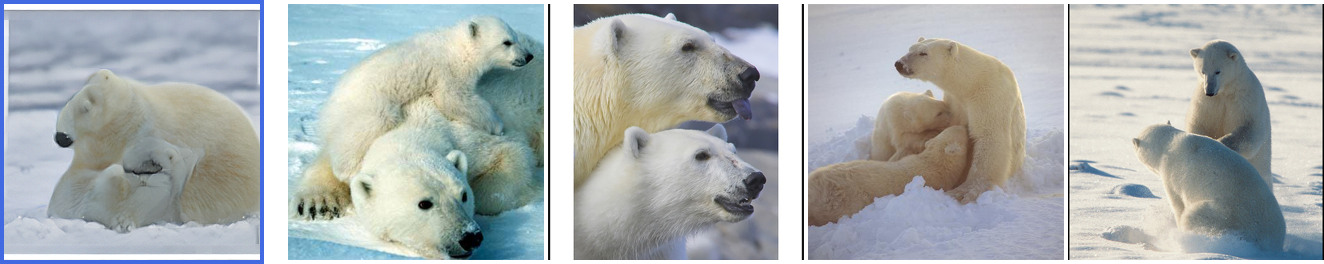}\par\vspace{1pt}
\makebox[\linewidth][l]{\fontsize{5.5}{6}\selectfont \makebox[0.1993958\linewidth][c]{\textcolor{xblue}{\textbf{Generated}}}%
\hspace{0.0181269\linewidth}%
\makebox[0.1933535\linewidth][c]{NN 1}%
\hspace{0.0030211\linewidth}%
\makebox[0.1933535\linewidth][c]{NN 2}%
\hspace{0.0030211\linewidth}%
\makebox[0.1933535\linewidth][c]{NN 3}%
\hspace{0.0030211\linewidth}%
\makebox[0.1933535\linewidth][c]{NN 4}}
\end{minipage}\end{minipage}\par\vspace{4pt}
\noindent\begin{minipage}{\linewidth}
\begin{minipage}[t]{0.485\linewidth}\centering
{\fontsize{7}{8}\selectfont zebra (class 340)\par}\vspace{1pt}
\includegraphics[width=\linewidth]{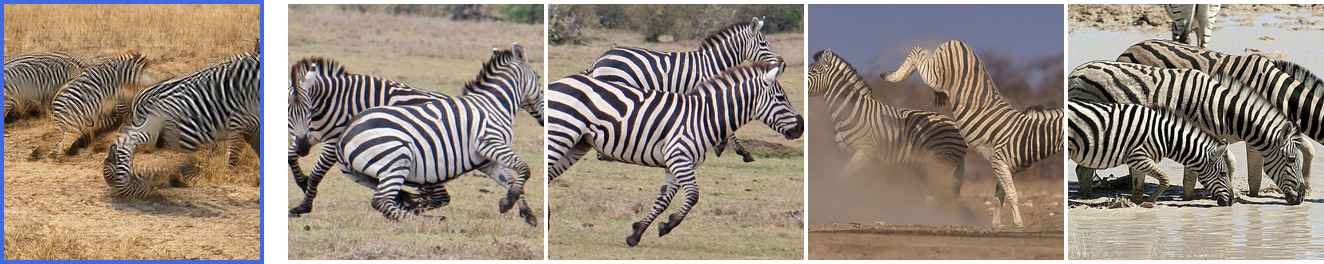}\par\vspace{1pt}
\makebox[\linewidth][l]{\fontsize{5.5}{6}\selectfont \makebox[0.1993958\linewidth][c]{\textcolor{xblue}{\textbf{Generated}}}%
\hspace{0.0181269\linewidth}%
\makebox[0.1933535\linewidth][c]{NN 1}%
\hspace{0.0030211\linewidth}%
\makebox[0.1933535\linewidth][c]{NN 2}%
\hspace{0.0030211\linewidth}%
\makebox[0.1933535\linewidth][c]{NN 3}%
\hspace{0.0030211\linewidth}%
\makebox[0.1933535\linewidth][c]{NN 4}}
\end{minipage}\hfill%
\begin{minipage}[t]{0.485\linewidth}\centering
{\fontsize{7}{8}\selectfont otter (class 360)\par}\vspace{1pt}
\includegraphics[width=\linewidth]{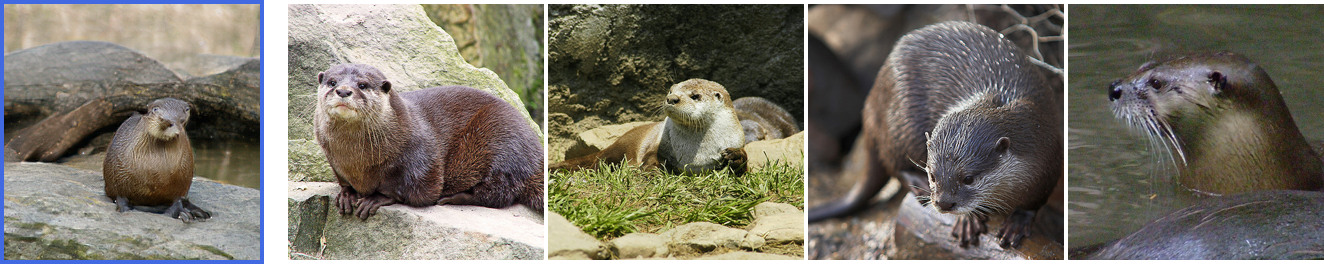}\par\vspace{1pt}
\makebox[\linewidth][l]{\fontsize{5.5}{6}\selectfont \makebox[0.1993958\linewidth][c]{\textcolor{xblue}{\textbf{Generated}}}%
\hspace{0.0181269\linewidth}%
\makebox[0.1933535\linewidth][c]{NN 1}%
\hspace{0.0030211\linewidth}%
\makebox[0.1933535\linewidth][c]{NN 2}%
\hspace{0.0030211\linewidth}%
\makebox[0.1933535\linewidth][c]{NN 3}%
\hspace{0.0030211\linewidth}%
\makebox[0.1933535\linewidth][c]{NN 4}}
\end{minipage}\end{minipage}\par\vspace{4pt}
\noindent\begin{minipage}{\linewidth}
\begin{minipage}[t]{0.485\linewidth}\centering
{\fontsize{7}{8}\selectfont African elephant (class 386)\par}\vspace{1pt}
\includegraphics[width=\linewidth]{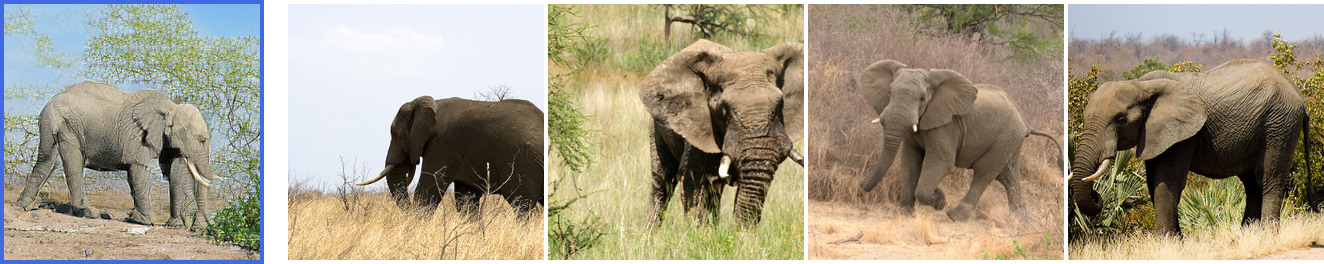}\par\vspace{1pt}
\makebox[\linewidth][l]{\fontsize{5.5}{6}\selectfont \makebox[0.1993958\linewidth][c]{\textcolor{xblue}{\textbf{Generated}}}%
\hspace{0.0181269\linewidth}%
\makebox[0.1933535\linewidth][c]{NN 1}%
\hspace{0.0030211\linewidth}%
\makebox[0.1933535\linewidth][c]{NN 2}%
\hspace{0.0030211\linewidth}%
\makebox[0.1933535\linewidth][c]{NN 3}%
\hspace{0.0030211\linewidth}%
\makebox[0.1933535\linewidth][c]{NN 4}}
\end{minipage}\hfill%
\begin{minipage}[t]{0.485\linewidth}\centering
{\fontsize{7}{8}\selectfont lesser panda (class 387)\par}\vspace{1pt}
\includegraphics[width=\linewidth]{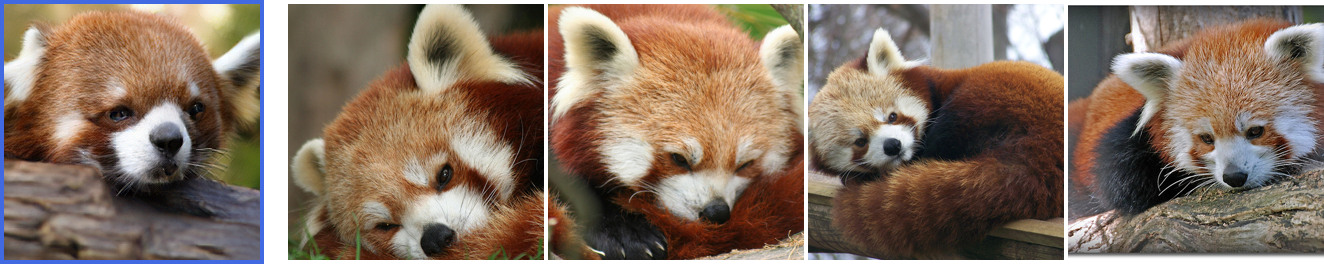}\par\vspace{1pt}
\makebox[\linewidth][l]{\fontsize{5.5}{6}\selectfont \makebox[0.1993958\linewidth][c]{\textcolor{xblue}{\textbf{Generated}}}%
\hspace{0.0181269\linewidth}%
\makebox[0.1933535\linewidth][c]{NN 1}%
\hspace{0.0030211\linewidth}%
\makebox[0.1933535\linewidth][c]{NN 2}%
\hspace{0.0030211\linewidth}%
\makebox[0.1933535\linewidth][c]{NN 3}%
\hspace{0.0030211\linewidth}%
\makebox[0.1933535\linewidth][c]{NN 4}}
\end{minipage}\end{minipage}\par\vspace{4pt}
\noindent\begin{minipage}{\linewidth}
\begin{minipage}[t]{0.485\linewidth}\centering
{\fontsize{7}{8}\selectfont giant panda (class 388)\par}\vspace{1pt}
\includegraphics[width=\linewidth]{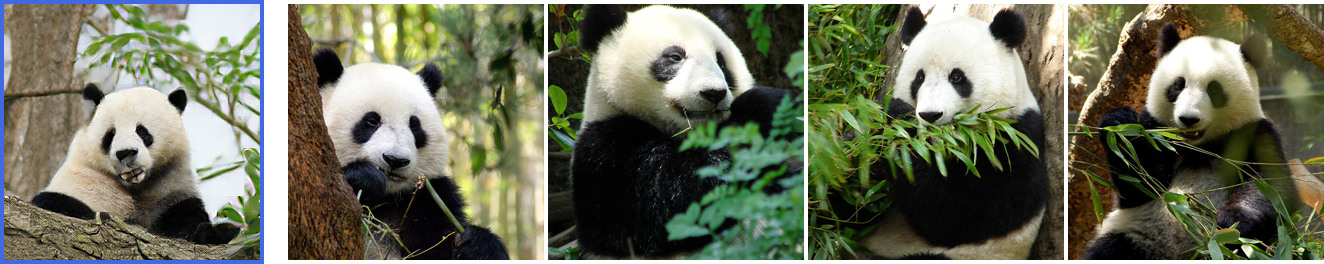}\par\vspace{1pt}
\makebox[\linewidth][l]{\fontsize{5.5}{6}\selectfont \makebox[0.1993958\linewidth][c]{\textcolor{xblue}{\textbf{Generated}}}%
\hspace{0.0181269\linewidth}%
\makebox[0.1933535\linewidth][c]{NN 1}%
\hspace{0.0030211\linewidth}%
\makebox[0.1933535\linewidth][c]{NN 2}%
\hspace{0.0030211\linewidth}%
\makebox[0.1933535\linewidth][c]{NN 3}%
\hspace{0.0030211\linewidth}%
\makebox[0.1933535\linewidth][c]{NN 4}}
\end{minipage}\hfill%
\begin{minipage}[t]{0.485\linewidth}\centering
{\fontsize{7}{8}\selectfont balloon (class 417)\par}\vspace{1pt}
\includegraphics[width=\linewidth]{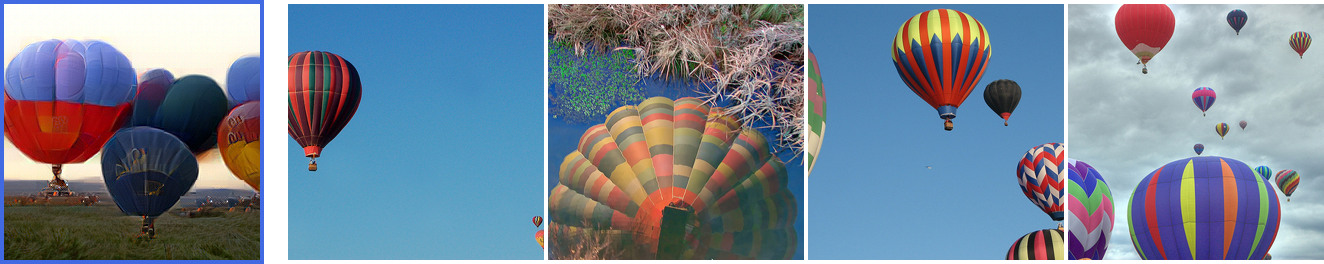}\par\vspace{1pt}
\makebox[\linewidth][l]{\fontsize{5.5}{6}\selectfont \makebox[0.1993958\linewidth][c]{\textcolor{xblue}{\textbf{Generated}}}%
\hspace{0.0181269\linewidth}%
\makebox[0.1933535\linewidth][c]{NN 1}%
\hspace{0.0030211\linewidth}%
\makebox[0.1933535\linewidth][c]{NN 2}%
\hspace{0.0030211\linewidth}%
\makebox[0.1933535\linewidth][c]{NN 3}%
\hspace{0.0030211\linewidth}%
\makebox[0.1933535\linewidth][c]{NN 4}}
\end{minipage}\end{minipage}\par\vspace{4pt}
\noindent\begin{minipage}{\linewidth}
\begin{minipage}[t]{0.485\linewidth}\centering
{\fontsize{7}{8}\selectfont sports car (class 817)\par}\vspace{1pt}
\includegraphics[width=\linewidth]{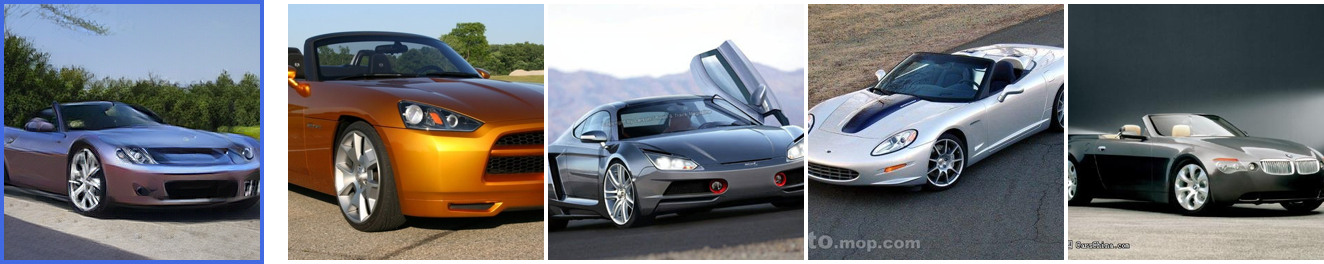}\par\vspace{1pt}
\makebox[\linewidth][l]{\fontsize{5.5}{6}\selectfont \makebox[0.1993958\linewidth][c]{\textcolor{xblue}{\textbf{Generated}}}%
\hspace{0.0181269\linewidth}%
\makebox[0.1933535\linewidth][c]{NN 1}%
\hspace{0.0030211\linewidth}%
\makebox[0.1933535\linewidth][c]{NN 2}%
\hspace{0.0030211\linewidth}%
\makebox[0.1933535\linewidth][c]{NN 3}%
\hspace{0.0030211\linewidth}%
\makebox[0.1933535\linewidth][c]{NN 4}}
\end{minipage}\hfill%
\begin{minipage}[t]{0.485\linewidth}\centering
{\fontsize{7}{8}\selectfont geyser (class 974)\par}\vspace{1pt}
\includegraphics[width=\linewidth]{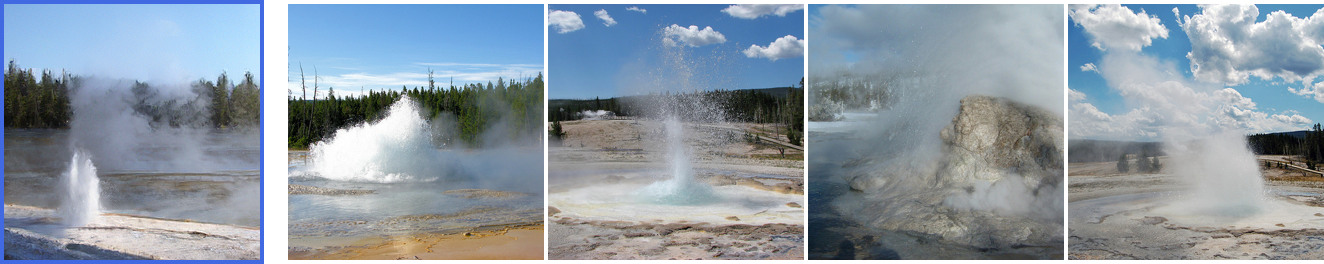}\par\vspace{1pt}
\makebox[\linewidth][l]{\fontsize{5.5}{6}\selectfont \makebox[0.1993958\linewidth][c]{\textcolor{xblue}{\textbf{Generated}}}%
\hspace{0.0181269\linewidth}%
\makebox[0.1933535\linewidth][c]{NN 1}%
\hspace{0.0030211\linewidth}%
\makebox[0.1933535\linewidth][c]{NN 2}%
\hspace{0.0030211\linewidth}%
\makebox[0.1933535\linewidth][c]{NN 3}%
\hspace{0.0030211\linewidth}%
\makebox[0.1933535\linewidth][c]{NN 4}}
\end{minipage}\end{minipage}\par\vspace{4pt}
\noindent\begin{minipage}{\linewidth}
\begin{minipage}[t]{0.485\linewidth}\centering
{\fontsize{7}{8}\selectfont volcano (class 980)\par}\vspace{1pt}
\includegraphics[width=\linewidth]{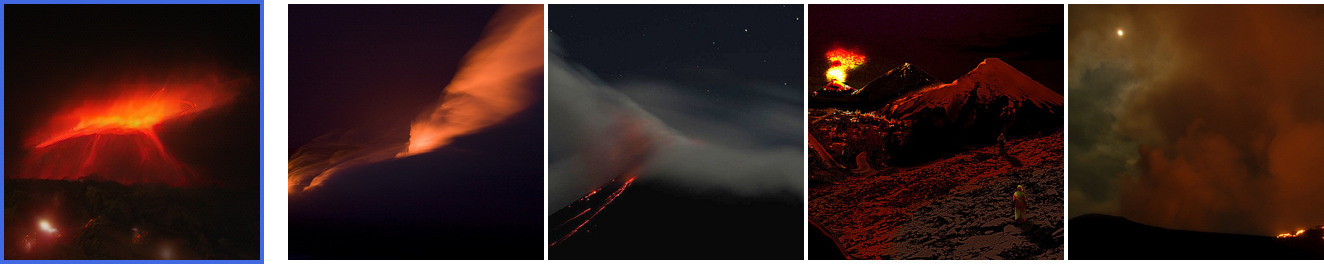}\par\vspace{1pt}
\makebox[\linewidth][l]{\fontsize{5.5}{6}\selectfont \makebox[0.1993958\linewidth][c]{\textcolor{xblue}{\textbf{Generated}}}%
\hspace{0.0181269\linewidth}%
\makebox[0.1933535\linewidth][c]{NN 1}%
\hspace{0.0030211\linewidth}%
\makebox[0.1933535\linewidth][c]{NN 2}%
\hspace{0.0030211\linewidth}%
\makebox[0.1933535\linewidth][c]{NN 3}%
\hspace{0.0030211\linewidth}%
\makebox[0.1933535\linewidth][c]{NN 4}}
\end{minipage}\hfill%
\begin{minipage}[t]{0.485\linewidth}\centering
{\fontsize{7}{8}\selectfont peacock (class 84)\par}\vspace{1pt}
\includegraphics[width=\linewidth]{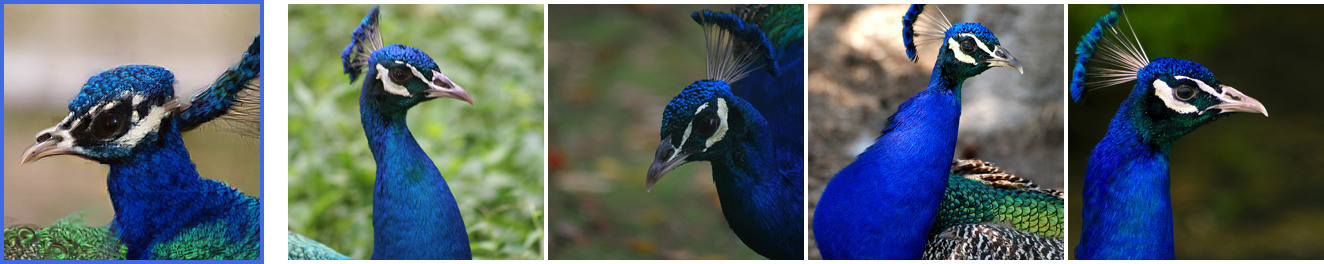}\par\vspace{1pt}
\makebox[\linewidth][l]{\fontsize{5.5}{6}\selectfont \makebox[0.1993958\linewidth][c]{\textcolor{xblue}{\textbf{Generated}}}%
\hspace{0.0181269\linewidth}%
\makebox[0.1933535\linewidth][c]{NN 1}%
\hspace{0.0030211\linewidth}%
\makebox[0.1933535\linewidth][c]{NN 2}%
\hspace{0.0030211\linewidth}%
\makebox[0.1933535\linewidth][c]{NN 3}%
\hspace{0.0030211\linewidth}%
\makebox[0.1933535\linewidth][c]{NN 4}}
\end{minipage}\end{minipage}\par\vspace{4pt}
\noindent\begin{minipage}{\linewidth}
\textbf{Priming-L (160 ep.)}\par\vspace{3pt}
\begin{minipage}[t]{0.485\linewidth}\centering
{\fontsize{7}{8}\selectfont goldfish (class 1)\par}\vspace{1pt}
\includegraphics[width=\linewidth]{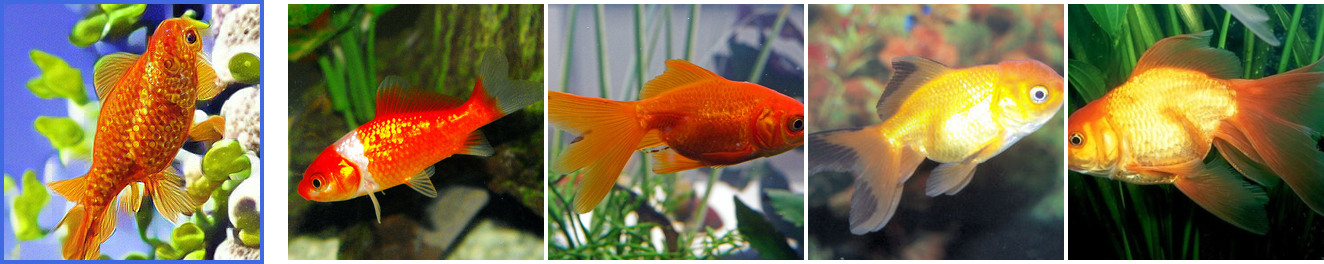}\par\vspace{1pt}
\makebox[\linewidth][l]{\fontsize{5.5}{6}\selectfont \makebox[0.1993958\linewidth][c]{\textcolor{xblue}{\textbf{Generated}}}%
\hspace{0.0181269\linewidth}%
\makebox[0.1933535\linewidth][c]{NN 1}%
\hspace{0.0030211\linewidth}%
\makebox[0.1933535\linewidth][c]{NN 2}%
\hspace{0.0030211\linewidth}%
\makebox[0.1933535\linewidth][c]{NN 3}%
\hspace{0.0030211\linewidth}%
\makebox[0.1933535\linewidth][c]{NN 4}}
\end{minipage}\hfill%
\begin{minipage}[t]{0.485\linewidth}\centering
{\fontsize{7}{8}\selectfont bald eagle (class 22)\par}\vspace{1pt}
\includegraphics[width=\linewidth]{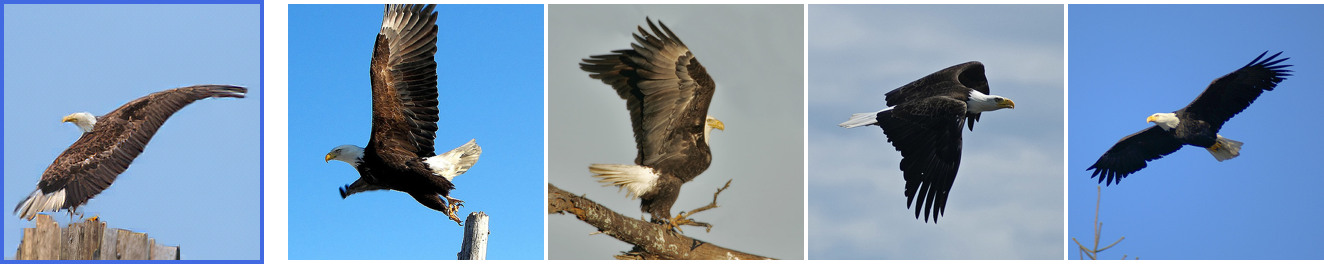}\par\vspace{1pt}
\makebox[\linewidth][l]{\fontsize{5.5}{6}\selectfont \makebox[0.1993958\linewidth][c]{\textcolor{xblue}{\textbf{Generated}}}%
\hspace{0.0181269\linewidth}%
\makebox[0.1933535\linewidth][c]{NN 1}%
\hspace{0.0030211\linewidth}%
\makebox[0.1933535\linewidth][c]{NN 2}%
\hspace{0.0030211\linewidth}%
\makebox[0.1933535\linewidth][c]{NN 3}%
\hspace{0.0030211\linewidth}%
\makebox[0.1933535\linewidth][c]{NN 4}}
\end{minipage}\end{minipage}\par\vspace{4pt}
\noindent\begin{minipage}{\linewidth}
\begin{minipage}[t]{0.485\linewidth}\centering
{\fontsize{7}{8}\selectfont macaw (class 88)\par}\vspace{1pt}
\includegraphics[width=\linewidth]{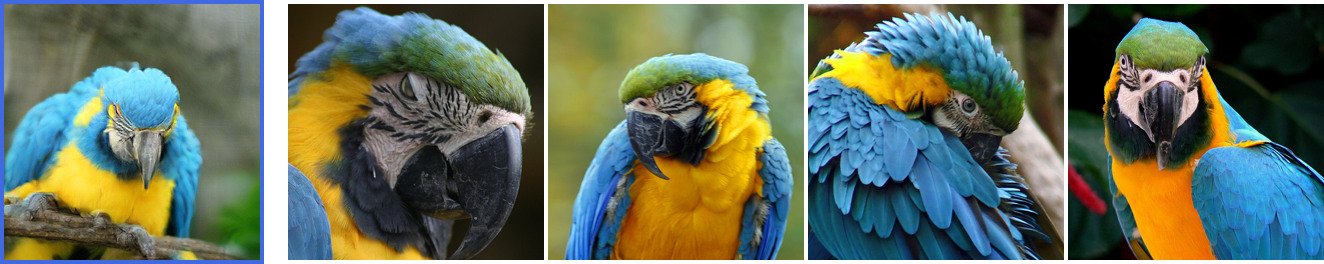}\par\vspace{1pt}
\makebox[\linewidth][l]{\fontsize{5.5}{6}\selectfont \makebox[0.1993958\linewidth][c]{\textcolor{xblue}{\textbf{Generated}}}%
\hspace{0.0181269\linewidth}%
\makebox[0.1933535\linewidth][c]{NN 1}%
\hspace{0.0030211\linewidth}%
\makebox[0.1933535\linewidth][c]{NN 2}%
\hspace{0.0030211\linewidth}%
\makebox[0.1933535\linewidth][c]{NN 3}%
\hspace{0.0030211\linewidth}%
\makebox[0.1933535\linewidth][c]{NN 4}}
\end{minipage}\hfill%
\begin{minipage}[t]{0.485\linewidth}\centering
{\fontsize{7}{8}\selectfont flamingo (class 130)\par}\vspace{1pt}
\includegraphics[width=\linewidth]{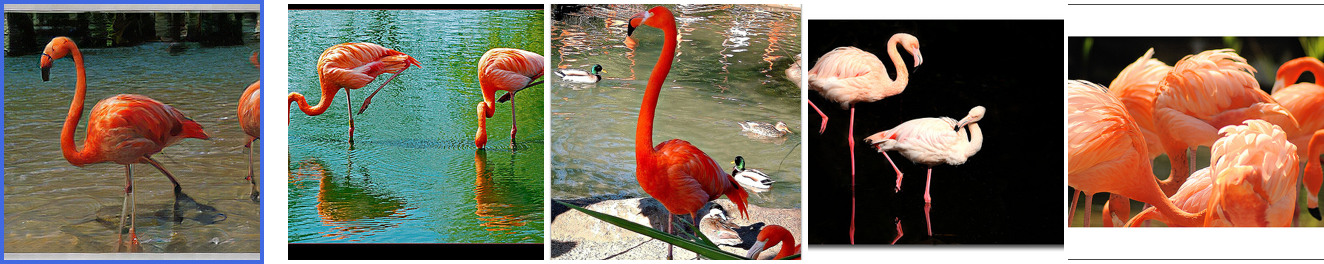}\par\vspace{1pt}
\makebox[\linewidth][l]{\fontsize{5.5}{6}\selectfont \makebox[0.1993958\linewidth][c]{\textcolor{xblue}{\textbf{Generated}}}%
\hspace{0.0181269\linewidth}%
\makebox[0.1933535\linewidth][c]{NN 1}%
\hspace{0.0030211\linewidth}%
\makebox[0.1933535\linewidth][c]{NN 2}%
\hspace{0.0030211\linewidth}%
\makebox[0.1933535\linewidth][c]{NN 3}%
\hspace{0.0030211\linewidth}%
\makebox[0.1933535\linewidth][c]{NN 4}}
\end{minipage}\end{minipage}\par\vspace{4pt}
\noindent\begin{minipage}{\linewidth}
\begin{minipage}[t]{0.485\linewidth}\centering
{\fontsize{7}{8}\selectfont golden retriever (class 207)\par}\vspace{1pt}
\includegraphics[width=\linewidth]{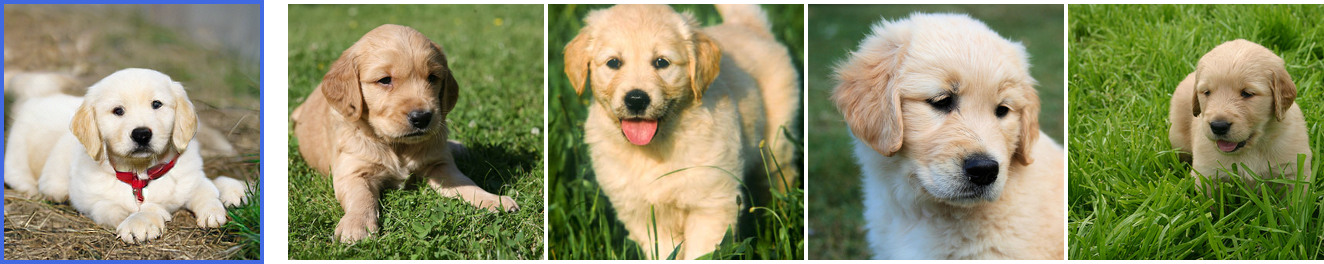}\par\vspace{1pt}
\makebox[\linewidth][l]{\fontsize{5.5}{6}\selectfont \makebox[0.1993958\linewidth][c]{\textcolor{xblue}{\textbf{Generated}}}%
\hspace{0.0181269\linewidth}%
\makebox[0.1933535\linewidth][c]{NN 1}%
\hspace{0.0030211\linewidth}%
\makebox[0.1933535\linewidth][c]{NN 2}%
\hspace{0.0030211\linewidth}%
\makebox[0.1933535\linewidth][c]{NN 3}%
\hspace{0.0030211\linewidth}%
\makebox[0.1933535\linewidth][c]{NN 4}}
\end{minipage}\hfill%
\begin{minipage}[t]{0.485\linewidth}\centering
{\fontsize{7}{8}\selectfont Siberian husky (class 250)\par}\vspace{1pt}
\includegraphics[width=\linewidth]{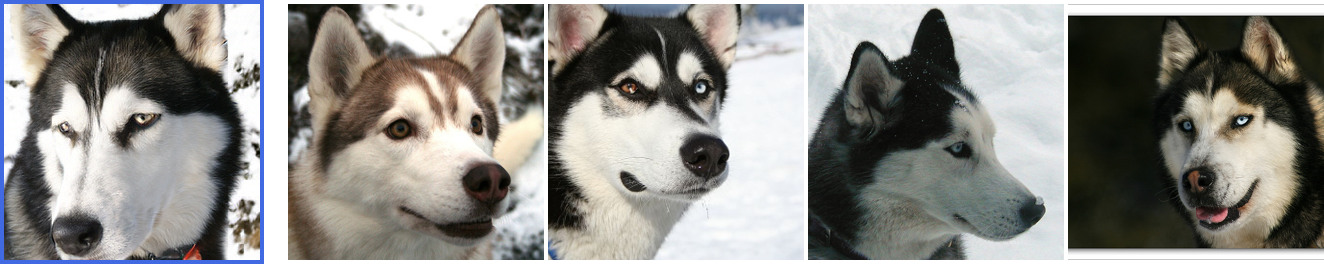}\par\vspace{1pt}
\makebox[\linewidth][l]{\fontsize{5.5}{6}\selectfont \makebox[0.1993958\linewidth][c]{\textcolor{xblue}{\textbf{Generated}}}%
\hspace{0.0181269\linewidth}%
\makebox[0.1933535\linewidth][c]{NN 1}%
\hspace{0.0030211\linewidth}%
\makebox[0.1933535\linewidth][c]{NN 2}%
\hspace{0.0030211\linewidth}%
\makebox[0.1933535\linewidth][c]{NN 3}%
\hspace{0.0030211\linewidth}%
\makebox[0.1933535\linewidth][c]{NN 4}}
\end{minipage}\end{minipage}\par\vspace{4pt}
\noindent\begin{minipage}{\linewidth}
\begin{minipage}[t]{0.485\linewidth}\centering
{\fontsize{7}{8}\selectfont tabby (class 281)\par}\vspace{1pt}
\includegraphics[width=\linewidth]{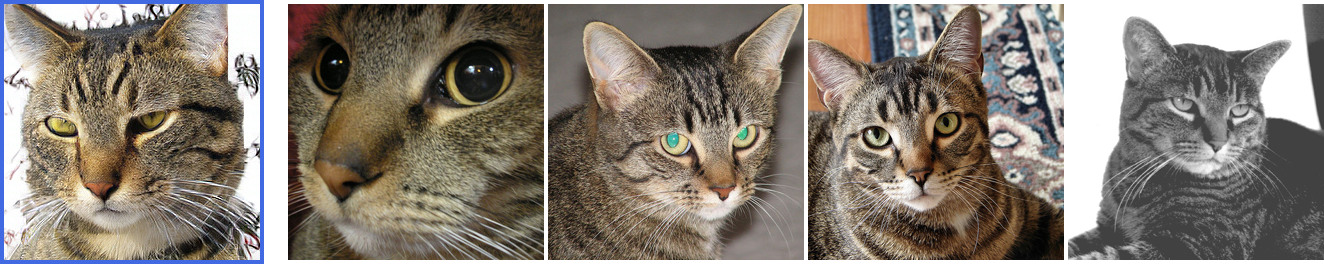}\par\vspace{1pt}
\makebox[\linewidth][l]{\fontsize{5.5}{6}\selectfont \makebox[0.1993958\linewidth][c]{\textcolor{xblue}{\textbf{Generated}}}%
\hspace{0.0181269\linewidth}%
\makebox[0.1933535\linewidth][c]{NN 1}%
\hspace{0.0030211\linewidth}%
\makebox[0.1933535\linewidth][c]{NN 2}%
\hspace{0.0030211\linewidth}%
\makebox[0.1933535\linewidth][c]{NN 3}%
\hspace{0.0030211\linewidth}%
\makebox[0.1933535\linewidth][c]{NN 4}}
\end{minipage}\hfill%
\begin{minipage}[t]{0.485\linewidth}\centering
{\fontsize{7}{8}\selectfont lion (class 291)\par}\vspace{1pt}
\includegraphics[width=\linewidth]{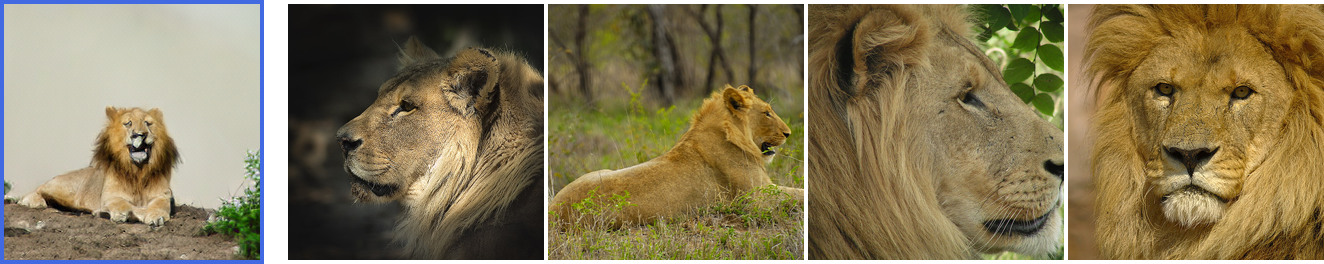}\par\vspace{1pt}
\makebox[\linewidth][l]{\fontsize{5.5}{6}\selectfont \makebox[0.1993958\linewidth][c]{\textcolor{xblue}{\textbf{Generated}}}%
\hspace{0.0181269\linewidth}%
\makebox[0.1933535\linewidth][c]{NN 1}%
\hspace{0.0030211\linewidth}%
\makebox[0.1933535\linewidth][c]{NN 2}%
\hspace{0.0030211\linewidth}%
\makebox[0.1933535\linewidth][c]{NN 3}%
\hspace{0.0030211\linewidth}%
\makebox[0.1933535\linewidth][c]{NN 4}}
\end{minipage}\end{minipage}\par\vspace{4pt}
\noindent\begin{minipage}{\linewidth}
\begin{minipage}[t]{0.485\linewidth}\centering
{\fontsize{7}{8}\selectfont tiger (class 292)\par}\vspace{1pt}
\includegraphics[width=\linewidth]{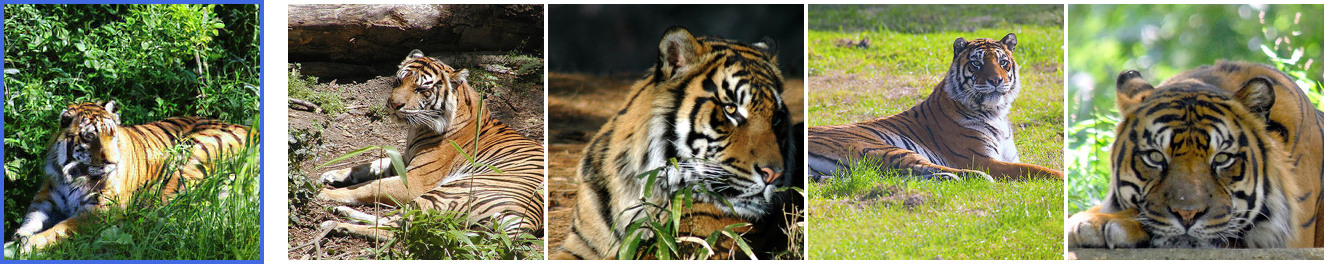}\par\vspace{1pt}
\makebox[\linewidth][l]{\fontsize{5.5}{6}\selectfont \makebox[0.1993958\linewidth][c]{\textcolor{xblue}{\textbf{Generated}}}%
\hspace{0.0181269\linewidth}%
\makebox[0.1933535\linewidth][c]{NN 1}%
\hspace{0.0030211\linewidth}%
\makebox[0.1933535\linewidth][c]{NN 2}%
\hspace{0.0030211\linewidth}%
\makebox[0.1933535\linewidth][c]{NN 3}%
\hspace{0.0030211\linewidth}%
\makebox[0.1933535\linewidth][c]{NN 4}}
\end{minipage}\hfill%
\begin{minipage}[t]{0.485\linewidth}\centering
{\fontsize{7}{8}\selectfont ice bear (class 296)\par}\vspace{1pt}
\includegraphics[width=\linewidth]{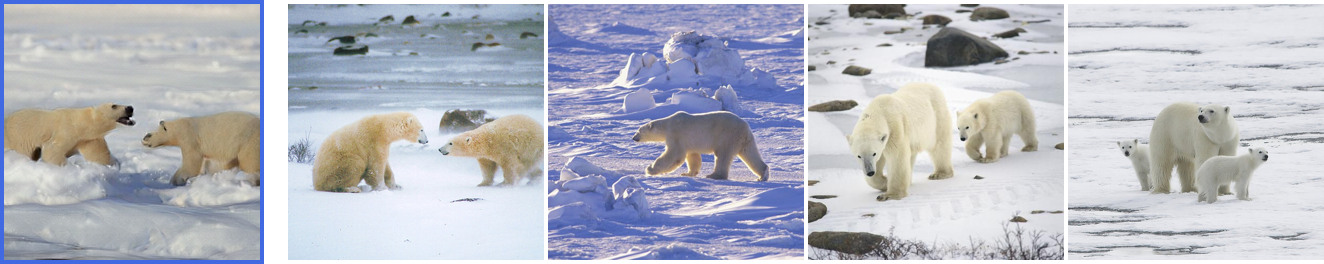}\par\vspace{1pt}
\makebox[\linewidth][l]{\fontsize{5.5}{6}\selectfont \makebox[0.1993958\linewidth][c]{\textcolor{xblue}{\textbf{Generated}}}%
\hspace{0.0181269\linewidth}%
\makebox[0.1933535\linewidth][c]{NN 1}%
\hspace{0.0030211\linewidth}%
\makebox[0.1933535\linewidth][c]{NN 2}%
\hspace{0.0030211\linewidth}%
\makebox[0.1933535\linewidth][c]{NN 3}%
\hspace{0.0030211\linewidth}%
\makebox[0.1933535\linewidth][c]{NN 4}}
\end{minipage}\end{minipage}\par\vspace{4pt}
\noindent\begin{minipage}{\linewidth}
\begin{minipage}[t]{0.485\linewidth}\centering
{\fontsize{7}{8}\selectfont zebra (class 340)\par}\vspace{1pt}
\includegraphics[width=\linewidth]{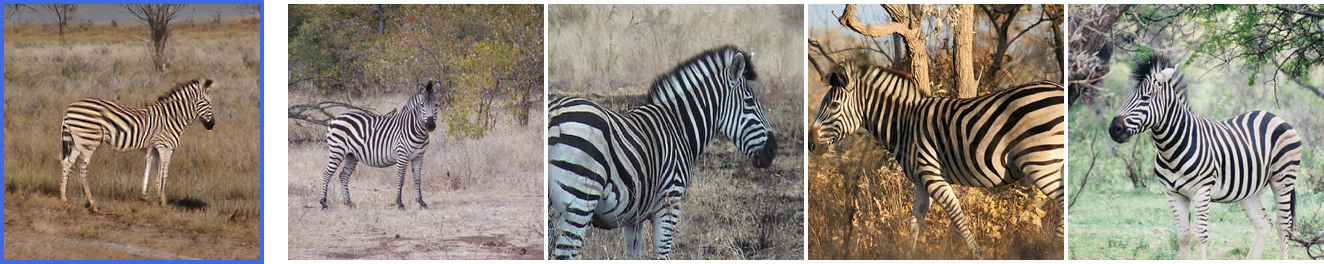}\par\vspace{1pt}
\makebox[\linewidth][l]{\fontsize{5.5}{6}\selectfont \makebox[0.1993958\linewidth][c]{\textcolor{xblue}{\textbf{Generated}}}%
\hspace{0.0181269\linewidth}%
\makebox[0.1933535\linewidth][c]{NN 1}%
\hspace{0.0030211\linewidth}%
\makebox[0.1933535\linewidth][c]{NN 2}%
\hspace{0.0030211\linewidth}%
\makebox[0.1933535\linewidth][c]{NN 3}%
\hspace{0.0030211\linewidth}%
\makebox[0.1933535\linewidth][c]{NN 4}}
\end{minipage}\hfill%
\begin{minipage}[t]{0.485\linewidth}\centering
{\fontsize{7}{8}\selectfont otter (class 360)\par}\vspace{1pt}
\includegraphics[width=\linewidth]{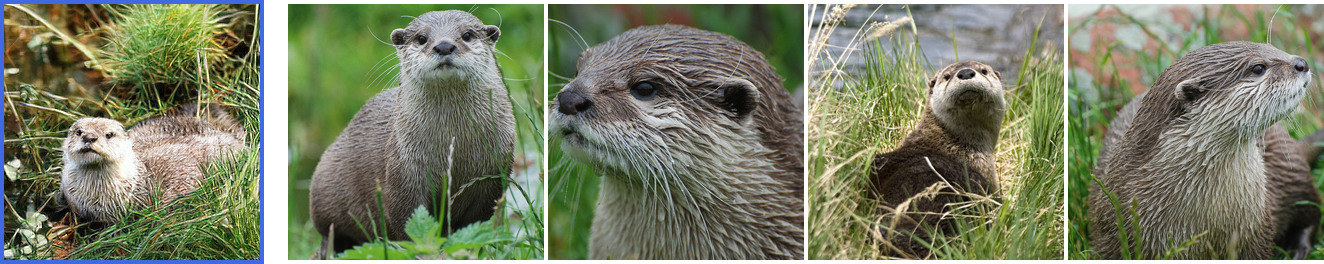}\par\vspace{1pt}
\makebox[\linewidth][l]{\fontsize{5.5}{6}\selectfont \makebox[0.1993958\linewidth][c]{\textcolor{xblue}{\textbf{Generated}}}%
\hspace{0.0181269\linewidth}%
\makebox[0.1933535\linewidth][c]{NN 1}%
\hspace{0.0030211\linewidth}%
\makebox[0.1933535\linewidth][c]{NN 2}%
\hspace{0.0030211\linewidth}%
\makebox[0.1933535\linewidth][c]{NN 3}%
\hspace{0.0030211\linewidth}%
\makebox[0.1933535\linewidth][c]{NN 4}}
\end{minipage}\end{minipage}\par\vspace{4pt}
\noindent\begin{minipage}{\linewidth}
\begin{minipage}[t]{0.485\linewidth}\centering
{\fontsize{7}{8}\selectfont African elephant (class 386)\par}\vspace{1pt}
\includegraphics[width=\linewidth]{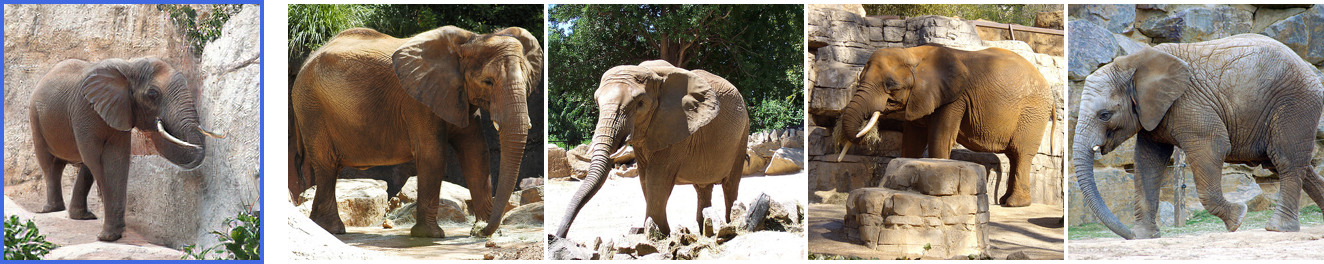}\par\vspace{1pt}
\makebox[\linewidth][l]{\fontsize{5.5}{6}\selectfont \makebox[0.1993958\linewidth][c]{\textcolor{xblue}{\textbf{Generated}}}%
\hspace{0.0181269\linewidth}%
\makebox[0.1933535\linewidth][c]{NN 1}%
\hspace{0.0030211\linewidth}%
\makebox[0.1933535\linewidth][c]{NN 2}%
\hspace{0.0030211\linewidth}%
\makebox[0.1933535\linewidth][c]{NN 3}%
\hspace{0.0030211\linewidth}%
\makebox[0.1933535\linewidth][c]{NN 4}}
\end{minipage}\hfill%
\begin{minipage}[t]{0.485\linewidth}\centering
{\fontsize{7}{8}\selectfont lesser panda (class 387)\par}\vspace{1pt}
\includegraphics[width=\linewidth]{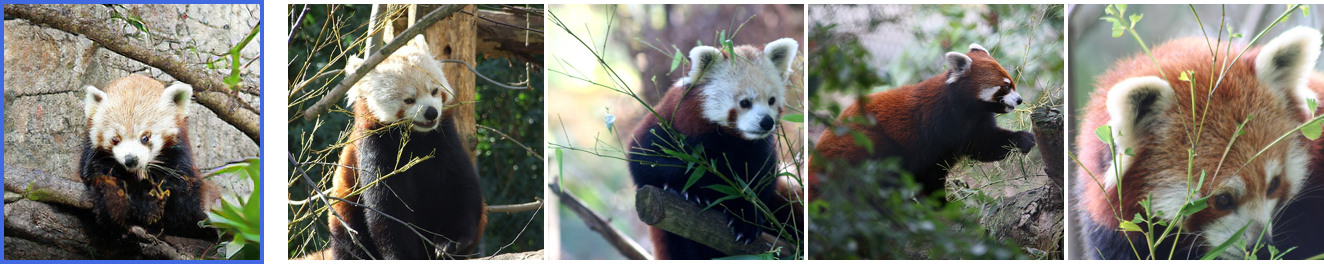}\par\vspace{1pt}
\makebox[\linewidth][l]{\fontsize{5.5}{6}\selectfont \makebox[0.1993958\linewidth][c]{\textcolor{xblue}{\textbf{Generated}}}%
\hspace{0.0181269\linewidth}%
\makebox[0.1933535\linewidth][c]{NN 1}%
\hspace{0.0030211\linewidth}%
\makebox[0.1933535\linewidth][c]{NN 2}%
\hspace{0.0030211\linewidth}%
\makebox[0.1933535\linewidth][c]{NN 3}%
\hspace{0.0030211\linewidth}%
\makebox[0.1933535\linewidth][c]{NN 4}}
\end{minipage}\end{minipage}\par\vspace{4pt}
\noindent\begin{minipage}{\linewidth}
\begin{minipage}[t]{0.485\linewidth}\centering
{\fontsize{7}{8}\selectfont giant panda (class 388)\par}\vspace{1pt}
\includegraphics[width=\linewidth]{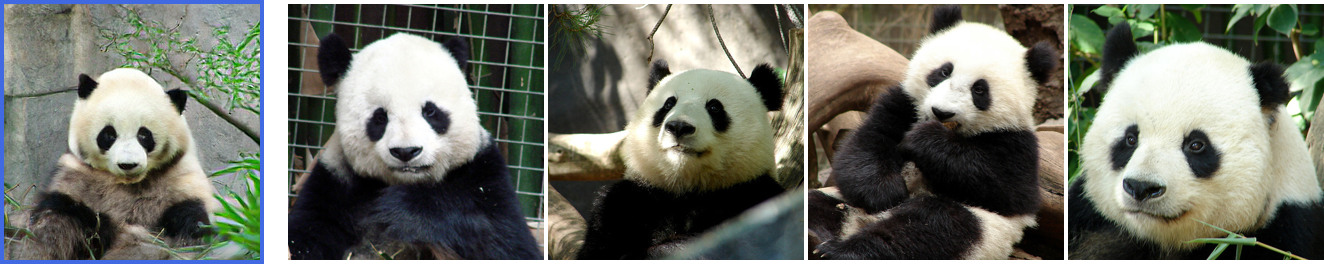}\par\vspace{1pt}
\makebox[\linewidth][l]{\fontsize{5.5}{6}\selectfont \makebox[0.1993958\linewidth][c]{\textcolor{xblue}{\textbf{Generated}}}%
\hspace{0.0181269\linewidth}%
\makebox[0.1933535\linewidth][c]{NN 1}%
\hspace{0.0030211\linewidth}%
\makebox[0.1933535\linewidth][c]{NN 2}%
\hspace{0.0030211\linewidth}%
\makebox[0.1933535\linewidth][c]{NN 3}%
\hspace{0.0030211\linewidth}%
\makebox[0.1933535\linewidth][c]{NN 4}}
\end{minipage}\hfill%
\begin{minipage}[t]{0.485\linewidth}\centering
{\fontsize{7}{8}\selectfont balloon (class 417)\par}\vspace{1pt}
\includegraphics[width=\linewidth]{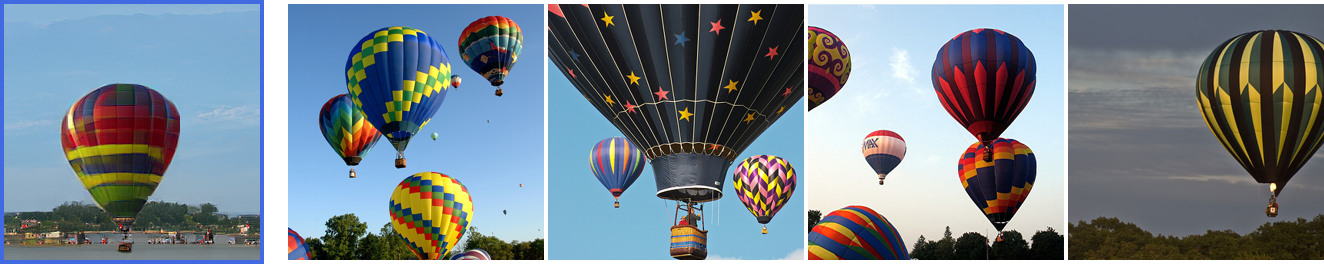}\par\vspace{1pt}
\makebox[\linewidth][l]{\fontsize{5.5}{6}\selectfont \makebox[0.1993958\linewidth][c]{\textcolor{xblue}{\textbf{Generated}}}%
\hspace{0.0181269\linewidth}%
\makebox[0.1933535\linewidth][c]{NN 1}%
\hspace{0.0030211\linewidth}%
\makebox[0.1933535\linewidth][c]{NN 2}%
\hspace{0.0030211\linewidth}%
\makebox[0.1933535\linewidth][c]{NN 3}%
\hspace{0.0030211\linewidth}%
\makebox[0.1933535\linewidth][c]{NN 4}}
\end{minipage}\end{minipage}\par\vspace{4pt}
\noindent\begin{minipage}{\linewidth}
\begin{minipage}[t]{0.485\linewidth}\centering
{\fontsize{7}{8}\selectfont sports car (class 817)\par}\vspace{1pt}
\includegraphics[width=\linewidth]{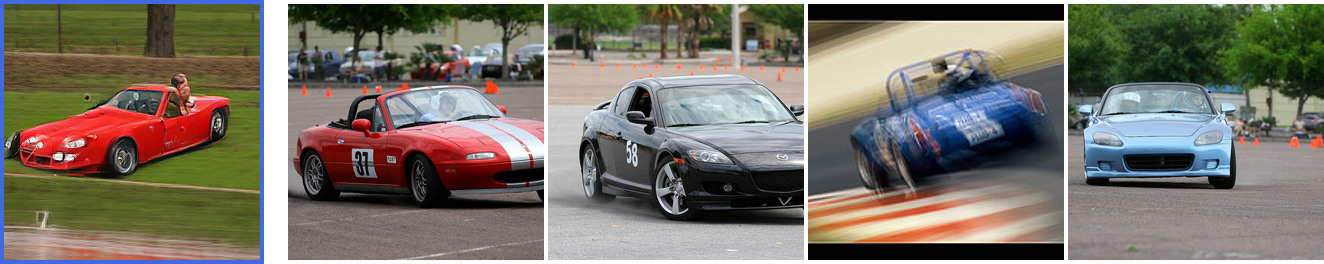}\par\vspace{1pt}
\makebox[\linewidth][l]{\fontsize{5.5}{6}\selectfont \makebox[0.1993958\linewidth][c]{\textcolor{xblue}{\textbf{Generated}}}%
\hspace{0.0181269\linewidth}%
\makebox[0.1933535\linewidth][c]{NN 1}%
\hspace{0.0030211\linewidth}%
\makebox[0.1933535\linewidth][c]{NN 2}%
\hspace{0.0030211\linewidth}%
\makebox[0.1933535\linewidth][c]{NN 3}%
\hspace{0.0030211\linewidth}%
\makebox[0.1933535\linewidth][c]{NN 4}}
\end{minipage}\hfill%
\begin{minipage}[t]{0.485\linewidth}\centering
{\fontsize{7}{8}\selectfont geyser (class 974)\par}\vspace{1pt}
\includegraphics[width=\linewidth]{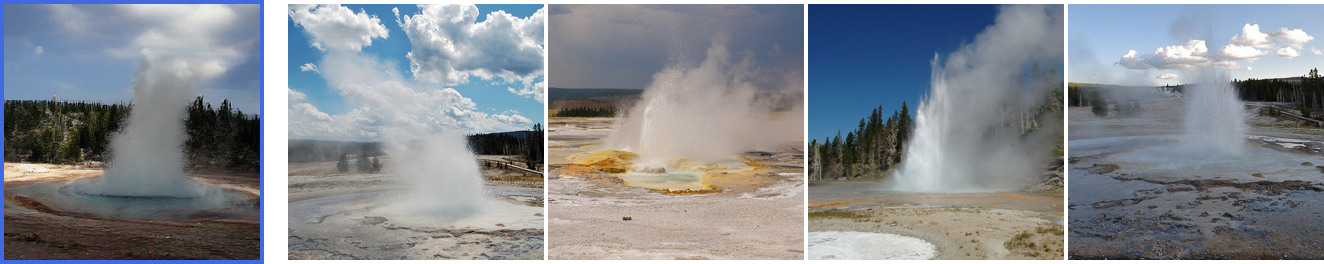}\par\vspace{1pt}
\makebox[\linewidth][l]{\fontsize{5.5}{6}\selectfont \makebox[0.1993958\linewidth][c]{\textcolor{xblue}{\textbf{Generated}}}%
\hspace{0.0181269\linewidth}%
\makebox[0.1933535\linewidth][c]{NN 1}%
\hspace{0.0030211\linewidth}%
\makebox[0.1933535\linewidth][c]{NN 2}%
\hspace{0.0030211\linewidth}%
\makebox[0.1933535\linewidth][c]{NN 3}%
\hspace{0.0030211\linewidth}%
\makebox[0.1933535\linewidth][c]{NN 4}}
\end{minipage}\end{minipage}\par\vspace{4pt}
\noindent\begin{minipage}{\linewidth}
\begin{minipage}[t]{0.485\linewidth}\centering
{\fontsize{7}{8}\selectfont volcano (class 980)\par}\vspace{1pt}
\includegraphics[width=\linewidth]{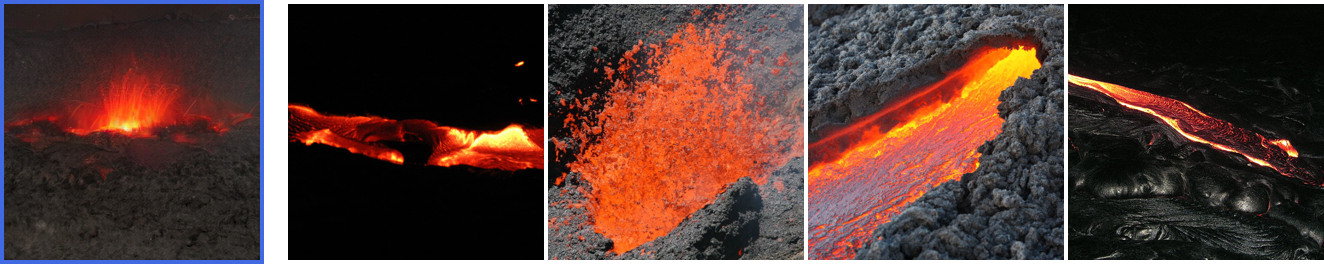}\par\vspace{1pt}
\makebox[\linewidth][l]{\fontsize{5.5}{6}\selectfont \makebox[0.1993958\linewidth][c]{\textcolor{xblue}{\textbf{Generated}}}%
\hspace{0.0181269\linewidth}%
\makebox[0.1933535\linewidth][c]{NN 1}%
\hspace{0.0030211\linewidth}%
\makebox[0.1933535\linewidth][c]{NN 2}%
\hspace{0.0030211\linewidth}%
\makebox[0.1933535\linewidth][c]{NN 3}%
\hspace{0.0030211\linewidth}%
\makebox[0.1933535\linewidth][c]{NN 4}}
\end{minipage}\hfill%
\begin{minipage}[t]{0.485\linewidth}\centering
{\fontsize{7}{8}\selectfont peacock (class 84)\par}\vspace{1pt}
\includegraphics[width=\linewidth]{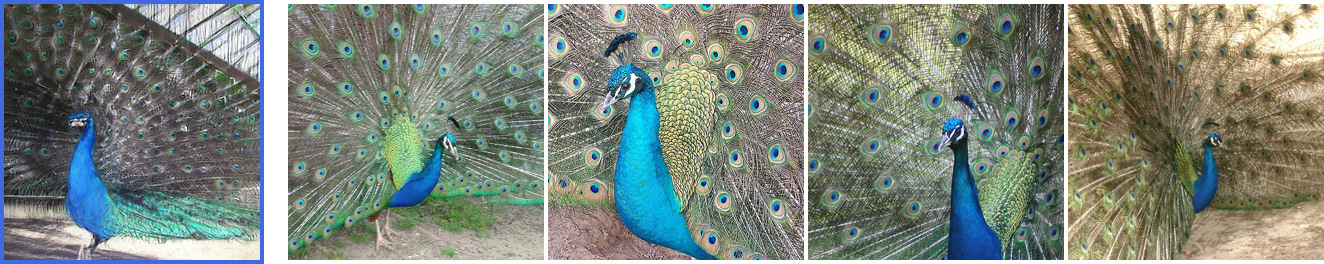}\par\vspace{1pt}
\makebox[\linewidth][l]{\fontsize{5.5}{6}\selectfont \makebox[0.1993958\linewidth][c]{\textcolor{xblue}{\textbf{Generated}}}%
\hspace{0.0181269\linewidth}%
\makebox[0.1933535\linewidth][c]{NN 1}%
\hspace{0.0030211\linewidth}%
\makebox[0.1933535\linewidth][c]{NN 2}%
\hspace{0.0030211\linewidth}%
\makebox[0.1933535\linewidth][c]{NN 3}%
\hspace{0.0030211\linewidth}%
\makebox[0.1933535\linewidth][c]{NN 4}}
\end{minipage}\end{minipage}\par\vspace{4pt}
\endgroup

%% file: Appendix_B_qualitative.tex
\section{qualitative examples}

\subsection{Imagenet}
\label{appendix:examples_imagenet}
In this section, we provide qualitative examples on ImageNet for Priming-B and Priming-L, both pretrained for $160$ epochs. For each ImageNet class, we use the same fixed set of $10$ randomly sampled Gaussian noise vectors.
\input{imagenet_qualitative}
\clearpage
\subsection{Text-to-image generation}
\label{appendix:examples_t2i}

We compare one-step samples from Priming with logit-normal pretraining against
the released MiniT2I MeanFlow baseline sampled in one step with classifier-free
guidance scale $12.0$ and guidance interval $[0.2,0.8]$. For each selected
caption, the two models are shown at the corresponding sample index within their
four-sample groups.

\newcommand{\tTwoOursRoot}{figures/t2i_qualitative/priming_b}
\newcommand{\tTwoMeanFlowRoot}{figures/t2i_qualitative/meanflow}
\newcommand{\tTwoQualExample}[3]{%
    \begin{minipage}[t]{0.49\textwidth}
        \centering
        \setlength{\tabcolsep}{1pt}
        \begin{tabular}{@{}cc@{}}
            \scriptsize Priming (ours) & \scriptsize MeanFlow \\
            \includegraphics[width=0.48\linewidth]{\tTwoOursRoot/#2_#3.jpg} &
            \includegraphics[width=0.48\linewidth]{\tTwoMeanFlowRoot/#2_#3.jpg}
        \end{tabular}\par
        \vspace{2pt}
        \parbox[t][2em][t]{0.97\linewidth}{\raggedright\scriptsize\itshape #1}
    \end{minipage}%
}

\begin{figure*}[!h]
    \centering
    \tTwoQualExample{A young botanist in a sunlit greenhouse, holding a tray of tiny seedlings, natural editorial photograph.}{00001}{00002}\hfill
    \tTwoQualExample{A street musician playing a golden saxophone beneath a red awning on a rainy evening.}{00005}{00000}

    \vspace{16pt}
    \tTwoQualExample{A pastry chef dusted with flour presenting a perfect raspberry tart in a quiet bakery kitchen.}{00003}{00003}\hfill
    \tTwoQualExample{A joyful ceramic artist holding a freshly glazed turquoise vase in a bright workshop.}{00006}{00002}

    \vspace{16pt}
    \tTwoQualExample{A mountain guide wearing an orange jacket and round snow goggles, dramatic snowy peaks in the background.}{00007}{00000}\hfill
    \tTwoQualExample{A fashionable woman in a white suit and wide-brimmed hat standing against a cobalt blue wall, magazine portrait.}{00009}{00000}
    \caption{\textbf{Text-to-image qualitative comparison (1/6).} One-step samples from Priming with logit-normal pretraining and one-step MeanFlow with CFG scale $12.0$.}
    \label{fig:qualitative_t2i_1}
\end{figure*}

\clearpage
\begin{figure*}[!h]
    \centering
    \tTwoQualExample{A red panda wearing a bright blue baseball cap, sitting on a moss-covered log.}{00010}{00001}\hfill
    \tTwoQualExample{A curious baby elephant spraying water beside its mother at a sunlit watering hole.}{00012}{00002}

    \vspace{16pt}
    \tTwoQualExample{Two emperor penguins standing together beneath a green aurora on an Antarctic ice field.}{00013}{00000}\hfill
    \tTwoQualExample{A tiny tree frog resting on a glossy leaf covered with morning raindrops, macro photograph.}{00014}{00000}

    \vspace{16pt}
    \tTwoQualExample{A barn owl gliding silently through an old library filled with floating dust motes.}{00015}{00000}\hfill
    \tTwoQualExample{A fluffy white alpaca peeking over a wooden fence in a field of purple lupines.}{00016}{00001}

    \vspace{16pt}
    \tTwoQualExample{A sea turtle swimming above a colorful coral reef with shafts of sunlight in clear water.}{00017}{00002}\hfill
    \tTwoQualExample{A fox curled asleep inside a hollow tree while gentle snow falls outside.}{00018}{00003}
    \caption{\textbf{Text-to-image qualitative comparison (2/6).} One-step samples from Priming with logit-normal pretraining and one-step MeanFlow with CFG scale $12.0$.}
    \label{fig:qualitative_t2i_2}
\end{figure*}

\clearpage
\begin{figure*}[!h]
    \centering
    \tTwoQualExample{Three playful otters floating on their backs in calm blue water, each holding a small seashell.}{00019}{00001}\hfill
    \tTwoQualExample{A polished glass teapot filled with yellow wildflowers on a rustic wooden table, soft morning light.}{00020}{00003}

    \vspace{16pt}
    \tTwoQualExample{A plate of blueberry pancakes topped with strawberries and maple syrup, inviting food photography.}{00021}{00000}\hfill
    \tTwoQualExample{A transparent perfume bottle on black stone surrounded by jasmine petals and silver reflections, luxury product photo.}{00022}{00001}

    \vspace{16pt}
    \tTwoQualExample{A bowl of ramen with a soft-boiled egg, mushrooms, scallions, and rising steam on a dark wooden counter.}{00023}{00002}\hfill
    \tTwoQualExample{An antique silver pocket watch resting on an open handwritten letter with a sprig of lavender.}{00028}{00002}

    \vspace{16pt}
    \tTwoQualExample{A delicate matcha parfait layered with cream and red beans in a tall glass, Japanese cafe photography.}{00027}{00000}\hfill
    \tTwoQualExample{A pair of red running shoes splashing through a shallow puddle, crisp commercial product photograph.}{00026}{00003}
    \caption{\textbf{Text-to-image qualitative comparison (3/6).} One-step samples from Priming with logit-normal pretraining and one-step MeanFlow with CFG scale $12.0$.}
    \label{fig:qualitative_t2i_3}
\end{figure*}

\clearpage
\begin{figure*}[!h]
    \centering
    \tTwoQualExample{A misty waterfall descending through a lush Icelandic canyon at sunrise, ultra-wide landscape photograph.}{00031}{00001}\hfill
    \tTwoQualExample{A winding river reflecting golden autumn trees in a quiet mountain valley, aerial view.}{00033}{00001}

    \vspace{16pt}
    \tTwoQualExample{A lone acacia tree silhouetted against a fiery orange sunset on the African savanna.}{00035}{00001}\hfill
    \tTwoQualExample{A crystal-clear alpine lake reflecting snowcapped mountains and a single red canoe.}{00037}{00003}

    \vspace{16pt}
    \tTwoQualExample{A premium toy-like 3D render of a tiny orange camper van beside a teal lake and rounded mountains.}{00099}{00003}\hfill
    \tTwoQualExample{A cozy low-poly campsite with a green tent, pine trees, and a glowing fire beneath a purple night sky.}{00095}{00000}

    \vspace{16pt}
    \tTwoQualExample{An isometric miniature city powered by wind turbines and rooftop gardens, clean 3D render.}{00090}{00002}\hfill
    \tTwoQualExample{A vintage travel-poster illustration of a red cable car climbing through the Swiss Alps.}{00088}{00000}
    \caption{\textbf{Text-to-image qualitative comparison (4/6).} One-step samples from Priming with logit-normal pretraining and one-step MeanFlow with CFG scale $12.0$.}
    \label{fig:qualitative_t2i_4}
\end{figure*}

\clearpage
\begin{figure*}[!h]
    \centering
    \tTwoQualExample{A charcoal drawing of two jazz musicians performing in a smoky basement club.}{00089}{00003}\hfill
    \tTwoQualExample{A stained-glass window depicting a peacock beneath flowering vines, jewel-toned light.}{00086}{00002}

    \vspace{16pt}
    \tTwoQualExample{A richly textured pastel drawing of peaches and wildflowers on a checkered tablecloth.}{00084}{00001}\hfill
    \tTwoQualExample{A watercolor illustration of a lighthouse standing against enormous storm waves.}{00080}{00000}

    \vspace{16pt}
    \tTwoQualExample{A mechanical hummingbird with brass wings hovering beside a luminous orchid.}{00076}{00000}\hfill
    \tTwoQualExample{Three ceramic owls lined up on a bookshelf, each wearing a different colored scarf.}{00061}{00002}

    \vspace{16pt}
    \tTwoQualExample{A small red cube to the left of a large blue sphere on a clean white studio floor.}{00060}{00001}\hfill
    \tTwoQualExample{Two sailboats on a calm lake, the orange boat closer to the viewer and the blue boat near the mountains.}{00065}{00001}
    \caption{\textbf{Text-to-image qualitative comparison (5/6).} One-step samples from Priming with logit-normal pretraining and one-step MeanFlow with CFG scale $12.0$.}
    \label{fig:qualitative_t2i_5}
\end{figure*}

\clearpage
\begin{figure*}[!h]
    \centering
    \vspace{16pt}
    \tTwoQualExample{A firefighter rescuing a small dog from a flooded street during heavy rain, documentary photograph.}{00059}{00002}\hfill
    \tTwoQualExample{A surfer carving through a translucent turquoise wave at golden hour, water droplets frozen in motion.}{00052}{00003}

    \vspace{16pt}
    \tTwoQualExample{A tiny cabin with warm windows nestled among snow-covered pine trees at blue hour.}{00045}{00001}\hfill
    \tTwoQualExample{A narrow street in Kyoto after rain, glowing paper lanterns reflected on wet pavement, cinematic photograph.}{00042}{00001}
    \caption{\textbf{Text-to-image qualitative comparison (6/6).} One-step samples from Priming with logit-normal pretraining and one-step MeanFlow with CFG scale $12.0$.}
    \label{fig:qualitative_t2i_6}
\end{figure*}

\clearpage
\let\tTwoQualExample\relax
\let\tTwoOursRoot\relax
\let\tTwoMeanFlowRoot\relax

\clearpage
\subsection{world models}
\label{appendix:examples_wm}

\begin{figure*}[!h]
    \centering
    \setlength{\tabcolsep}{1.5pt}
    \renewcommand{\arraystretch}{0.98}
    \textbf{RT-1}\par\vspace{1pt}
    \begin{tabular}{@{}>{\bfseries}c*{6}{c}@{}}
        & \multicolumn{6}{c}{\scriptsize Example 1: final context frame $\longrightarrow$ autoregressive rollout} \\
        GT &
        \includegraphics[width=0.122\textwidth]{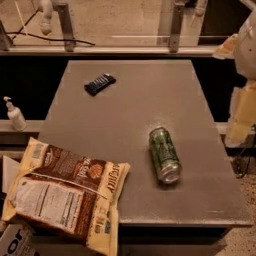} &
        \includegraphics[width=0.122\textwidth]{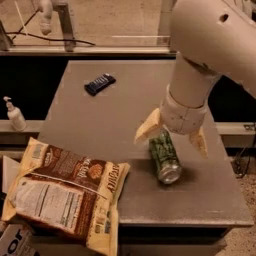} &
        \includegraphics[width=0.122\textwidth]{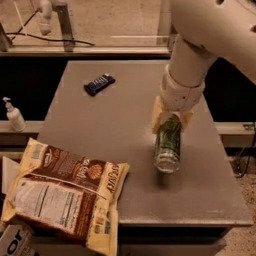} &
        \includegraphics[width=0.122\textwidth]{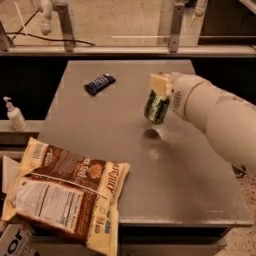} &
        \includegraphics[width=0.122\textwidth]{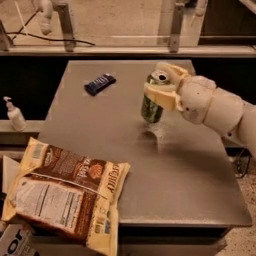} &
        \includegraphics[width=0.122\textwidth]{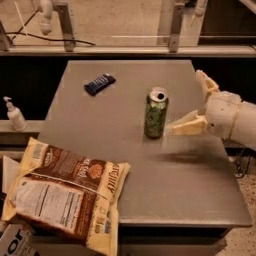} \\
        Ours &
        \includegraphics[width=0.122\textwidth]{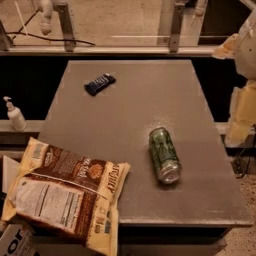} &
        \includegraphics[width=0.122\textwidth]{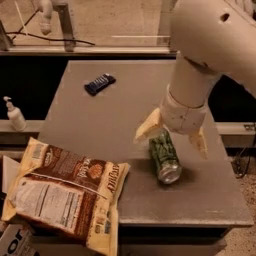} &
        \includegraphics[width=0.122\textwidth]{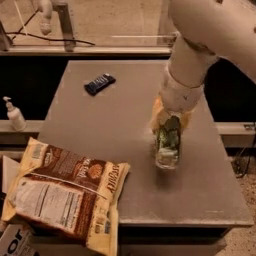} &
        \includegraphics[width=0.122\textwidth]{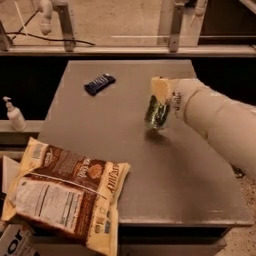} &
        \includegraphics[width=0.122\textwidth]{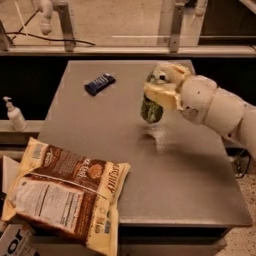} &
        \includegraphics[width=0.122\textwidth]{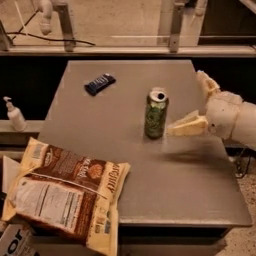} \\[3pt]
        & \multicolumn{6}{c}{\scriptsize Example 2: final context frame $\longrightarrow$ autoregressive rollout} \\
        GT &
        \includegraphics[width=0.122\textwidth]{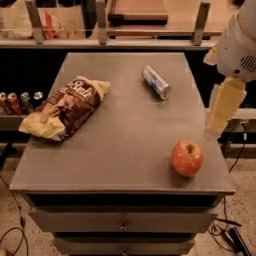} &
        \includegraphics[width=0.122\textwidth]{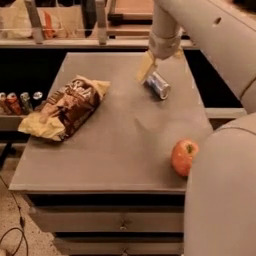} &
        \includegraphics[width=0.122\textwidth]{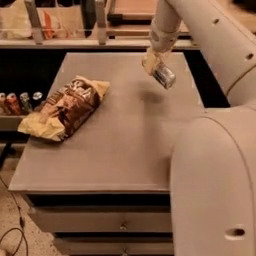} &
        \includegraphics[width=0.122\textwidth]{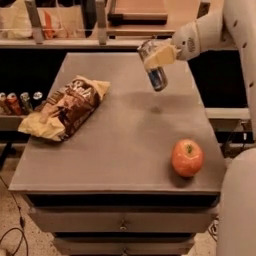} &
        \includegraphics[width=0.122\textwidth]{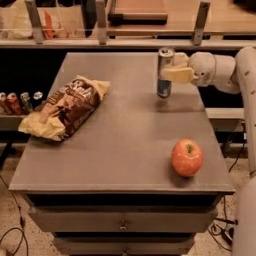} &
        \includegraphics[width=0.122\textwidth]{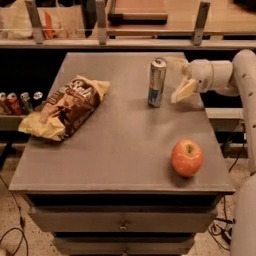} \\
        Ours &
        \includegraphics[width=0.122\textwidth]{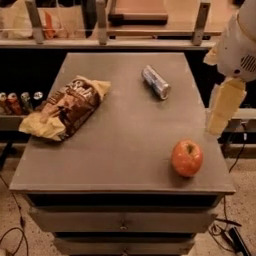} &
        \includegraphics[width=0.122\textwidth]{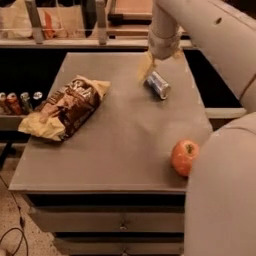} &
        \includegraphics[width=0.122\textwidth]{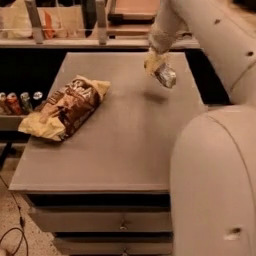} &
        \includegraphics[width=0.122\textwidth]{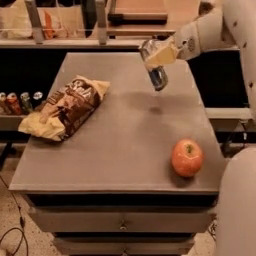} &
        \includegraphics[width=0.122\textwidth]{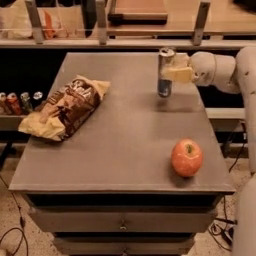} &
        \includegraphics[width=0.122\textwidth]{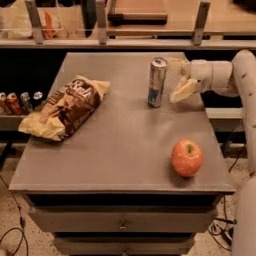}
    \end{tabular}

    \vspace{5pt}
    \textbf{Bridge-V2}\par\vspace{1pt}
    \begin{tabular}{@{}>{\bfseries}c*{6}{c}@{}}
        & \multicolumn{6}{c}{\scriptsize Example 1: final context frame $\longrightarrow$ autoregressive rollout} \\
        GT &
        \includegraphics[width=0.122\textwidth]{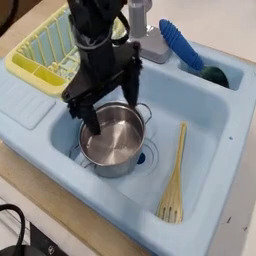} &
        \includegraphics[width=0.122\textwidth]{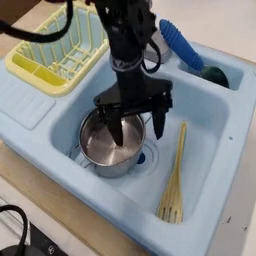} &
        \includegraphics[width=0.122\textwidth]{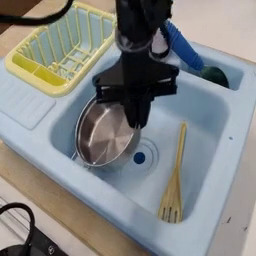} &
        \includegraphics[width=0.122\textwidth]{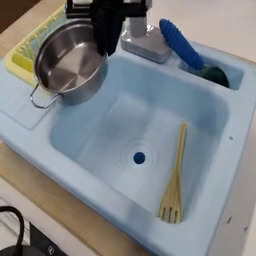} &
        \includegraphics[width=0.122\textwidth]{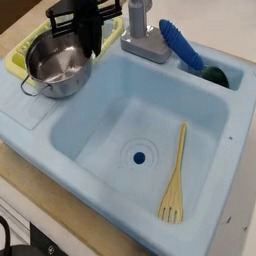} &
        \includegraphics[width=0.122\textwidth]{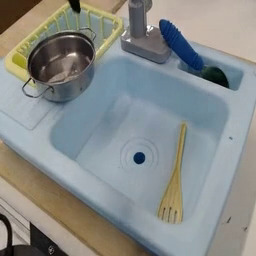} \\
        Ours &
        \includegraphics[width=0.122\textwidth]{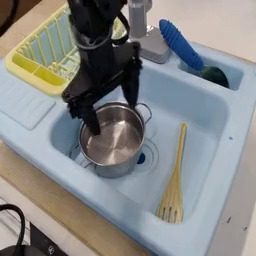} &
        \includegraphics[width=0.122\textwidth]{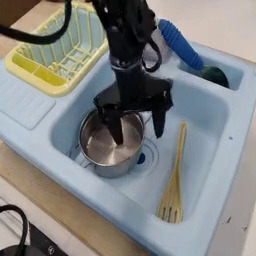} &
        \includegraphics[width=0.122\textwidth]{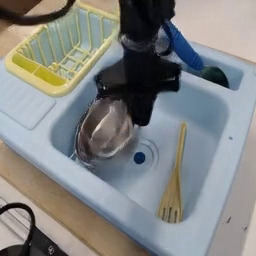} &
        \includegraphics[width=0.122\textwidth]{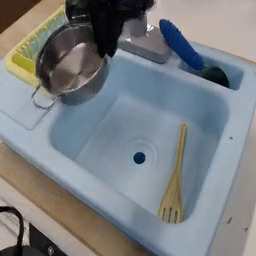} &
        \includegraphics[width=0.122\textwidth]{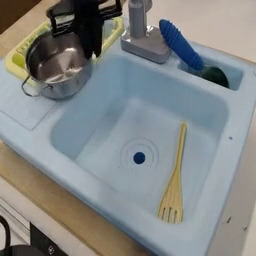} &
        \includegraphics[width=0.122\textwidth]{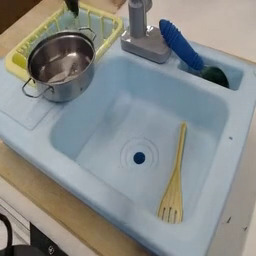} \\[3pt]
        & \multicolumn{6}{c}{\scriptsize Example 2: final context frame $\longrightarrow$ autoregressive rollout} \\
        GT &
        \includegraphics[width=0.122\textwidth]{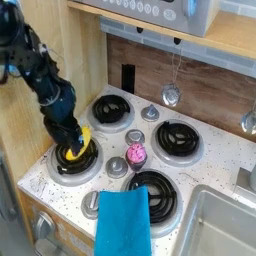} &
        \includegraphics[width=0.122\textwidth]{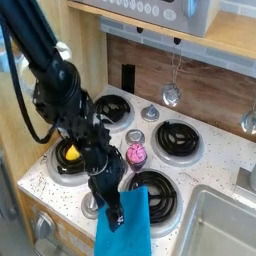} &
        \includegraphics[width=0.122\textwidth]{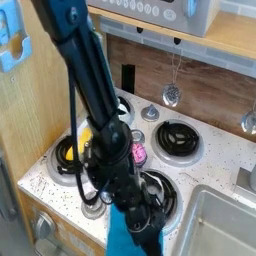} &
        \includegraphics[width=0.122\textwidth]{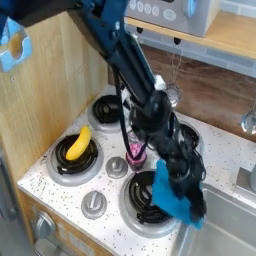} &
        \includegraphics[width=0.122\textwidth]{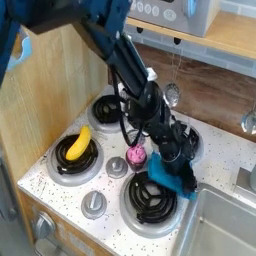} &
        \includegraphics[width=0.122\textwidth]{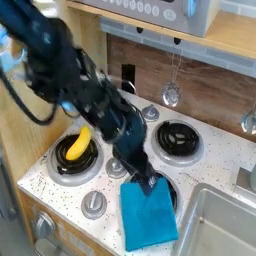} \\
        Ours &
        \includegraphics[width=0.122\textwidth]{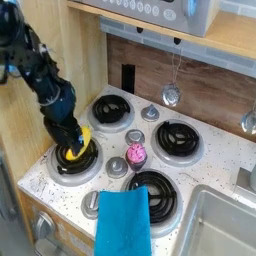} &
        \includegraphics[width=0.122\textwidth]{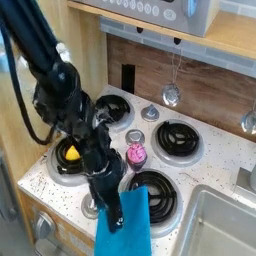} &
        \includegraphics[width=0.122\textwidth]{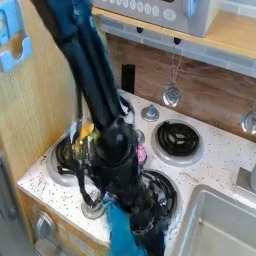} &
        \includegraphics[width=0.122\textwidth]{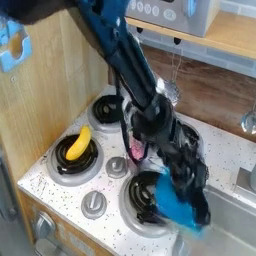} &
        \includegraphics[width=0.122\textwidth]{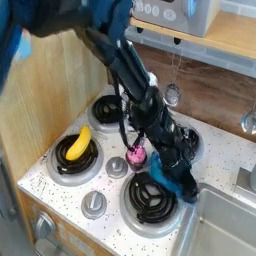} &
        \includegraphics[width=0.122\textwidth]{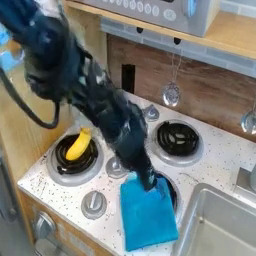}
    \end{tabular}

    \caption{\textbf{Autoregressive world-model rollouts.} Qualitative examples on RT-1 and Bridge-V2. For each rollout, the top row shows ground-truth frames and the bottom row shows predictions from our model at the same time points. The leftmost column is the final observed context frame; the remaining columns are uniformly spaced over the predicted rollout. Predictions are generated autoregressively while conditioning on the provided action sequence.}
    \label{fig:qualitative_world_model}
\end{figure*}
\clearpage

\subsection{style embedding}
\label{appendix:style_embedding}

\begin{figure*}[!h]
    \centering
    \setlength{\tabcolsep}{2pt}
    \renewcommand{\arraystretch}{0.92}
    \textbf{Drifting Pixel-L}\par\vspace{1pt}
    \begin{tabular}{@{}r@{\hspace{3pt}}c@{}}
        & \scriptsize Independent Gaussian inputs $\epsilon_0,\ldots,\epsilon_9$ \\
        \scriptsize macaw & \includegraphics[width=0.86\textwidth]{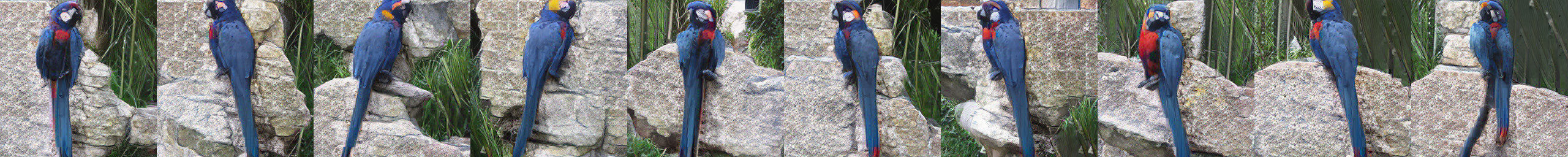} \\
        \scriptsize golden retriever & \includegraphics[width=0.86\textwidth]{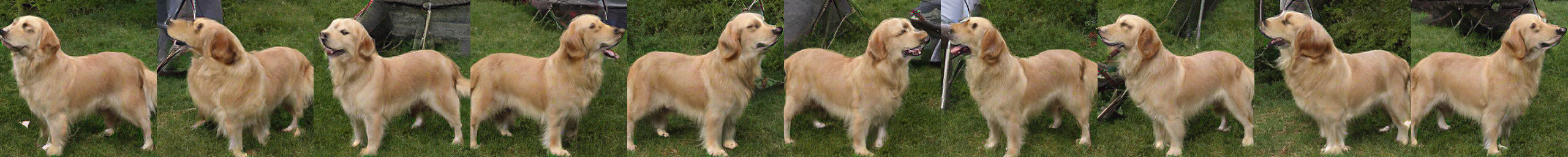} \\
        \scriptsize tiger & \includegraphics[width=0.86\textwidth]{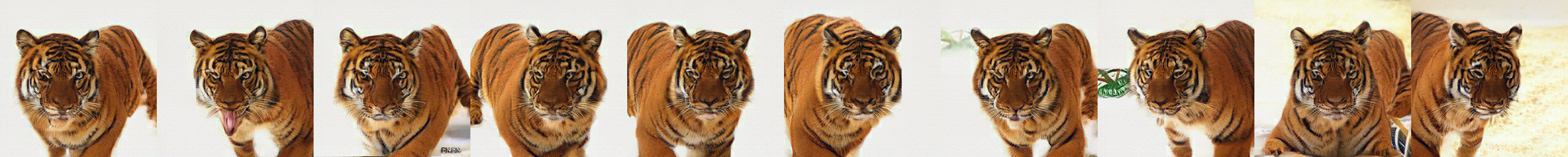} \\
        \scriptsize balloon & \includegraphics[width=0.86\textwidth]{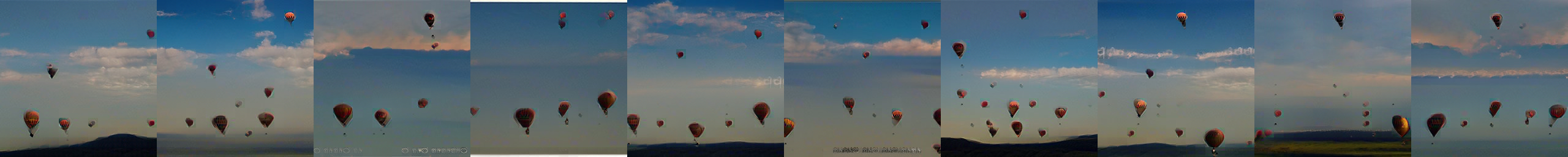} \\
        \scriptsize volcano & \includegraphics[width=0.86\textwidth]{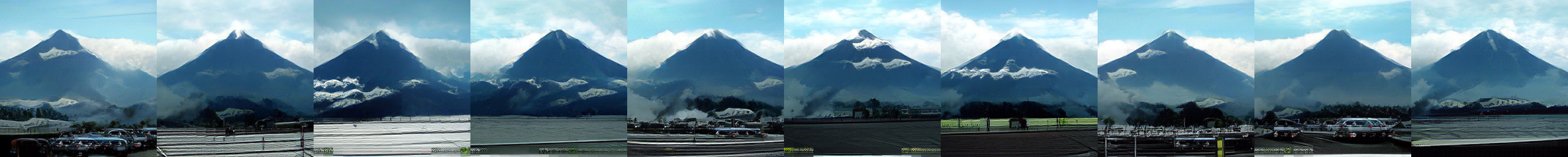}
    \end{tabular}

    \vspace{4pt}
    \textbf{W-Flow latent XL/2}\par\vspace{1pt}
    \begin{tabular}{@{}r@{\hspace{3pt}}c@{}}
        & \scriptsize Independent Gaussian inputs $\epsilon_0,\ldots,\epsilon_9$ \\
        \scriptsize macaw & \includegraphics[width=0.86\textwidth]{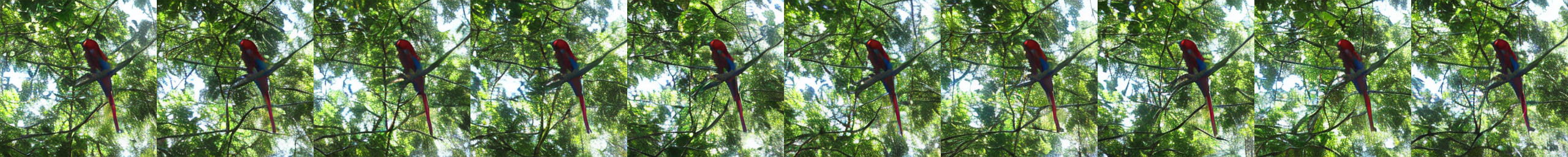} \\
        \scriptsize golden retriever & \includegraphics[width=0.86\textwidth]{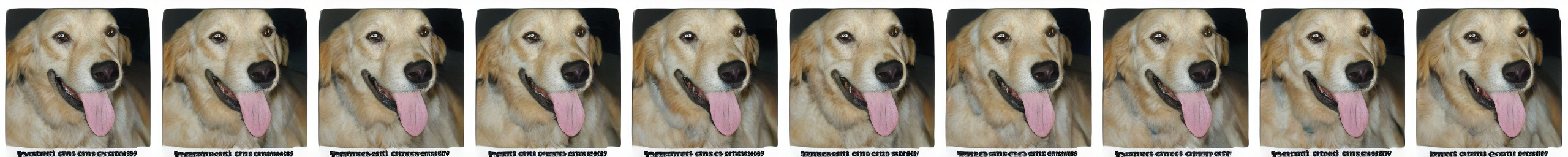} \\
        \scriptsize tiger & \includegraphics[width=0.86\textwidth]{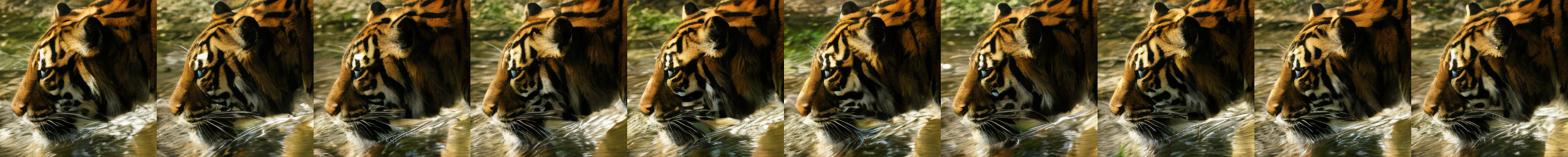} \\
        \scriptsize balloon & \includegraphics[width=0.86\textwidth]{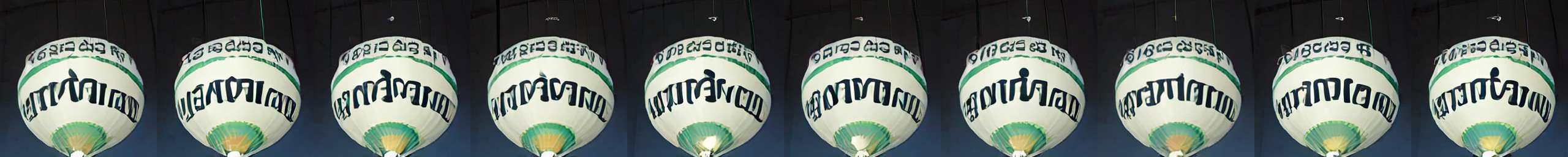} \\
        \scriptsize volcano & \includegraphics[width=0.86\textwidth]{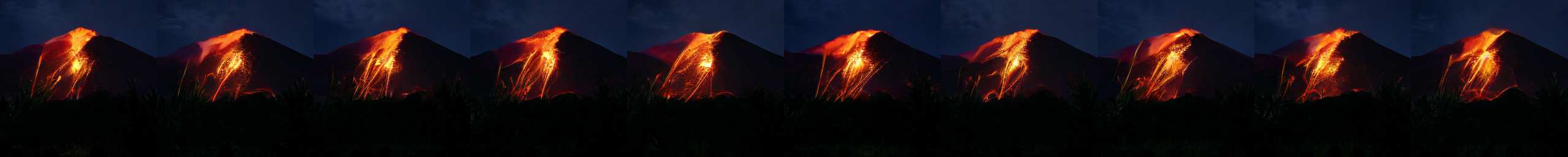}
    \end{tabular}

    \caption{\textbf{Fixing the style embedding limits sample diversity.} For each row, we fix the ImageNet class and learned style embedding while varying only the Gaussian input across ten independent samples. For both Drifting Pixel-L and W-Flow latent XL/2, the samples retain highly similar appearance and composition across columns, indicating that the learned style embedding accounts for much of the variation that would otherwise be attributed to the Gaussian input.}
    \label{fig:style_embedding_fixed}
\end{figure*}

%% file: imagenet_qualitative.tex
% Generated by final_paper/build_imagenet_appendix.py
\par\begingroup\centering\fontsize{7}{8}\selectfont
\setlength{\parskip}{0pt}
\par\medskip\textbf{\normalsize Priming-B (160 ep.)}\par\nobreak\vspace{2pt}
\noindent\makebox[\textwidth][c]{%
\begin{minipage}[t]{0.41\textwidth}\centering
{\scriptsize class 12: house finch\par}\vspace{1pt}
\includegraphics[width=\linewidth]{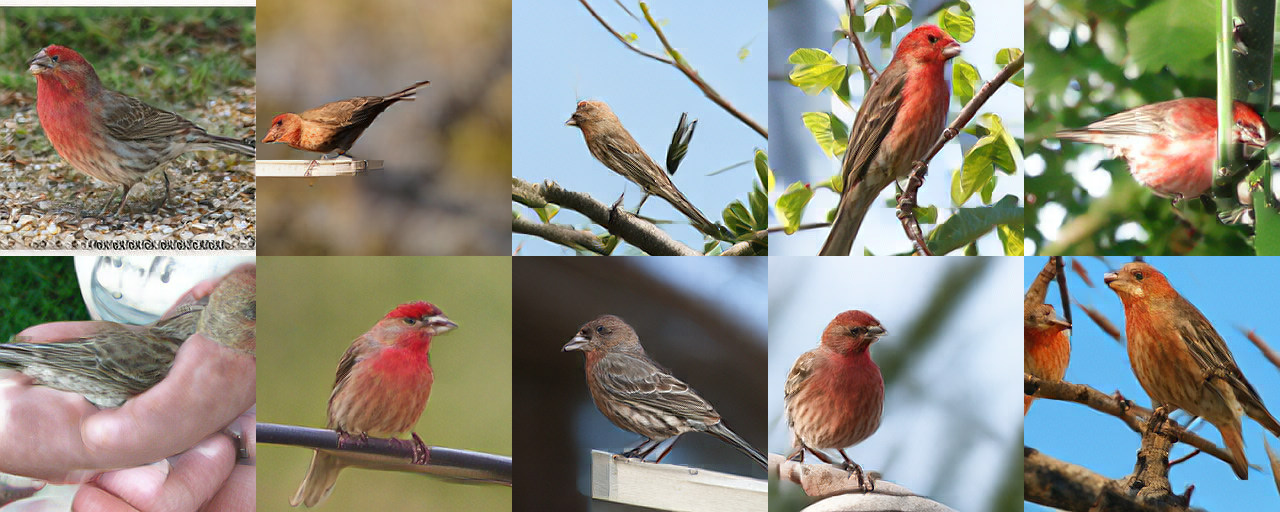}
\end{minipage}\hspace{0.025\textwidth}%
\begin{minipage}[t]{0.41\textwidth}\centering
{\scriptsize class 14: indigo bunting\par}\vspace{1pt}
\includegraphics[width=\linewidth]{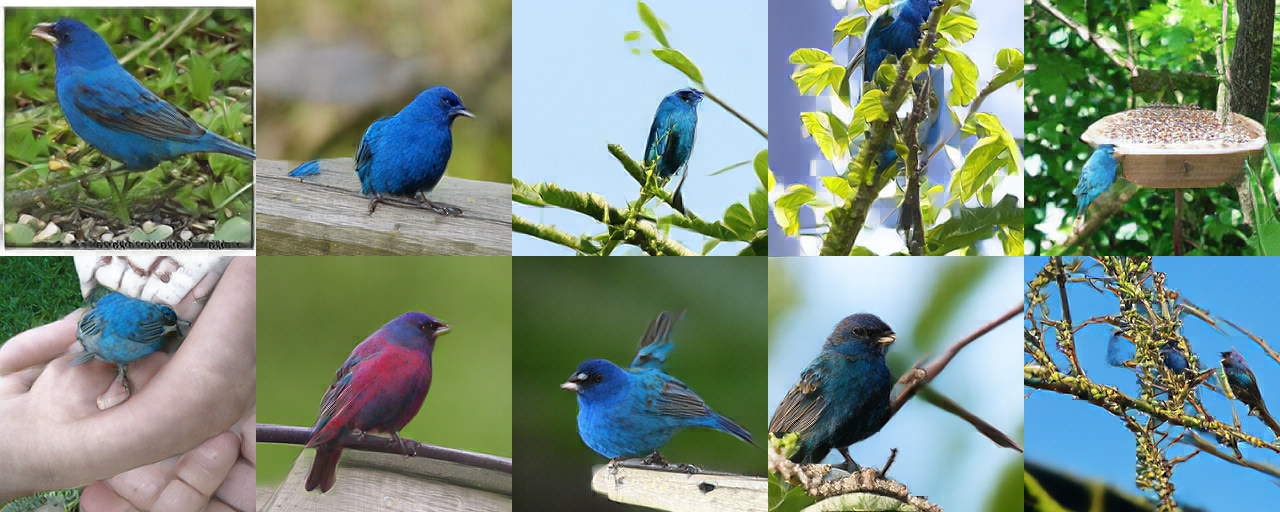}
\end{minipage}%
}\par\vspace{2pt}
\noindent\makebox[\textwidth][c]{%
\begin{minipage}[t]{0.41\textwidth}\centering
{\scriptsize class 17: jay\par}\vspace{1pt}
\includegraphics[width=\linewidth]{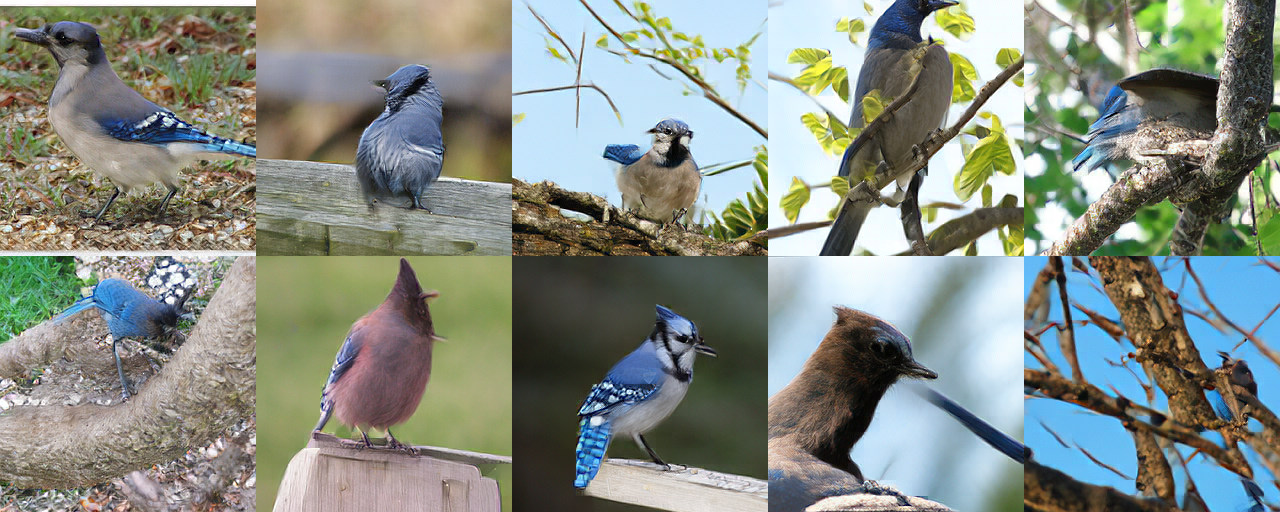}
\end{minipage}\hspace{0.025\textwidth}%
\begin{minipage}[t]{0.41\textwidth}\centering
{\scriptsize class 21: kite\par}\vspace{1pt}
\includegraphics[width=\linewidth]{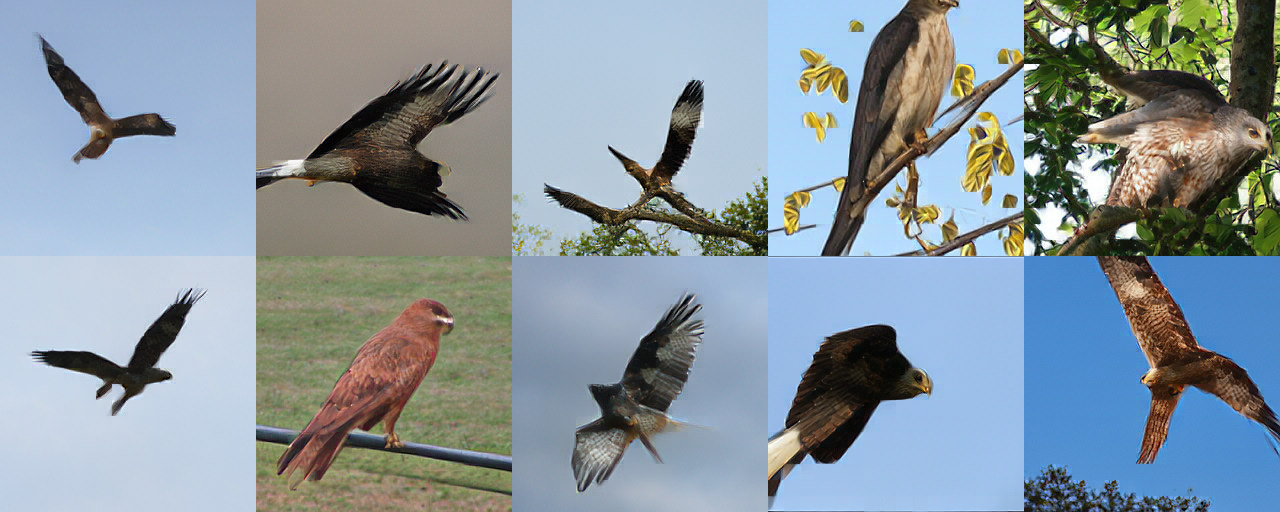}
\end{minipage}%
}\par\vspace{2pt}
\noindent\makebox[\textwidth][c]{%
\begin{minipage}[t]{0.41\textwidth}\centering
{\scriptsize class 22: bald eagle\par}\vspace{1pt}
\includegraphics[width=\linewidth]{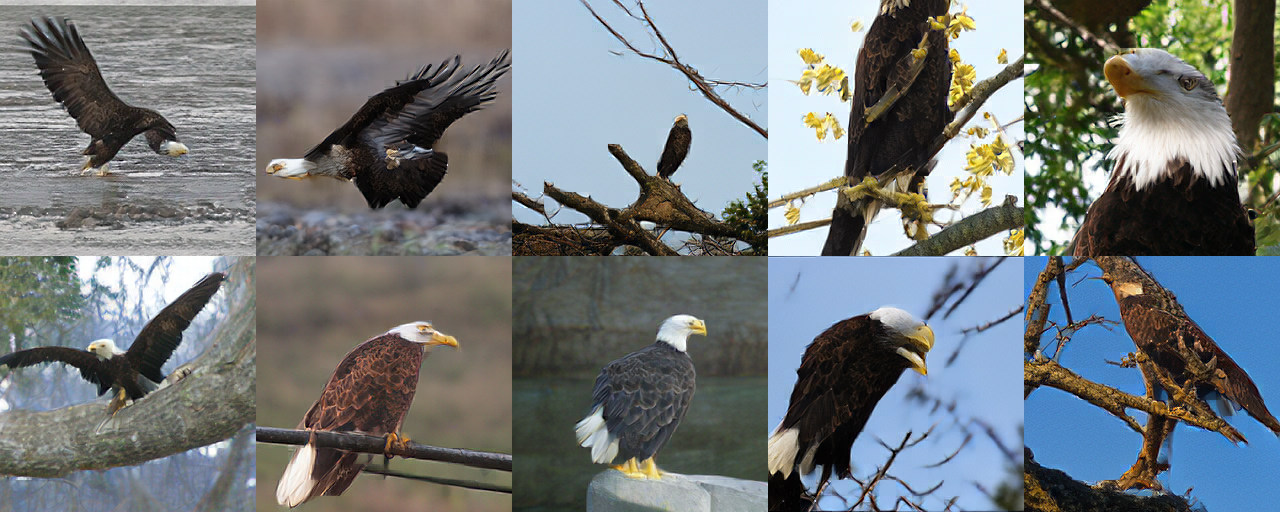}
\end{minipage}\hspace{0.025\textwidth}%
\begin{minipage}[t]{0.41\textwidth}\centering
{\scriptsize class 24: great grey owl\par}\vspace{1pt}
\includegraphics[width=\linewidth]{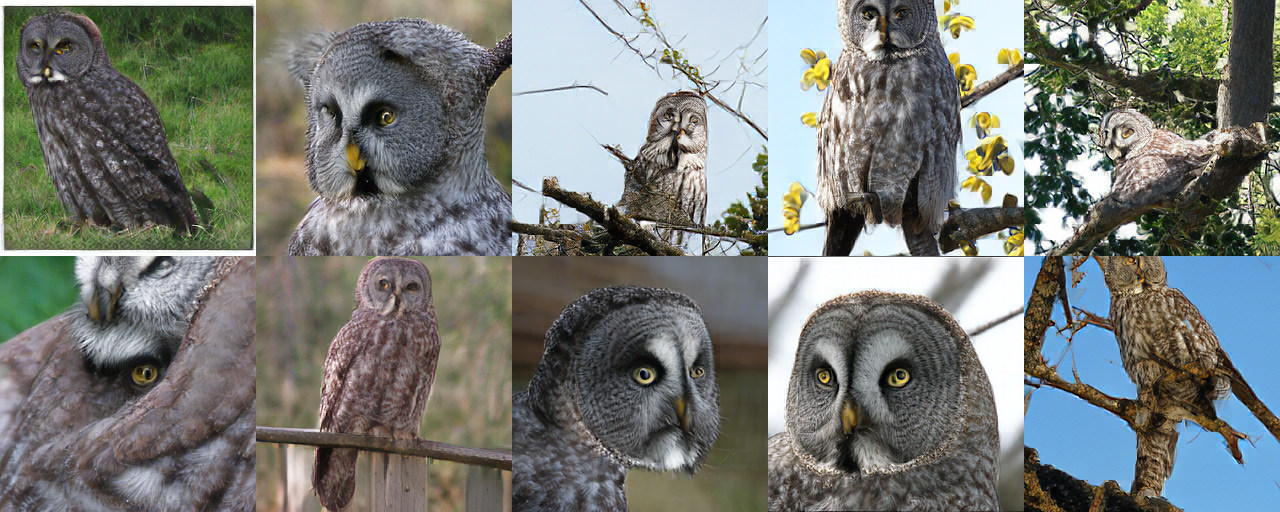}
\end{minipage}%
}\par\vspace{2pt}
\noindent\makebox[\textwidth][c]{%
\begin{minipage}[t]{0.41\textwidth}\centering
{\scriptsize class 84: peacock\par}\vspace{1pt}
\includegraphics[width=\linewidth]{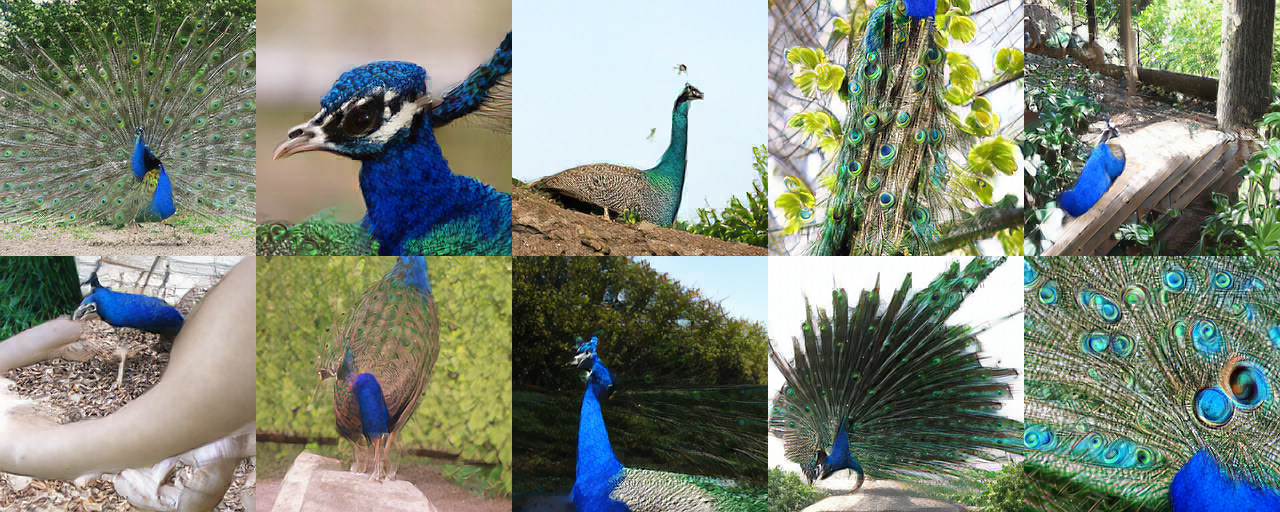}
\end{minipage}\hspace{0.025\textwidth}%
\begin{minipage}[t]{0.41\textwidth}\centering
{\scriptsize class 88: macaw\par}\vspace{1pt}
\includegraphics[width=\linewidth]{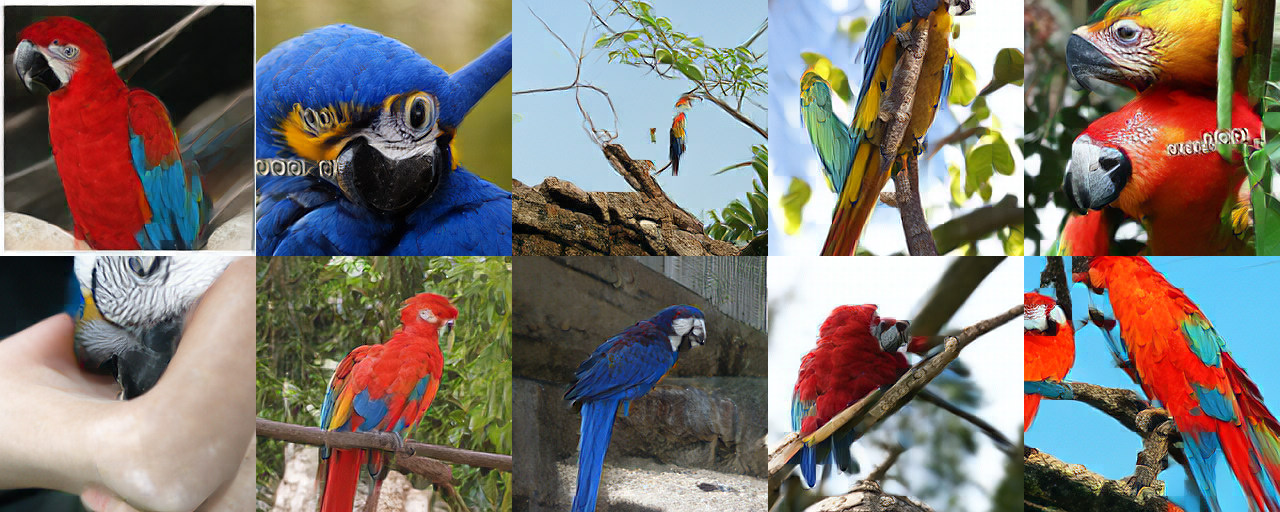}
\end{minipage}%
}\par\vspace{2pt}
\noindent\makebox[\textwidth][c]{%
\begin{minipage}[t]{0.41\textwidth}\centering
{\scriptsize class 90: lorikeet\par}\vspace{1pt}
\includegraphics[width=\linewidth]{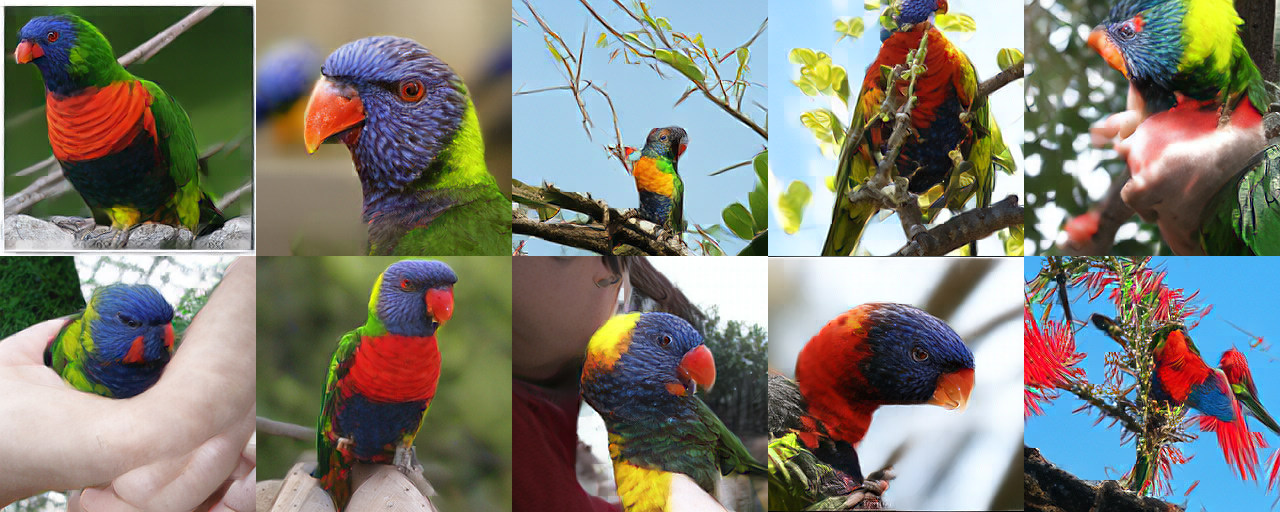}
\end{minipage}\hspace{0.025\textwidth}%
\begin{minipage}[t]{0.41\textwidth}\centering
{\scriptsize class 92: bee eater\par}\vspace{1pt}
\includegraphics[width=\linewidth]{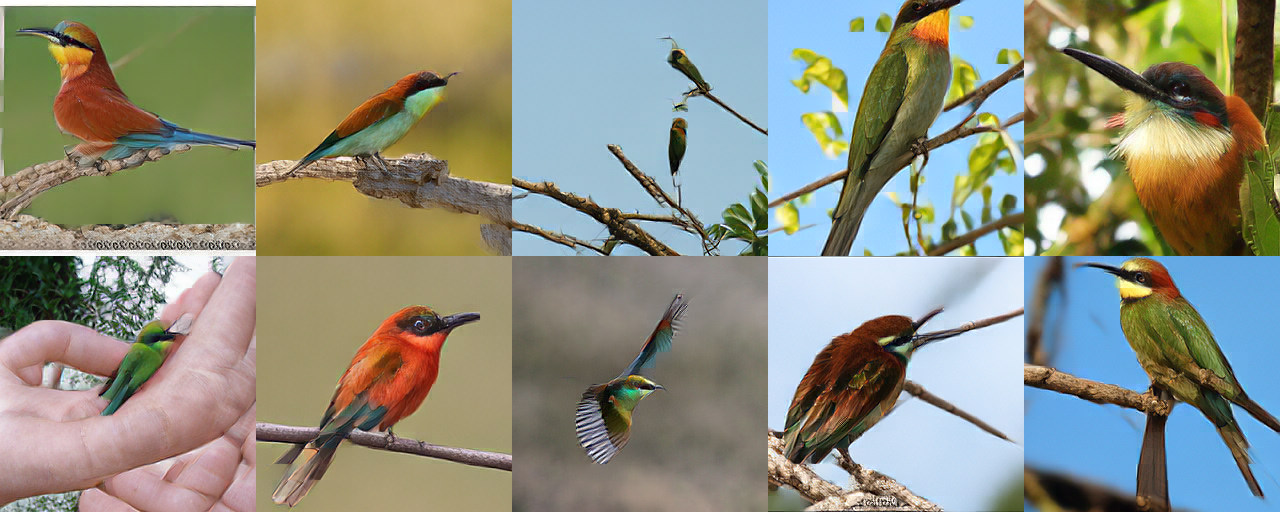}
\end{minipage}%
}\par\vspace{2pt}
\noindent\makebox[\textwidth][c]{%
\begin{minipage}[t]{0.41\textwidth}\centering
{\scriptsize class 94: hummingbird\par}\vspace{1pt}
\includegraphics[width=\linewidth]{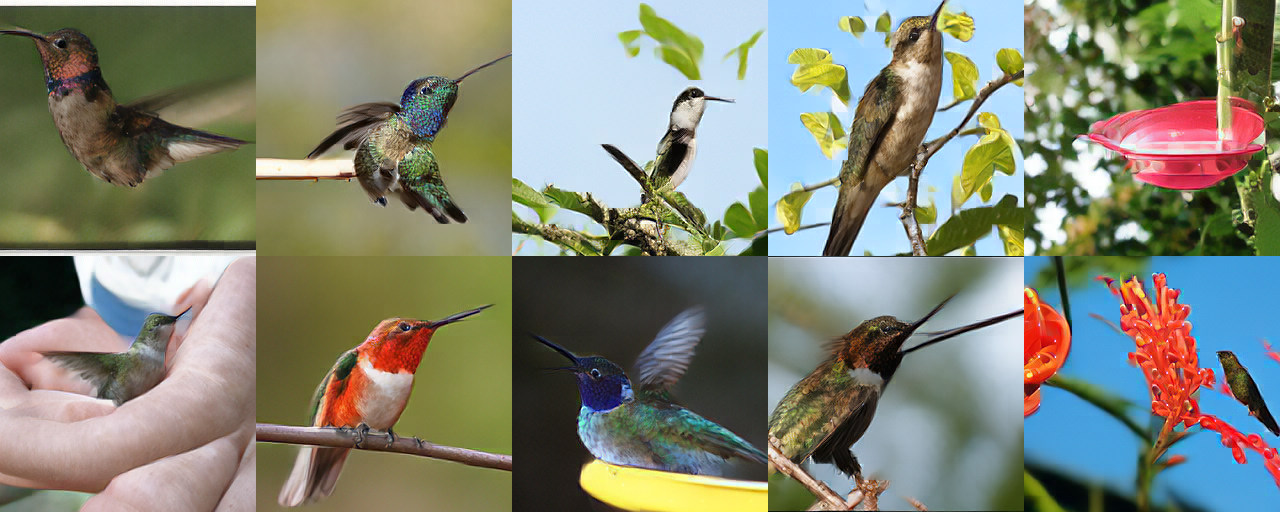}
\end{minipage}\hspace{0.025\textwidth}%
\begin{minipage}[t]{0.41\textwidth}\centering
{\scriptsize class 95: jacamar\par}\vspace{1pt}
\includegraphics[width=\linewidth]{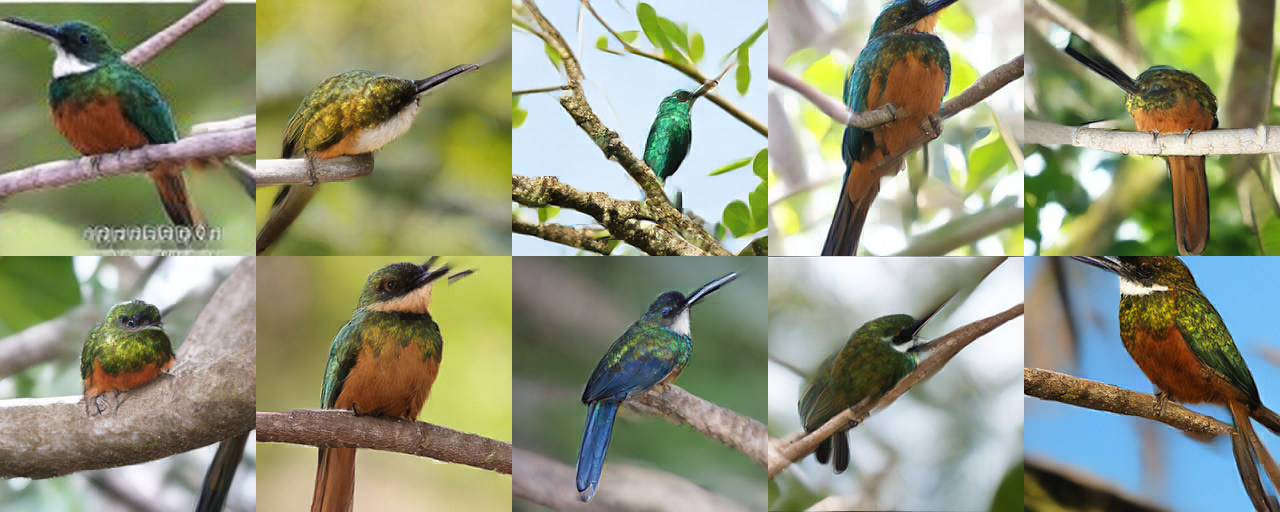}
\end{minipage}%
}\par\vspace{2pt}
\noindent\makebox[\textwidth][c]{%
\begin{minipage}[t]{0.41\textwidth}\centering
{\scriptsize class 96: toucan\par}\vspace{1pt}
\includegraphics[width=\linewidth]{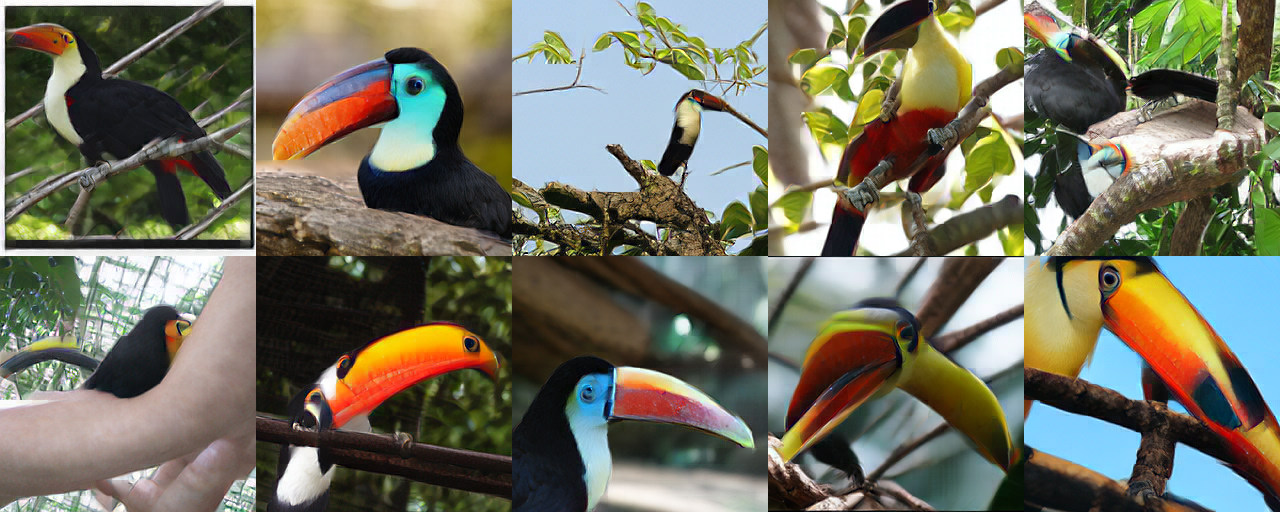}
\end{minipage}\hspace{0.025\textwidth}%
\begin{minipage}[t]{0.41\textwidth}\centering
{\scriptsize class 130: flamingo\par}\vspace{1pt}
\includegraphics[width=\linewidth]{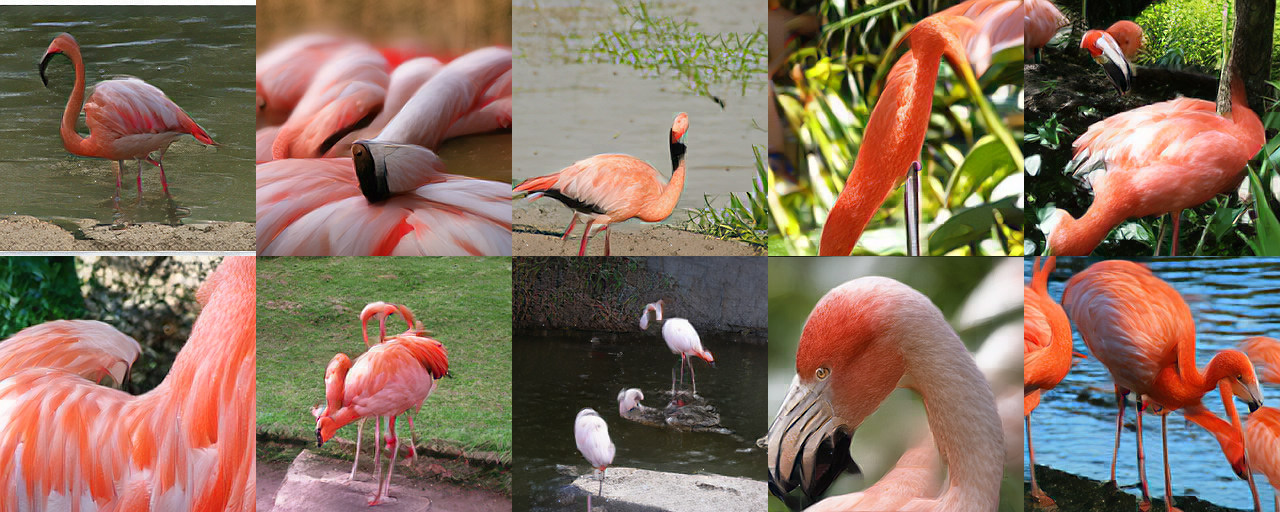}
\end{minipage}%
}\par\vspace{2pt}
\noindent\makebox[\textwidth][c]{%
\begin{minipage}[t]{0.41\textwidth}\centering
{\scriptsize class 145: king penguin\par}\vspace{1pt}
\includegraphics[width=\linewidth]{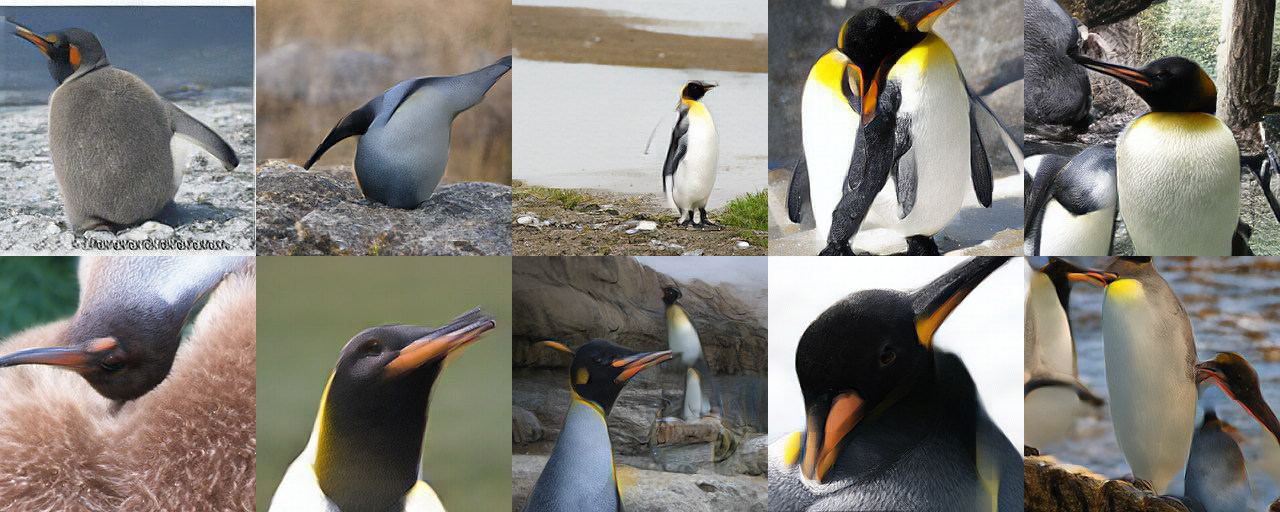}
\end{minipage}\hspace{0.025\textwidth}%
\begin{minipage}[t]{0.41\textwidth}\centering
{\scriptsize class 207: golden retriever\par}\vspace{1pt}
\includegraphics[width=\linewidth]{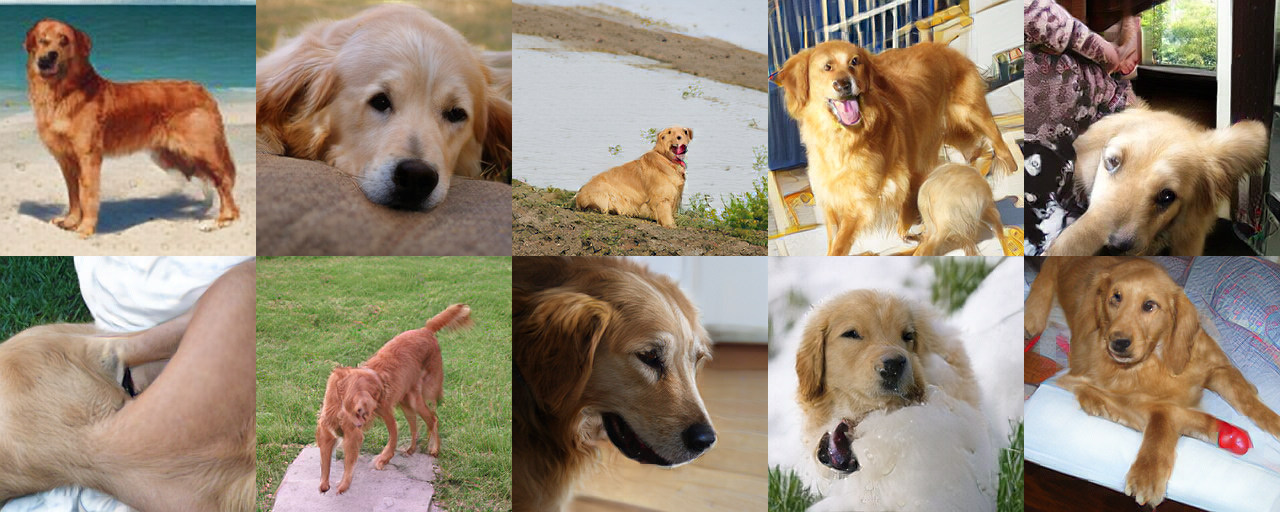}
\end{minipage}%
}\par\vspace{2pt}
\noindent\makebox[\textwidth][c]{%
\begin{minipage}[t]{0.41\textwidth}\centering
{\scriptsize class 208: Labrador retriever\par}\vspace{1pt}
\includegraphics[width=\linewidth]{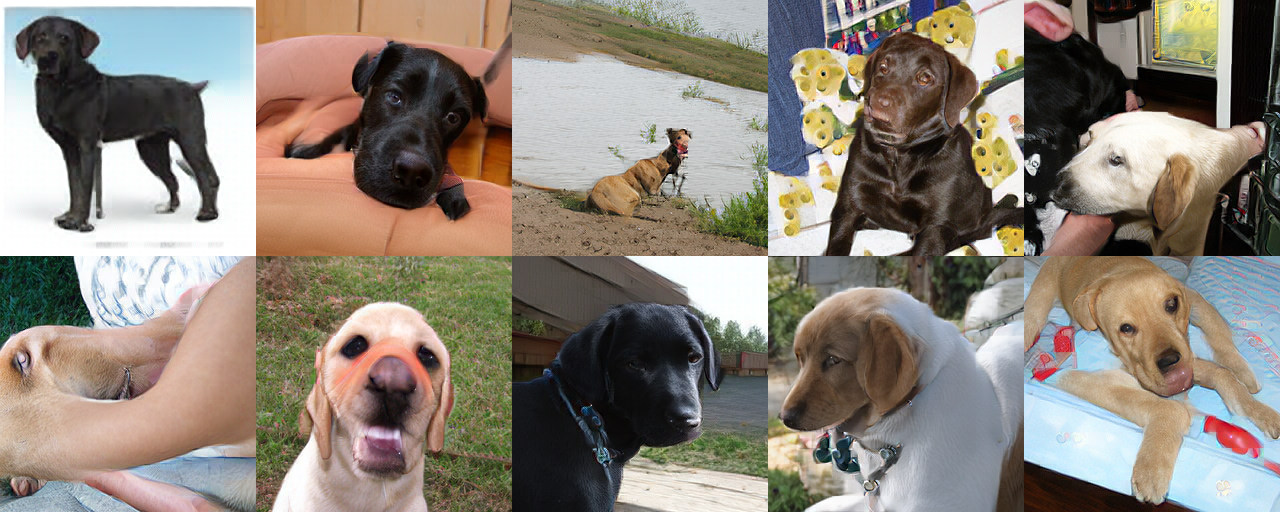}
\end{minipage}\hspace{0.025\textwidth}%
\begin{minipage}[t]{0.41\textwidth}\centering
{\scriptsize class 250: Siberian husky\par}\vspace{1pt}
\includegraphics[width=\linewidth]{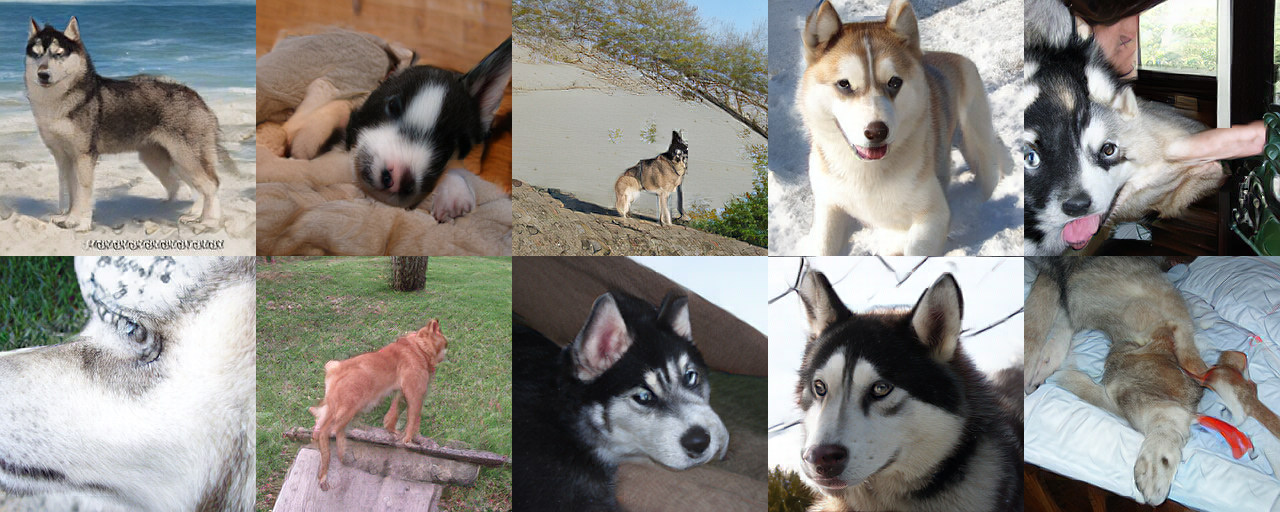}
\end{minipage}%
}\par\vspace{2pt}
\noindent\makebox[\textwidth][c]{%
\begin{minipage}[t]{0.41\textwidth}\centering
{\scriptsize class 270: white wolf\par}\vspace{1pt}
\includegraphics[width=\linewidth]{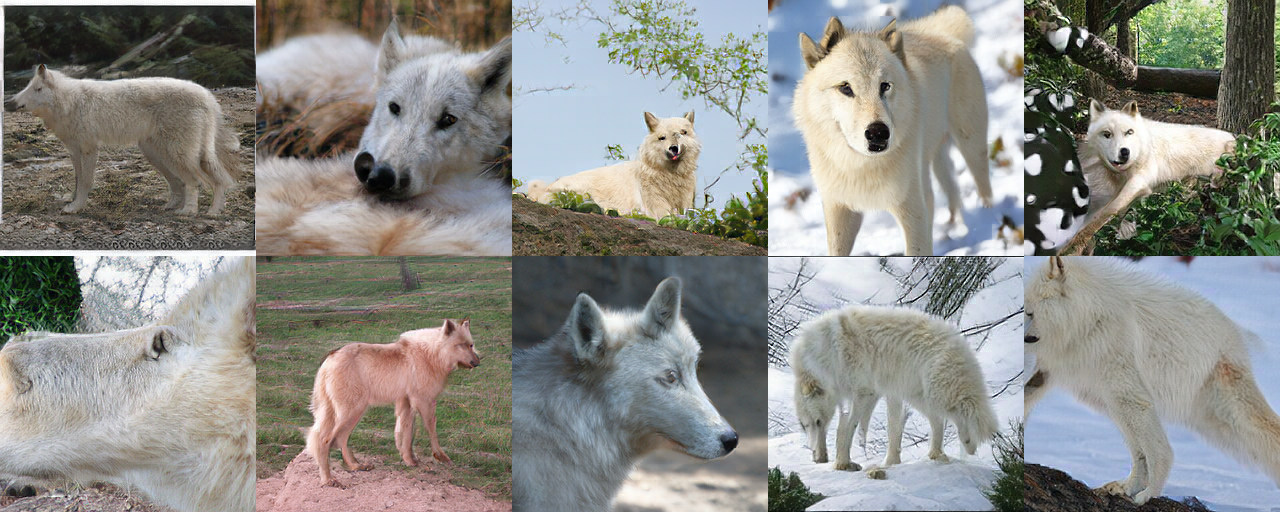}
\end{minipage}\hspace{0.025\textwidth}%
\begin{minipage}[t]{0.41\textwidth}\centering
{\scriptsize class 279: Arctic fox\par}\vspace{1pt}
\includegraphics[width=\linewidth]{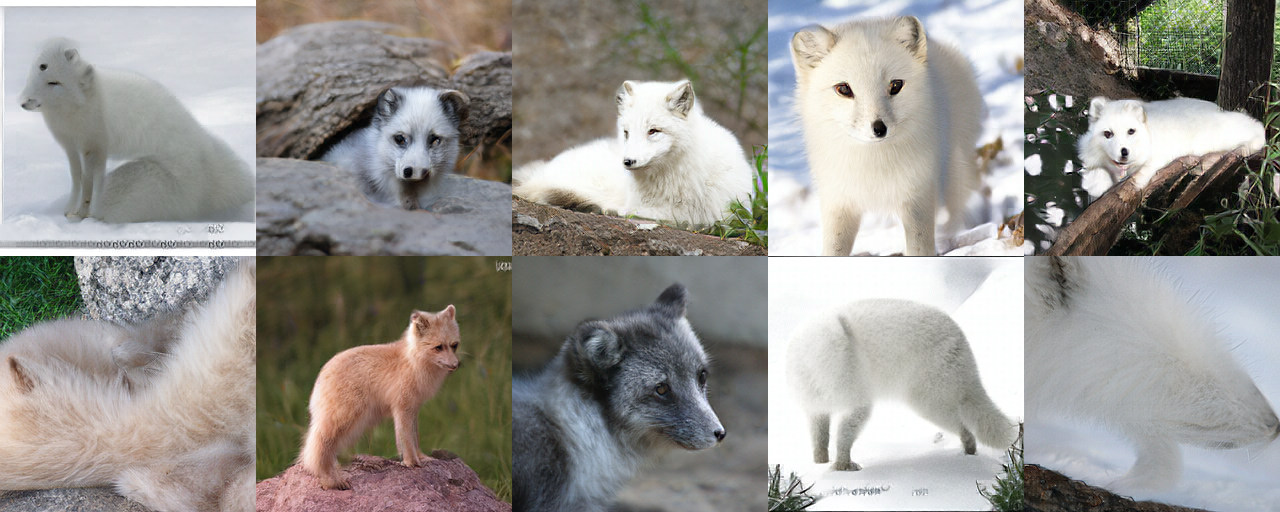}
\end{minipage}%
}\par\vspace{2pt}
\noindent\makebox[\textwidth][c]{%
\begin{minipage}[t]{0.41\textwidth}\centering
{\scriptsize class 281: tabby\par}\vspace{1pt}
\includegraphics[width=\linewidth]{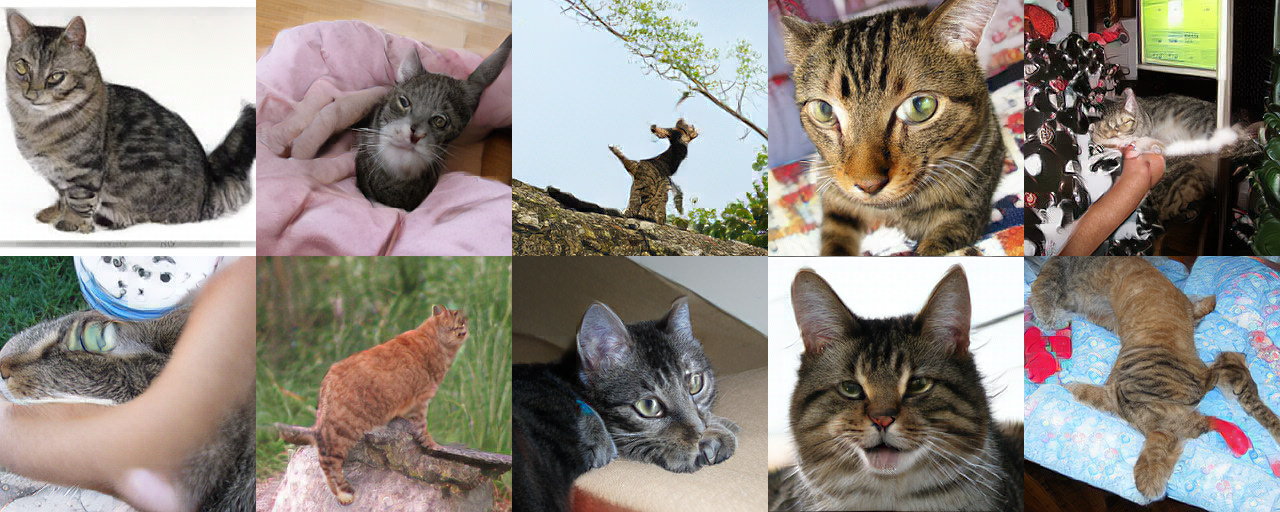}
\end{minipage}\hspace{0.025\textwidth}%
\begin{minipage}[t]{0.41\textwidth}\centering
{\scriptsize class 283: Persian cat\par}\vspace{1pt}
\includegraphics[width=\linewidth]{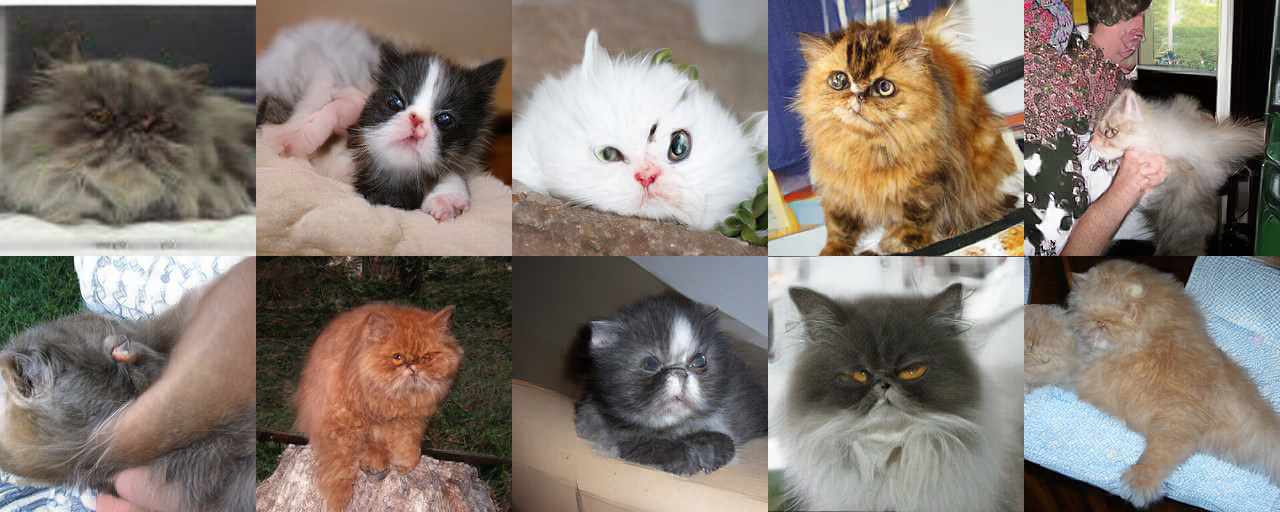}
\end{minipage}%
}\par\vspace{2pt}
\noindent\makebox[\textwidth][c]{%
\begin{minipage}[t]{0.41\textwidth}\centering
{\scriptsize class 285: Egyptian cat\par}\vspace{1pt}
\includegraphics[width=\linewidth]{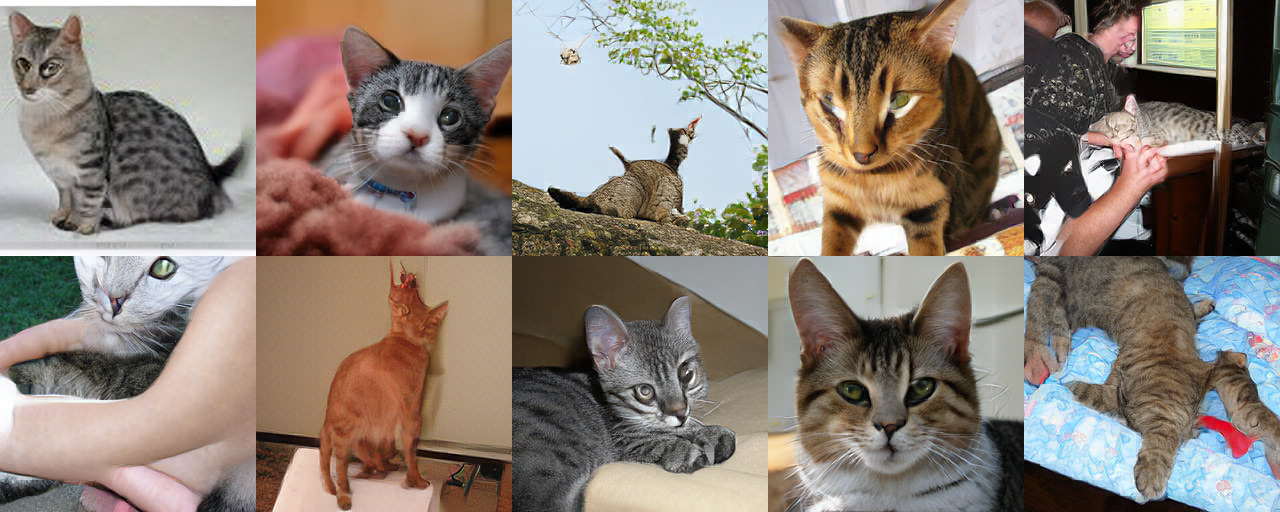}
\end{minipage}\hspace{0.025\textwidth}%
\begin{minipage}[t]{0.41\textwidth}\centering
{\scriptsize class 288: leopard\par}\vspace{1pt}
\includegraphics[width=\linewidth]{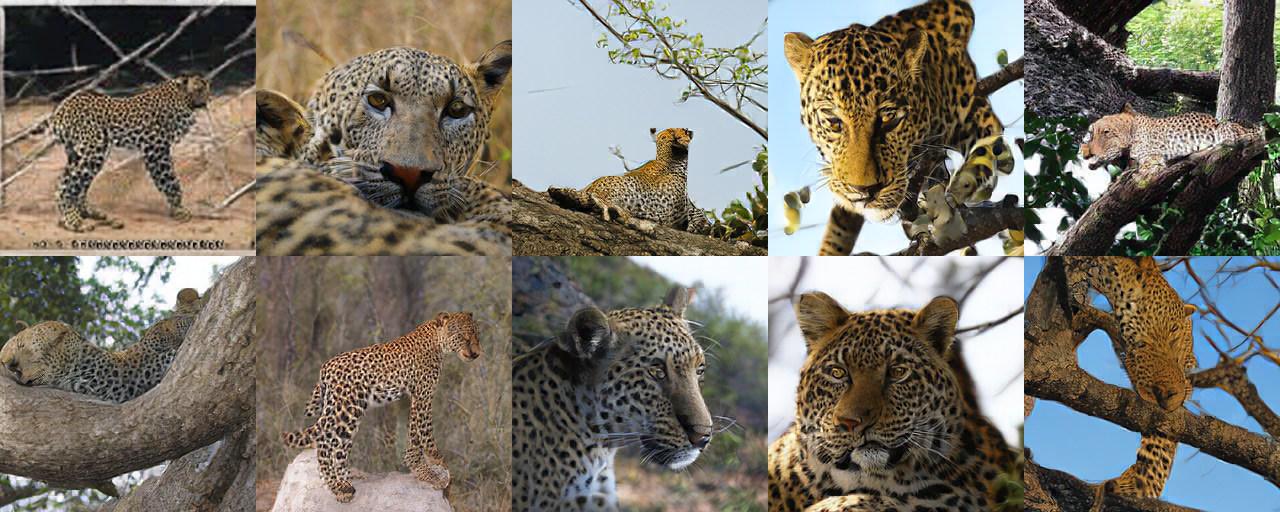}
\end{minipage}%
}\par\vspace{2pt}
\noindent\makebox[\textwidth][c]{%
\begin{minipage}[t]{0.41\textwidth}\centering
{\scriptsize class 291: lion\par}\vspace{1pt}
\includegraphics[width=\linewidth]{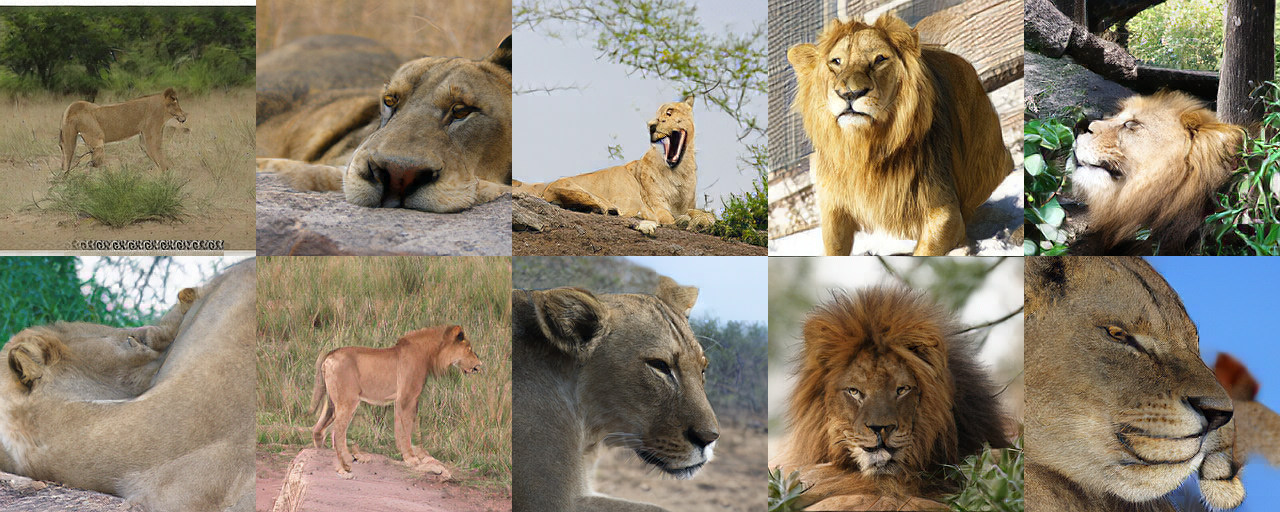}
\end{minipage}\hspace{0.025\textwidth}%
\begin{minipage}[t]{0.41\textwidth}\centering
{\scriptsize class 292: tiger\par}\vspace{1pt}
\includegraphics[width=\linewidth]{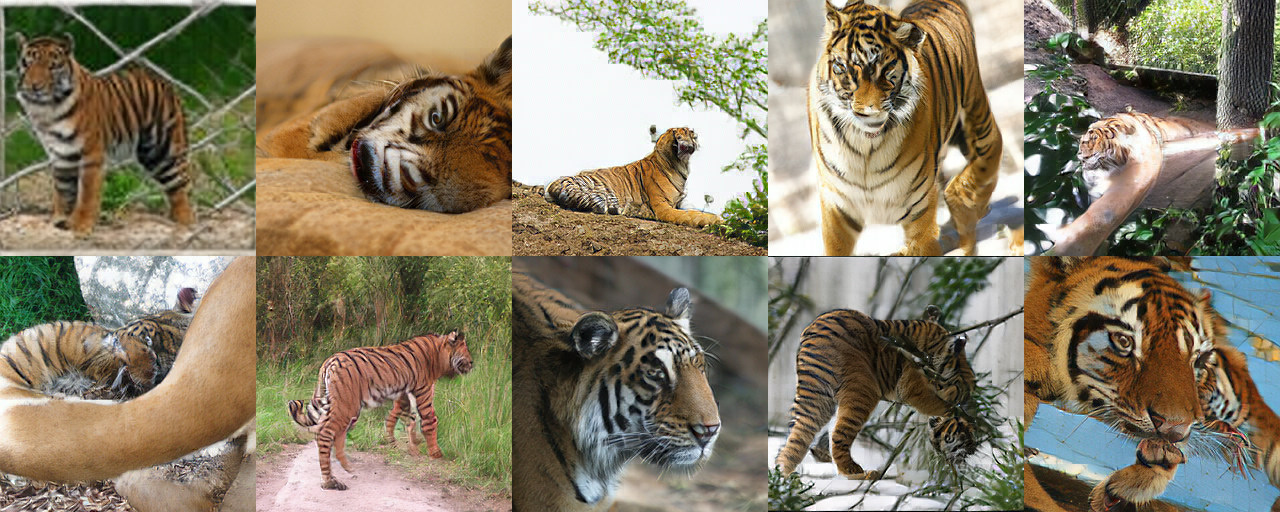}
\end{minipage}%
}\par\vspace{2pt}
\noindent\makebox[\textwidth][c]{%
\begin{minipage}[t]{0.41\textwidth}\centering
{\scriptsize class 296: ice bear\par}\vspace{1pt}
\includegraphics[width=\linewidth]{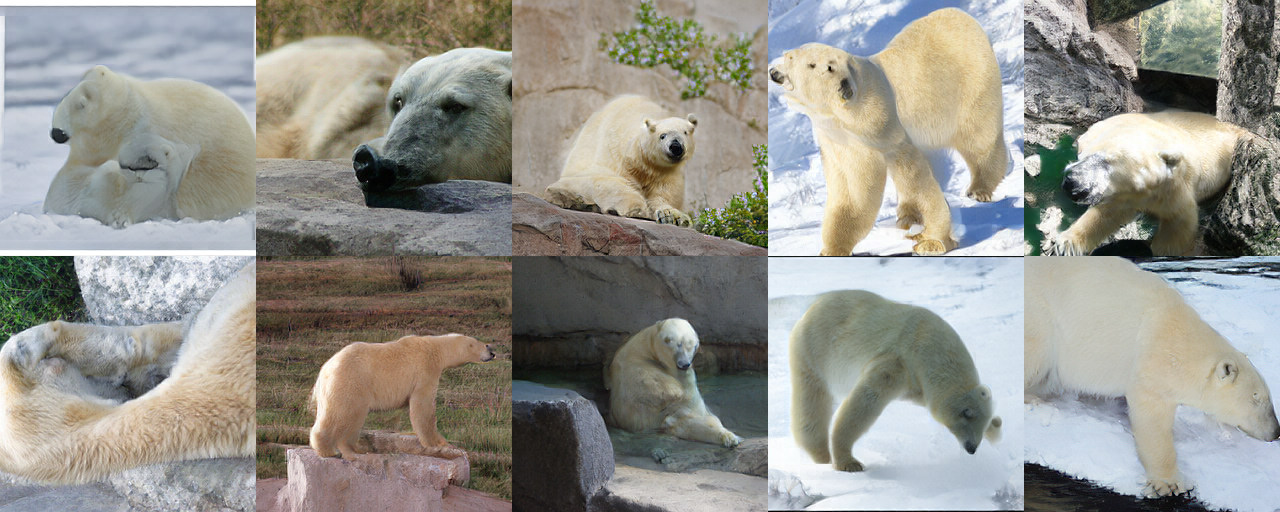}
\end{minipage}\hspace{0.025\textwidth}%
\begin{minipage}[t]{0.41\textwidth}\centering
{\scriptsize class 340: zebra\par}\vspace{1pt}
\includegraphics[width=\linewidth]{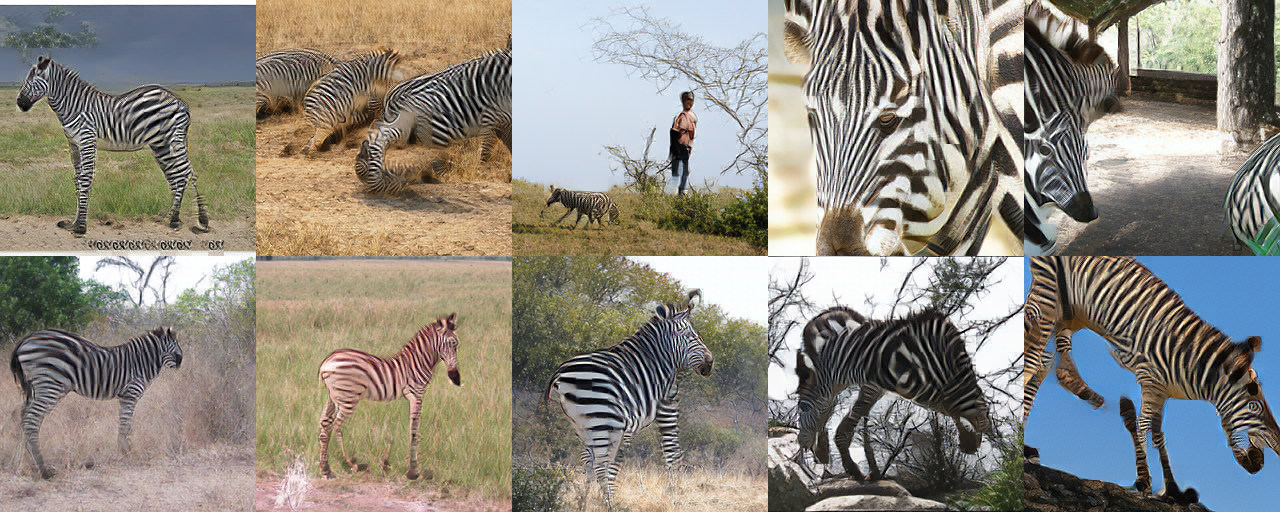}
\end{minipage}%
}\par\vspace{2pt}
\noindent\makebox[\textwidth][c]{%
\begin{minipage}[t]{0.41\textwidth}\centering
{\scriptsize class 349: bighorn\par}\vspace{1pt}
\includegraphics[width=\linewidth]{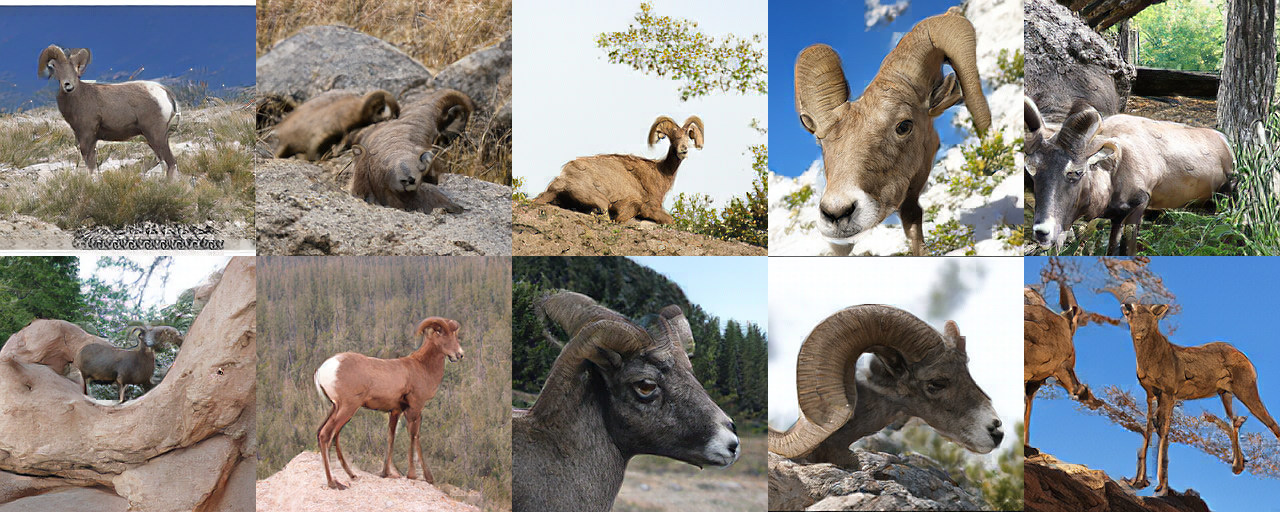}
\end{minipage}\hspace{0.025\textwidth}%
\begin{minipage}[t]{0.41\textwidth}\centering
{\scriptsize class 360: otter\par}\vspace{1pt}
\includegraphics[width=\linewidth]{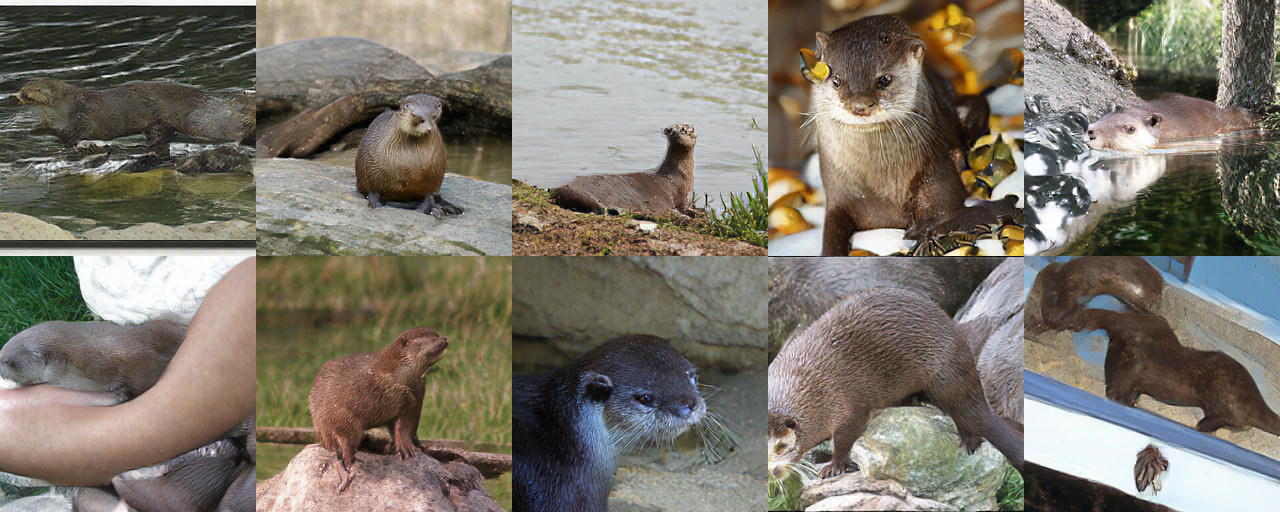}
\end{minipage}%
}\par\vspace{2pt}
\noindent\makebox[\textwidth][c]{%
\begin{minipage}[t]{0.41\textwidth}\centering
{\scriptsize class 386: African elephant\par}\vspace{1pt}
\includegraphics[width=\linewidth]{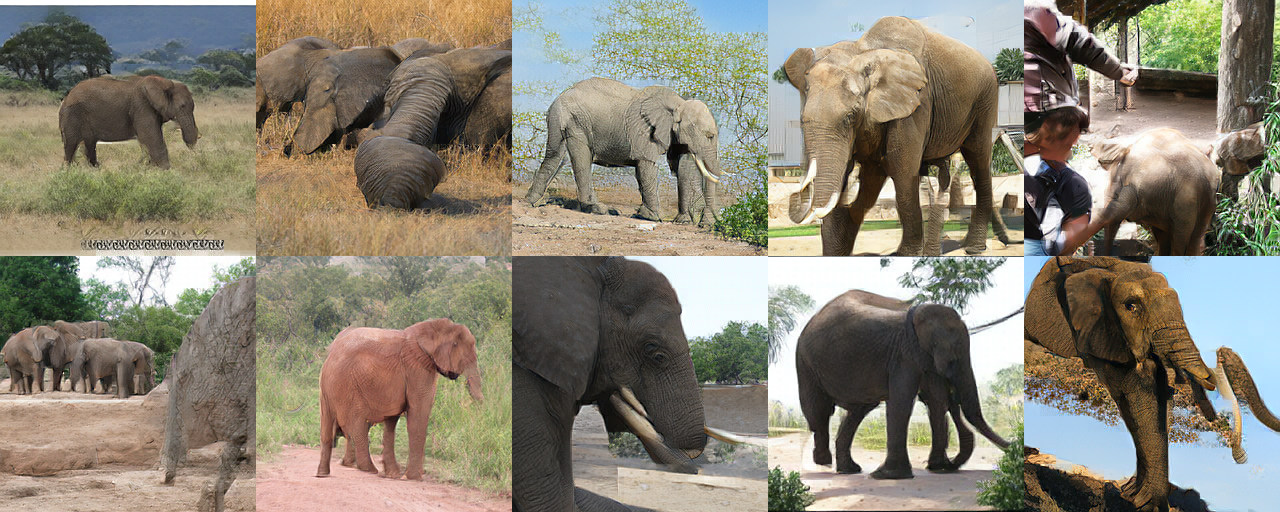}
\end{minipage}\hspace{0.025\textwidth}%
\begin{minipage}[t]{0.41\textwidth}\centering
{\scriptsize class 387: lesser panda\par}\vspace{1pt}
\includegraphics[width=\linewidth]{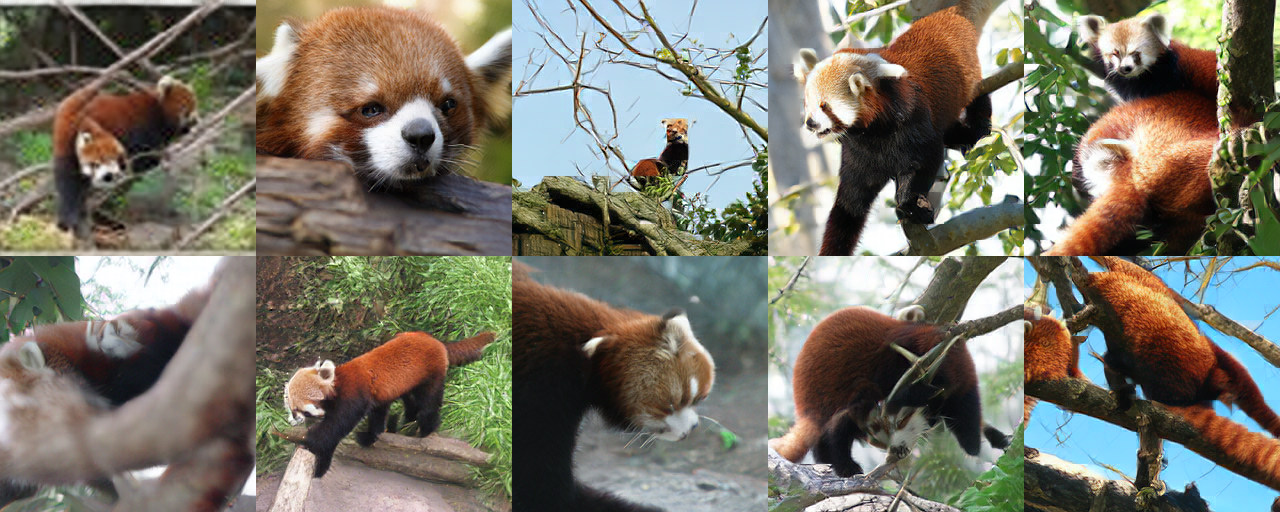}
\end{minipage}%
}\par\vspace{2pt}
\noindent\makebox[\textwidth][c]{%
\begin{minipage}[t]{0.41\textwidth}\centering
{\scriptsize class 388: giant panda\par}\vspace{1pt}
\includegraphics[width=\linewidth]{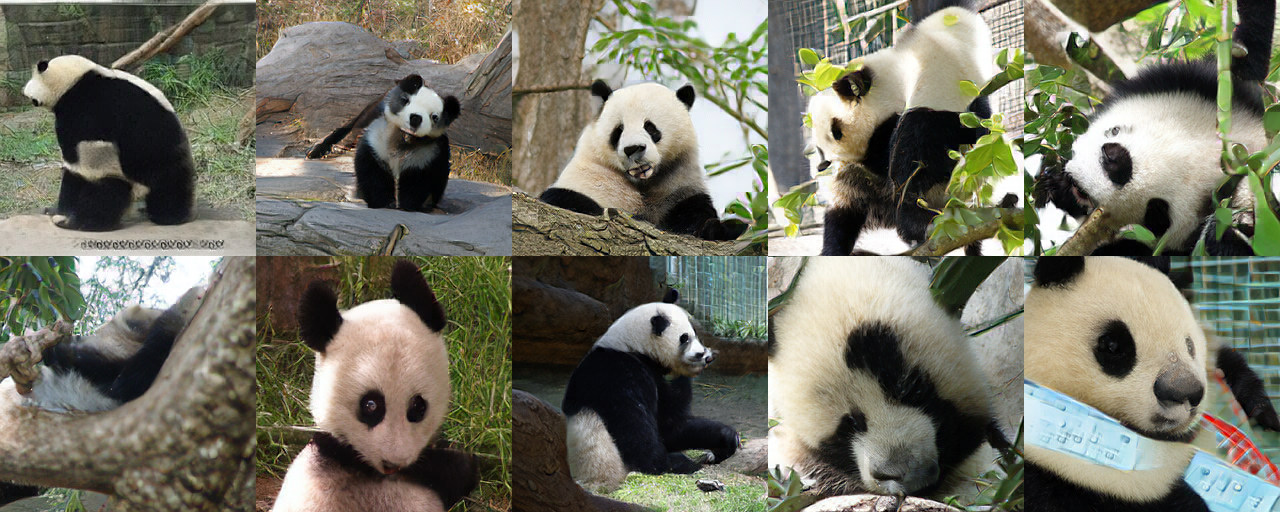}
\end{minipage}\hspace{0.025\textwidth}%
\begin{minipage}[t]{0.41\textwidth}\centering
{\scriptsize class 1: goldfish\par}\vspace{1pt}
\includegraphics[width=\linewidth]{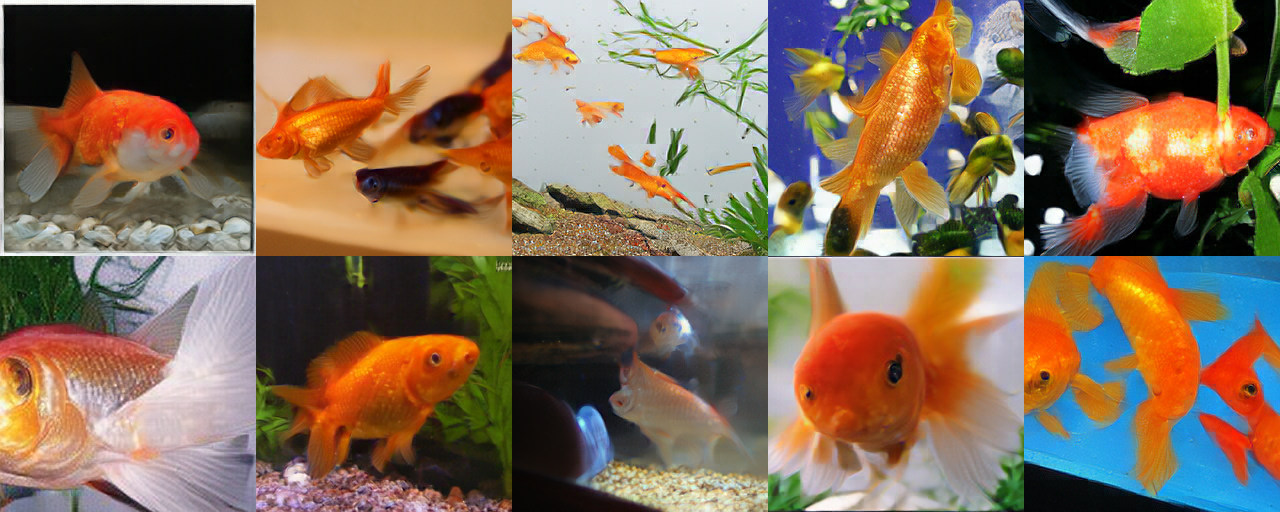}
\end{minipage}%
}\par\vspace{2pt}
\noindent\makebox[\textwidth][c]{%
\begin{minipage}[t]{0.41\textwidth}\centering
{\scriptsize class 31: tree frog\par}\vspace{1pt}
\includegraphics[width=\linewidth]{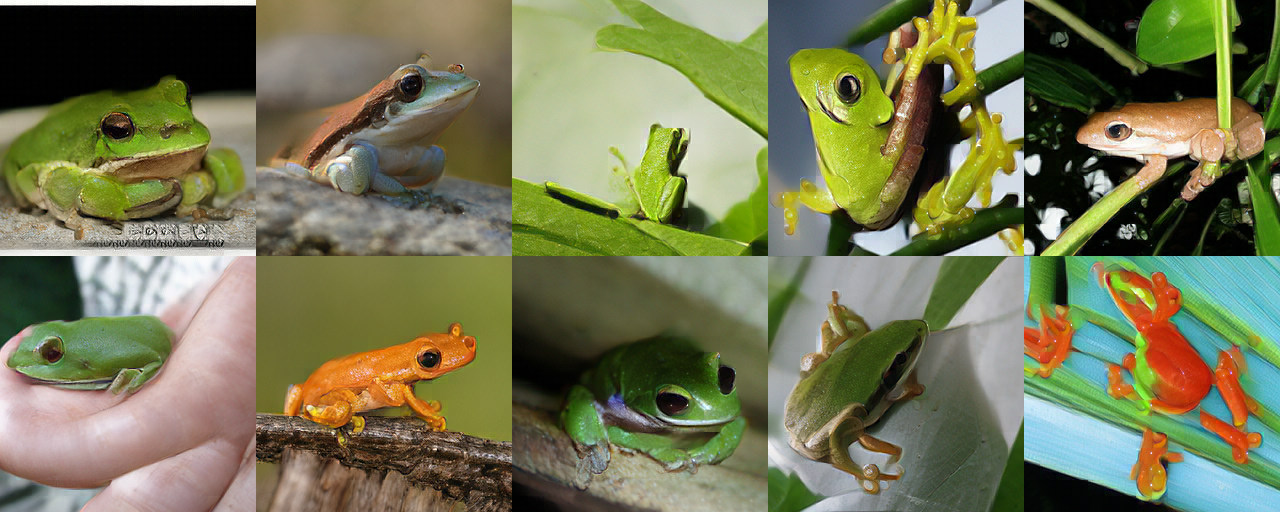}
\end{minipage}\hspace{0.025\textwidth}%
\begin{minipage}[t]{0.41\textwidth}\centering
{\scriptsize class 107: jellyfish\par}\vspace{1pt}
\includegraphics[width=\linewidth]{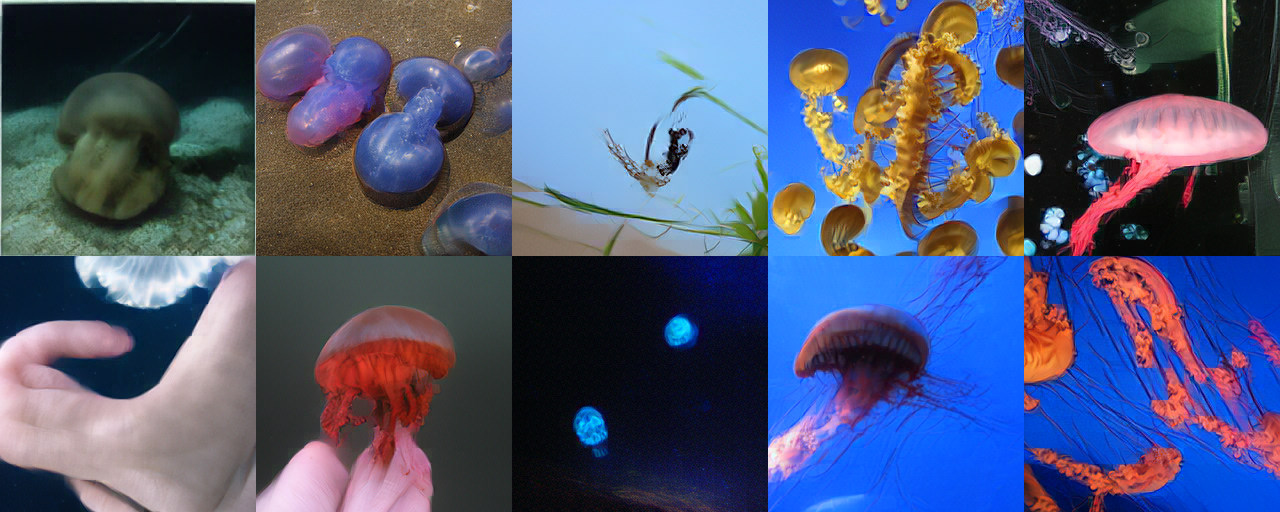}
\end{minipage}%
}\par\vspace{2pt}
\noindent\makebox[\textwidth][c]{%
\begin{minipage}[t]{0.41\textwidth}\centering
{\scriptsize class 108: sea anemone\par}\vspace{1pt}
\includegraphics[width=\linewidth]{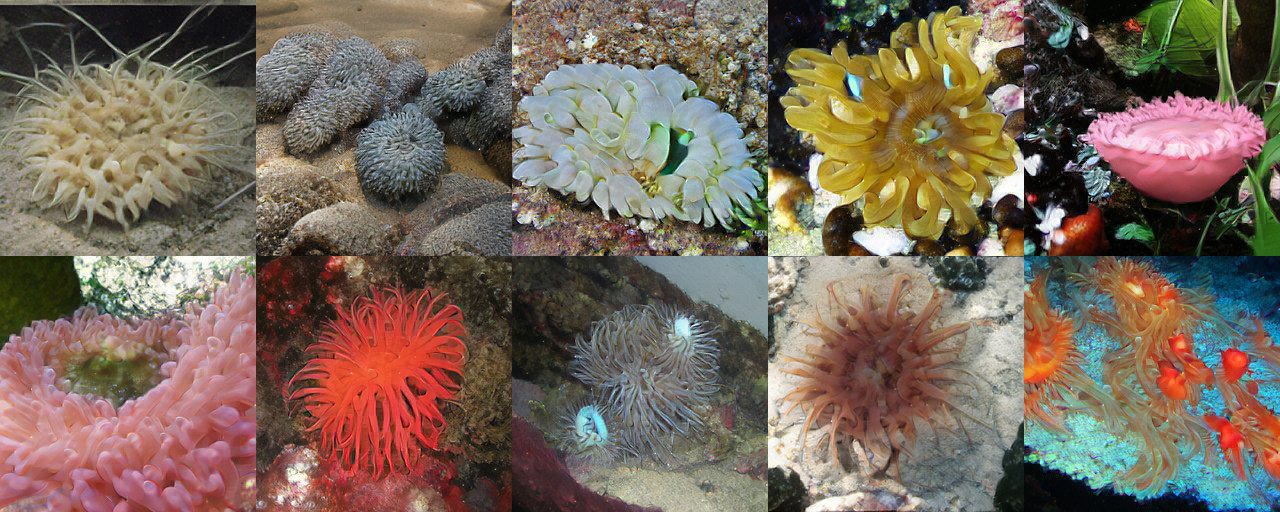}
\end{minipage}\hspace{0.025\textwidth}%
\begin{minipage}[t]{0.41\textwidth}\centering
{\scriptsize class 323: monarch\par}\vspace{1pt}
\includegraphics[width=\linewidth]{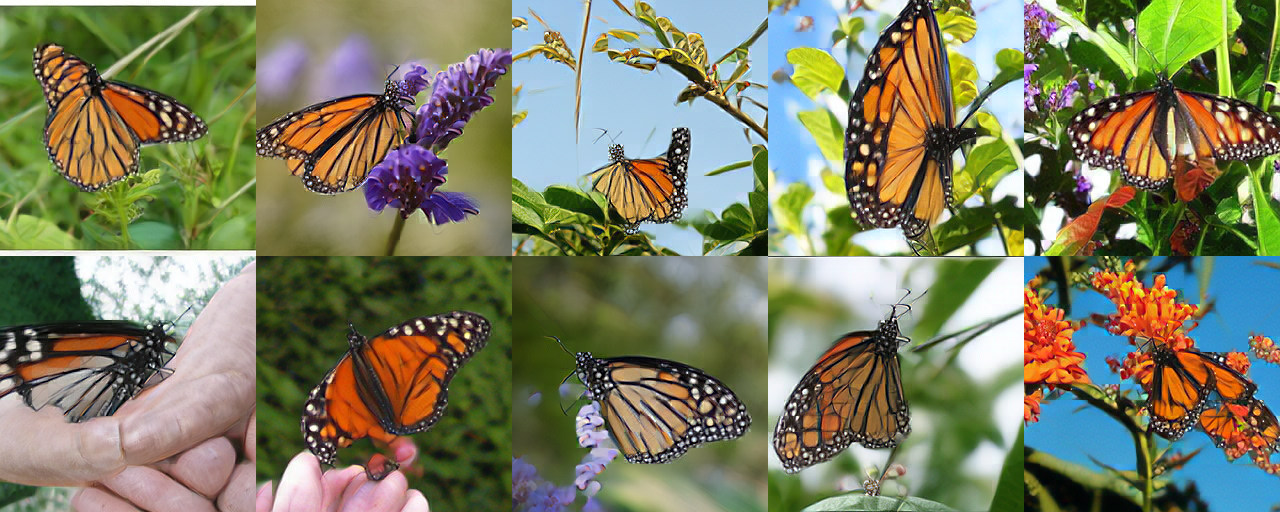}
\end{minipage}%
}\par\vspace{2pt}
\noindent\makebox[\textwidth][c]{%
\begin{minipage}[t]{0.41\textwidth}\centering
{\scriptsize class 973: coral reef\par}\vspace{1pt}
\includegraphics[width=\linewidth]{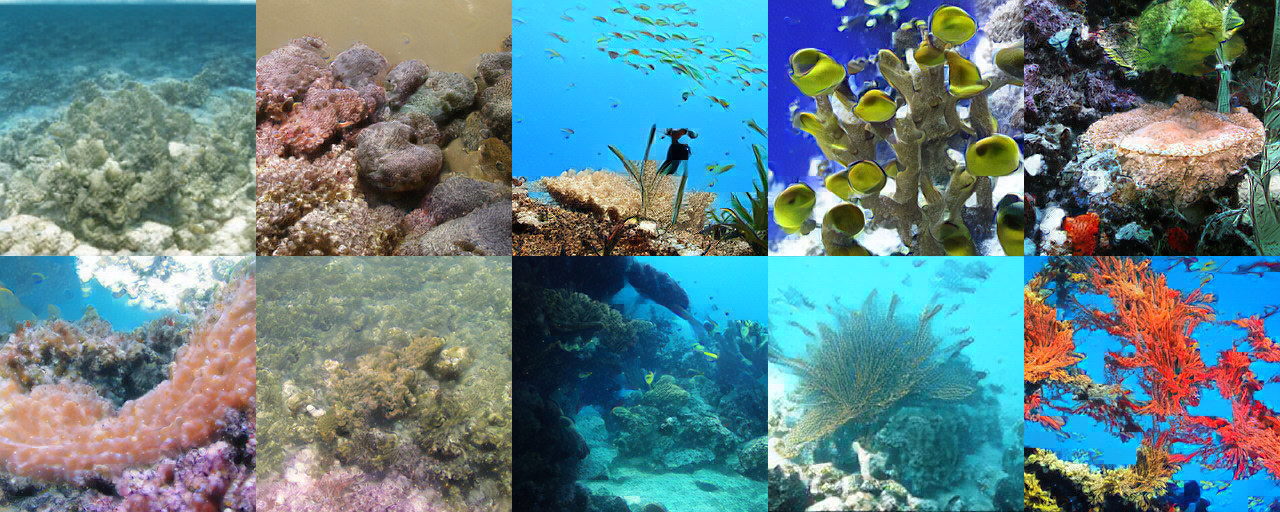}
\end{minipage}\hspace{0.025\textwidth}%
\begin{minipage}[t]{0.41\textwidth}\centering
{\scriptsize class 985: daisy\par}\vspace{1pt}
\includegraphics[width=\linewidth]{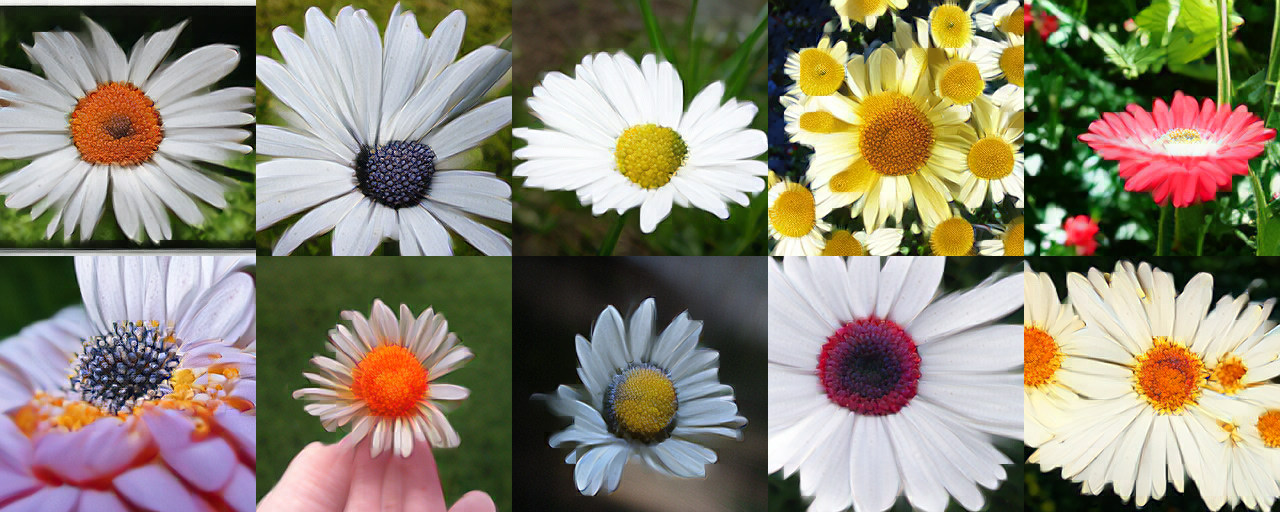}
\end{minipage}%
}\par\vspace{2pt}
\noindent\makebox[\textwidth][c]{%
\begin{minipage}[t]{0.41\textwidth}\centering
{\scriptsize class 992: agaric\par}\vspace{1pt}
\includegraphics[width=\linewidth]{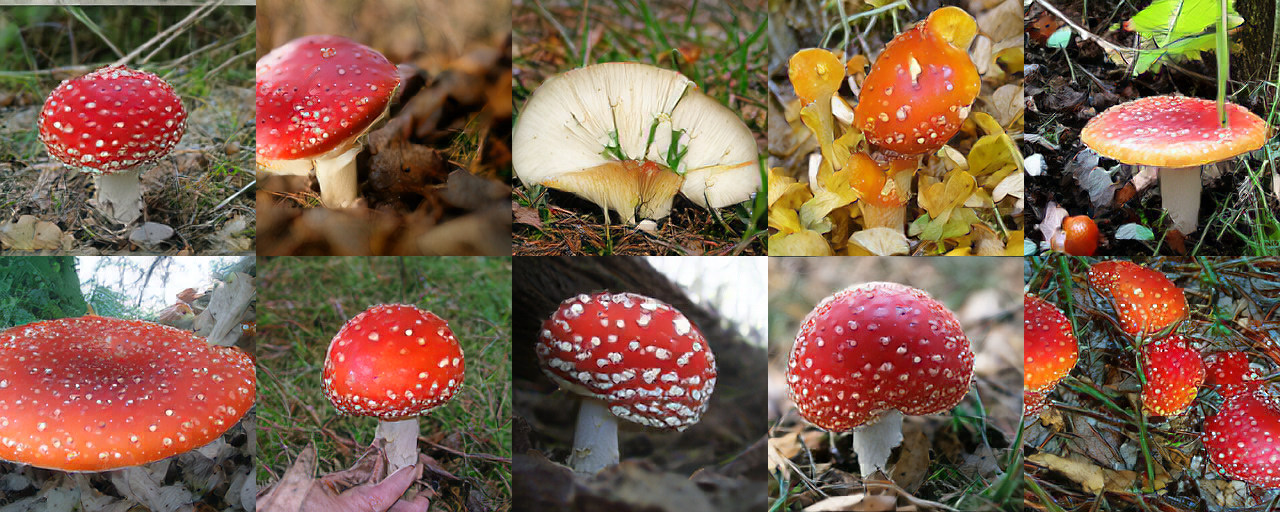}
\end{minipage}\hspace{0.025\textwidth}%
\begin{minipage}[t]{0.41\textwidth}\centering
{\scriptsize class 417: balloon\par}\vspace{1pt}
\includegraphics[width=\linewidth]{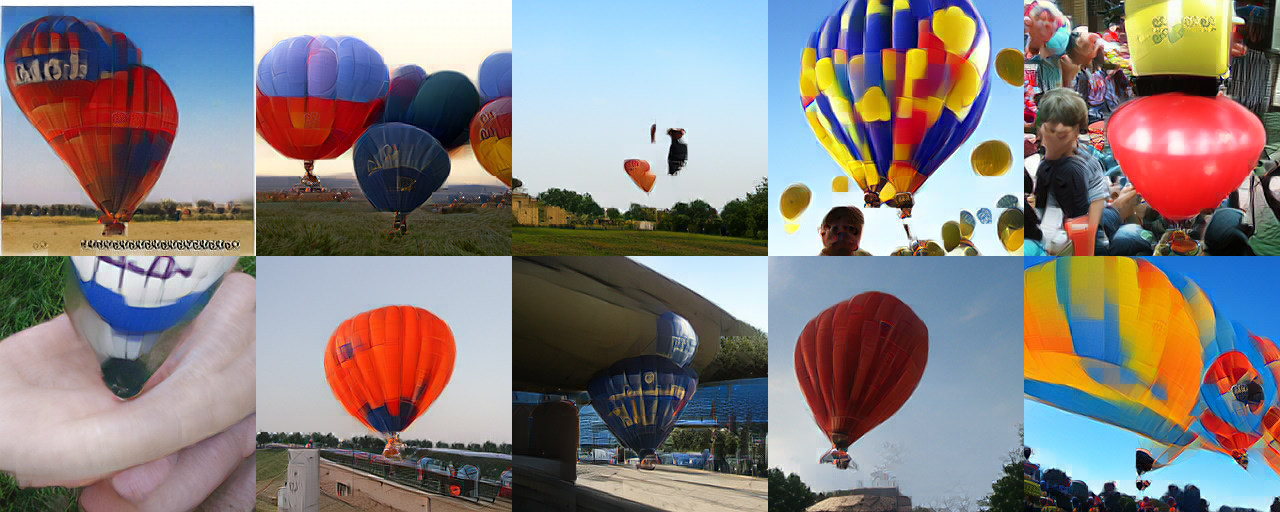}
\end{minipage}%
}\par\vspace{2pt}
\noindent\makebox[\textwidth][c]{%
\begin{minipage}[t]{0.41\textwidth}\centering
{\scriptsize class 425: barn\par}\vspace{1pt}
\includegraphics[width=\linewidth]{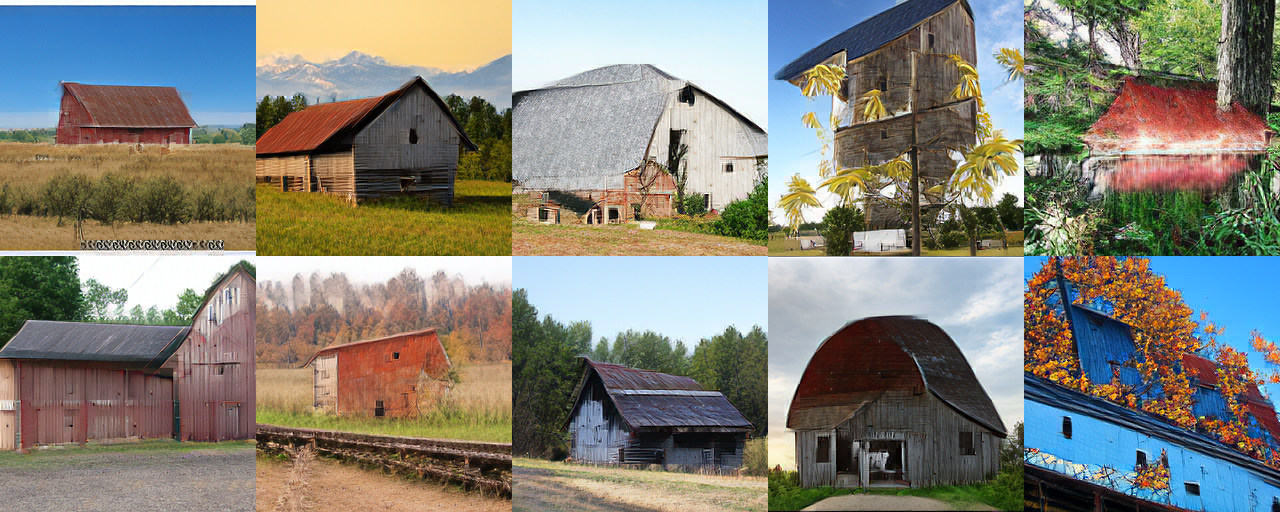}
\end{minipage}\hspace{0.025\textwidth}%
\begin{minipage}[t]{0.41\textwidth}\centering
{\scriptsize class 448: birdhouse\par}\vspace{1pt}
\includegraphics[width=\linewidth]{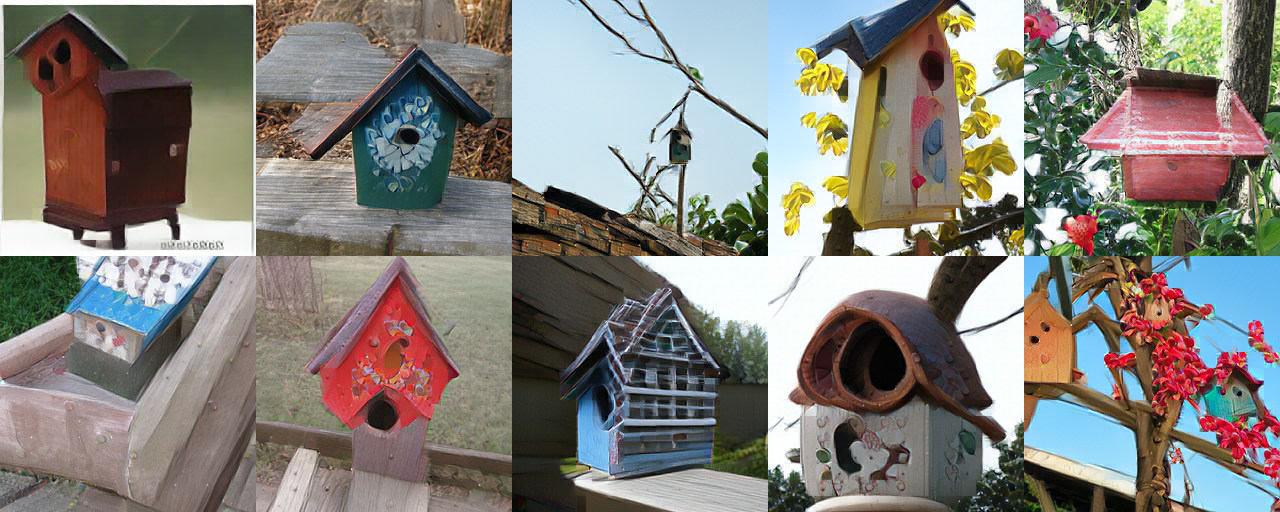}
\end{minipage}%
}\par\vspace{2pt}
\noindent\makebox[\textwidth][c]{%
\begin{minipage}[t]{0.41\textwidth}\centering
{\scriptsize class 483: castle\par}\vspace{1pt}
\includegraphics[width=\linewidth]{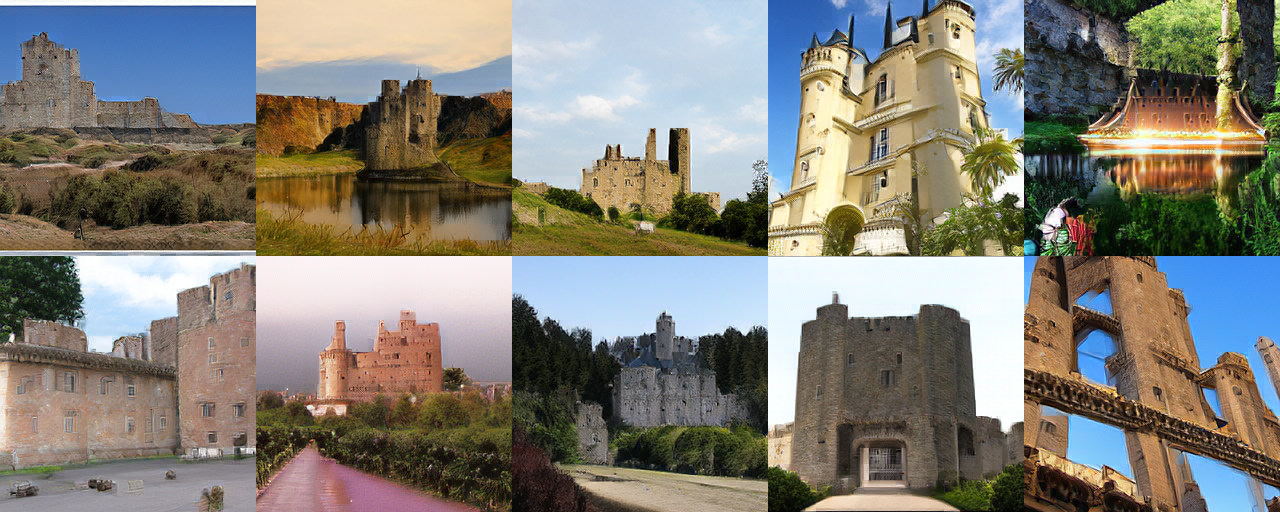}
\end{minipage}\hspace{0.025\textwidth}%
\begin{minipage}[t]{0.41\textwidth}\centering
{\scriptsize class 580: greenhouse\par}\vspace{1pt}
\includegraphics[width=\linewidth]{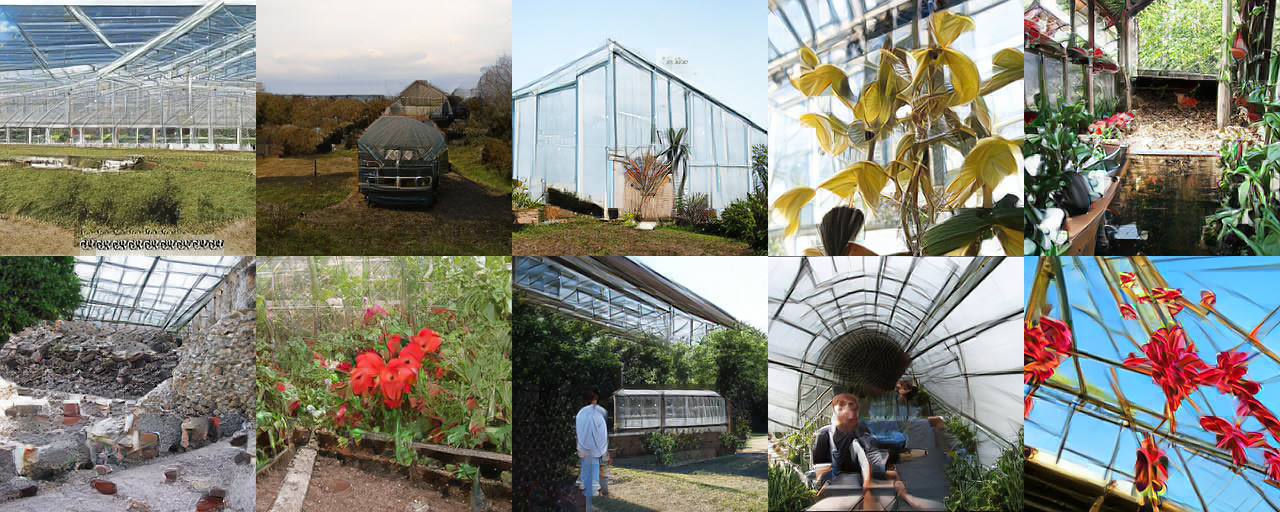}
\end{minipage}%
}\par\vspace{2pt}
\noindent\makebox[\textwidth][c]{%
\begin{minipage}[t]{0.41\textwidth}\centering
{\scriptsize class 649: megalith\par}\vspace{1pt}
\includegraphics[width=\linewidth]{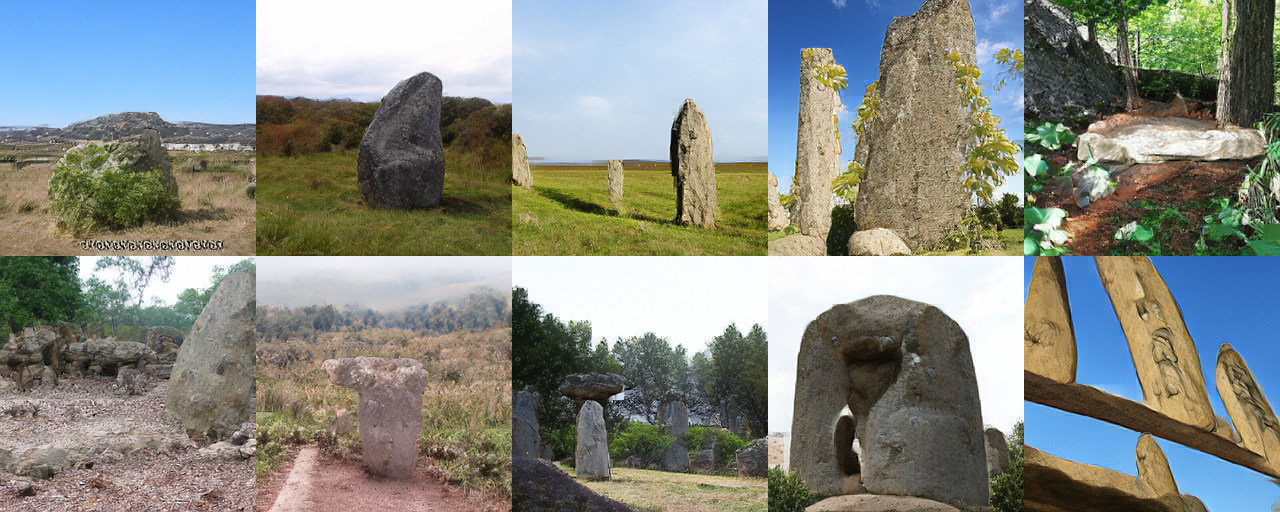}
\end{minipage}\hspace{0.025\textwidth}%
\begin{minipage}[t]{0.41\textwidth}\centering
{\scriptsize class 698: palace\par}\vspace{1pt}
\includegraphics[width=\linewidth]{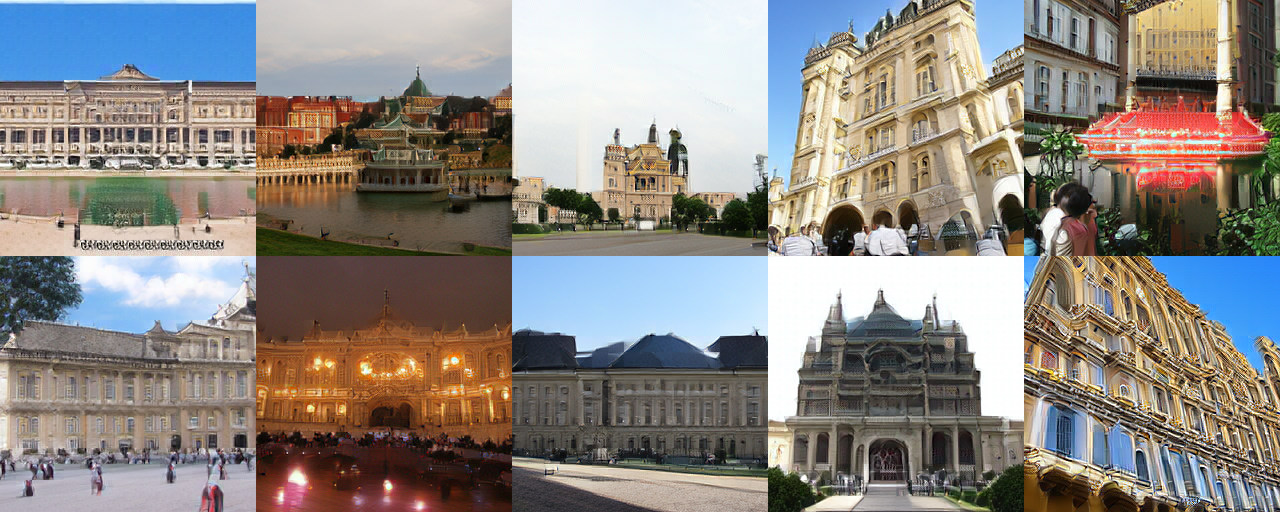}
\end{minipage}%
}\par\vspace{2pt}
\noindent\makebox[\textwidth][c]{%
\begin{minipage}[t]{0.41\textwidth}\centering
{\scriptsize class 718: pier\par}\vspace{1pt}
\includegraphics[width=\linewidth]{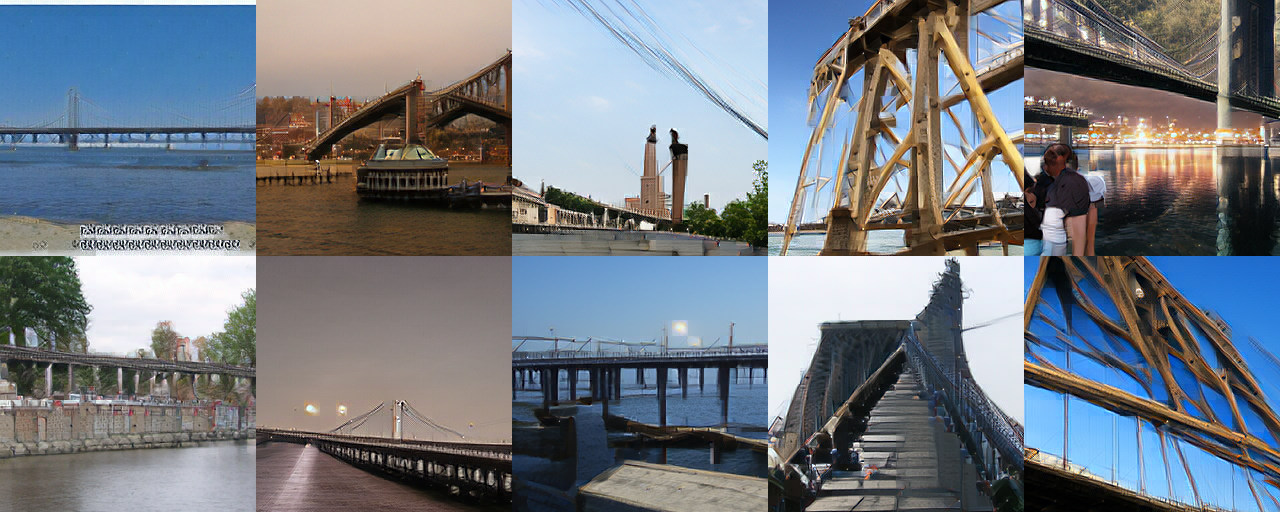}
\end{minipage}\hspace{0.025\textwidth}%
\begin{minipage}[t]{0.41\textwidth}\centering
{\scriptsize class 755: radio telescope\par}\vspace{1pt}
\includegraphics[width=\linewidth]{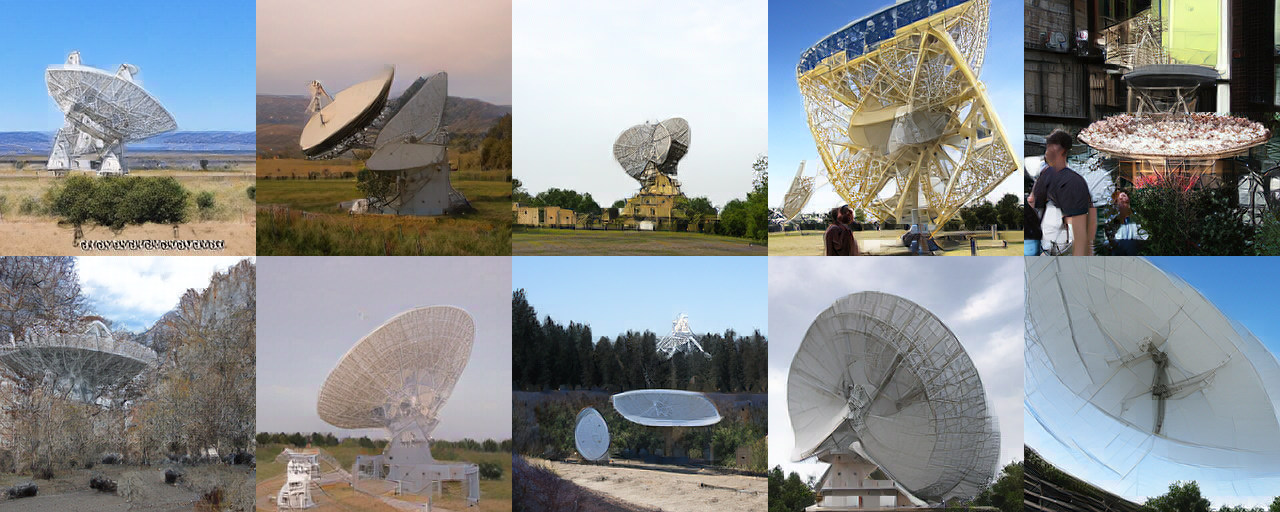}
\end{minipage}%
}\par\vspace{2pt}
\noindent\makebox[\textwidth][c]{%
\begin{minipage}[t]{0.41\textwidth}\centering
{\scriptsize class 780: schooner\par}\vspace{1pt}
\includegraphics[width=\linewidth]{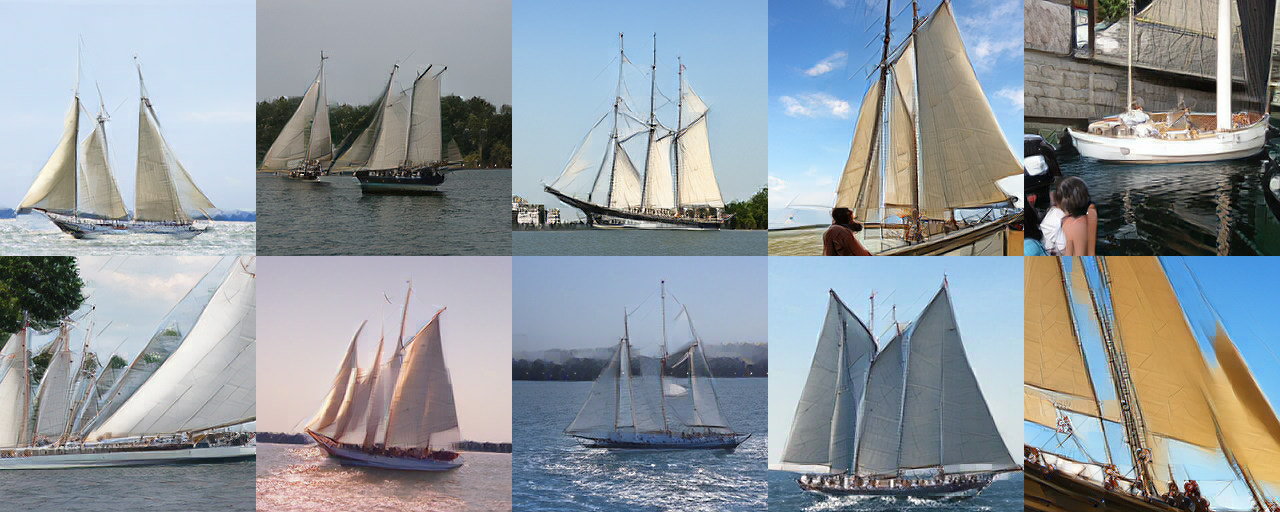}
\end{minipage}\hspace{0.025\textwidth}%
\begin{minipage}[t]{0.41\textwidth}\centering
{\scriptsize class 817: sports car\par}\vspace{1pt}
\includegraphics[width=\linewidth]{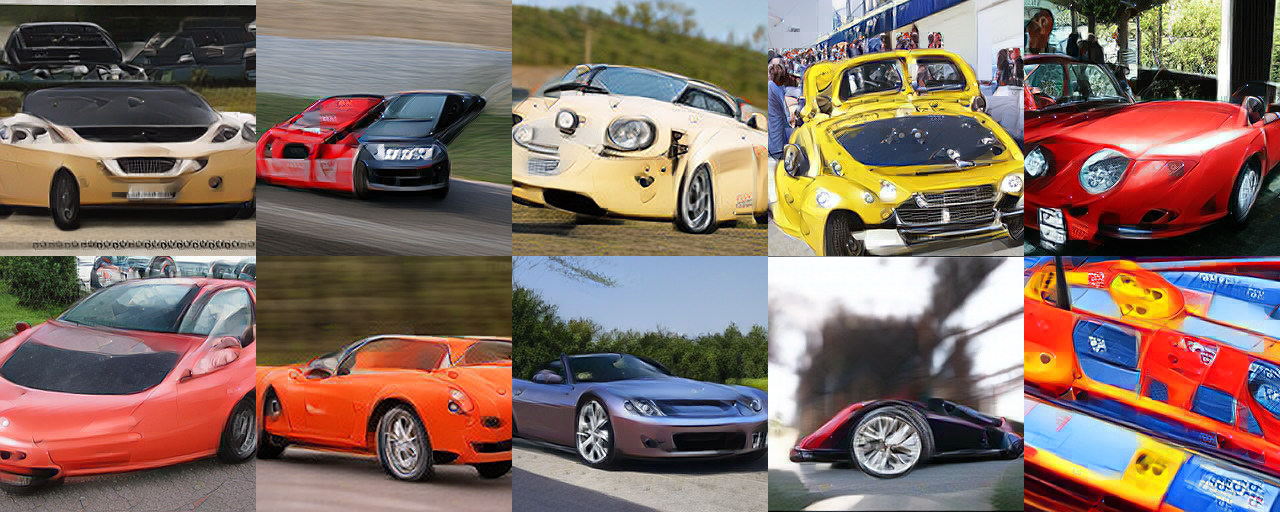}
\end{minipage}%
}\par\vspace{2pt}
\noindent\makebox[\textwidth][c]{%
\begin{minipage}[t]{0.41\textwidth}\centering
{\scriptsize class 829: streetcar\par}\vspace{1pt}
\includegraphics[width=\linewidth]{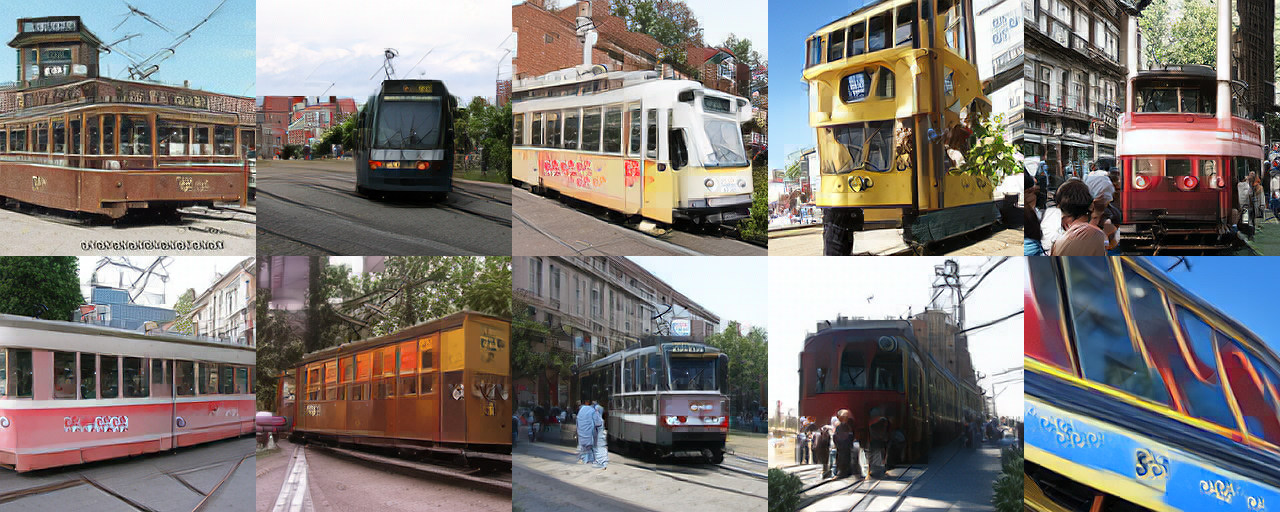}
\end{minipage}\hspace{0.025\textwidth}%
\begin{minipage}[t]{0.41\textwidth}\centering
{\scriptsize class 927: trifle\par}\vspace{1pt}
\includegraphics[width=\linewidth]{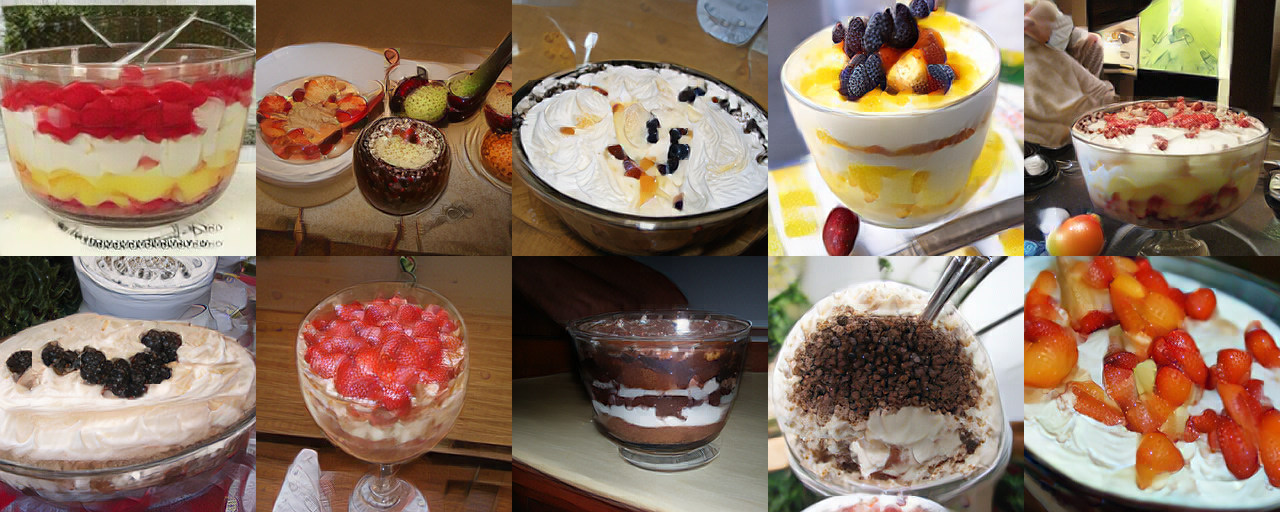}
\end{minipage}%
}\par\vspace{2pt}
\noindent\makebox[\textwidth][c]{%
\begin{minipage}[t]{0.41\textwidth}\centering
{\scriptsize class 958: hay\par}\vspace{1pt}
\includegraphics[width=\linewidth]{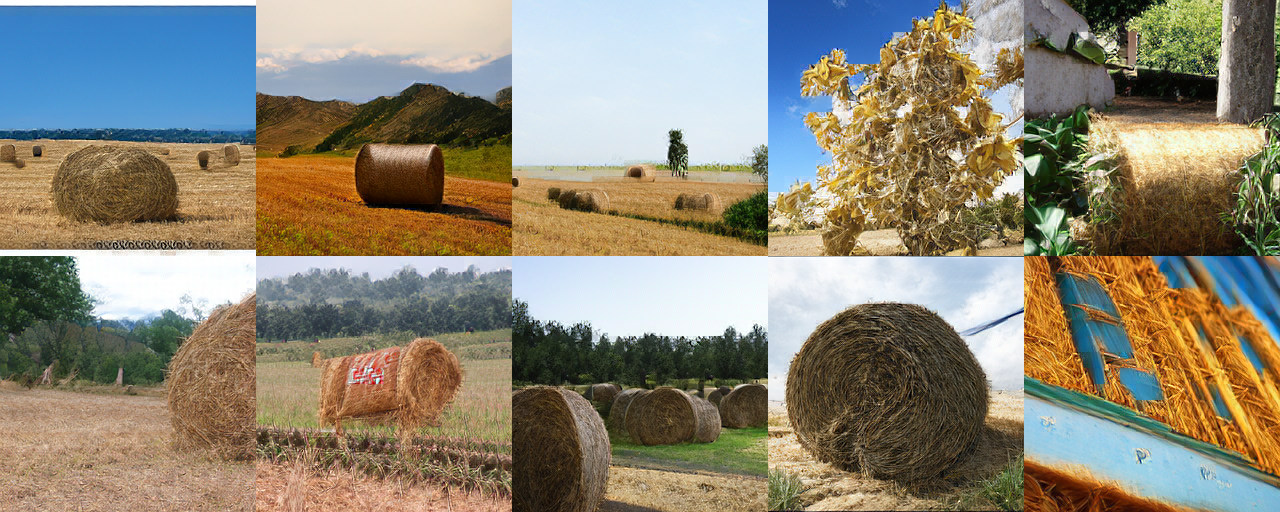}
\end{minipage}\hspace{0.025\textwidth}%
\begin{minipage}[t]{0.41\textwidth}\centering
{\scriptsize class 970: alp\par}\vspace{1pt}
\includegraphics[width=\linewidth]{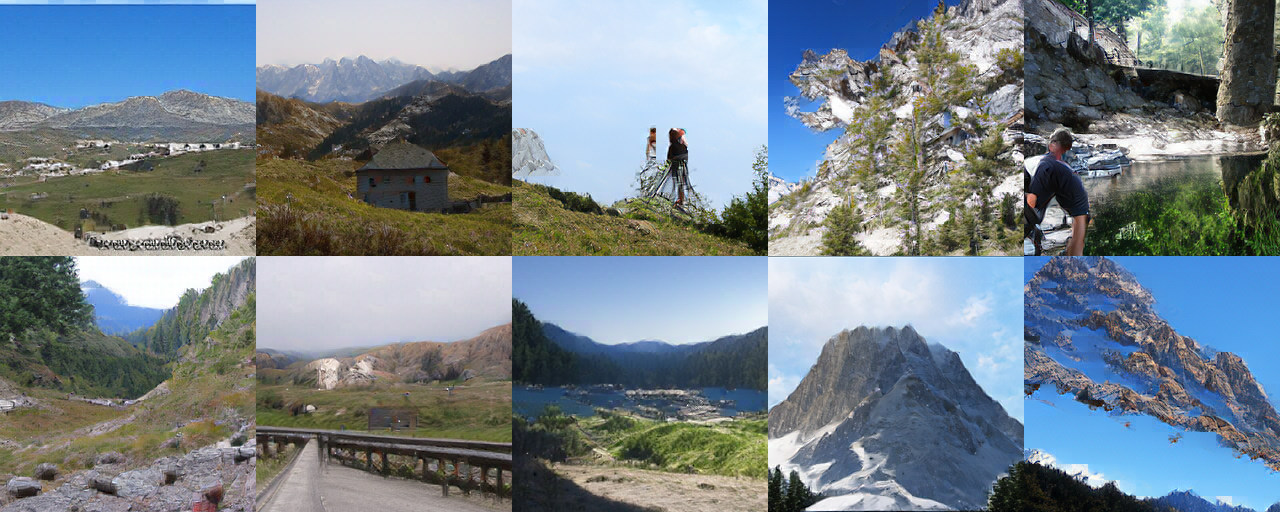}
\end{minipage}%
}\par\vspace{2pt}
\noindent\makebox[\textwidth][c]{%
\begin{minipage}[t]{0.41\textwidth}\centering
{\scriptsize class 974: geyser\par}\vspace{1pt}
\includegraphics[width=\linewidth]{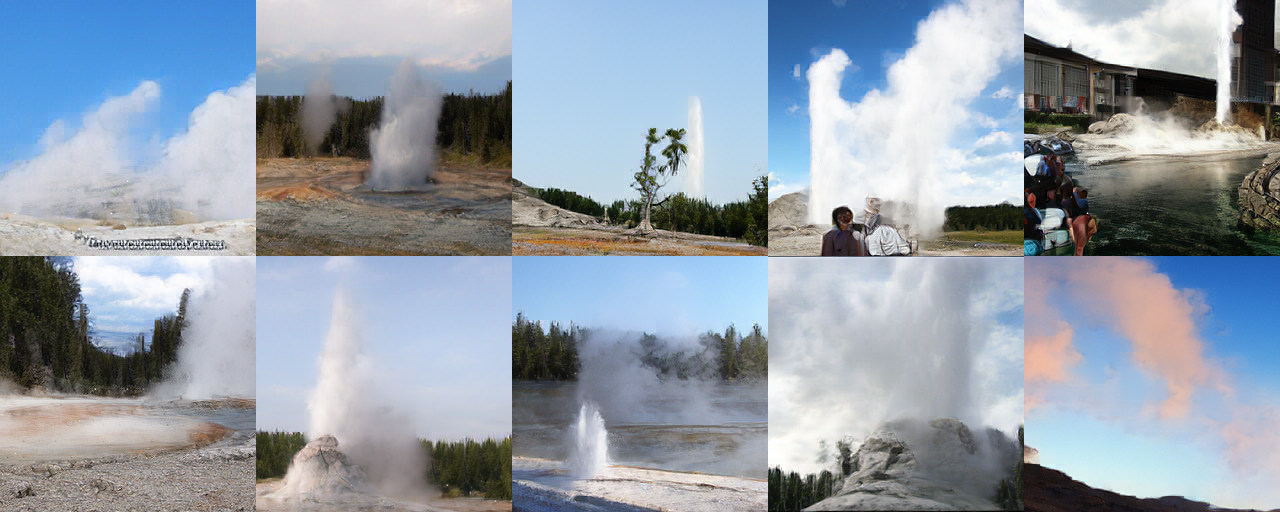}
\end{minipage}\hspace{0.025\textwidth}%
\begin{minipage}[t]{0.41\textwidth}\centering
{\scriptsize class 975: lakeside\par}\vspace{1pt}
\includegraphics[width=\linewidth]{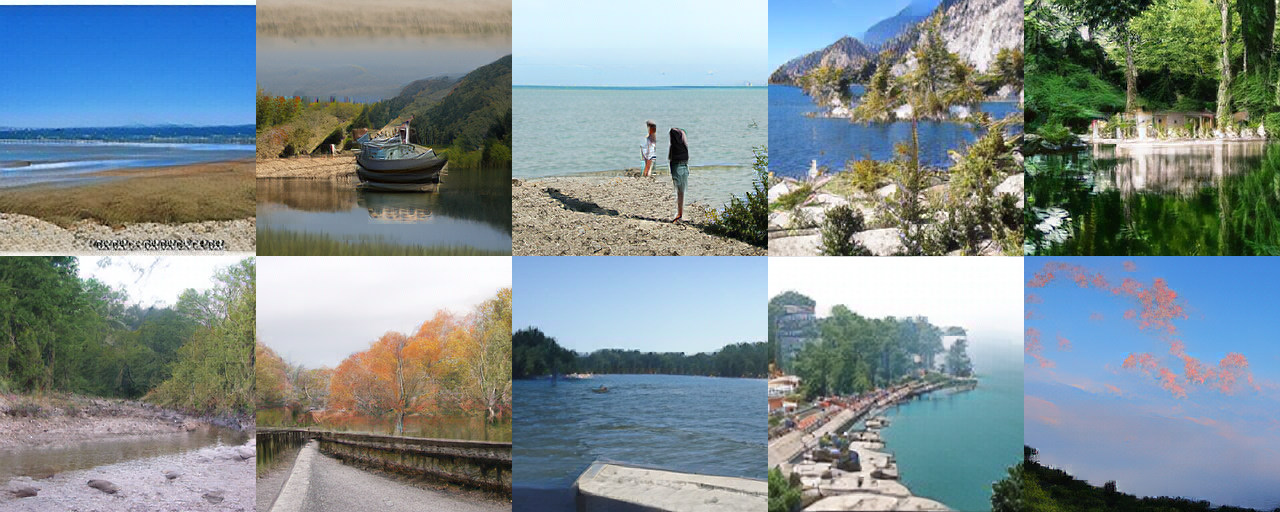}
\end{minipage}%
}\par\vspace{2pt}
\noindent\makebox[\textwidth][c]{%
\begin{minipage}[t]{0.41\textwidth}\centering
{\scriptsize class 979: valley\par}\vspace{1pt}
\includegraphics[width=\linewidth]{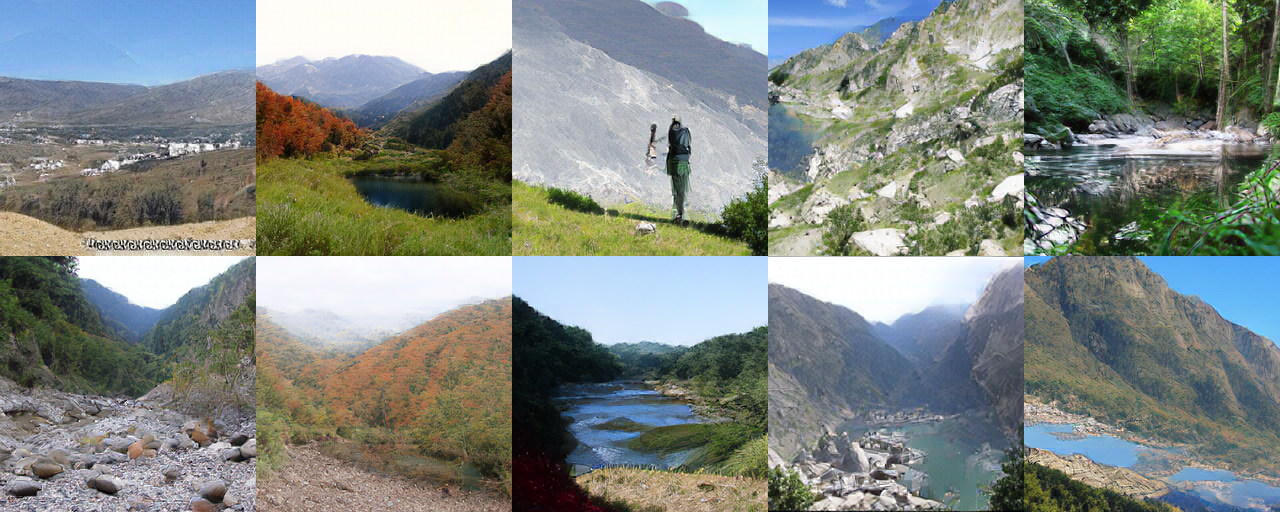}
\end{minipage}\hspace{0.025\textwidth}%
\begin{minipage}[t]{0.41\textwidth}\centering
{\scriptsize class 980: volcano\par}\vspace{1pt}
\includegraphics[width=\linewidth]{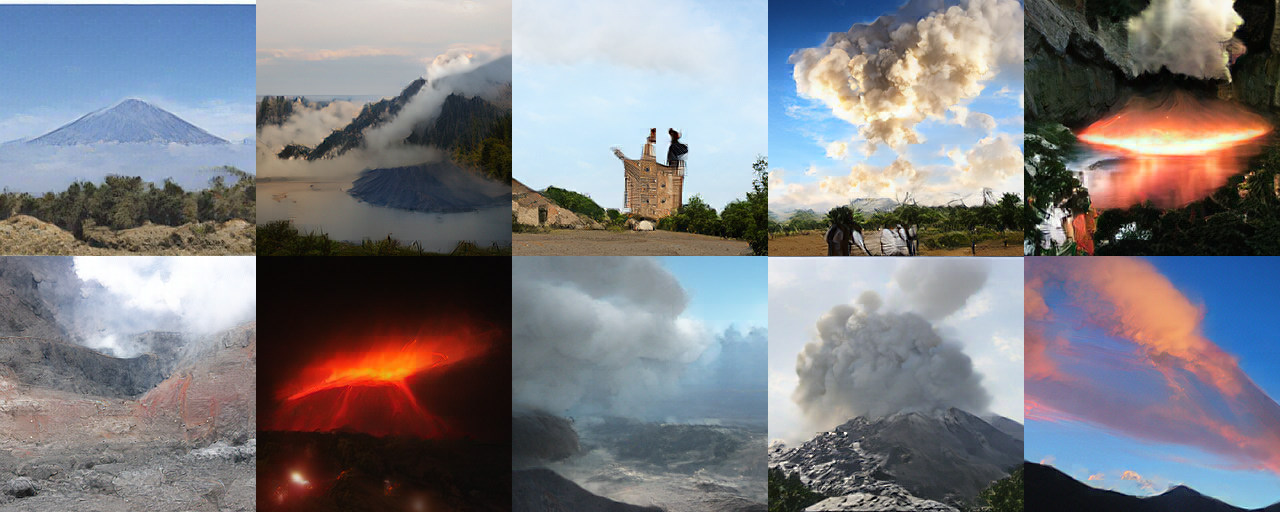}
\end{minipage}%
}\par\vspace{2pt}
\par\medskip\textbf{\normalsize Priming-L (160 ep.)}\par\nobreak\vspace{2pt}
\noindent\makebox[\textwidth][c]{%
\begin{minipage}[t]{0.41\textwidth}\centering
{\scriptsize class 12: house finch\par}\vspace{1pt}
\includegraphics[width=\linewidth]{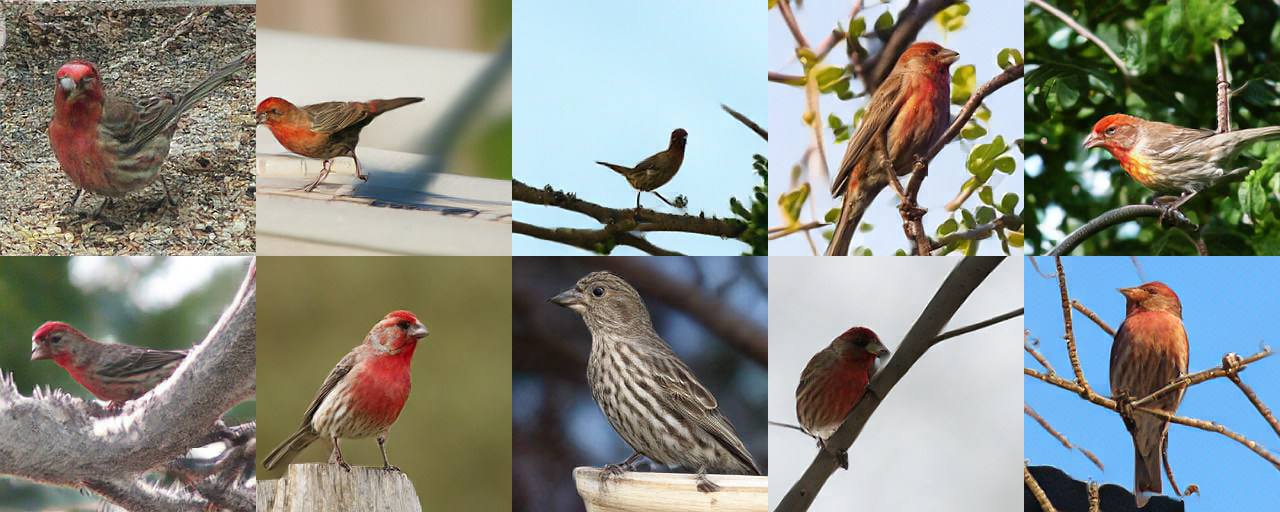}
\end{minipage}\hspace{0.025\textwidth}%
\begin{minipage}[t]{0.41\textwidth}\centering
{\scriptsize class 14: indigo bunting\par}\vspace{1pt}
\includegraphics[width=\linewidth]{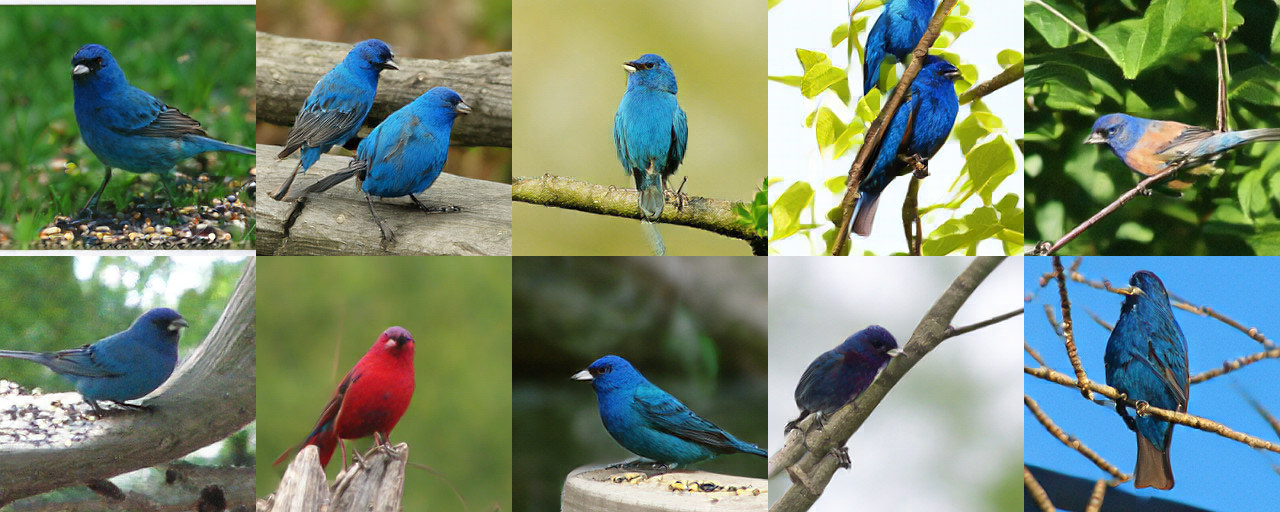}
\end{minipage}%
}\par\vspace{2pt}
\noindent\makebox[\textwidth][c]{%
\begin{minipage}[t]{0.41\textwidth}\centering
{\scriptsize class 17: jay\par}\vspace{1pt}
\includegraphics[width=\linewidth]{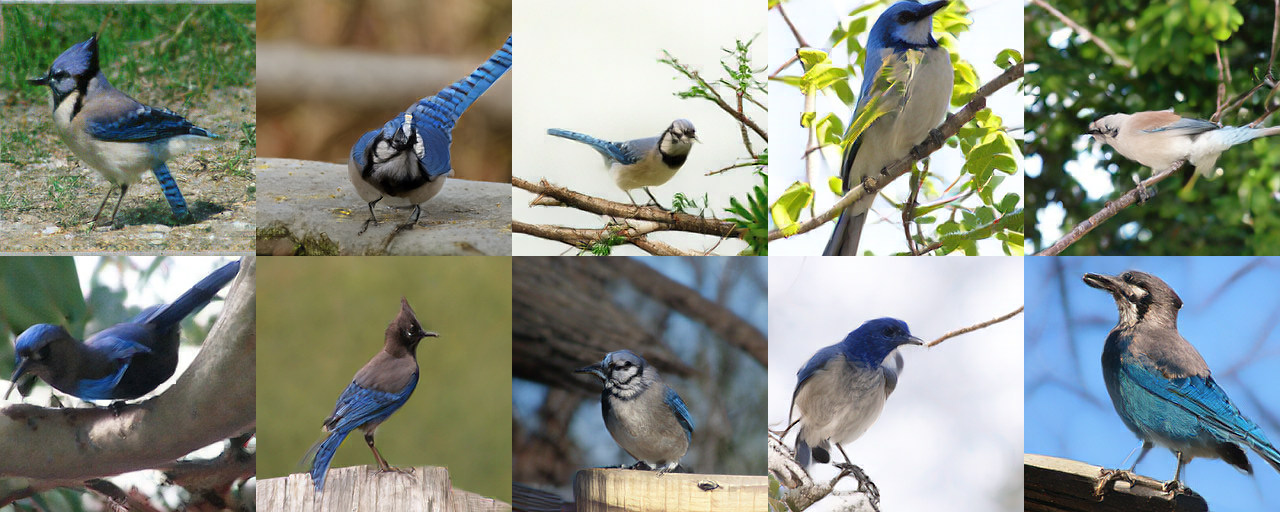}
\end{minipage}\hspace{0.025\textwidth}%
\begin{minipage}[t]{0.41\textwidth}\centering
{\scriptsize class 21: kite\par}\vspace{1pt}
\includegraphics[width=\linewidth]{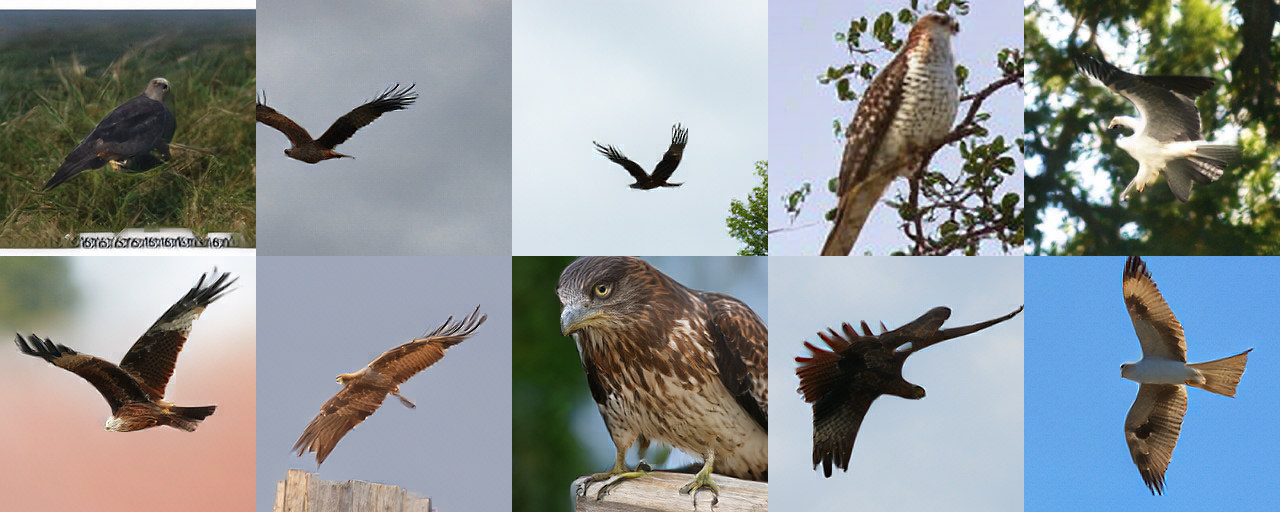}
\end{minipage}%
}\par\vspace{2pt}
\noindent\makebox[\textwidth][c]{%
\begin{minipage}[t]{0.41\textwidth}\centering
{\scriptsize class 22: bald eagle\par}\vspace{1pt}
\includegraphics[width=\linewidth]{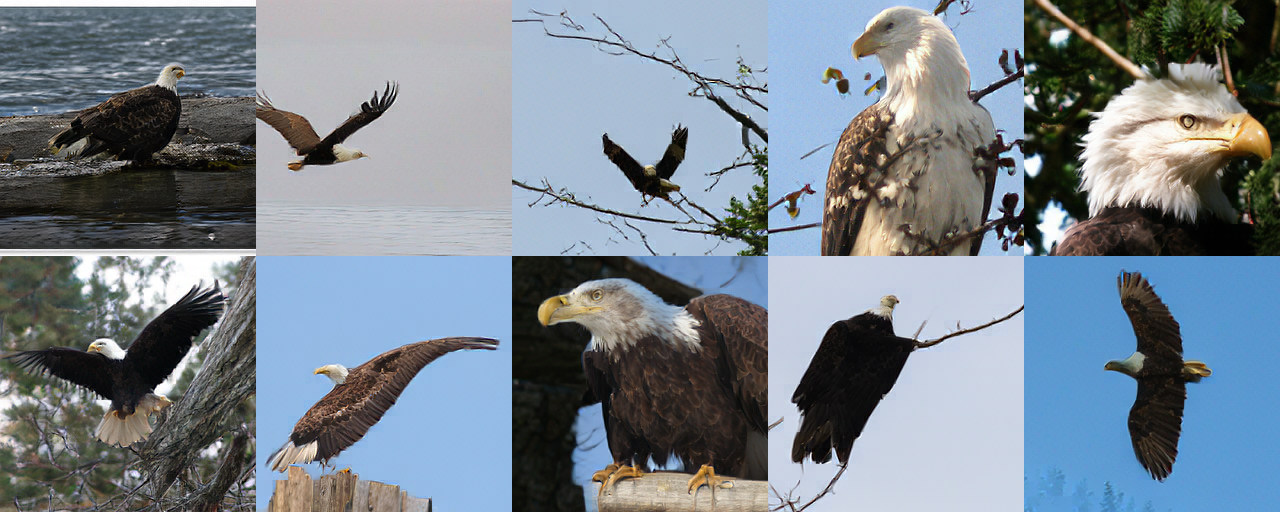}
\end{minipage}\hspace{0.025\textwidth}%
\begin{minipage}[t]{0.41\textwidth}\centering
{\scriptsize class 24: great grey owl\par}\vspace{1pt}
\includegraphics[width=\linewidth]{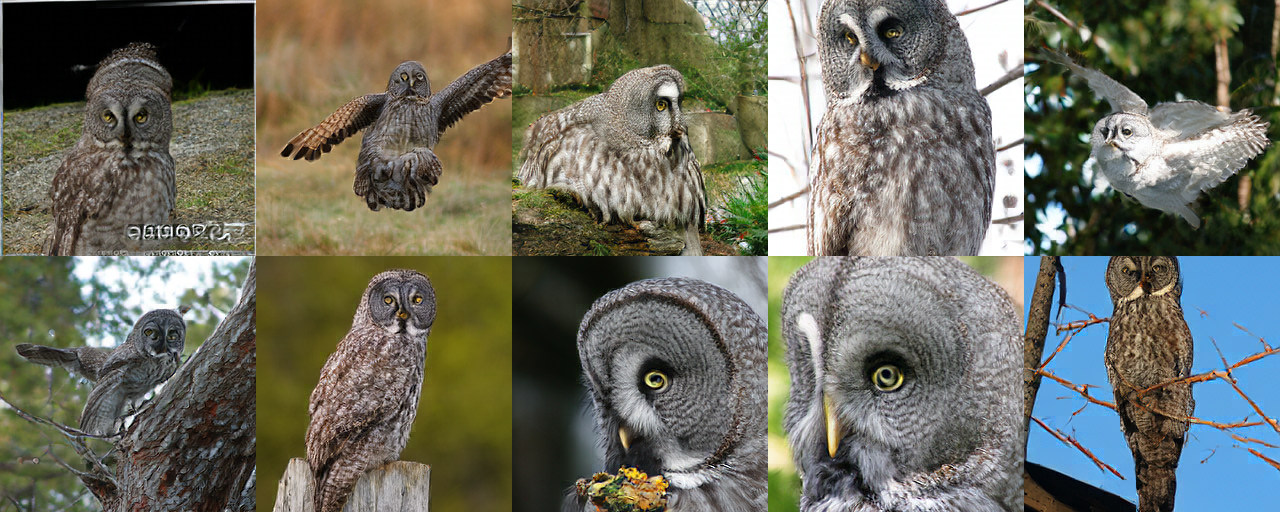}
\end{minipage}%
}\par\vspace{2pt}
\noindent\makebox[\textwidth][c]{%
\begin{minipage}[t]{0.41\textwidth}\centering
{\scriptsize class 84: peacock\par}\vspace{1pt}
\includegraphics[width=\linewidth]{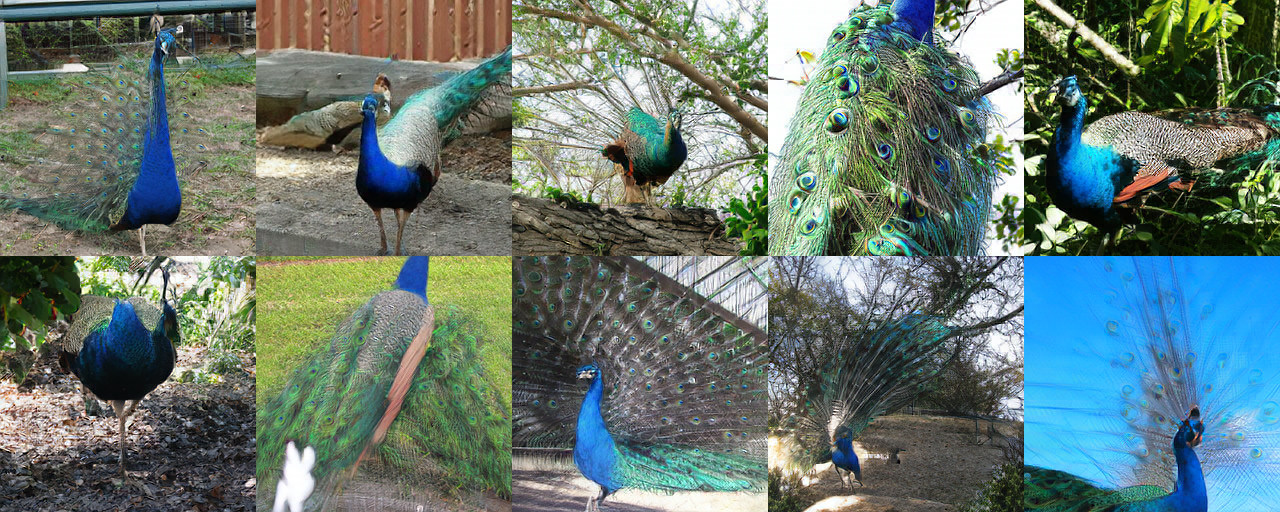}
\end{minipage}\hspace{0.025\textwidth}%
\begin{minipage}[t]{0.41\textwidth}\centering
{\scriptsize class 88: macaw\par}\vspace{1pt}
\includegraphics[width=\linewidth]{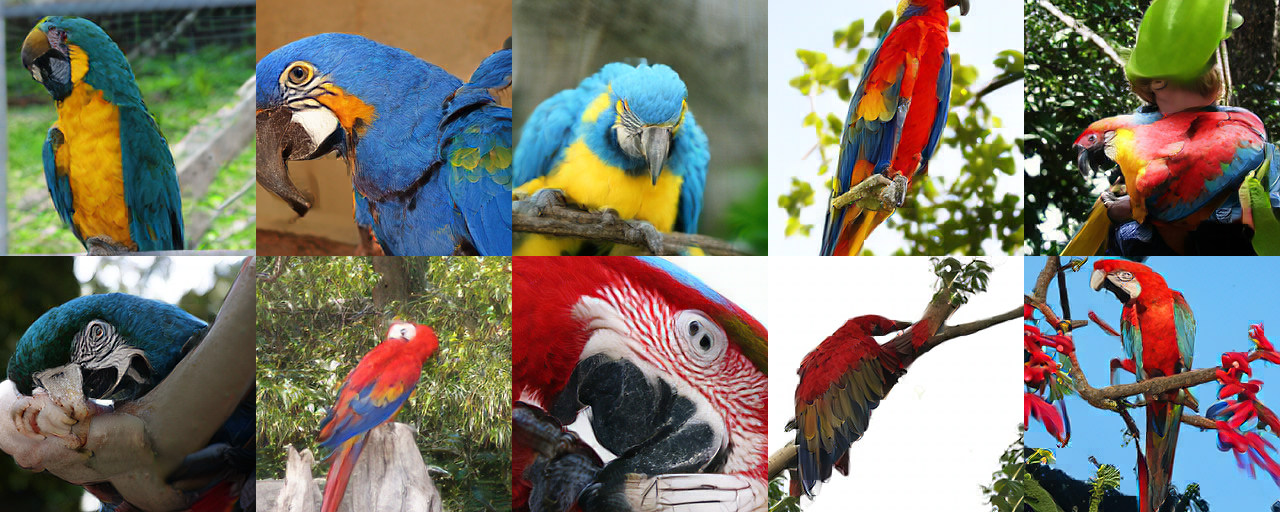}
\end{minipage}%
}\par\vspace{2pt}
\noindent\makebox[\textwidth][c]{%
\begin{minipage}[t]{0.41\textwidth}\centering
{\scriptsize class 90: lorikeet\par}\vspace{1pt}
\includegraphics[width=\linewidth]{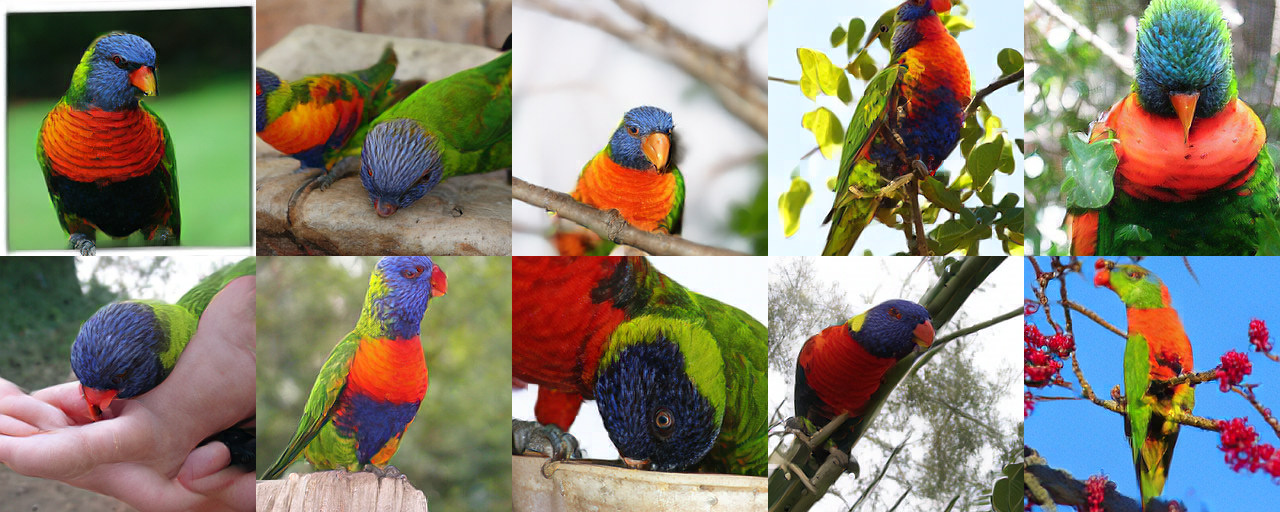}
\end{minipage}\hspace{0.025\textwidth}%
\begin{minipage}[t]{0.41\textwidth}\centering
{\scriptsize class 92: bee eater\par}\vspace{1pt}
\includegraphics[width=\linewidth]{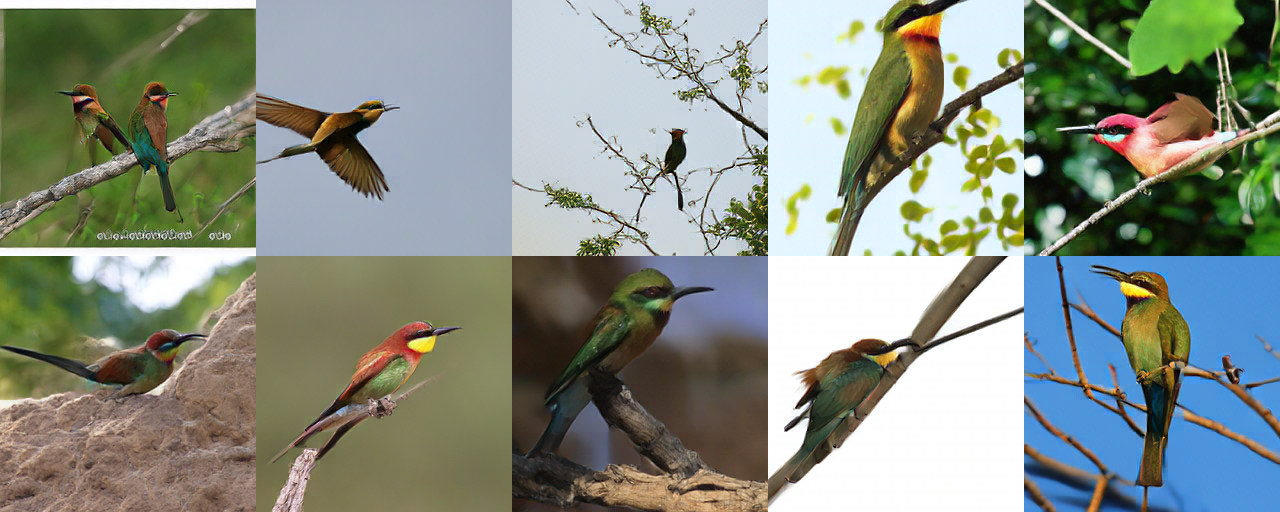}
\end{minipage}%
}\par\vspace{2pt}
\noindent\makebox[\textwidth][c]{%
\begin{minipage}[t]{0.41\textwidth}\centering
{\scriptsize class 94: hummingbird\par}\vspace{1pt}
\includegraphics[width=\linewidth]{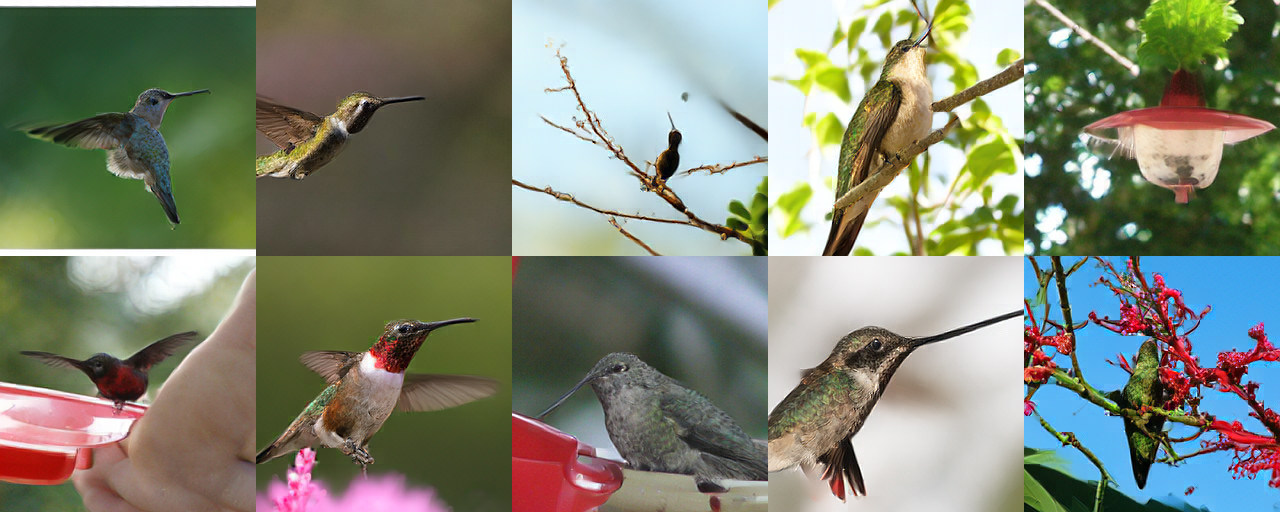}
\end{minipage}\hspace{0.025\textwidth}%
\begin{minipage}[t]{0.41\textwidth}\centering
{\scriptsize class 95: jacamar\par}\vspace{1pt}
\includegraphics[width=\linewidth]{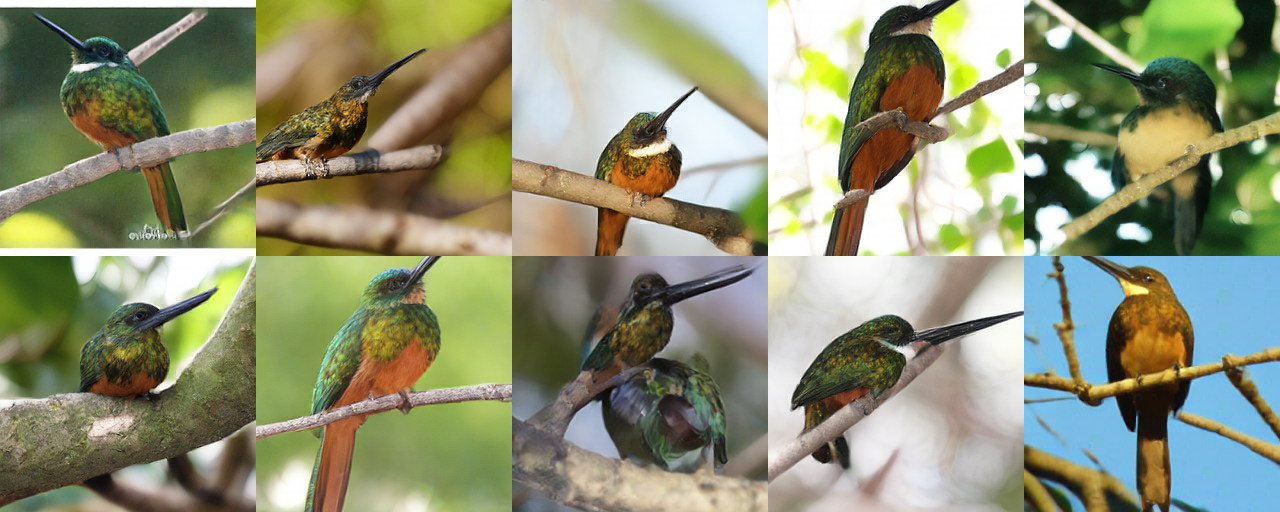}
\end{minipage}%
}\par\vspace{2pt}
\noindent\makebox[\textwidth][c]{%
\begin{minipage}[t]{0.41\textwidth}\centering
{\scriptsize class 96: toucan\par}\vspace{1pt}
\includegraphics[width=\linewidth]{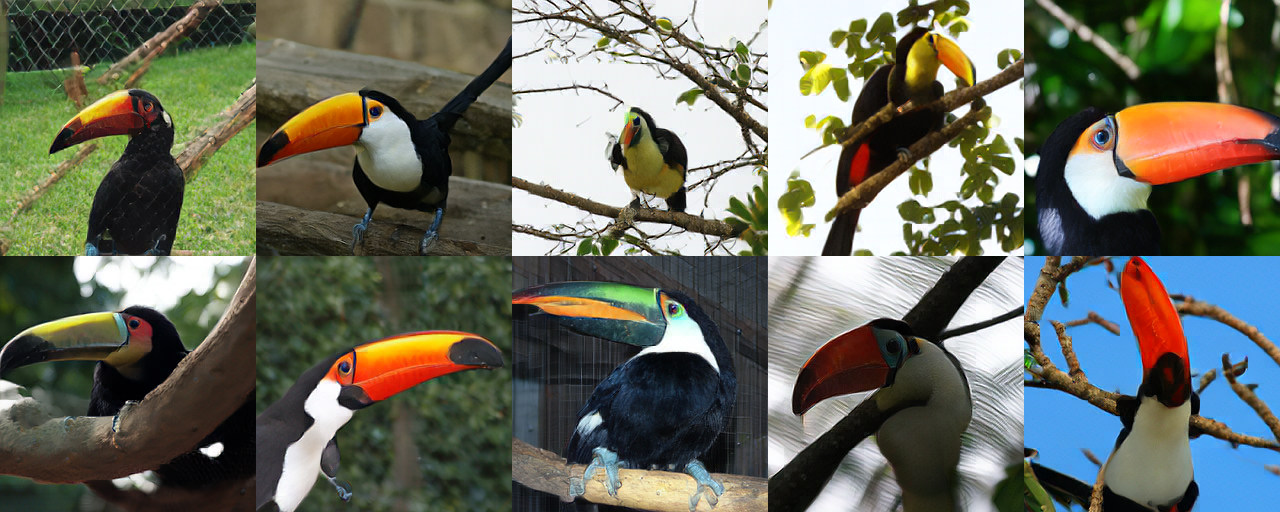}
\end{minipage}\hspace{0.025\textwidth}%
\begin{minipage}[t]{0.41\textwidth}\centering
{\scriptsize class 130: flamingo\par}\vspace{1pt}
\includegraphics[width=\linewidth]{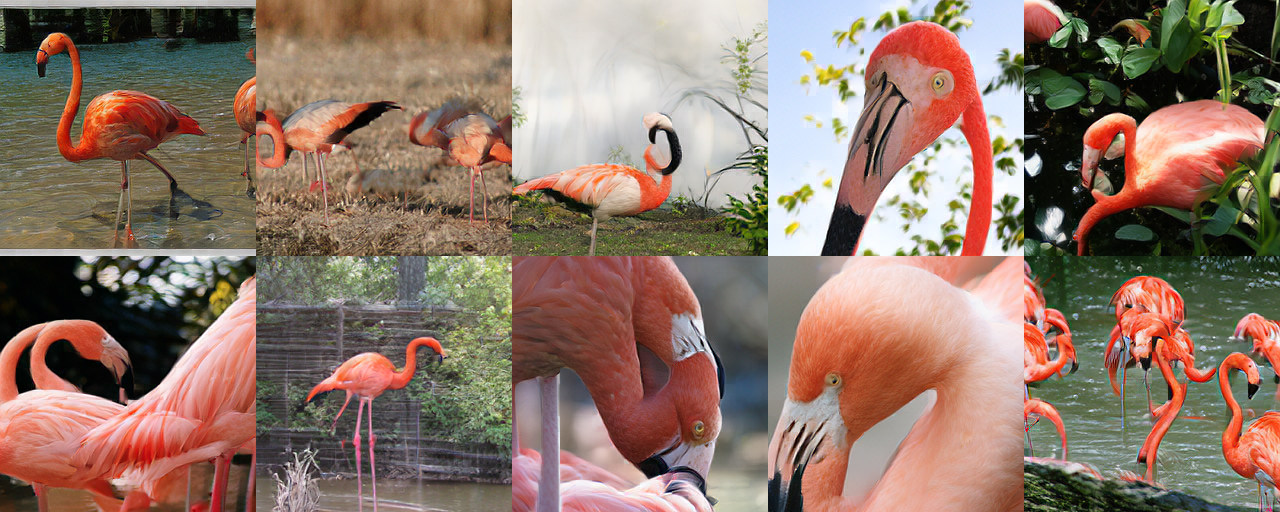}
\end{minipage}%
}\par\vspace{2pt}
\noindent\makebox[\textwidth][c]{%
\begin{minipage}[t]{0.41\textwidth}\centering
{\scriptsize class 145: king penguin\par}\vspace{1pt}
\includegraphics[width=\linewidth]{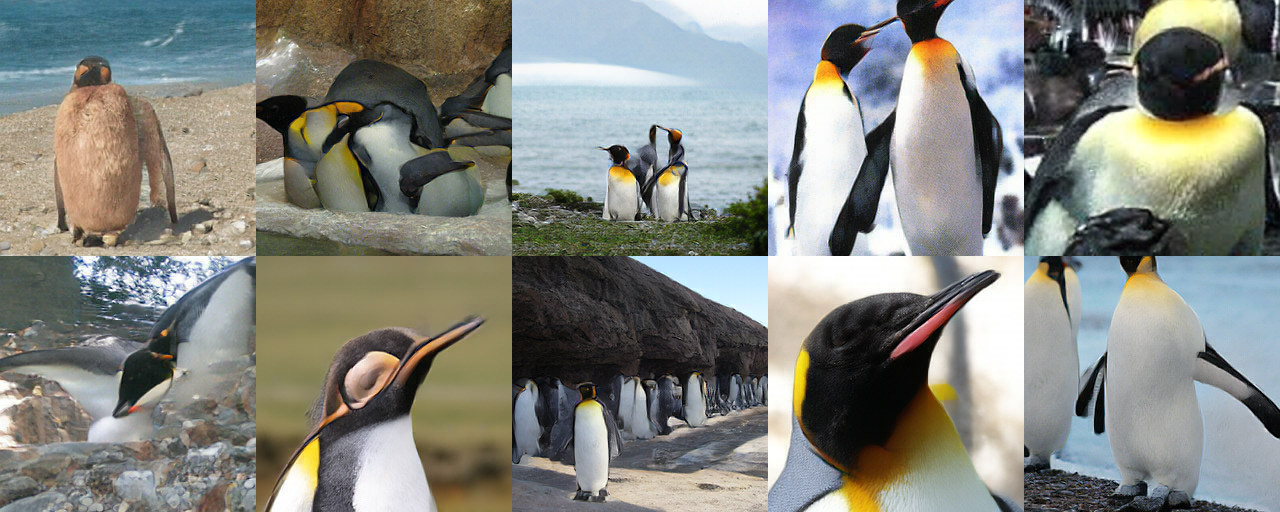}
\end{minipage}\hspace{0.025\textwidth}%
\begin{minipage}[t]{0.41\textwidth}\centering
{\scriptsize class 207: golden retriever\par}\vspace{1pt}
\includegraphics[width=\linewidth]{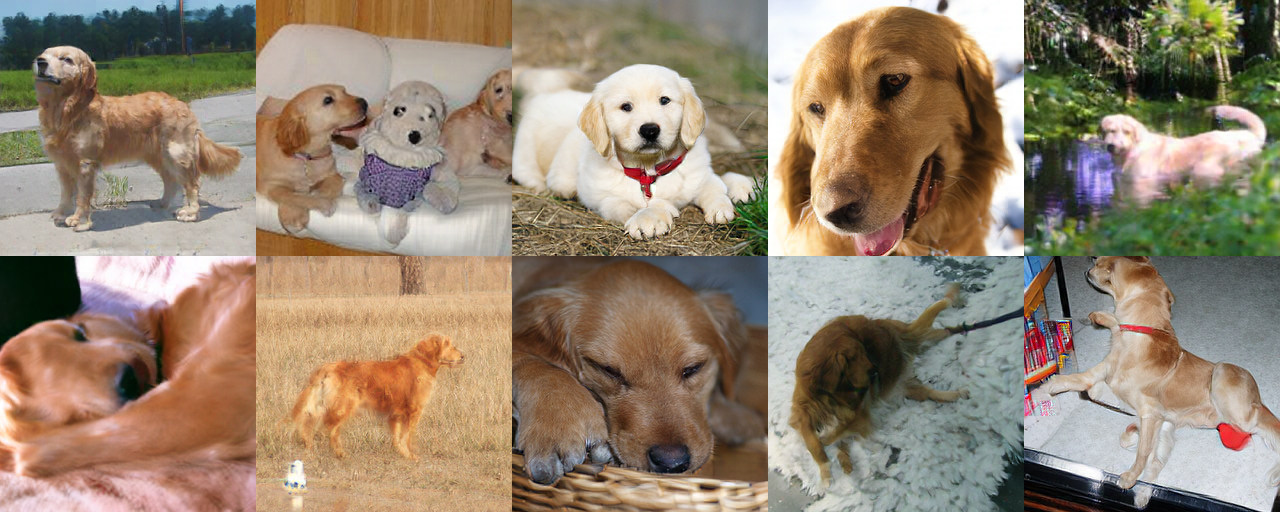}
\end{minipage}%
}\par\vspace{2pt}
\noindent\makebox[\textwidth][c]{%
\begin{minipage}[t]{0.41\textwidth}\centering
{\scriptsize class 208: Labrador retriever\par}\vspace{1pt}
\includegraphics[width=\linewidth]{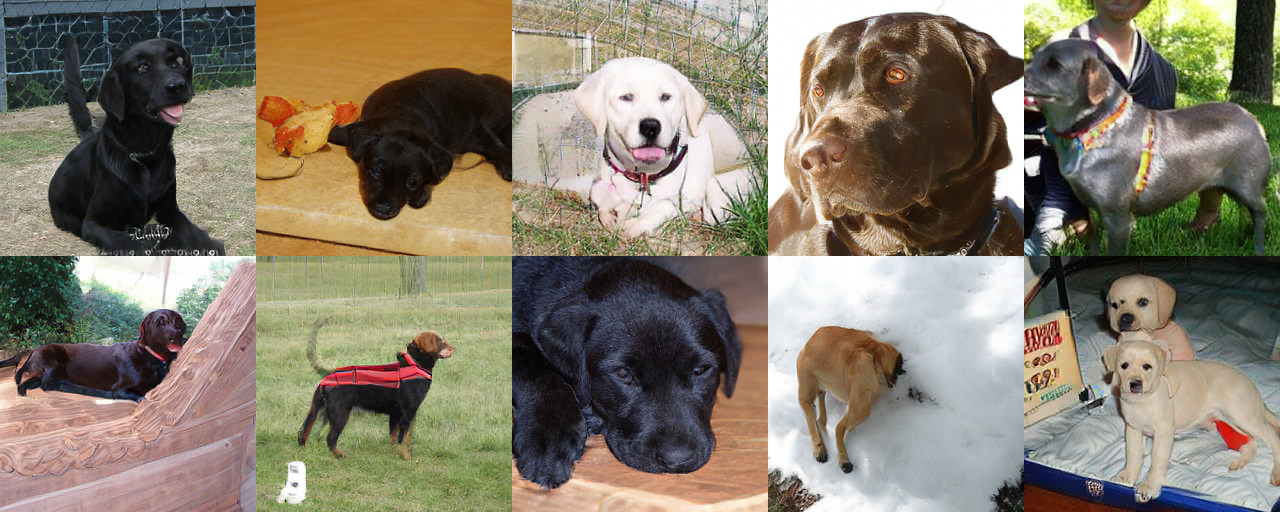}
\end{minipage}\hspace{0.025\textwidth}%
\begin{minipage}[t]{0.41\textwidth}\centering
{\scriptsize class 250: Siberian husky\par}\vspace{1pt}
\includegraphics[width=\linewidth]{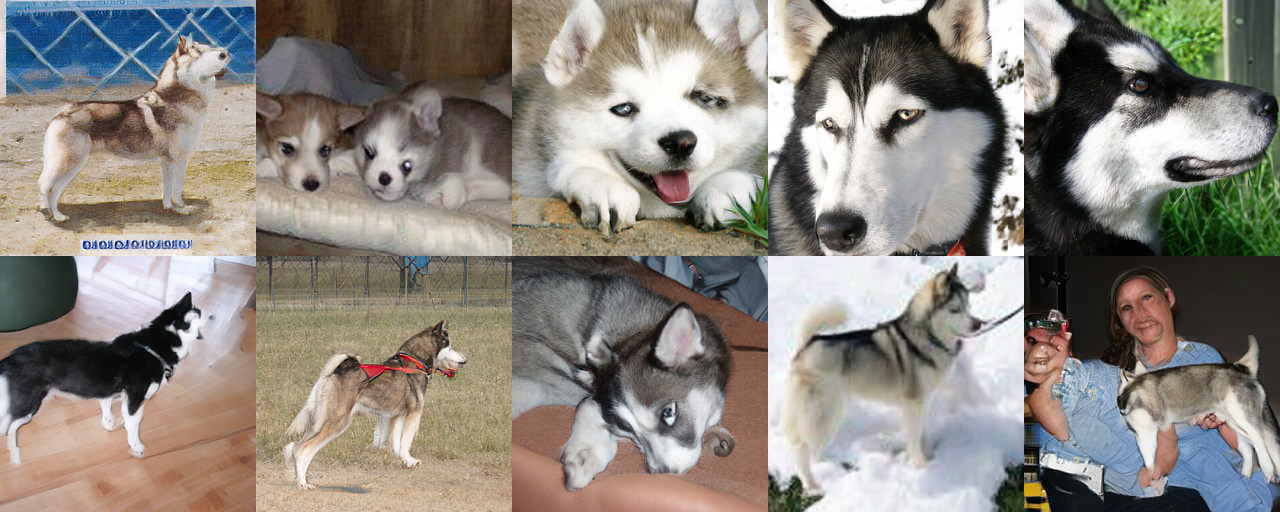}
\end{minipage}%
}\par\vspace{2pt}
\noindent\makebox[\textwidth][c]{%
\begin{minipage}[t]{0.41\textwidth}\centering
{\scriptsize class 270: white wolf\par}\vspace{1pt}
\includegraphics[width=\linewidth]{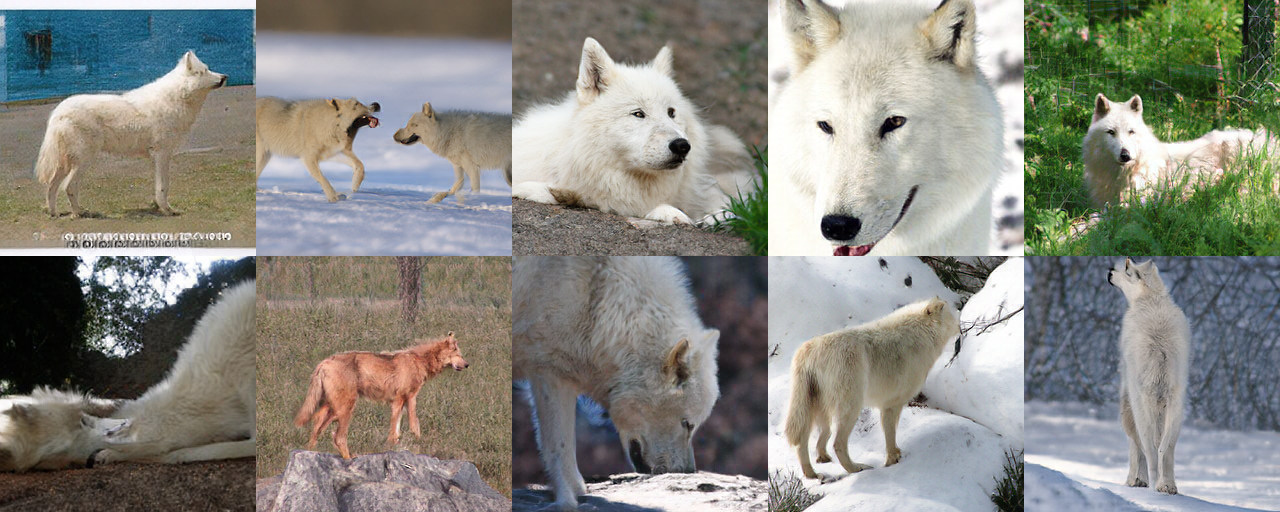}
\end{minipage}\hspace{0.025\textwidth}%
\begin{minipage}[t]{0.41\textwidth}\centering
{\scriptsize class 279: Arctic fox\par}\vspace{1pt}
\includegraphics[width=\linewidth]{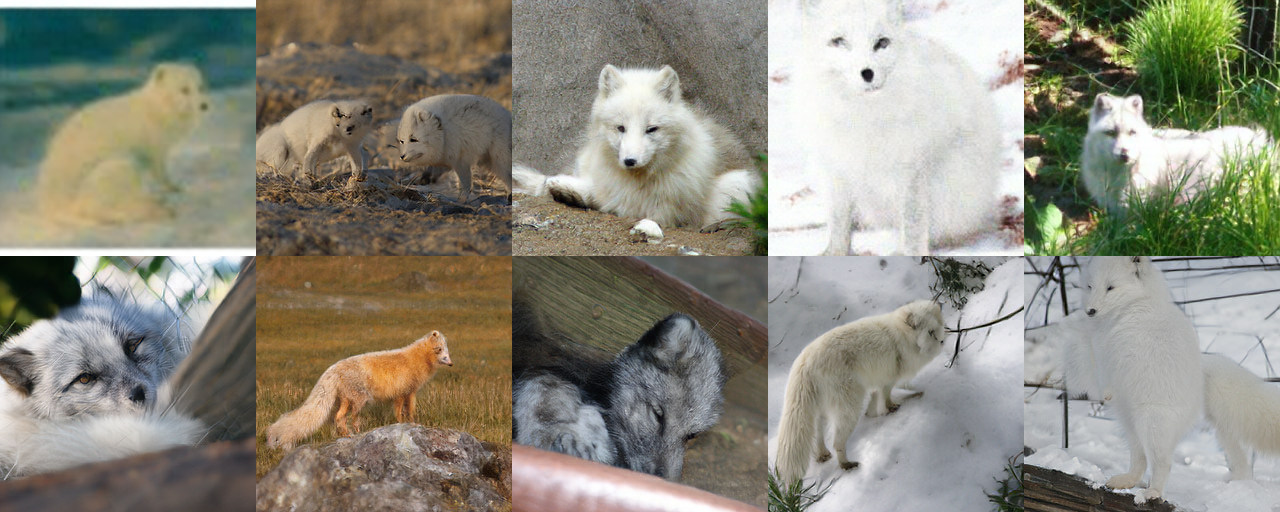}
\end{minipage}%
}\par\vspace{2pt}
\noindent\makebox[\textwidth][c]{%
\begin{minipage}[t]{0.41\textwidth}\centering
{\scriptsize class 281: tabby\par}\vspace{1pt}
\includegraphics[width=\linewidth]{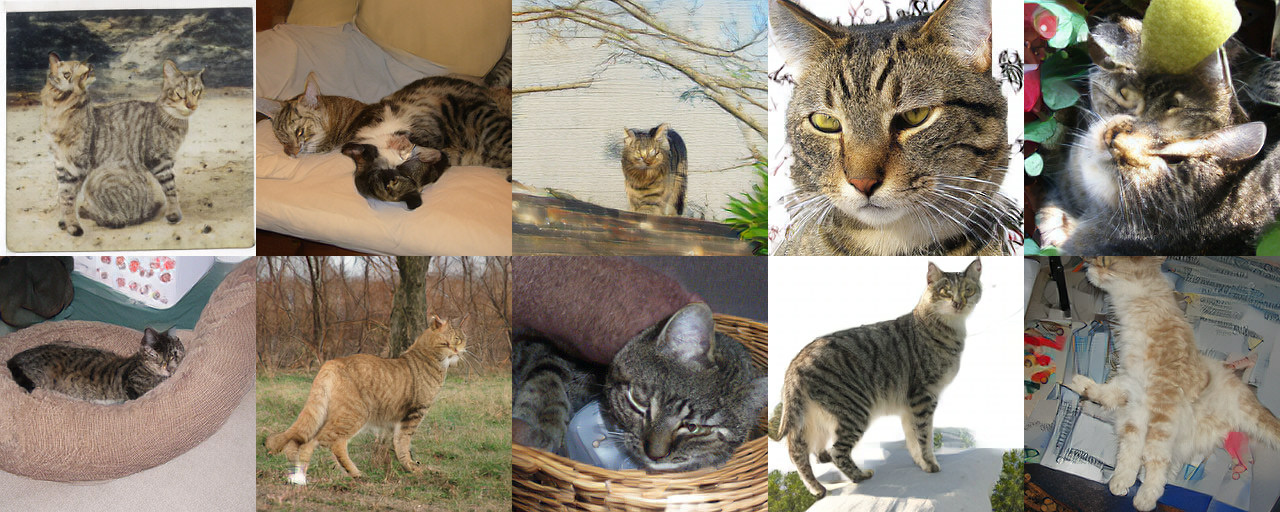}
\end{minipage}\hspace{0.025\textwidth}%
\begin{minipage}[t]{0.41\textwidth}\centering
{\scriptsize class 283: Persian cat\par}\vspace{1pt}
\includegraphics[width=\linewidth]{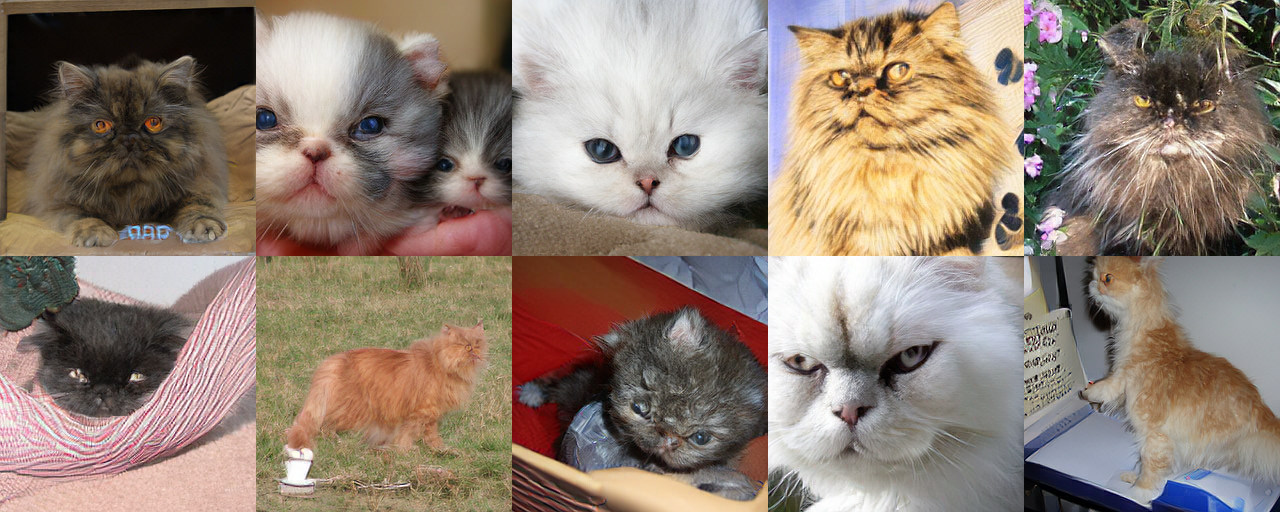}
\end{minipage}%
}\par\vspace{2pt}
\noindent\makebox[\textwidth][c]{%
\begin{minipage}[t]{0.41\textwidth}\centering
{\scriptsize class 285: Egyptian cat\par}\vspace{1pt}
\includegraphics[width=\linewidth]{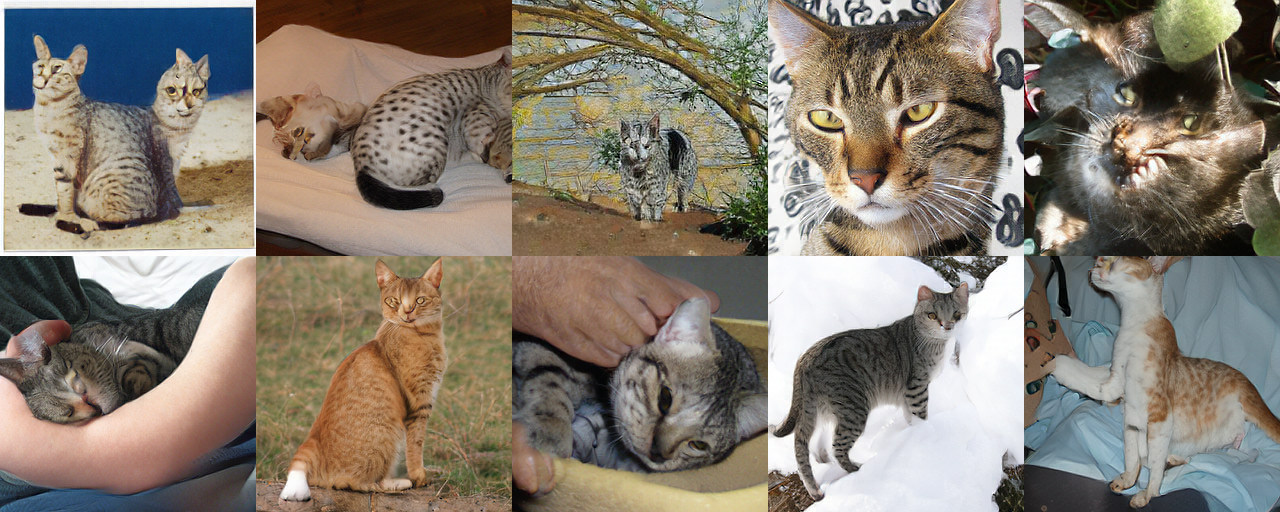}
\end{minipage}\hspace{0.025\textwidth}%
\begin{minipage}[t]{0.41\textwidth}\centering
{\scriptsize class 288: leopard\par}\vspace{1pt}
\includegraphics[width=\linewidth]{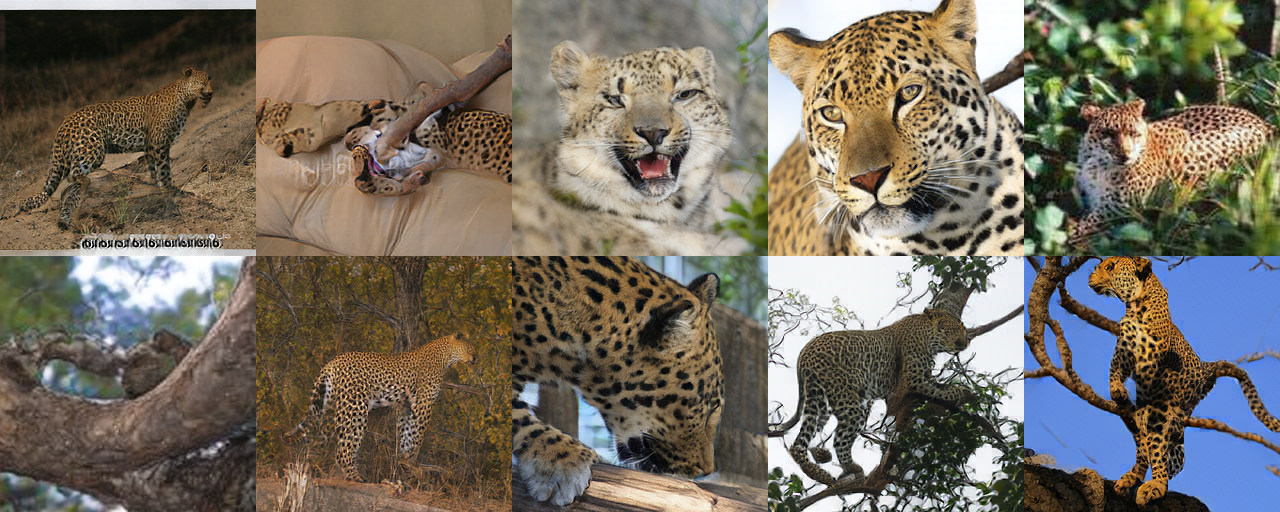}
\end{minipage}%
}\par\vspace{2pt}
\noindent\makebox[\textwidth][c]{%
\begin{minipage}[t]{0.41\textwidth}\centering
{\scriptsize class 291: lion\par}\vspace{1pt}
\includegraphics[width=\linewidth]{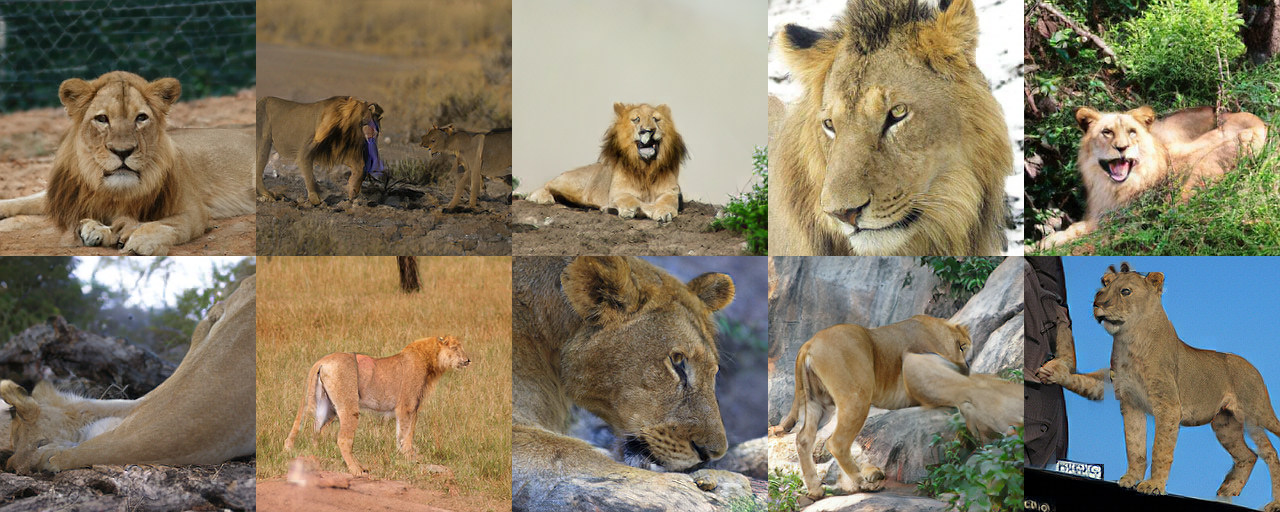}
\end{minipage}\hspace{0.025\textwidth}%
\begin{minipage}[t]{0.41\textwidth}\centering
{\scriptsize class 292: tiger\par}\vspace{1pt}
\includegraphics[width=\linewidth]{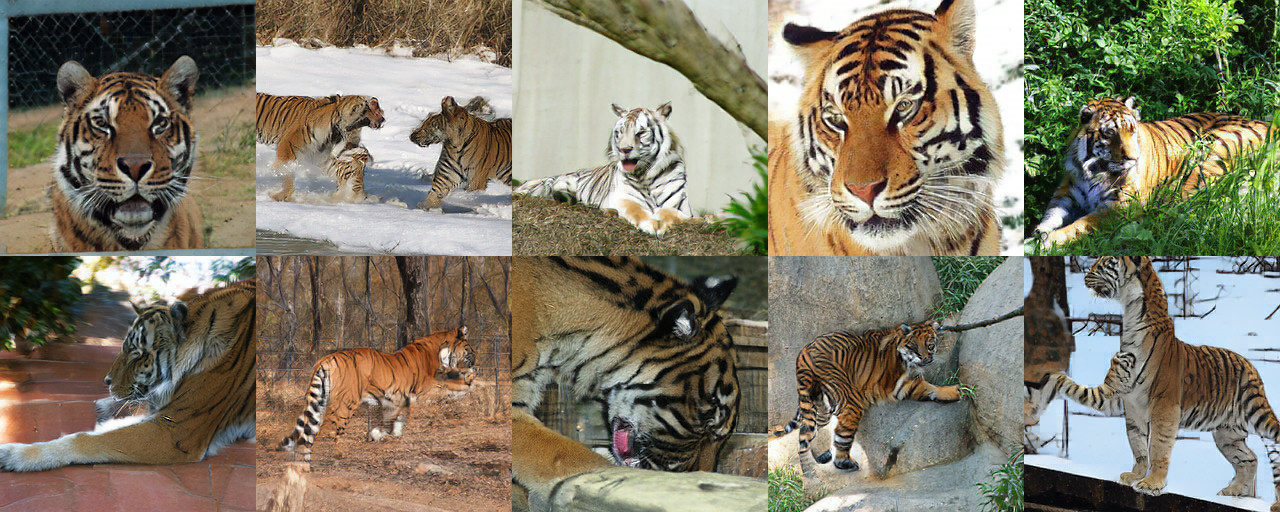}
\end{minipage}%
}\par\vspace{2pt}
\noindent\makebox[\textwidth][c]{%
\begin{minipage}[t]{0.41\textwidth}\centering
{\scriptsize class 296: ice bear\par}\vspace{1pt}
\includegraphics[width=\linewidth]{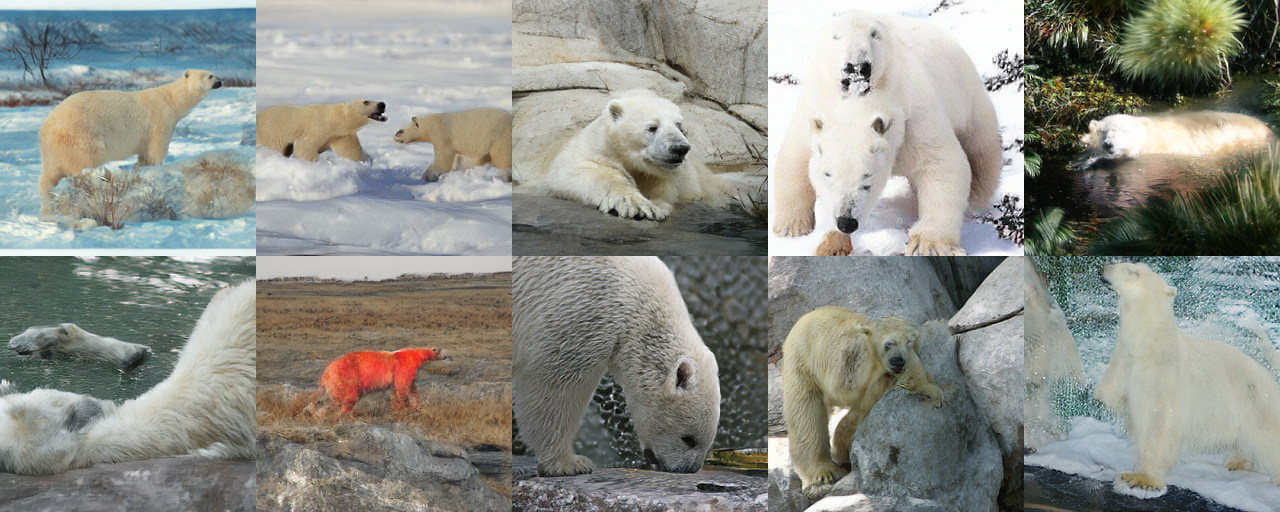}
\end{minipage}\hspace{0.025\textwidth}%
\begin{minipage}[t]{0.41\textwidth}\centering
{\scriptsize class 340: zebra\par}\vspace{1pt}
\includegraphics[width=\linewidth]{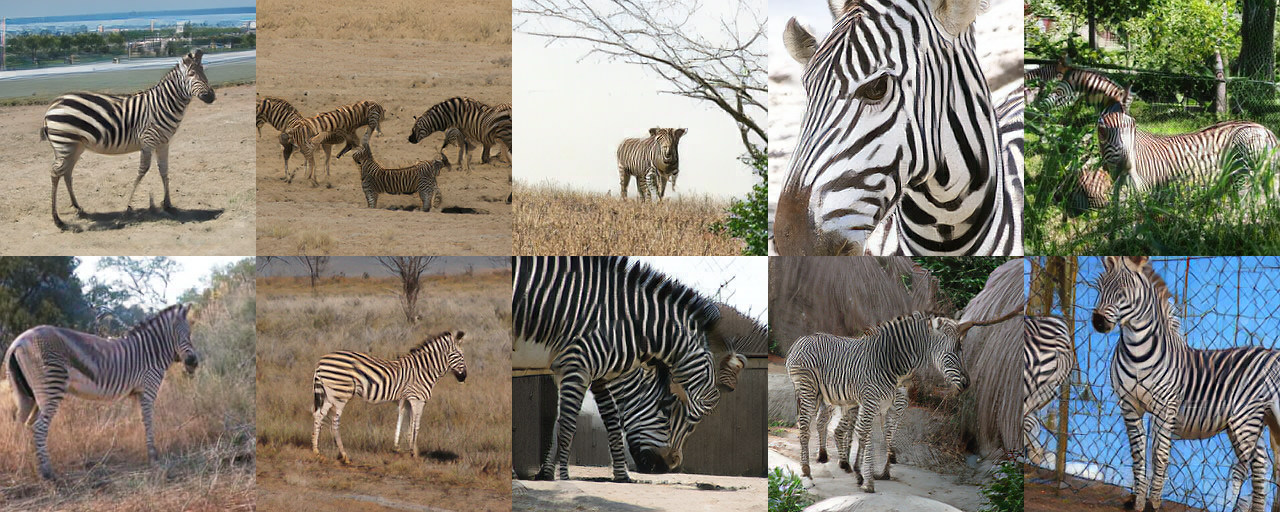}
\end{minipage}%
}\par\vspace{2pt}
\noindent\makebox[\textwidth][c]{%
\begin{minipage}[t]{0.41\textwidth}\centering
{\scriptsize class 349: bighorn\par}\vspace{1pt}
\includegraphics[width=\linewidth]{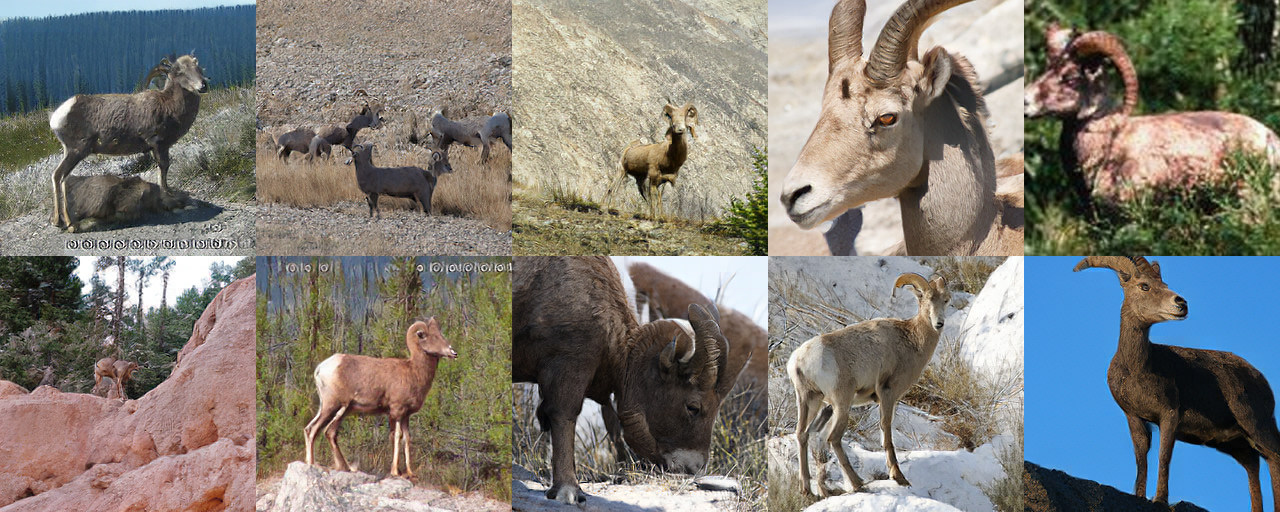}
\end{minipage}\hspace{0.025\textwidth}%
\begin{minipage}[t]{0.41\textwidth}\centering
{\scriptsize class 360: otter\par}\vspace{1pt}
\includegraphics[width=\linewidth]{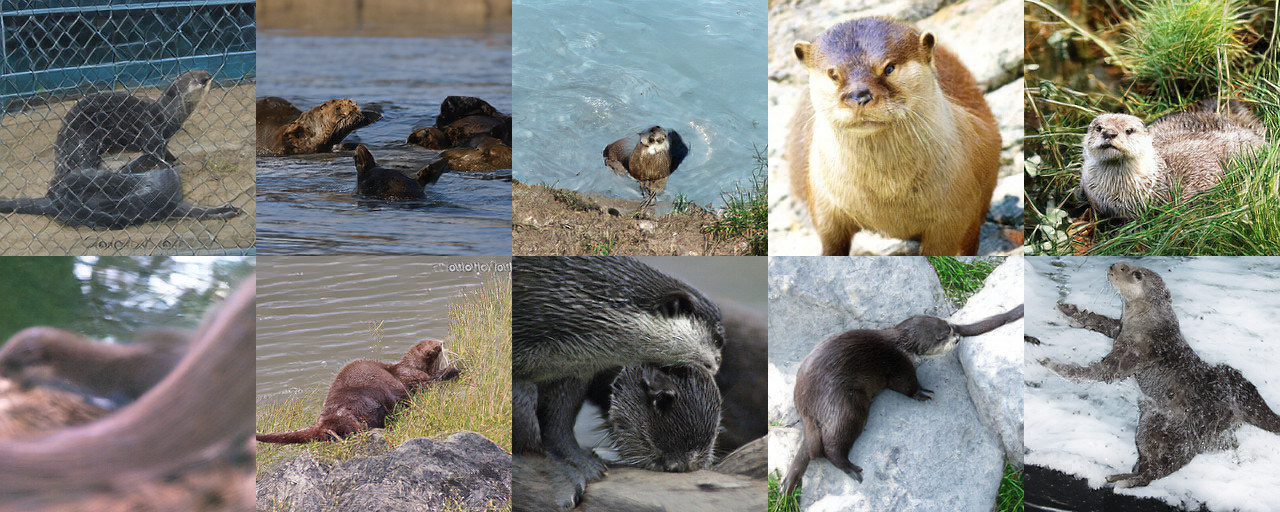}
\end{minipage}%
}\par\vspace{2pt}
\noindent\makebox[\textwidth][c]{%
\begin{minipage}[t]{0.41\textwidth}\centering
{\scriptsize class 386: African elephant\par}\vspace{1pt}
\includegraphics[width=\linewidth]{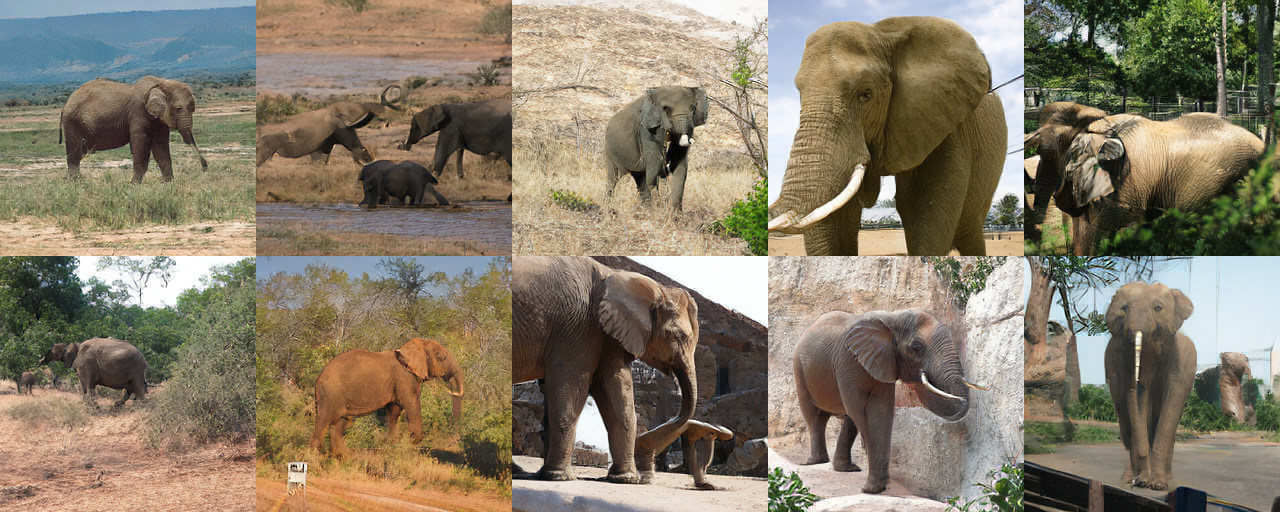}
\end{minipage}\hspace{0.025\textwidth}%
\begin{minipage}[t]{0.41\textwidth}\centering
{\scriptsize class 387: lesser panda\par}\vspace{1pt}
\includegraphics[width=\linewidth]{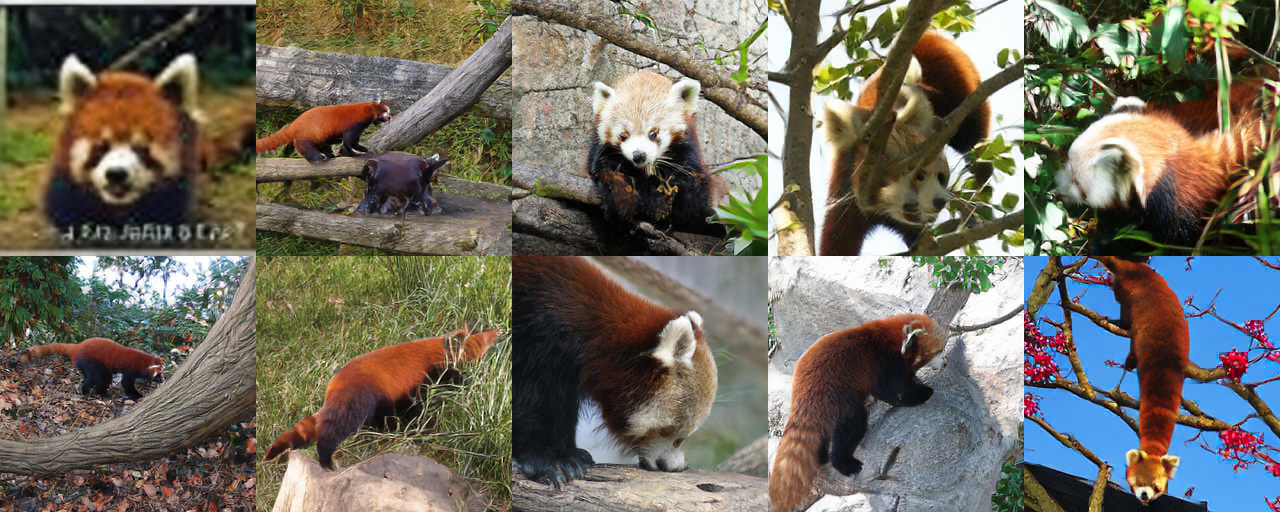}
\end{minipage}%
}\par\vspace{2pt}
\noindent\makebox[\textwidth][c]{%
\begin{minipage}[t]{0.41\textwidth}\centering
{\scriptsize class 388: giant panda\par}\vspace{1pt}
\includegraphics[width=\linewidth]{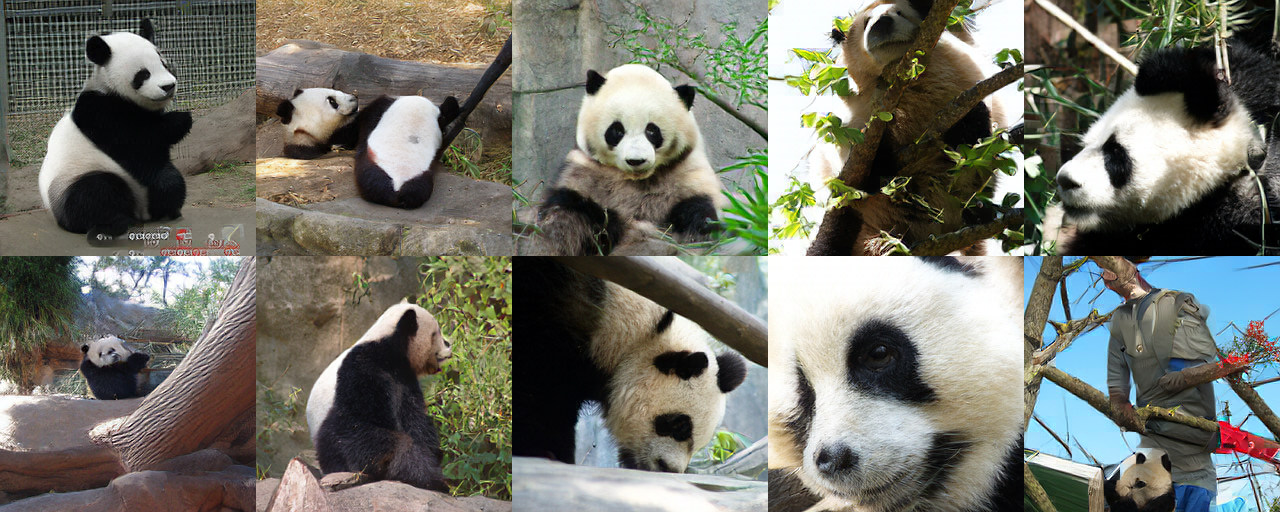}
\end{minipage}\hspace{0.025\textwidth}%
\begin{minipage}[t]{0.41\textwidth}\centering
{\scriptsize class 1: goldfish\par}\vspace{1pt}
\includegraphics[width=\linewidth]{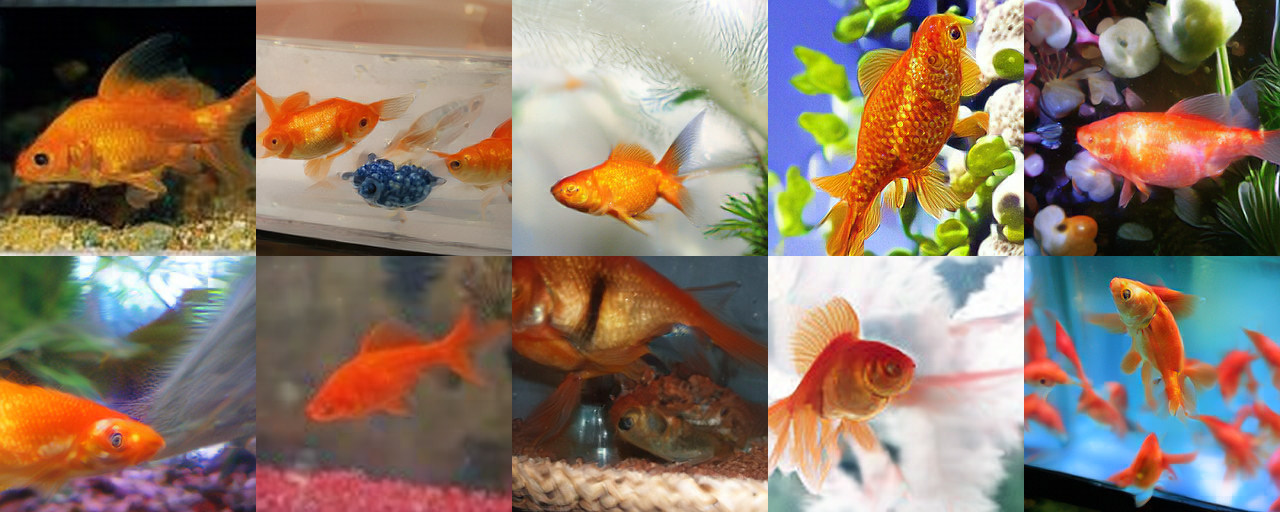}
\end{minipage}%
}\par\vspace{2pt}
\noindent\makebox[\textwidth][c]{%
\begin{minipage}[t]{0.41\textwidth}\centering
{\scriptsize class 31: tree frog\par}\vspace{1pt}
\includegraphics[width=\linewidth]{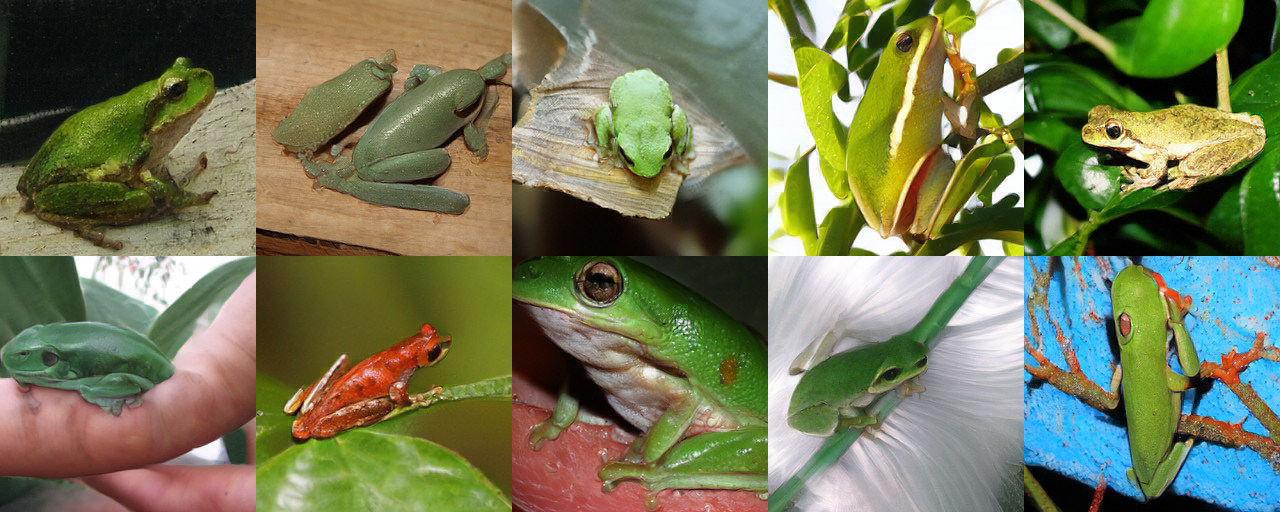}
\end{minipage}\hspace{0.025\textwidth}%
\begin{minipage}[t]{0.41\textwidth}\centering
{\scriptsize class 107: jellyfish\par}\vspace{1pt}
\includegraphics[width=\linewidth]{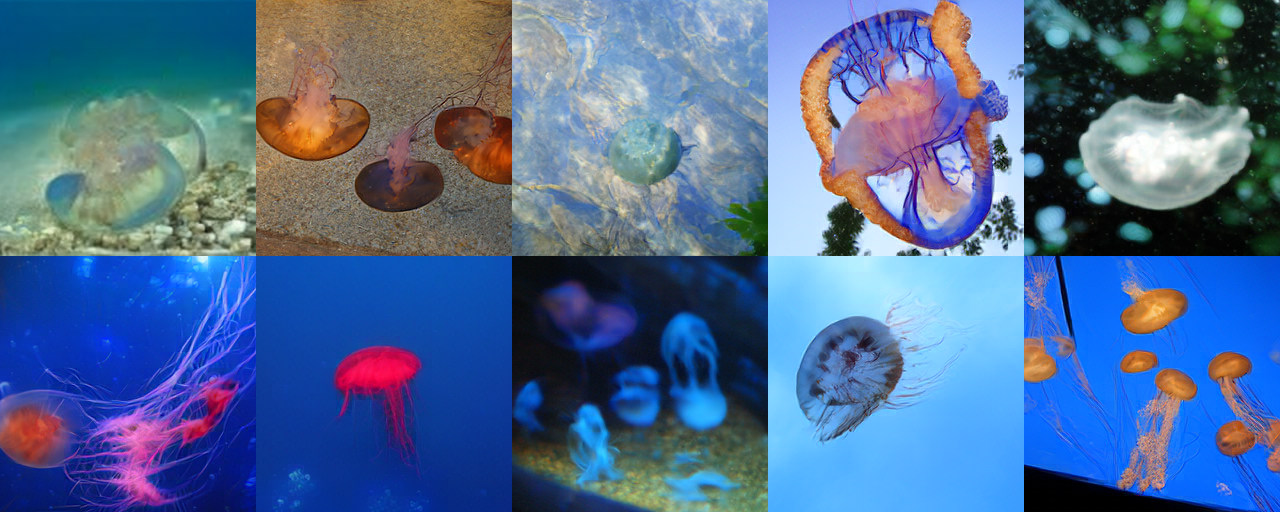}
\end{minipage}%
}\par\vspace{2pt}
\noindent\makebox[\textwidth][c]{%
\begin{minipage}[t]{0.41\textwidth}\centering
{\scriptsize class 108: sea anemone\par}\vspace{1pt}
\includegraphics[width=\linewidth]{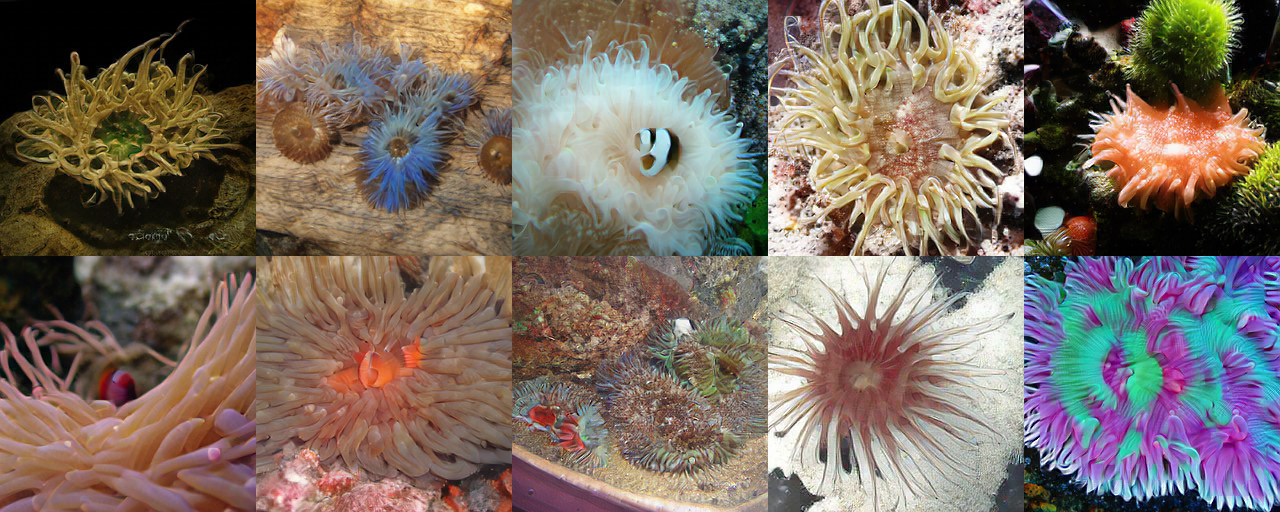}
\end{minipage}\hspace{0.025\textwidth}%
\begin{minipage}[t]{0.41\textwidth}\centering
{\scriptsize class 323: monarch\par}\vspace{1pt}
\includegraphics[width=\linewidth]{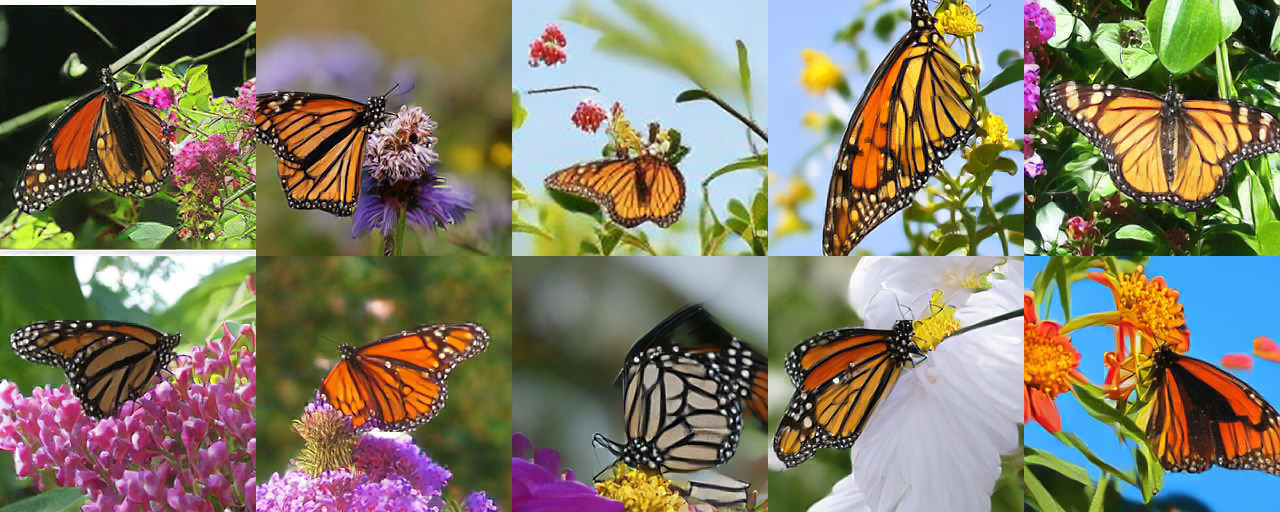}
\end{minipage}%
}\par\vspace{2pt}
\noindent\makebox[\textwidth][c]{%
\begin{minipage}[t]{0.41\textwidth}\centering
{\scriptsize class 973: coral reef\par}\vspace{1pt}
\includegraphics[width=\linewidth]{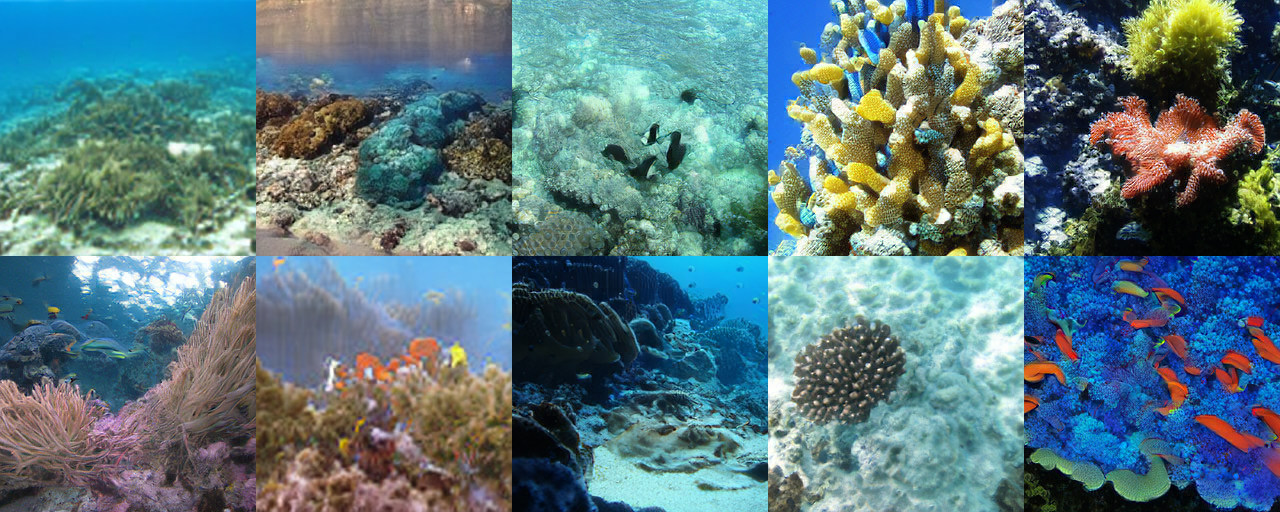}
\end{minipage}\hspace{0.025\textwidth}%
\begin{minipage}[t]{0.41\textwidth}\centering
{\scriptsize class 985: daisy\par}\vspace{1pt}
\includegraphics[width=\linewidth]{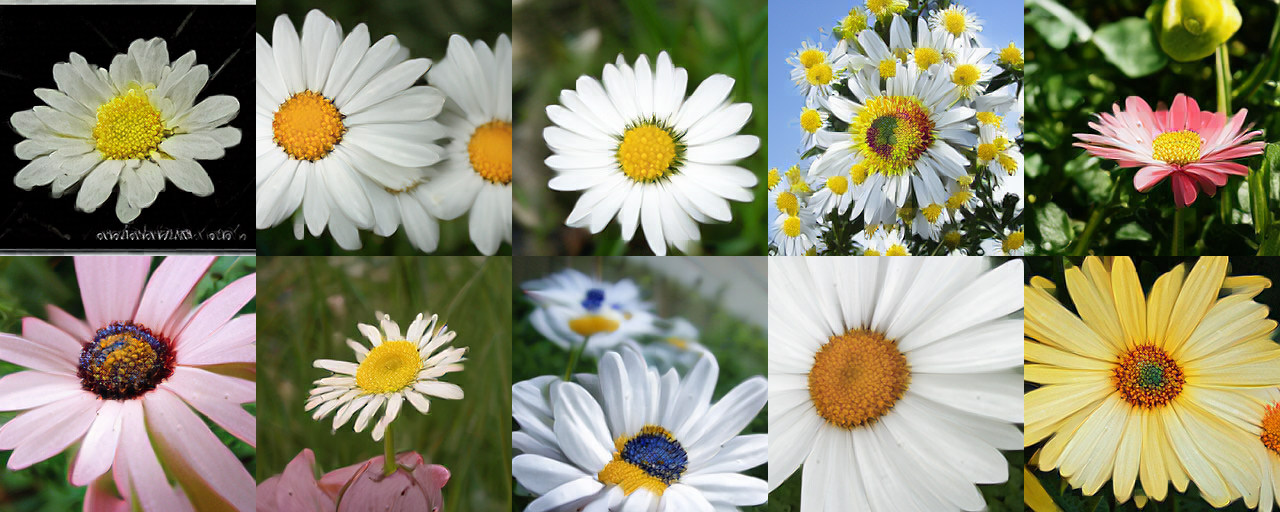}
\end{minipage}%
}\par\vspace{2pt}
\noindent\makebox[\textwidth][c]{%
\begin{minipage}[t]{0.41\textwidth}\centering
{\scriptsize class 992: agaric\par}\vspace{1pt}
\includegraphics[width=\linewidth]{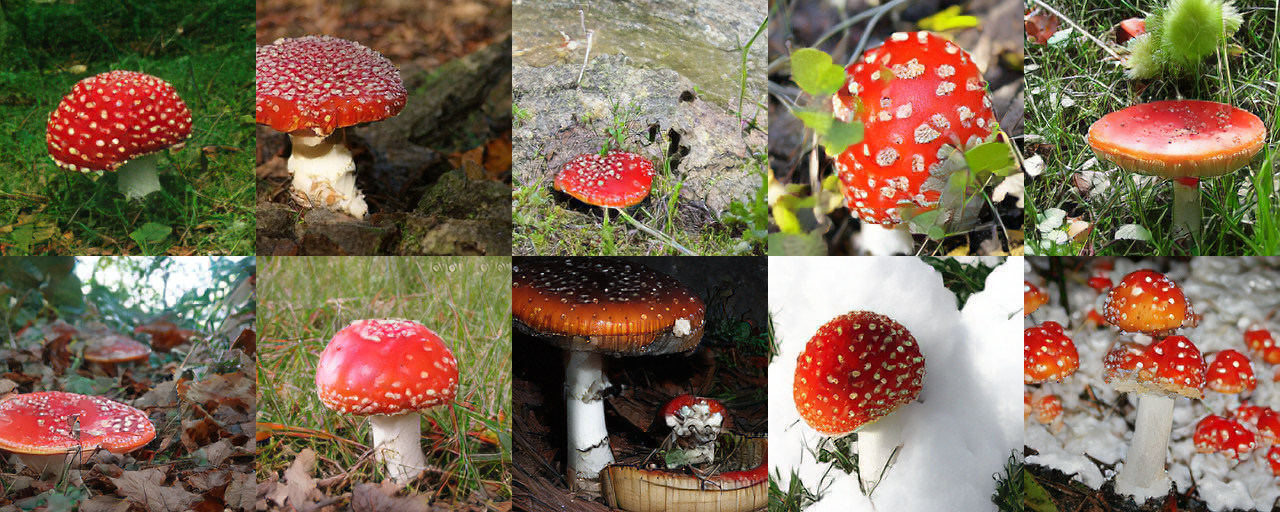}
\end{minipage}\hspace{0.025\textwidth}%
\begin{minipage}[t]{0.41\textwidth}\centering
{\scriptsize class 417: balloon\par}\vspace{1pt}
\includegraphics[width=\linewidth]{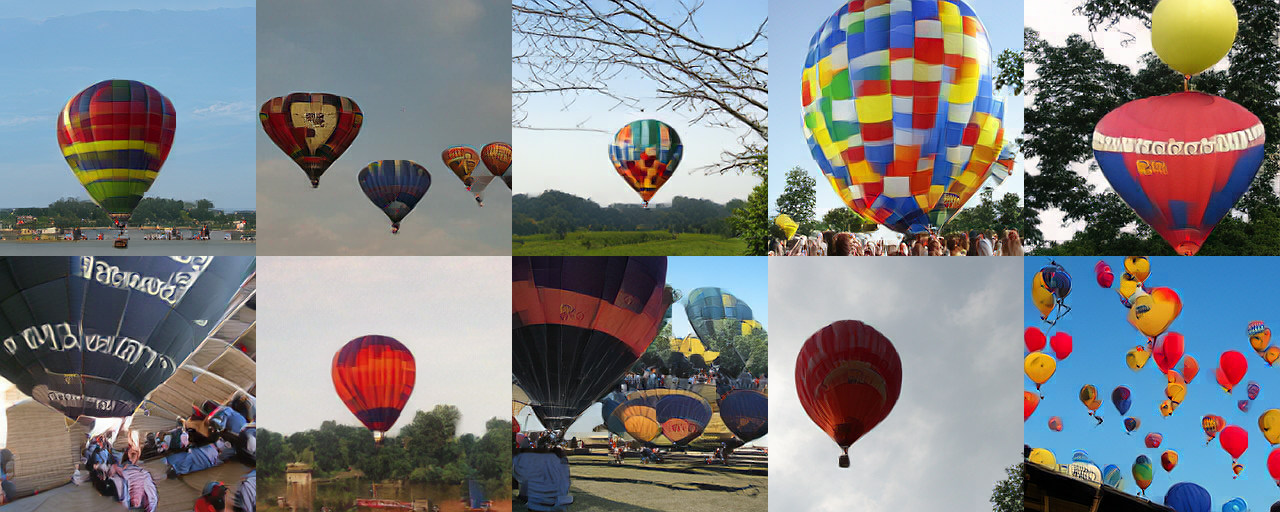}
\end{minipage}%
}\par\vspace{2pt}
\noindent\makebox[\textwidth][c]{%
\begin{minipage}[t]{0.41\textwidth}\centering
{\scriptsize class 425: barn\par}\vspace{1pt}
\includegraphics[width=\linewidth]{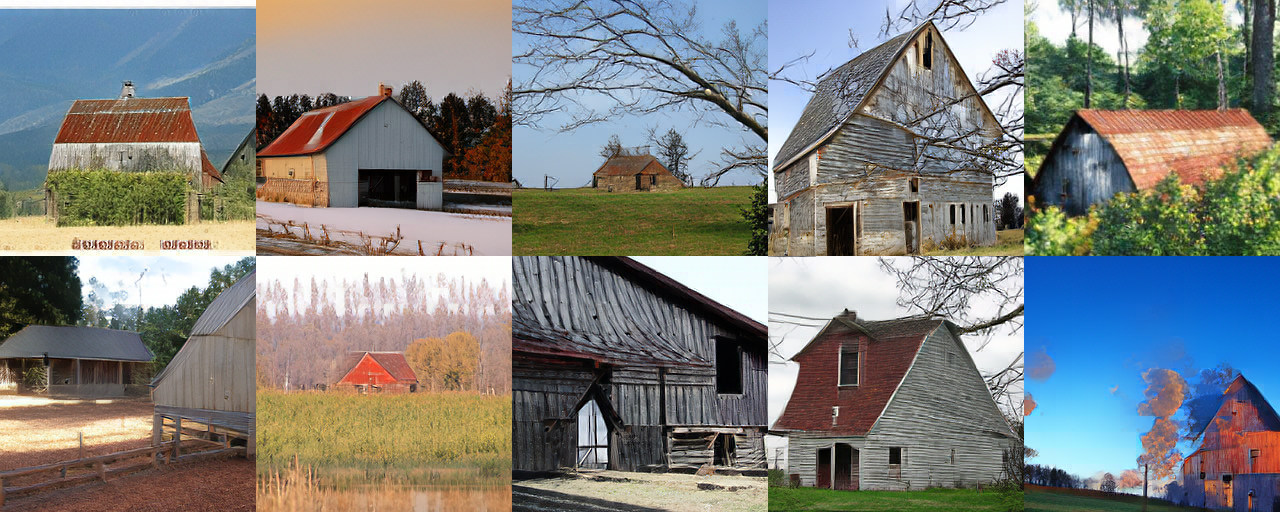}
\end{minipage}\hspace{0.025\textwidth}%
\begin{minipage}[t]{0.41\textwidth}\centering
{\scriptsize class 448: birdhouse\par}\vspace{1pt}
\includegraphics[width=\linewidth]{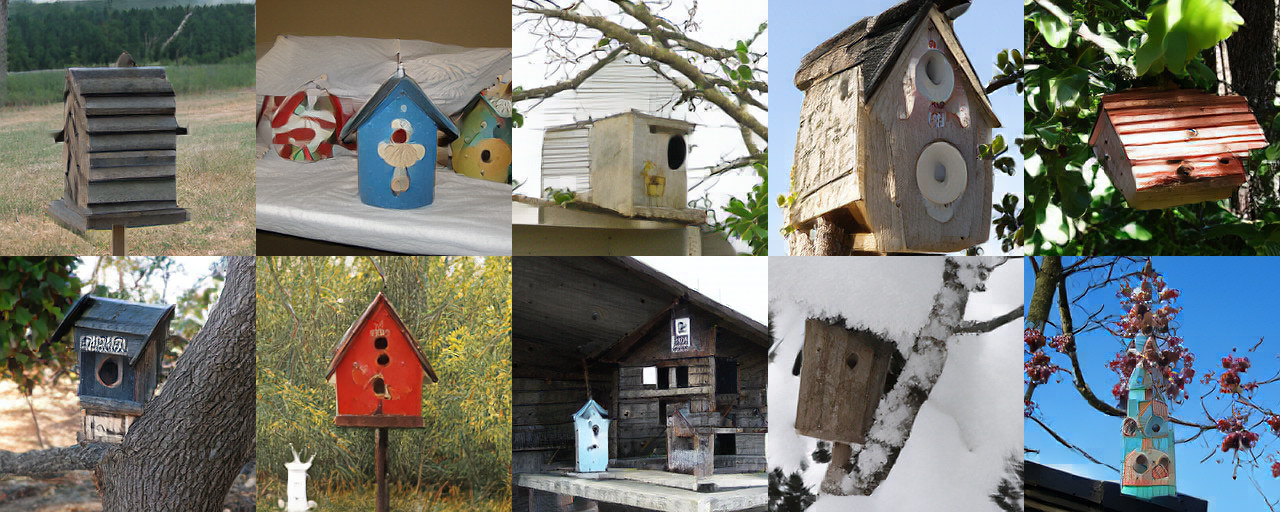}
\end{minipage}%
}\par\vspace{2pt}
\noindent\makebox[\textwidth][c]{%
\begin{minipage}[t]{0.41\textwidth}\centering
{\scriptsize class 483: castle\par}\vspace{1pt}
\includegraphics[width=\linewidth]{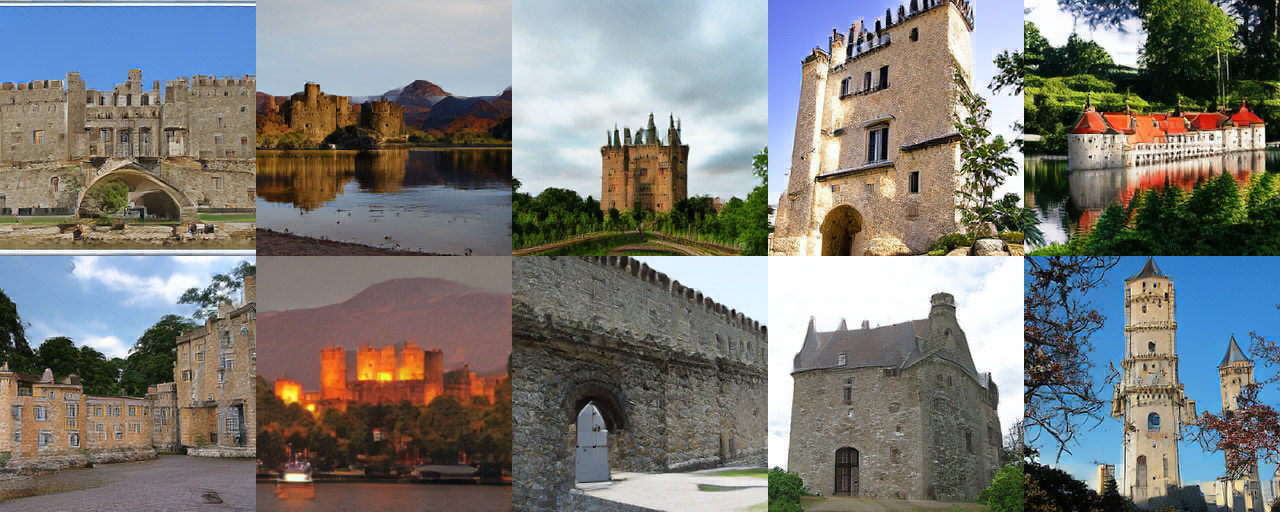}
\end{minipage}\hspace{0.025\textwidth}%
\begin{minipage}[t]{0.41\textwidth}\centering
{\scriptsize class 580: greenhouse\par}\vspace{1pt}
\includegraphics[width=\linewidth]{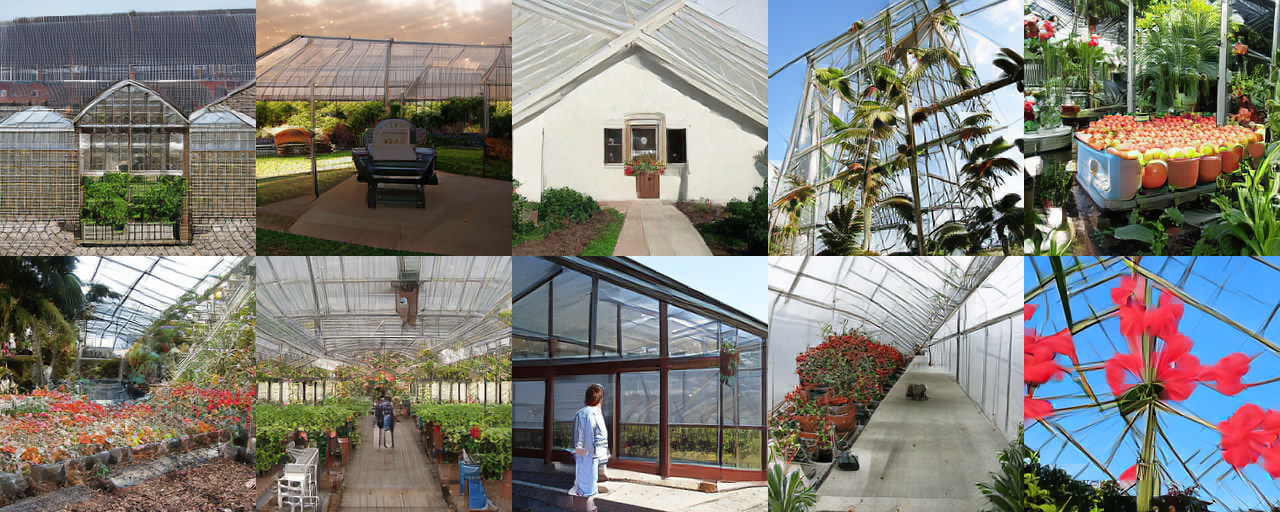}
\end{minipage}%
}\par\vspace{2pt}
\noindent\makebox[\textwidth][c]{%
\begin{minipage}[t]{0.41\textwidth}\centering
{\scriptsize class 649: megalith\par}\vspace{1pt}
\includegraphics[width=\linewidth]{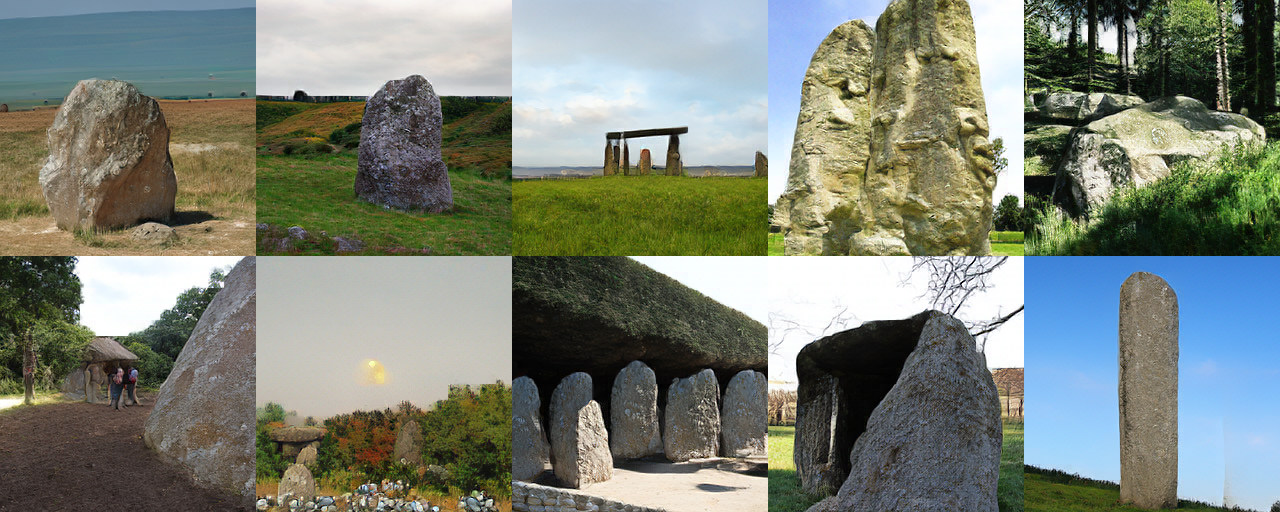}
\end{minipage}\hspace{0.025\textwidth}%
\begin{minipage}[t]{0.41\textwidth}\centering
{\scriptsize class 698: palace\par}\vspace{1pt}
\includegraphics[width=\linewidth]{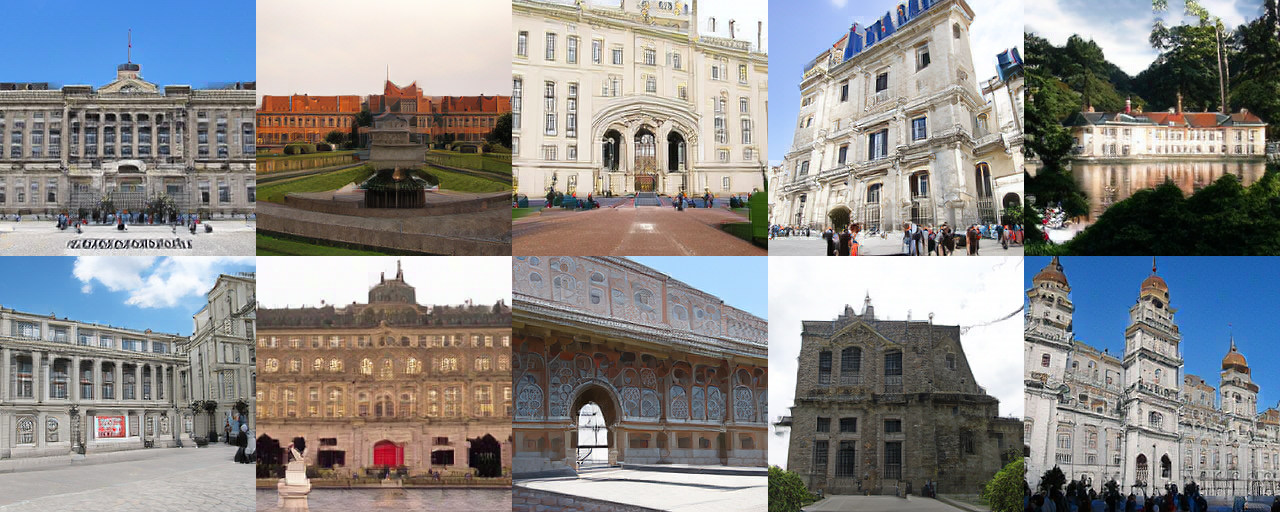}
\end{minipage}%
}\par\vspace{2pt}
\noindent\makebox[\textwidth][c]{%
\begin{minipage}[t]{0.41\textwidth}\centering
{\scriptsize class 718: pier\par}\vspace{1pt}
\includegraphics[width=\linewidth]{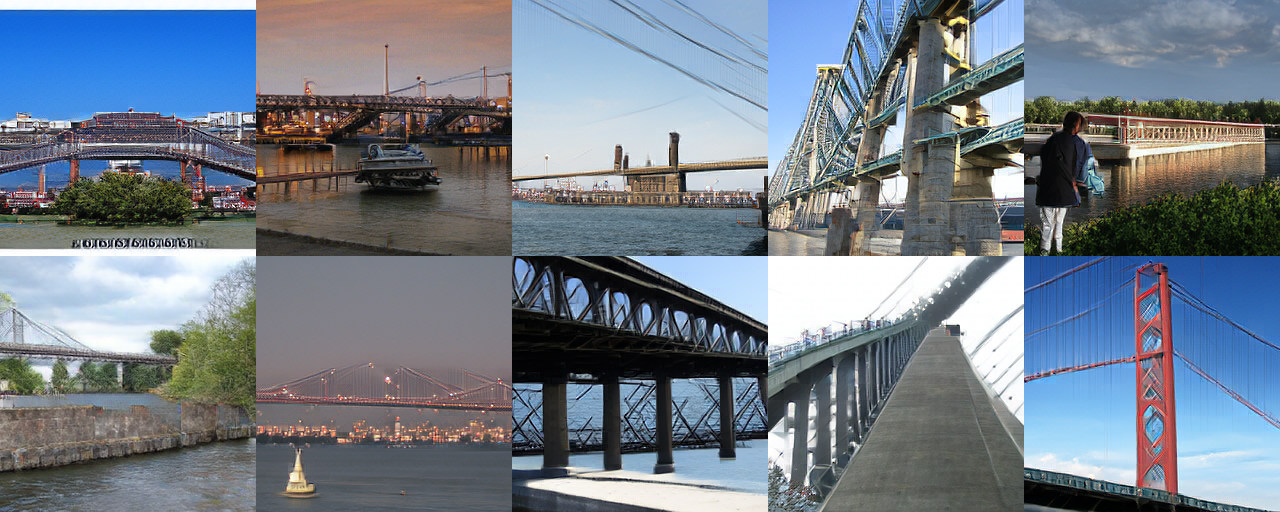}
\end{minipage}\hspace{0.025\textwidth}%
\begin{minipage}[t]{0.41\textwidth}\centering
{\scriptsize class 755: radio telescope\par}\vspace{1pt}
\includegraphics[width=\linewidth]{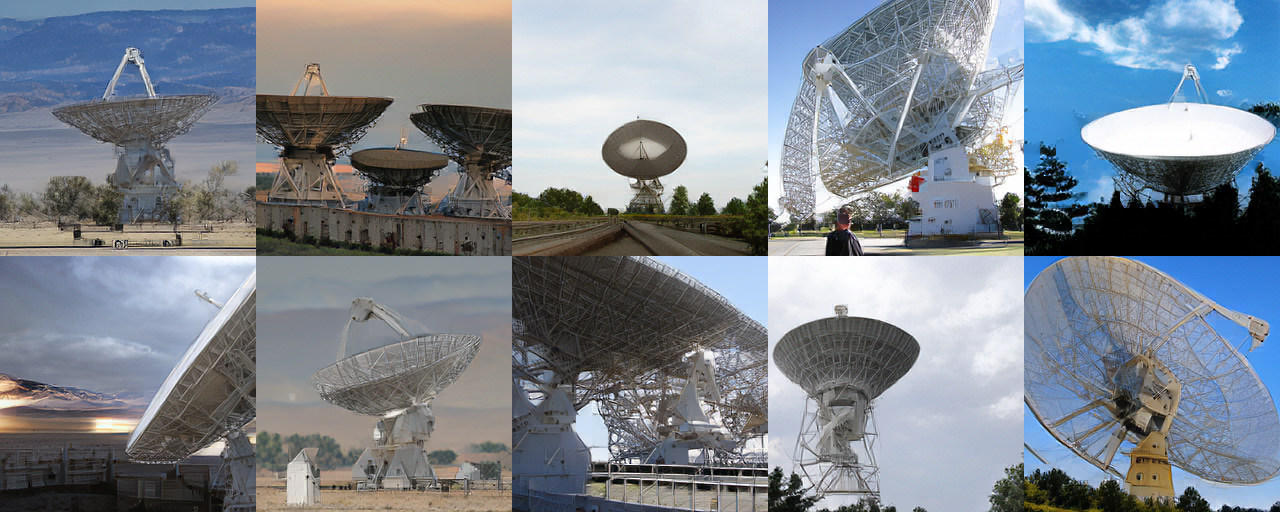}
\end{minipage}%
}\par\vspace{2pt}
\noindent\makebox[\textwidth][c]{%
\begin{minipage}[t]{0.41\textwidth}\centering
{\scriptsize class 780: schooner\par}\vspace{1pt}
\includegraphics[width=\linewidth]{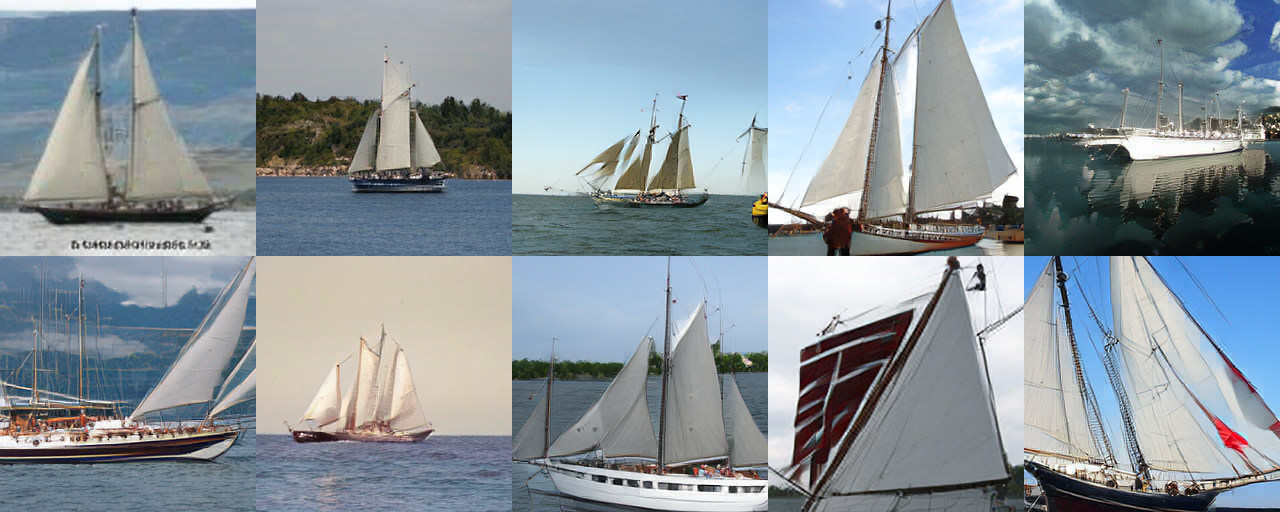}
\end{minipage}\hspace{0.025\textwidth}%
\begin{minipage}[t]{0.41\textwidth}\centering
{\scriptsize class 817: sports car\par}\vspace{1pt}
\includegraphics[width=\linewidth]{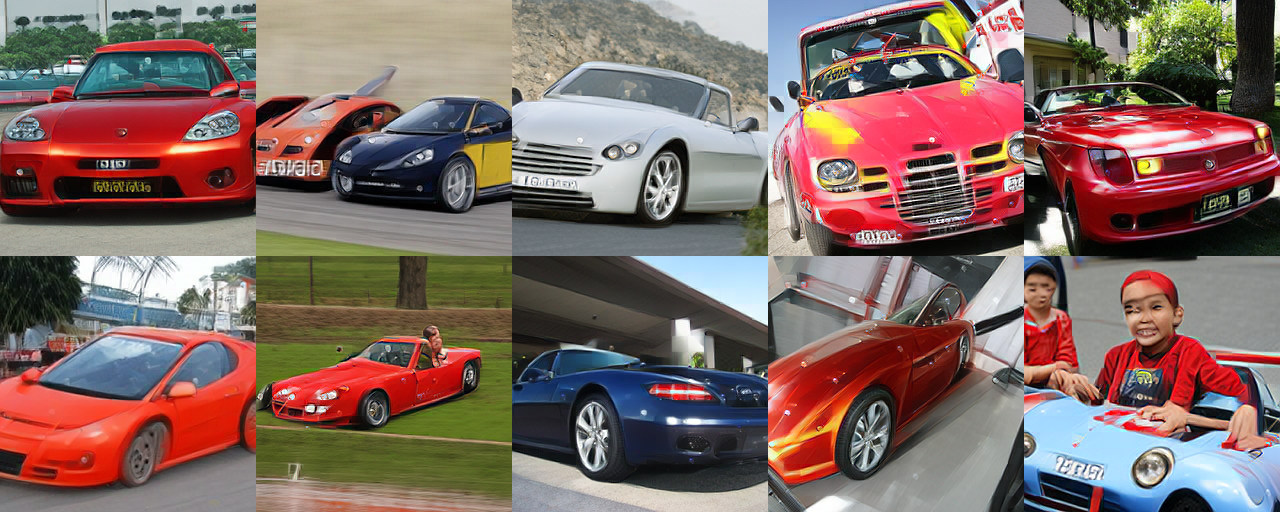}
\end{minipage}%
}\par\vspace{2pt}
\noindent\makebox[\textwidth][c]{%
\begin{minipage}[t]{0.41\textwidth}\centering
{\scriptsize class 829: streetcar\par}\vspace{1pt}
\includegraphics[width=\linewidth]{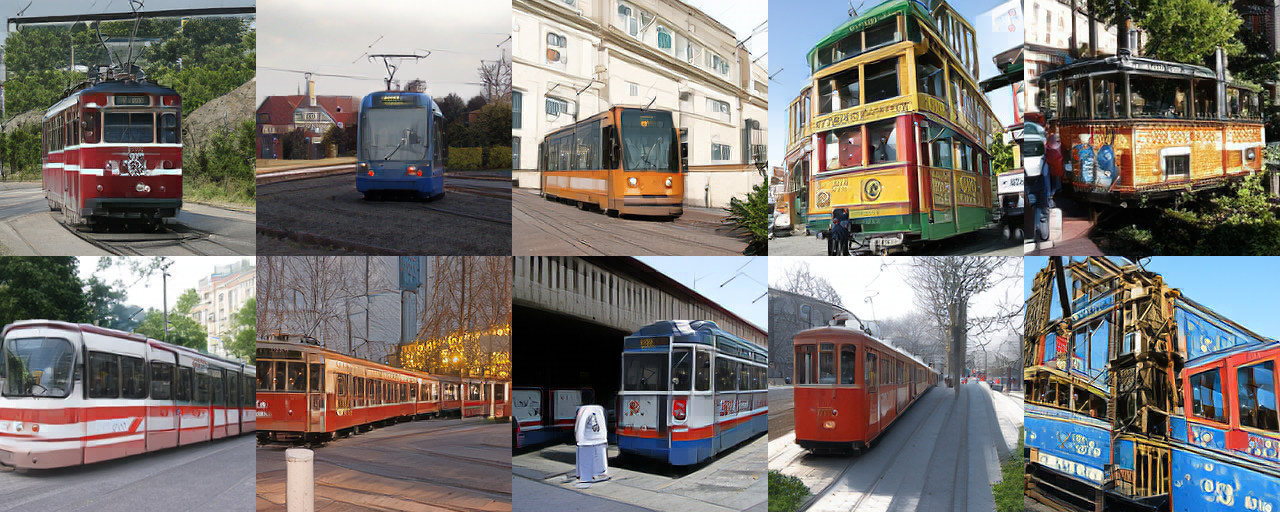}
\end{minipage}\hspace{0.025\textwidth}%
\begin{minipage}[t]{0.41\textwidth}\centering
{\scriptsize class 927: trifle\par}\vspace{1pt}
\includegraphics[width=\linewidth]{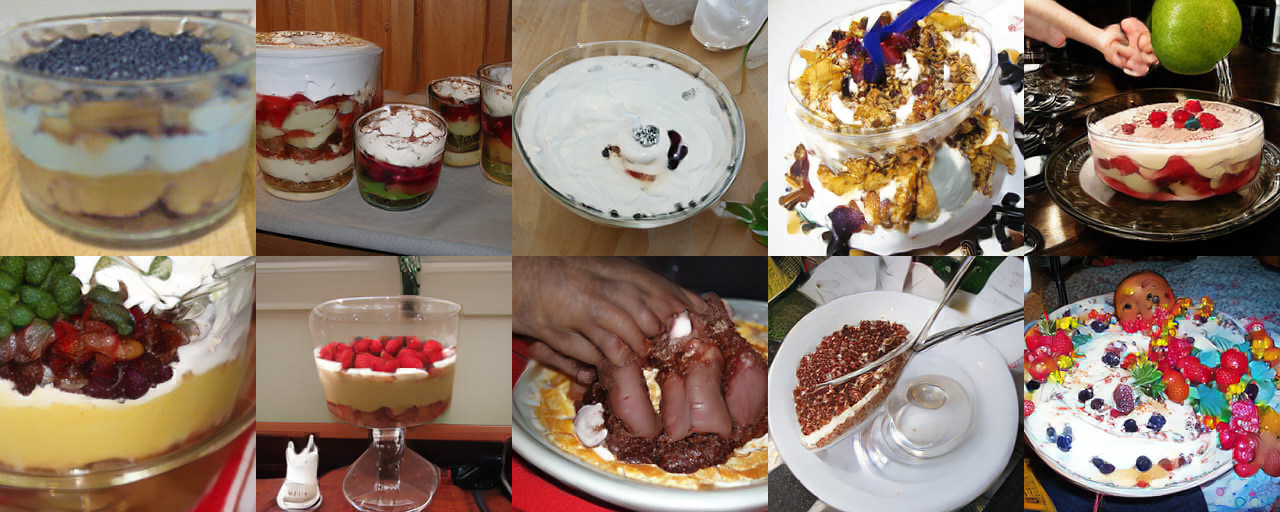}
\end{minipage}%
}\par\vspace{2pt}
\noindent\makebox[\textwidth][c]{%
\begin{minipage}[t]{0.41\textwidth}\centering
{\scriptsize class 958: hay\par}\vspace{1pt}
\includegraphics[width=\linewidth]{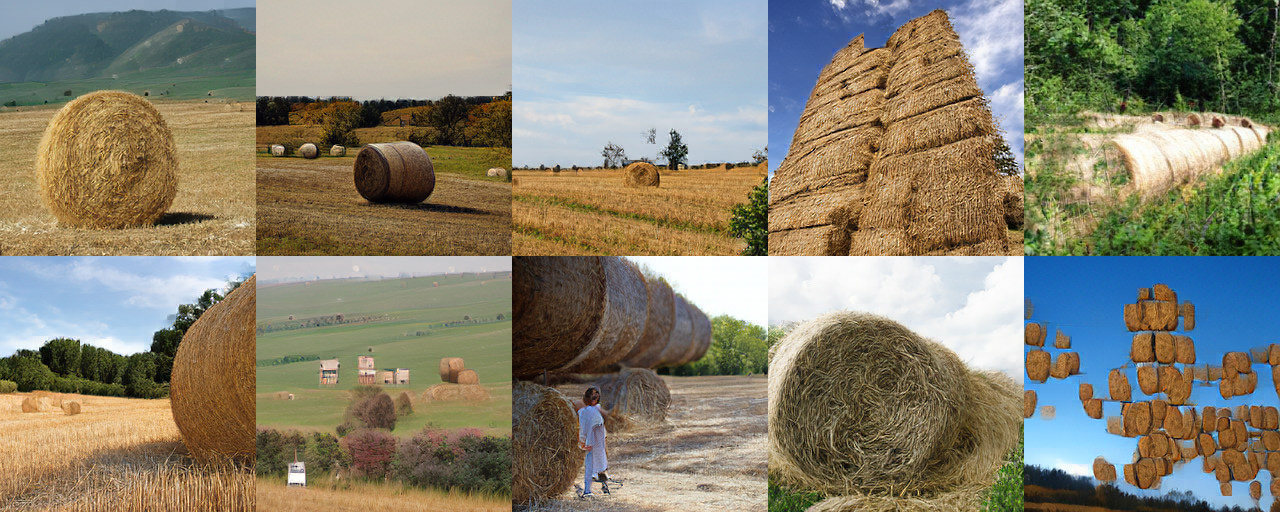}
\end{minipage}\hspace{0.025\textwidth}%
\begin{minipage}[t]{0.41\textwidth}\centering
{\scriptsize class 970: alp\par}\vspace{1pt}
\includegraphics[width=\linewidth]{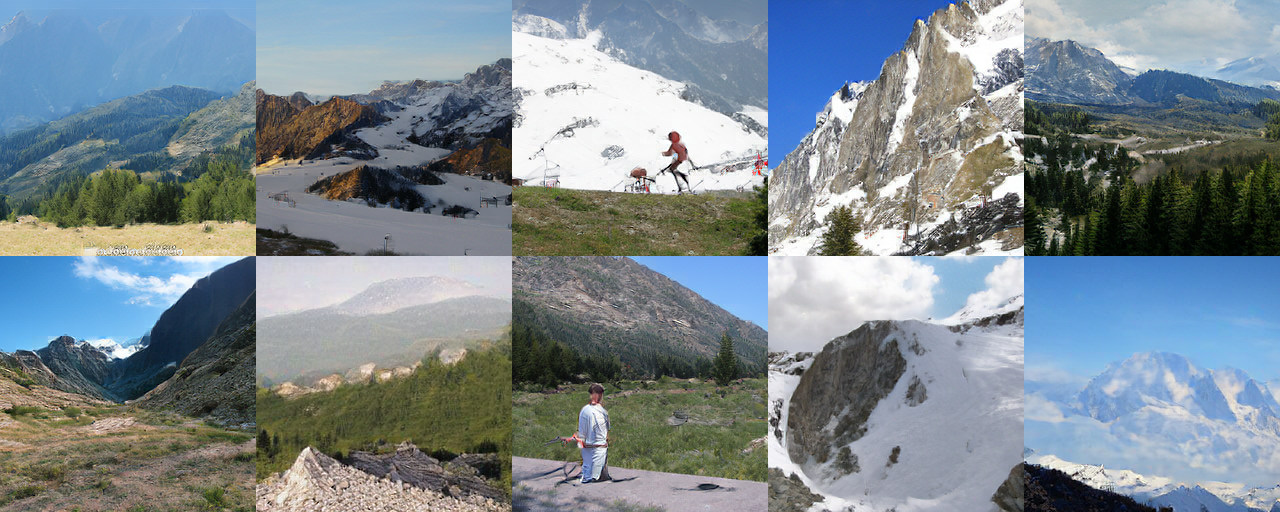}
\end{minipage}%
}\par\vspace{2pt}
\noindent\makebox[\textwidth][c]{%
\begin{minipage}[t]{0.41\textwidth}\centering
{\scriptsize class 974: geyser\par}\vspace{1pt}
\includegraphics[width=\linewidth]{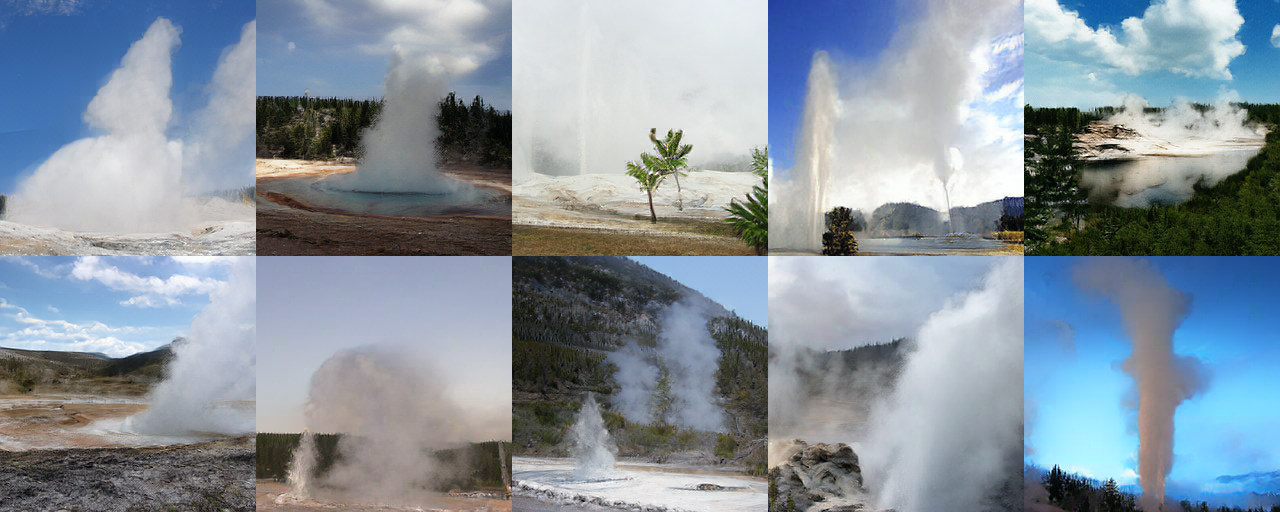}
\end{minipage}\hspace{0.025\textwidth}%
\begin{minipage}[t]{0.41\textwidth}\centering
{\scriptsize class 975: lakeside\par}\vspace{1pt}
\includegraphics[width=\linewidth]{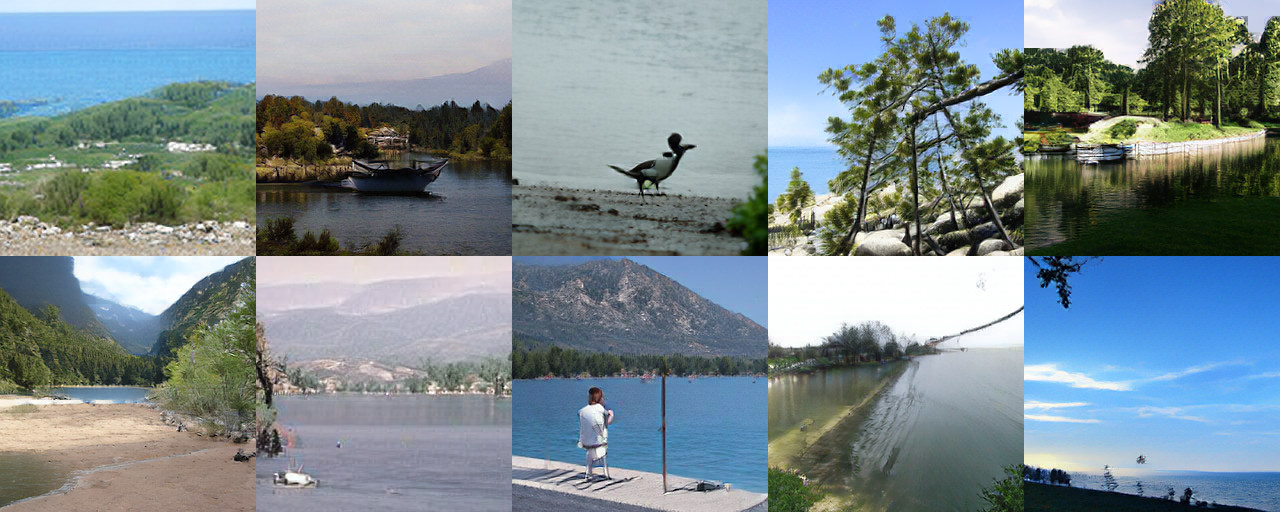}
\end{minipage}%
}\par\vspace{2pt}
\noindent\makebox[\textwidth][c]{%
\begin{minipage}[t]{0.41\textwidth}\centering
{\scriptsize class 979: valley\par}\vspace{1pt}
\includegraphics[width=\linewidth]{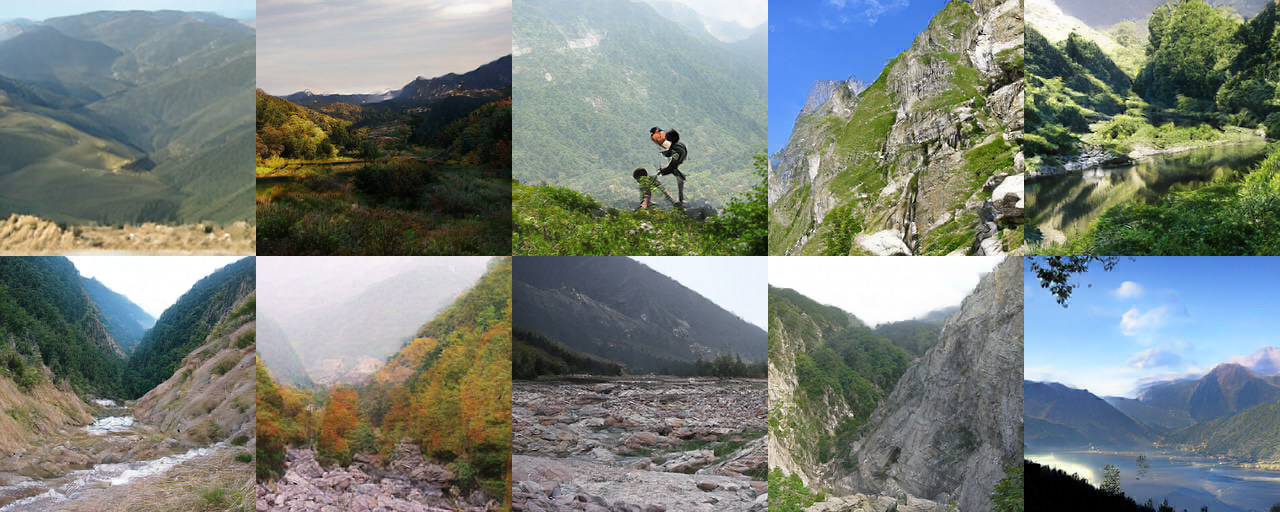}
\end{minipage}\hspace{0.025\textwidth}%
\begin{minipage}[t]{0.41\textwidth}\centering
{\scriptsize class 980: volcano\par}\vspace{1pt}
\includegraphics[width=\linewidth]{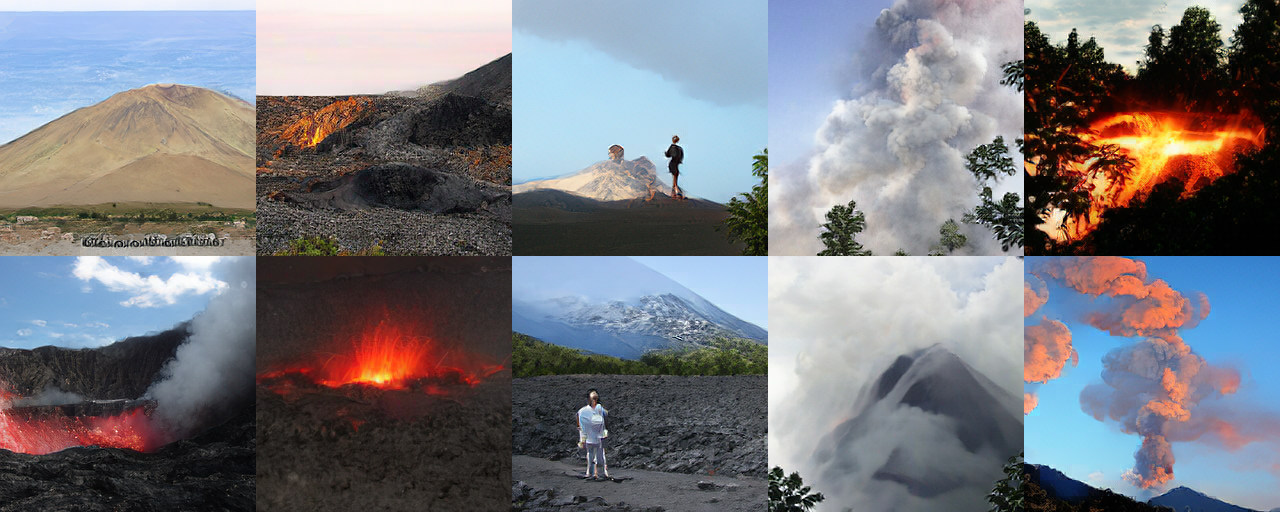}
\end{minipage}%
}\par\vspace{2pt}
\par\endgroup